%% file: main.tex
\documentclass{article}

\PassOptionsToPackage{sort,numbers}{natbib}
\usepackage[final]{neurips_data_2022}

\usepackage[utf8]{inputenc} % allow utf-8 input
\usepackage[T1]{fontenc}    % use 8-bit T1 fonts
\usepackage{hyperref}       % hyperlinks
\usepackage{url}            % simple URL typesetting
\usepackage{booktabs}       % professional-quality tables
\usepackage{amsfonts}       % blackboard math symbols
\usepackage{nicefrac}       % compact symbols for 1/2, etc.
\usepackage{microtype}      % microtypography
\usepackage{xcolor}         % colors
\usepackage{multirow}

\input{macros}

\newcommand\blfootnote[1]{%
  \begingroup
  \renewcommand\thefootnote{}\footnote{#1}%
  \addtocounter{footnote}{-1}%
  \endgroup
}

\title{NAS-Bench-Suite-Zero: \\Accelerating Research on Zero Cost Proxies}

\author{Arjun Krishnakumar$^{*1}$, 
    Colin White$^{*2}$, Arber Zela$^{*1}$, Renbo Tu$^{*3}$, \\
    \textbf{Mahmoud Safari$^1$, Frank Hutter$^{1,4}$}
    \vspace*{1mm}\\
    $^1$University of Freiburg,  $^2$Abacus.AI, 
    $^3$University of Toronto, \\
    $^4$Bosch Center for Artificial Intelligence
  }

\begin{document}
\maketitle

\input{0_abstract}

\input{1_introduction}

\input{2_background_and_related}
\input{3_suite_overview}

\input{4_analysis}
\input{5_nas}

\input{99_conclusions}

% acknowledgments
\begin{ack}
This research was supported by the following sources: 
Robert Bosch GmbH is acknowledged for financial support;
the German Federal Ministry of Education and Research (BMBF, grant RenormalizedFlows 01IS19077C);
TAILOR, a project funded by EU Horizon 2020 research and innovation programme under GA No 952215; the Deutsche Forschungsgemeinschaft (DFG, German Research Foundation) under grant number 417962828; the European Research Council (ERC) Consolidator Grant ``Deep Learning 2.0'' (grant no.\ 101045765). Funded by the European Union. Views and opinions expressed are however those of the author(s) only and do not necessarily reflect those of the European Union or the ERC. Neither the European Union nor the ERC can be held responsible for them.
\begin{center}\includegraphics[width=0.3\textwidth]{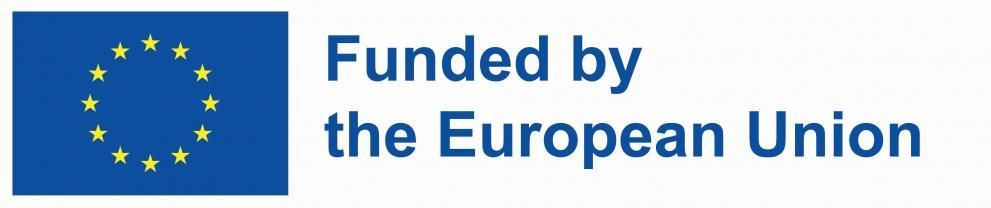}\end{center}
\end{ack}

\clearpage
\bibliography{main}
\bibliographystyle{plain}

% remove for arxiv:
%\input{A_checklist}

\clearpage
\appendix

% remove for arxiv:
%\input{B_nas_checklist}

\input{C_dataset_documentation}

\input{D_appendix_related_work}

\input{E_appendix}

\end{document}

%% file: macros.tex
\usepackage{longtable}

\usepackage{adjustbox}
\usepackage{array}

\usepackage{comment}
\usepackage{amsmath,amsthm,amssymb}
\usepackage{algorithm}
\usepackage{algorithmic}
\usepackage{multicol}
\usepackage{mathtools}
\usepackage{caption}
\usepackage{float}
\usepackage{graphicx}
\usepackage{subfigure}
\usepackage{wrapfig}
\usepackage{multirow}
\usepackage{mathrsfs}
\usepackage{newfloat}
\usepackage{relsize}
\usepackage{enumitem}
\usepackage{pifont}
\usepackage{bm}
\usepackage{lipsum}

\newcolumntype{R}[2]{%
    >{\adjustbox{angle=#1,lap=\width-(#2)}\bgroup}%
    l%
    <{\egroup}%
}

\newcolumntype{T}{>$l<$}
\newcolumntype{L}{>{\arraybackslash}m{4cm}}
\newcolumntype{S}{>{\arraybackslash}m{3cm}}

\newcommand{\cmark}{\color{green}\ding{51}}%
\newcommand{\xmark}{\color{red}\ding{55}}%

\newcommand{\Y}{\mathcal{Y}}
\newcommand{\Z}{\mathcal{Z}}
\newcommand{\suite}{\texttt{NAS-Bench-Suite-Zero}}%

\newcommand{\nproxies}{13}%
\newcommand{\ntasks}{28}%
\newcommand{\nevals}{1.5 million}%

%% file: 0_abstract.tex
\begin{abstract}
Zero-cost proxies (ZC proxies) are a recent architecture performance prediction technique aiming to significantly speed up algorithms for neural architecture search (NAS). 
Recent work has shown that these techniques show great promise, but certain aspects, such as evaluating and exploiting their complementary strengths, are under-studied.
In this work, we create \suite: we evaluate \nproxies{} ZC proxies across \ntasks{} tasks, creating by far the largest dataset (and unified codebase) for ZC proxies, enabling orders-of-magnitude faster experiments on ZC proxies, while avoiding confounding factors stemming from different implementations. To demonstrate the usefulness of \suite, we run a large-scale analysis of ZC proxies, including a bias analysis, and the first information-theoretic analysis which concludes that ZC proxies capture substantial complementary information. 
Motivated by these findings, we present a procedure to improve the performance of ZC proxies by reducing biases such as cell size, and we also show that incorporating all \nproxies{} ZC proxies into the surrogate models used by NAS algorithms can improve their predictive performance by up to 42\%.
Our code and datasets are available at \url{https://github.com/automl/naslib/tree/zerocost}.
\end{abstract}

%Zero-cost proxies (ZC proxies) are a recent architecture performance prediction technique aiming to significantly speed up algorithms for neural architecture search (NAS). Recent work has shown that these techniques show great promise, but certain aspects, such as evaluating and exploiting their complementary strengths, are under-studied. In this work, we create NAS-Bench-Suite: we evaluate 13 ZC proxies across 28 tasks, creating by far the largest dataset (and unified codebase) for ZC proxies, enabling orders-of-magnitude faster experiments on ZC proxies, while avoiding confounding factors stemming from different implementations. To demonstrate the usefulness of NAS-Bench-Suite, we run a large-scale analysis of ZC proxies, including a bias analysis, and the first information-theoretic analysis which concludes that ZC proxies capture substantial complementary information. Motivated by these findings, we present a procedure to improve the performance of ZC proxies by reducing biases such as cell size, and we also show that incorporating all 13 ZC proxies into the surrogate models used by NAS algorithms can improve their predictive performance by up to 42%. Our code and datasets are available at https://github.com/automl/naslib/tree/zerocost.

%% file: 1_introduction.tex
%\vspace{-1mm}
\section{Introduction} \label{sec:introduction}
\vspace{-2mm}

Algorithms for neural architecture search (NAS) seek to automate the design
of high-performing neural architectures for a given dataset.
NAS has successfully been used to discover architectures with better
accuracy/latency tradeoffs than the best human-designed architectures
\citep{dai2021fbnetv3, efficientnets, real2019regularized, nas-survey}.
Since early NAS algorithms were prohibitively expensive to run
\citep{zoph2017neural}, a long line of recent work has focused on
improving the runtime and efficiency of NAS methods (see \citep{nas-survey,wistuba2019survey} for recent surveys).
\blfootnote{$^*$Equal contribution. Work done while RT was part-time at Abacus.AI.
    Email to: \\
        \texttt{\{krishnan, zelaa, fh\}@cs.uni-freiburg.de},
        \texttt{colin@abacus.ai},
        \texttt{renbo.tu@mail.utoronto.ca},    
        \texttt{safarim@informatik.uni-freiburg.de}.
}

A recent thread of research within NAS focuses on
\emph{zero-cost proxies} (ZC proxies) \citep{mellor2021neural, abdelfattah2021zerocost}.
These novel techniques aim
to give an estimate of the (relative) performance
of neural architectures from just a \emph{single minibatch of data}.
Often taking just five seconds to run, these techniques are essentially
``zero cost'' compared to training an architecture or to any other method of predicting the performance of neural architectures \citep{white2021powerful}.
Since the initial ZC proxy was introduced \citep{mellor2021neural},
there have been many follow-up methods \citep{abdelfattah2021zerocost, lin2021zen}.
However, several recent works have shown that simple baselines such as
``number of parameters'' and ``FLOPS'' are competitive with all existing
ZC proxies across most settings, and that most ZC proxies do not generalize well across different benchmarks, thus requiring broader large-scale evaluations in order to assess their strengths \citep{ning2021evaluating, chen2021bench}.
A recent landscape overview concluded that ZC proxies show great promise, but certain aspects are under-studied and their true potential has not been realized thus far \citep{colin2022adeeperlook}.
In particular, it is still largely unknown whether ZC proxies can be effectively
combined, and how best to integrate ZC proxies into NAS algorithms.
%into existing NAS techniques.

\begin{figure}[t]
    \centering
    \includegraphics[width=.9\linewidth]{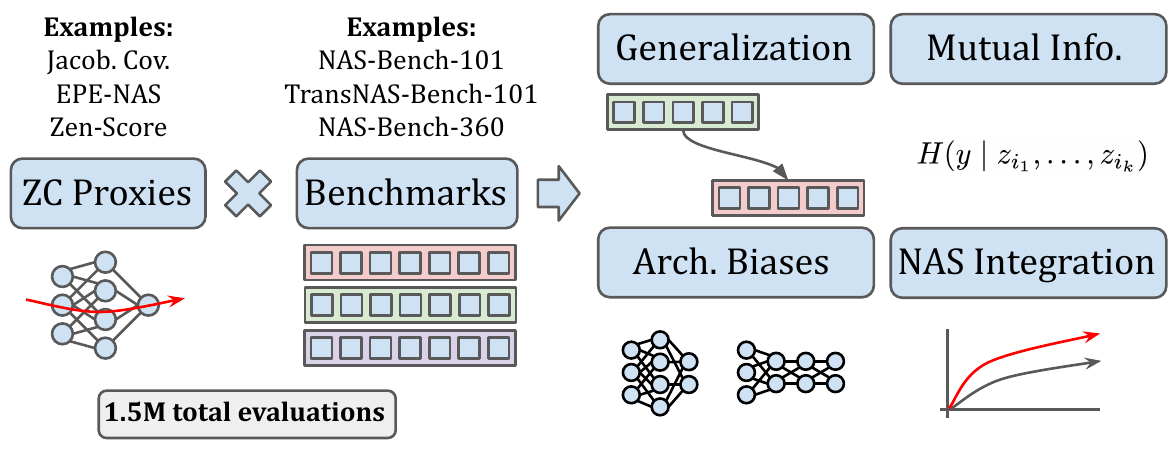}
    \caption{Overview of \suite. We implement and pre-compute \nproxies{} ZC proxies on \ntasks{} tasks in a unified framework, and then use this dataset to analyze the generalizability, complementary information, biases, and NAS integration of ZC proxies. 
    %Motivated by these analyses, we show that NAS algorithms can benefit from an ensemble of \nproxies{} ZC proxies.
     }%\vspace{-9mm}
    \label{fig:overview}
\end{figure}

In this work, we introduce \suite: a unified and extensible collection of \nproxies{} ZC proxies, accessible through a unified interface, which can be evaluated on a suite of \ntasks{} tasks through \texttt{NASLib} \citep{ruchte2020naslib} (see Figure \ref{fig:overview}).
In addition to the codebase itself, we release precomputed ZC proxy scores across all \nproxies{} ZC proxies and \ntasks{} tasks, which can be used to speed up ZC proxy experiments. Specifically, we show that the runtime of ZC proxy experiments such as NAS analyses and bias analyses are shortened by a factor of at least $10^3$ when using the precomputed ZC proxies in \suite.  By providing a unified framework with ready-to-use scripts to run large-scale experiments, \suite{} eliminates the overhead for researchers to compare against many other methods and across all popular NAS benchmark search spaces, helping the community to rapidly increase the speed of research in this promising direction. 
Our benchmark suite was very recently used successfully in the Zero Cost NAS Competition at AutoML-Conf 2022. See Appendix \ref{app:competition} for more details.
In Appendix \ref{app:documentation}, we give detailed documentation, including a datasheet \citep{gebru2021datasheets}, license, author responsibility, code of conduct, and maintenance plan.
We welcome contributions from the community and hope to grow the repository and benchmark suite as more ZC proxies and NAS benchmarks are released.

To demonstrate the usefulness of \suite, we run a large-scale analysis of ZC proxies: we give a thorough study of generalizability and biases, and we give the first information-theoretic analysis. 
Interestingly, based on the bias study, \textbf{we present a concrete method for improving the performance of a ZC proxy by reducing biases} (such as the tendency to favor larger architectures or architectures with more \texttt{conv} operations).  
This may have important consequences for the future design of ZC proxies.
Furthermore, based on the information-theoretic analysis, we find that there is high information gain of the validation accuracy when conditioned on multiple ZC proxies, suggesting that ZC proxies do indeed compute substantial complementary information.
Motivated by these findings, we incorporate all \nproxies{} proxies into the surrogate models used by NAS algorithms \citep{bananas, npenas}, showing that the Spearman rank correlation of the surrogate predictions can \emph{increase by up to 42\%}.
We show that this results in improved performance for two predictor-based NAS algorithms: BANANAS \citep{bananas} and NPENAS \citep{npenas}.
%, two predictor-guided NAS algorithms based on Bayesian optimization and evolution, respectively. 

\noindent\textbf{Our contributions.}
We summarize our main contributions below.
\begin{itemize}[topsep=0pt, itemsep=2pt, parsep=0pt, leftmargin=5mm]
    \item We release \suite, a collection of benchmarks and ZC proxies that unifies and accelerates research on ZC proxies -- a promising new sub-field of NAS -- by enabling orders-of-magnitude faster evaluations on a large suite of diverse benchmarks.   
    \item 
    We run a large-scale analysis of \nproxies{} ZC proxies across \ntasks{} different combinations of search spaces and tasks by studying the generalizability, bias, and mutual information among ZC proxies.
    \item 
    Motivated by our analysis, 
    we present a procedure to improve the performance of ZC proxies by reducing biases, and
    we show that the complementary information of ZC proxies can significantly improve the predictive power of surrogate models commonly used for NAS.
\end{itemize}

%% file: 2_background_and_related.tex
%\vspace{-1mm}
\section{Background and Related Work} \label{sec:background}
%\vspace{-1mm}
Given a dataset and a \emph{search space} -- a large set of neural architectures --
NAS seeks to find the architecture with the highest validation accuracy (or the best application-specific trade-off among accuracy, latency, size, and so on) on the dataset. 
NAS has been studied since the late 1980s \citep{tenorio1988self, miller1989designing}
and has seen a resurgence in the last few years \citep{zoph2017neural, pnas}, with over 1000 papers on NAS in the last two years alone.
%Common techniques include reinforcement learning \citep{zoph2017neural}, evolutionary search \citep{real2019regularized}, Bayesian optimization \citep{nasbot}, and gradient-based optimization \citep{darts, wang2021rethinking}.
For a survey of the different techniques used for NAS, see \citep{nas-survey,wistuba2019survey}.

Many NAS methods make use of performance prediction.
A \emph{performance prediction} method is any function which predicts the (relative) performance of architectures, without fully training the architectures \citep{white2021powerful}. 
BRP-NAS \citep{brpnas}, BONAS \citep{bonas}, and BANANAS \citep{bananas} are all examples of NAS methods that make use of performance prediction.
While performance prediction speeds up NAS algorithms by avoiding 
fully training neural networks, many still require non-trivial computation time.
%For example, although \emph{learning curve extrapolation} \citep{domhan2015speeding, baker2017accelerating} cuts down training time by around 90\%, predictions still take at least 10 minutes per architecture.
On the other hand, a recently-proposed line of techniques, \emph{zero-cost proxies} (ZC proxies) require just a single forward pass through the network, often taking just five seconds \citep{mellor2021neural}.

\begin{table}[t]
\caption{List of ZC proxies in \suite. 
%The meaning of columns are as follows.
%``Data-dependent'' denotes whether the ZC proxy depends on the training data; 
Note that ``neuron-wise'' denotes whether the total score is a sum of individual weights.
%``Type'': Type of ZC proxy.
}
%\vspace{1mm}
%\vspace{3mm}
\resizebox{\linewidth}{!}{%
\centering
%\begin{adjustbox}{width=0.8\columnwidth}
\begin{tabular}{@{}l|c|c|c|c@{}}
\toprule
\multicolumn{1}{l}{\textbf{Name}} & \multicolumn{1}{c}{\textbf{Data-dependent}} & \multicolumn{1}{c}{\textbf{Neuron-wise}} & \multicolumn{1}{c}{\textbf{Type}} &  \multicolumn{1}{c}{\textbf{In \suite}} \\
\midrule 
\texttt{epe-nas} \citep{lopes2021epe} & \cmark & \xmark & Jacobian & \cmark \\
\texttt{fisher} \citep{turner2020blockswap} & \cmark & \cmark & Pruning-at-init & \cmark\\
\texttt{flops} \citep{ning2021evaluating} & \cmark & \cmark & Baseline & \cmark \\
%\texttt{gen-nas} \citep{li2021generic} & \xmark & \cmark & \xmark & \cmark \\
\texttt{grad-norm} \citep{abdelfattah2021zerocost} & \cmark & \cmark & Pruning-at-init & \cmark \\
\texttt{grasp} \citep{wang2020picking} & \cmark & \cmark & Pruning-at-init & \cmark\\
\texttt{l2-norm} \citep{abdelfattah2021zerocost} & \xmark & \xmark & Baseline & \cmark\\
\texttt{jacov} \citep{mellor2021neural} & \cmark & \xmark & Jacobian & \cmark  \\
\texttt{nwot} \citep{mellor2021neural} & \cmark & \xmark & Jacobian & \cmark \\
\texttt{params} \citep{ning2021evaluating} & \xmark & \cmark & Baseline & \cmark  \\
\texttt{plain} \citep{abdelfattah2021zerocost} & \cmark & \cmark & Baseline & \cmark \\
\texttt{snip} \citep{lee2018snip} & \cmark & \cmark & Pruning-at-init & \cmark  \\
\texttt{synflow} \citep{tanaka2020pruning} & \xmark & \cmark & Pruning-at-init & \cmark  \\
\texttt{zen-score} \citep{lin2021zen} & \xmark & \xmark & Piece.\ Lin.\ & \cmark  \\
\bottomrule
\end{tabular}
%\end{adjustbox}
}
\label{tab:proxies}
\end{table}

\vspace{-2mm}
\paragraph{Zero-cost proxies.}
The original ZC proxy estimated the separability of the minibatch of data into different linear regions of the output space \citep{mellor2021neural}.
% todo: this was then updated to xxx.
Many other ZC proxies have been proposed since then, including data-independent ZC proxies \citep{abdelfattah2021zerocost, tanaka2020pruning, li2021generic, lin2021zen}, ZC proxies inspired by pruning-at-initialization techniques \citep{abdelfattah2021zerocost, lee2018snip, wang2020picking, tanaka2020pruning}, and ZC proxies inspired by neural tangent kernels \citep{shu2022nasi, chen2021neural}.
See Table \ref{tab:proxies} for a full list of the ZC proxies we use in this paper.
We describe theoretical ZC proxy results in Appendix \ref{app:zcp_theory}.

\vspace{-2mm}
\paragraph{Search spaces and tasks.}
In our experiments, we make use of several different NAS benchmark search spaces and tasks.
NAS-Bench-101 \citep{nasbench} is a popular cell-based search space for NAS research.
It consists of 423\,624 architectures trained on CIFAR-10. The cell-based search space is designed to model ResNet-like and Inception-like cells \citep{resnet, inceptionnet}.
NAS-Bench-201 \citep{nasbench201} is a cell-based search space consisting of 15\,625 architectures (6\,466 non-isomorphic) trained on CIFAR-10, CIFAR-100, and ImageNet16-120.
NAS-Bench-301 \citep{nasbench301} is a surrogate NAS benchmark for the DARTS search space \citep{darts}.
The search space consists of normal cell and reduction cells, with $10^{18}$ total architectures. 
%The surrogate model is created using XGBoost.
%
TransNAS-Bench-101 \citep{transnasbench} is a NAS benchmark consisting of two different search spaces:
a ``micro'' (cell-based) search space of size 4\,096, and a macro search space of size 3\,256.
The architectures are trained on seven different tasks from the Taskonomy dataset \citep{zamir2018taskonomy}.
%, which include object classification, scene classification, jigsaw solving, and image upscaling.
NAS-Bench-Suite \citep{nasbenchsuite} collects these search spaces and tasks within the unified framework of \texttt{NASLib} \citep{ruchte2020naslib}. 
In this work, we extend this collection by adding two datasets from NAS-Bench-360 \citep{nasbench360}, SVHN, and four datasets from Taskonomy.
NAS-Bench-360 is a collection of diverse tasks that are ready-to-use for NAS research.

\vspace{-2mm}
\paragraph{Large-scale studies of ZC proxies.}
A few recent works \citep{white2021powerful, ning2021evaluating, colin2022adeeperlook, chen2021bench} investigated the performance of ZC proxies in ranking architectures over different NAS benchmarks, showing that the relative performance highly depends on the search space, but none study more than 12 total tasks, and none make the ZC proxy values publicly available.
Two predictor-based NAS methods have recently been introduced: OMNI \citep{white2021powerful} and ProxyBO \citep{shen2021proxybo}. However, OMNI only uses a single ZC proxy, and while ProxyBO uses three, the algorithm dynamically chooses one in each iteration (so individual predictions are made using a single ZC proxy at a time). 
Recently, NAS-Bench-Zero was introduced \citep{chen2021bench}, a new benchmark based on popular computer vision models ResNet \citep{resnet} and MobileNetV2 \citep{sandler2018mobilenetv2}, which includes 10 ZC proxies. However, the NAS-Bench-Zero dataset is currently not publicly available.
For more related work details, see Appendix \ref{app:related_work}.
% todo: mention that they precomputed zc proxies on their three search spaces.

Only two prior works combine the information of multiple ZC proxies together in architecture predictions \citep{abdelfattah2021zerocost, chen2021bench} and both only use the \emph{voting} strategy to combine at most four ZC proxies. Our work is the first to publicly release ZC proxy values, combine ZC proxies in a nontrivial way, and exploit the complementary information of \nproxies{} ZC proxies simultaneously.

%% file: 3_suite_overview.tex
\vspace{-1mm}
\section{Overview of NAS-Bench-Suite-Zero} \label{sec:suite_overview}
\vspace{-1mm}
In this section, we give an overview of the \suite{} codebase and dataset, which allows researchers to quickly develop ZC proxies, compare against existing ZC proxies across diverse datasets, and integrate them into NAS algorithms, as shown in Sections \ref{sec:analysis} and \ref{sec:nas}. 
%Overall, \suite{} includes 1.5M ZC proxy evaluations with 1100 CPU hours of computation.

\begin{table}[t]
\caption{Overview of ZC proxy evaluations in \suite.
$^*$ Note that EPE-NAS is only defined for classification tasks \citep{lopes2021epe}.
}
%\vspace{1mm}
%\vspace{3mm}
\resizebox{\linewidth}{!}{%
\centering
%\begin{adjustbox}{width=0.8\columnwidth}
\begin{tabular}{@{}l|c|c|c|c@{}}
\toprule
\multicolumn{1}{l}{\textbf{Search space}} & \multicolumn{1}{c}{\textbf{Tasks}} & 
\multicolumn{1}{c}{\textbf{Num.\ ZC proxies}} & 
\multicolumn{1}{c}{\textbf{Num.\ architectures}} &
\multicolumn{1}{c}{\textbf{Total ZC proxy evaluations}} \\
\midrule
NAS-Bench-101 & 1 & 13 & 10\,000 & 130\,000 \\
NAS-Bench-201 & 3 & 13 & 15\,625 & 609\,375 \\
NAS-Bench-301 & 1 & 13 & 11\,221 & 145\,873 \\
TransNAS-Bench-101-Micro & 7 & 12$^*$ & 3\,256 & 273\,504 \\
TransNAS-Bench-101-Macro & 7 & 12$^*$ & 4\,096 & 344\,064 \\
Add'l.\ 201, 301, TNB-Micro & 9 & 13 & 600 & 23400 \\
%\textbf{Total} & 18 & 13 & 34\,198 & 1\,314\,720 \\
\textbf{Total} & 28 & 13 & 44\,798 & 1\,526\,216 \\
\bottomrule
\end{tabular}
%\end{adjustbox}
}
\label{tab:evaluations}
\end{table}

%\vspace{-3mm}
%\paragraph{Experimental setup.}
%We describe the process of creating the \suite{} dataset. 
We implement all ZC proxies from Table \ref{tab:proxies} in the same codebase (\texttt{NASLib} \citep{ruchte2020naslib}). For all ZC proxies, we use the default implementation from the original work.
While this list covers \nproxies{} ZC proxies, the majority of ZC proxies released to date, we did not yet include a few other ZC proxies, for example, due to requiring a trained supernetwork to make evaluations \citep{shu2022nasi, chen2021neural} (therefore needing to implement a supernetwork on \ntasks{} benchmarks), implementation in TensorFlow rather than PyTorch \citep{park2020towards}, or unreleased code.
%~\citep{Zhou2022Quasiorthogonality, Shu2022Unifying}.
Our modular framework easily allows additional ZC proxies to be added to \suite{} in the future.

%We run experiments using the NAS-Bench-Suite \citep{nasbenchsuite} collection of NAS benchmarks in NASLib, which we extended with six additional datasets. 
To build \suite, we extend the collection of \texttt{NASLib}'s publicly available benchmarks, known as NAS-Bench-Suite \citep{nasbenchsuite}.
This allows us to evaluate and fairly compare all ZC proxies in the same framework without confounding factors stemming from different implementations, software versions or training pipelines. 
Specifically, for the search spaces and tasks, we use 
NAS-Bench-101 (CIFAR-10),
NAS-Bench-201 (CIFAR-10, CIFAR-100, and ImageNet16-120), 
NAS-Bench-301 (CIFAR-10), and
TransNAS-Bench-101 Micro and Macro (Jigsaw, Object Classification, Scene Classification, Autoencoder) from NAS-Bench-Suite. 
We add the remaining tasks from TransNAS-Bench-101 (Room Layout, Surface Normal, Semantic Segmentation), and three tasks each for NAS-Bench-201, NAS-Bench-301, and TransNAS-Bench-101-Micro: Spherical-CIFAR-100, NinaPro, and SVHN. This yields a total of \ntasks{} benchmarks in our analysis.
For all NAS-Bench-201 and TransNAS-Bench-101 tasks, we evaluate all ZC proxy values and the respective runtimes, for all architectures.
For NAS-Bench-301, we evaluate on all 11\,221 randomly sampled architectures from the NAS-Bench-301 dataset, due to the computational infeasibility of exhaustively evaluating the full set of $10^{18}$ architectures. Similarly, we evaluate 10\,000 architectures from NAS-Bench-101.
Finally, for Spherical-CIFAR-100, NinaPro, and SVHN, we evaluate 200 architectures per search space, since only 200 architectures are fully trained for each of these tasks.
%since in these cases we had to train the architectures from scratch.
See Table \ref{tab:evaluations}.

We run all ZC proxies from Table \ref{tab:proxies} on Intel Xeon Gold 6242 CPUs and save their evaluations in order to create a queryable table with these pre-computed values.
We use a batch size of 64 for all ZC proxy evaluations, except for the case of TransNAS-Bench-101: due to the extreme memory usage of the Taskonomy tasks ($>30$GB memory), we used a batch size of 32. The total computation time for all 1.5M evaluations was 1100 CPU hours.

%\vspace{-3mm}
\paragraph{Speedups and recommended usage.}
The average time to compute a ZC proxy across all tasks is 2.6 seconds, and the maximum time (computing \texttt{grasp} on TNB-Macro Autoencoder) is 205 seconds, compared to $10^{-5}$ seconds when instead querying the \suite API.

When researchers evaluate ZC proxy-based NAS algorithms using queryable NAS benchmarks, the bottleneck is often (ironically) the ZC proxy evaluations. For example, for OMNI \citep{white2021powerful} or ProxyBO \citep{shen2021proxybo} running for 100 iterations and 100 candidates per iteration, the total evaluation time is roughly 9 hours, yet they can be run on \suite{} in under one minute.
Across all experiments done in this paper (mutual information study, bias study, NAS study, etc.), we calculate that using \suite{} decreases the computation time by at least three orders of magnitude. See Appendix \ref{app:speedup} for more details.

Since \suite{} reduces the runtime of experiments by at least three orders of magnitude (on queryable NAS benchmarks), we recommend researchers take advantage of \suite{} to \emph{(i)} run hundreds of trials of ZC proxy-based NAS algorithms, to reach statistically significant conclusions, \emph{(ii)} run extensive ablation studies, including the type and usage of ZC proxies, and \emph{(iii)} increase the total number of ZC proxies evaluated in the NAS algorithm.
Finally, when using \suite, researchers should report the real-world time NAS algorithms would take, by adding the time to run each ZC proxy evaluation (which can be queried in \suite) to the total runtime of the
NAS algorithm.
%architecture evaluations (which can also be queried for benchmarks in NAS-Bench-Suite), and other overheads of the NAS algorithm.

%% file: 4_analysis.tex
\vspace{-1mm}
\section{Generalizability, Mutual Information, and Bias of ZC Proxies} \label{sec:analysis}
\vspace{-1mm}

In this section, we use \suite{} to study concrete research questions relating to the generalizability, complementary information,  and bias of ZC proxies.
%across a diverse set of benchmarks. 
%With these experiments, we aim to answer the following concrete research questions:

\begin{comment}
\begin{itemize}[topsep=0pt, itemsep=2pt, parsep=0pt, leftmargin=5mm]
    \item \textbf{RQ 1}: \emph{How well do ZC proxies generalize across different benchmarks?}
    %Do they capture similarities among benchmarks?}    
    \item \textbf{RQ 2}: \emph{Are ZC proxies complementary with respect to explaining validation accuracy?}
    \item \textbf{RQ 3}: \emph{Do ZC proxies contain biases, such as a bias toward certain operations or sizes?}
%\item \textbf{RQ 3}: Are ZC proxies strongly correlated with each other or are they complementary?
    %Do (similar) ZC proxy predictions correlate to each other when evaluated on the same benchmark?
\end{itemize}
\end{comment}

\vspace{-1mm}
\subsection{RQ 1: How well do ZC proxies generalize across different benchmarks?} \label{subsec:generalization}
\vspace{-1mm}

\begin{figure}[t]
    \centering
    \includegraphics[width=\linewidth]{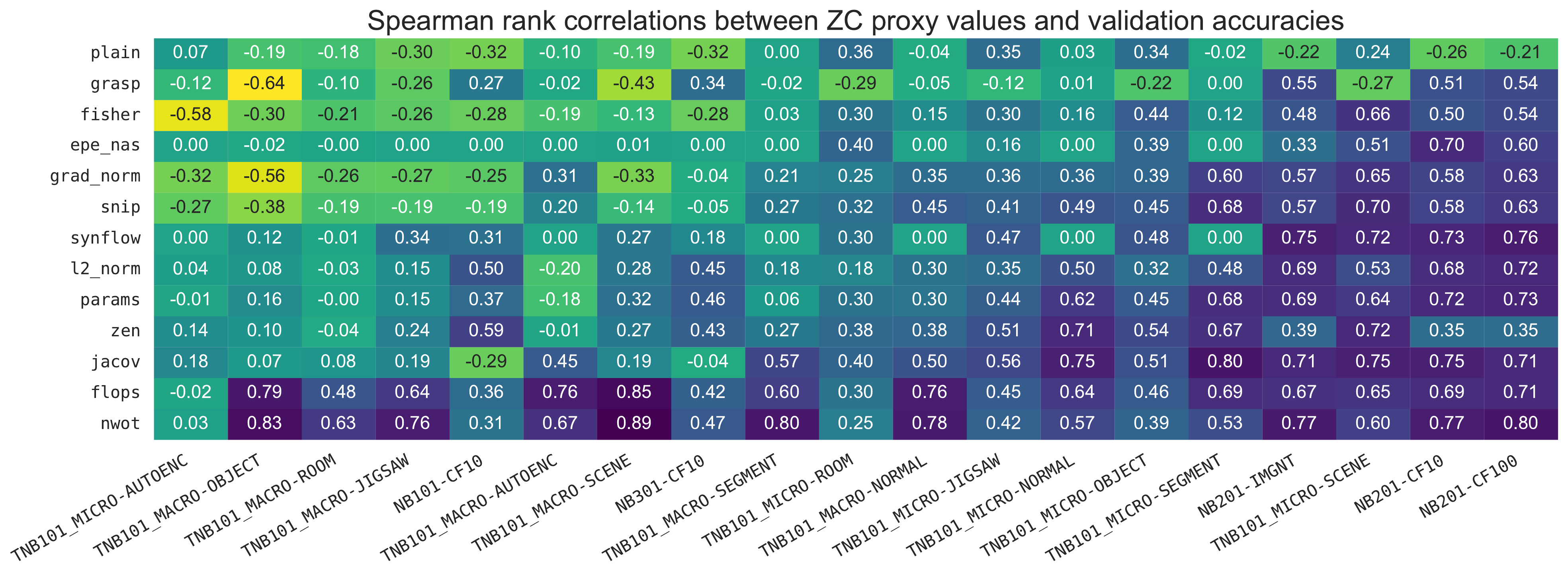}
    \caption{Spearman rank correlation coefficient between ZC proxy values and validation accuracies, for each ZC proxy and benchmark. 
    The rows and columns are ordered based on the mean scores across columns and rows, respectively. 
    %Note that some ZC proxies only work on image classification tasks. In such cases we provide an value of 0 in the plot.
    }
    \label{fig:corr_zcp}
\end{figure}

In Figure \ref{fig:corr_zcp}, for each ZC proxy and each benchmark, we compute the Spearman rank correlation between the ZC proxy values and the validation accuracies over a set of 1000 randomly drawn architectures (see Appendix \ref{app:analysis} for the full results on all benchmarks).
Out of all the ZC proxies, \texttt{nwot} and \texttt{flops} have the highest rank correlations across all benchmarks. On some of the benchmarks, such as TransNAS-Bench-101-Micro Autoencoder and Room Layout, all of the ZC proxies exhibit poor performance on average, while on the widely used NAS-Bench-201 benchmarks, almost all of them perform well. 
Several methods, such as \texttt{snip} and \texttt{grasp}, perform well on the NAS-Bench-201 tasks, but on average are outperformed by \texttt{params} and \texttt{flops} on the other benchmarks.

\begin{wrapfigure}{r}{0.7\textwidth}
\vspace{-3mm}
\begin{center}
    \centering
\includegraphics[width=0.7\textwidth]{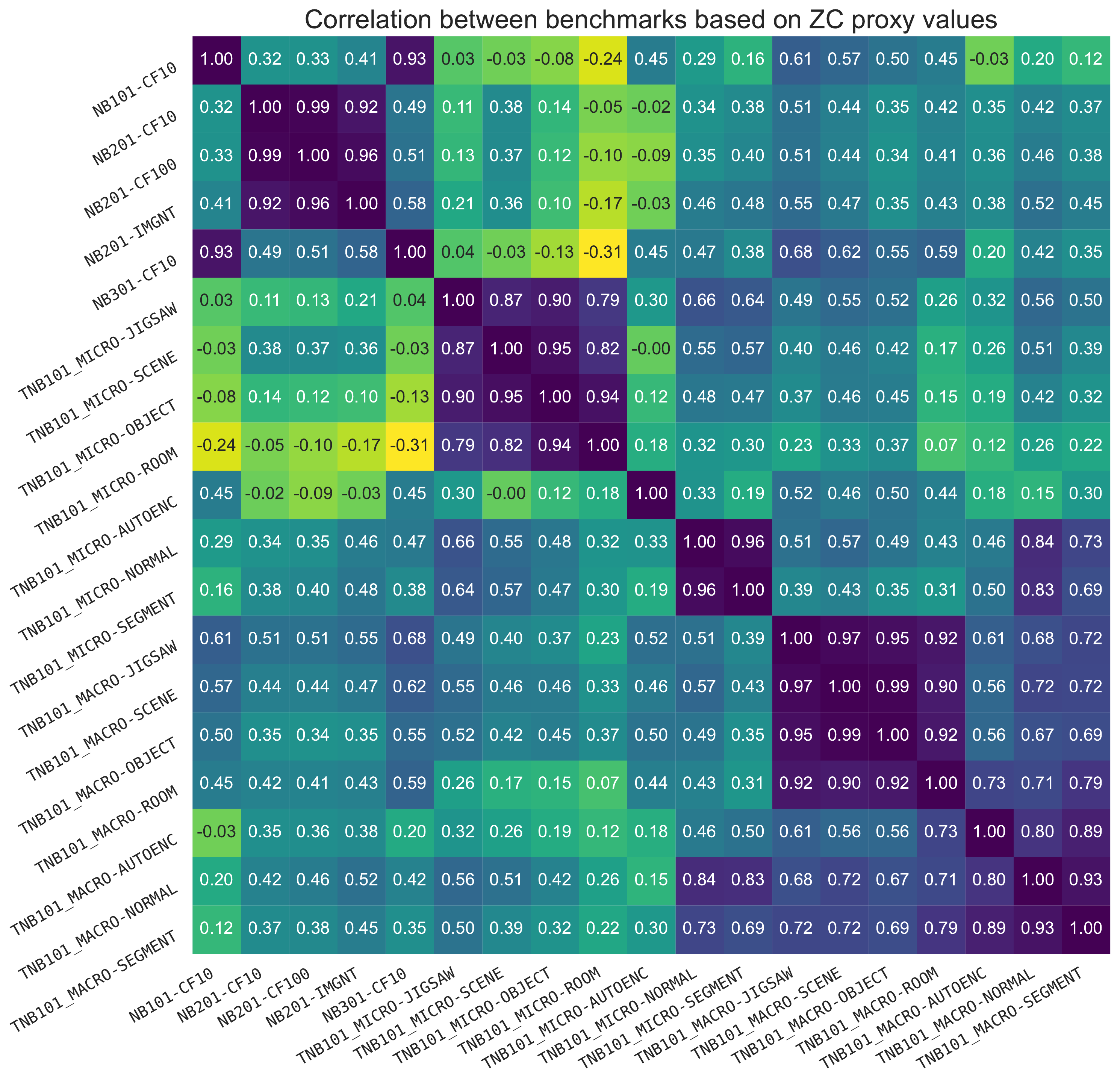}
\end{center}
\caption{Pearson correlation coefficient between ZC proxy scores on pairs of benchmarks. The entries in the plot are ordered based on the mean score across each row and column.}		\label{fig:xcorr_zcp_bench}
\vspace{-3mm}
\end{wrapfigure}

%todo: if short on space, we can summarize this result and put it in the appendix.
Although no ZC proxy performs consistently across all benchmarks, we may ask a related question: is the performance of all ZC proxies across benchmarks correlated enough to capture similarities among benchmarks? In other words, can we use ZC proxies as a tool to assess the similarities among tasks. This is particularly important in meta-learning or transfer learning, where a meta-algorithm aims to learn and transfer knowledge across a set of similar tasks. To answer this question, we compute the Pearson correlation of the ZC proxy scores on each pair of benchmarks. See Figure \ref{fig:xcorr_zcp_bench}. As expected, benchmarks that are based on the same or similar search spaces are highly correlated with respect to the ZC proxy scores. For example, we see clusters of high correlation for the Trans-NAS-Bench-101-Macro benchmarks, and the NAS-Bench-201 benchmarks. 
%Overall, this leads us to the following answer to RQ 1:

\noindent\textbf{Answer to RQ 1:} \emph{Only a few ZC proxies generalize well across most benchmarks and tasks. However, ZC proxies can be used to assess similarities across benchmarks.}
This suggests the potential future direction of incorporating them as task features in a meta-learning setting \citep{liu2019tpami}.

%\begin{comment}
%%%%%%%%%%%%%%%% wrapfig version

% todo: should be pearson?
%\end{comment}

\begin{comment}
%%%%%%%%%%%%%%%% full figure version

\begin{figure}[t]
    \centering
    \includegraphics[width=.9\linewidth]{plots/section_3/bench_xcorr.pdf}
    \caption{Spearman rank correlation coefficient between ZC proxy scores on pairs of benchmarks. 
    %The entries in the plot are ordered based on the mean score across each row and column.
    }
    \label{fig:xcorr_zcp_bench}
\end{figure}

\end{comment}

\vspace{-1mm}
\subsection{RQ 2: Are ZC proxies complementary with respect to explaining validation accuracy?} \label{subsec:info}
\vspace{-1mm}

While Figure \ref{fig:corr_zcp} shows the performance of each individual ZC proxy, now we consider the combined performance of multiple ZC proxies.
If ZC proxies measure different characteristics of architectures, then a NAS algorithm can exploit their complementary information in order to yield improved results. 
While prior work \citep{ning2021evaluating, colin2022adeeperlook} computes the correlation among pairs of ZC proxies,
%\footnote{
%For completeness, we re-run that experiment and include the results in Appendix \ref{app:analysis}.}
our true goal is to assess the complementary information of ZC proxies \emph{with respect to explaining the ground-truth validation accuracy} (But for completeness, we re-run that experiment and include the results in Appendix \ref{app:analysis}). Furthermore, we wish to measure the complementary information of more than just two ZC proxies at a time. For this, we turn to information theoretic measures: by treating the validation accuracy and ZC proxy values as random variables, we can measure the entropy of the validation accuracy conditioned on one or more ZC proxies, which intuitively tells us the information that one or more ZC proxies reveal 
about the validation accuracy.
%Intuitively, this tells us the amount of information that one or more ZC proxies reveal about the validation accuracy.

\begin{figure}[t]
    \centering
    \includegraphics[width=.48\linewidth]{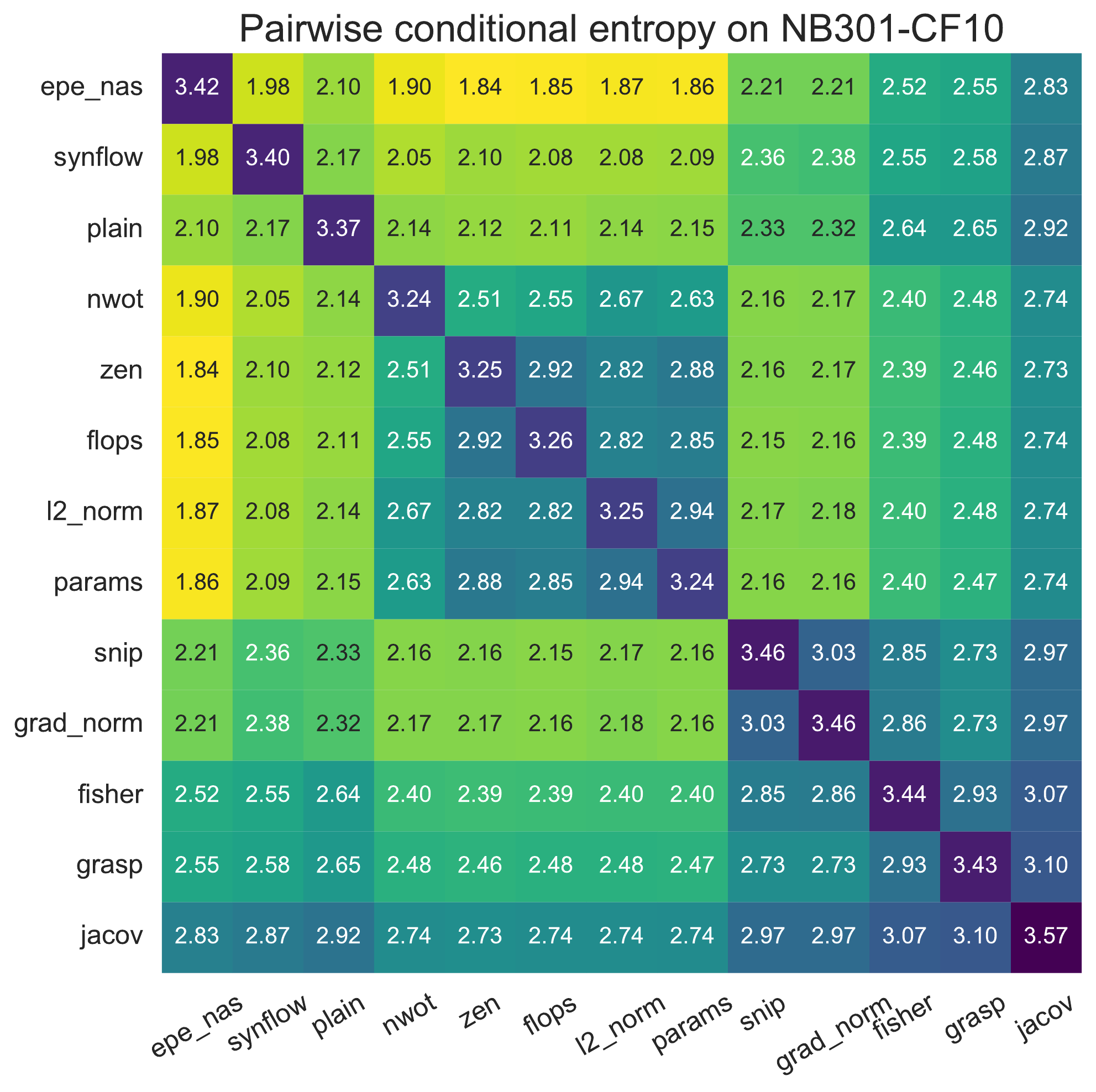}
    \includegraphics[width=.48\linewidth]{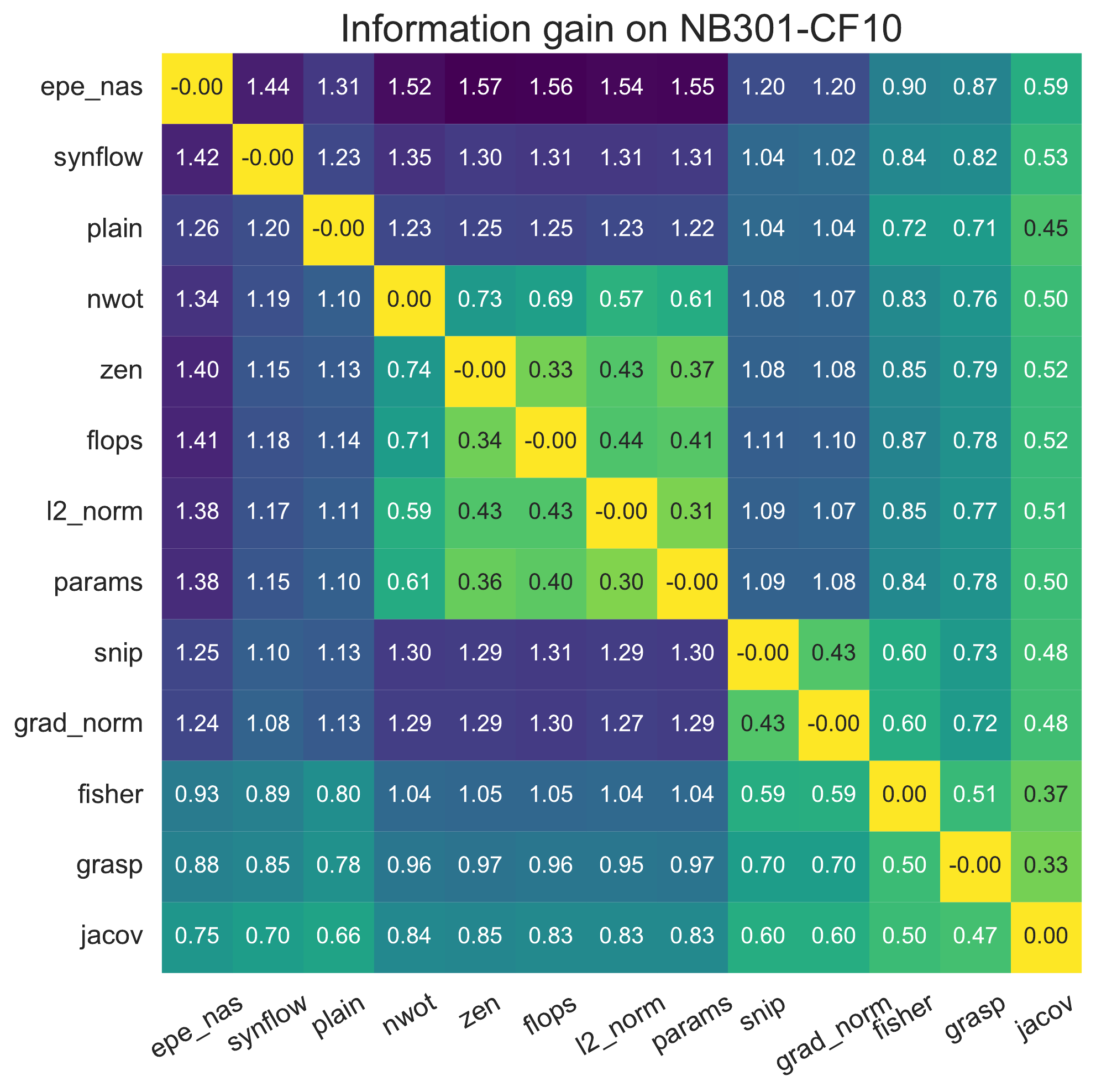} \\
    \includegraphics[width=.31\linewidth]{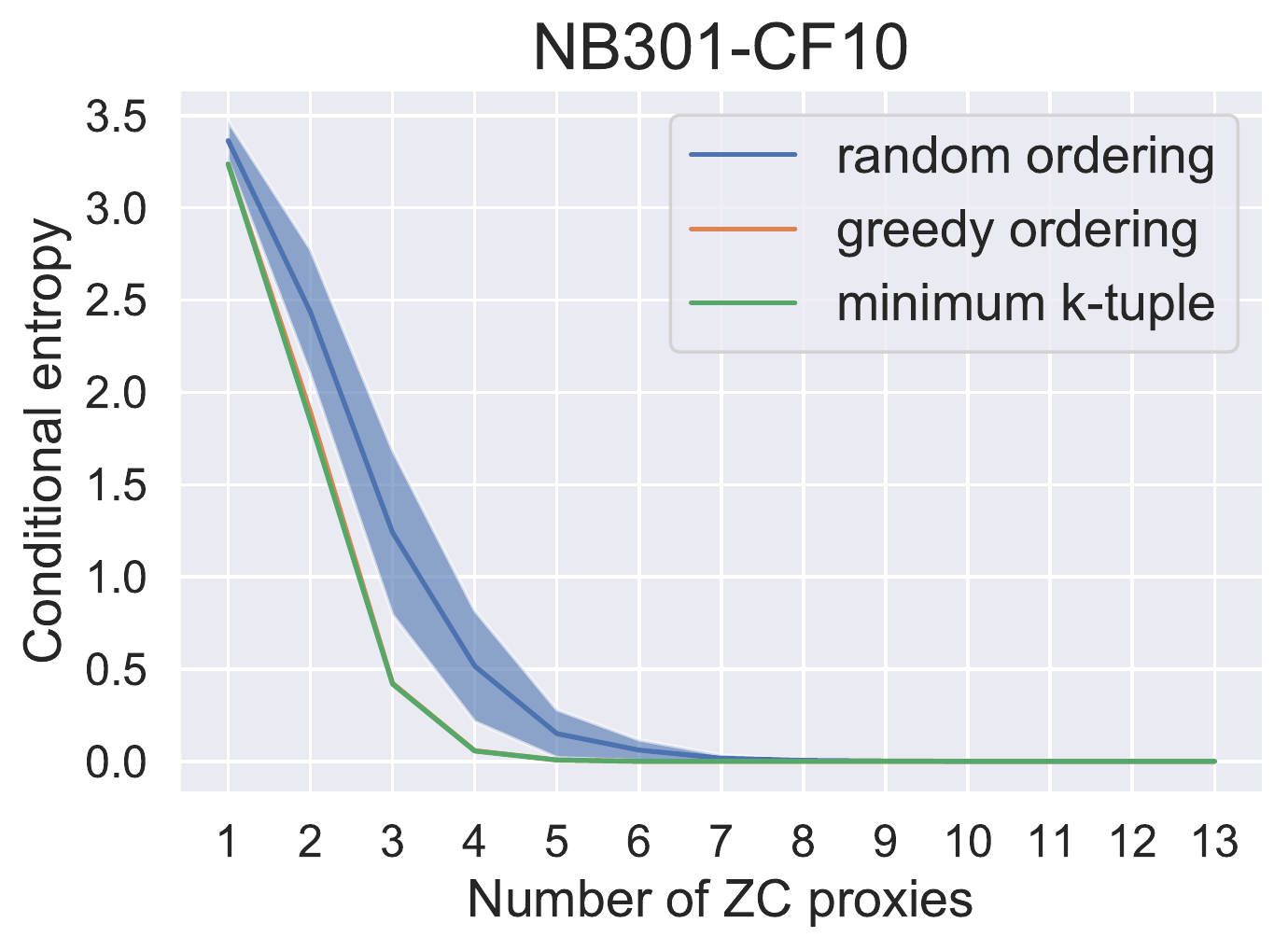}
    \includegraphics[width=.31\linewidth]{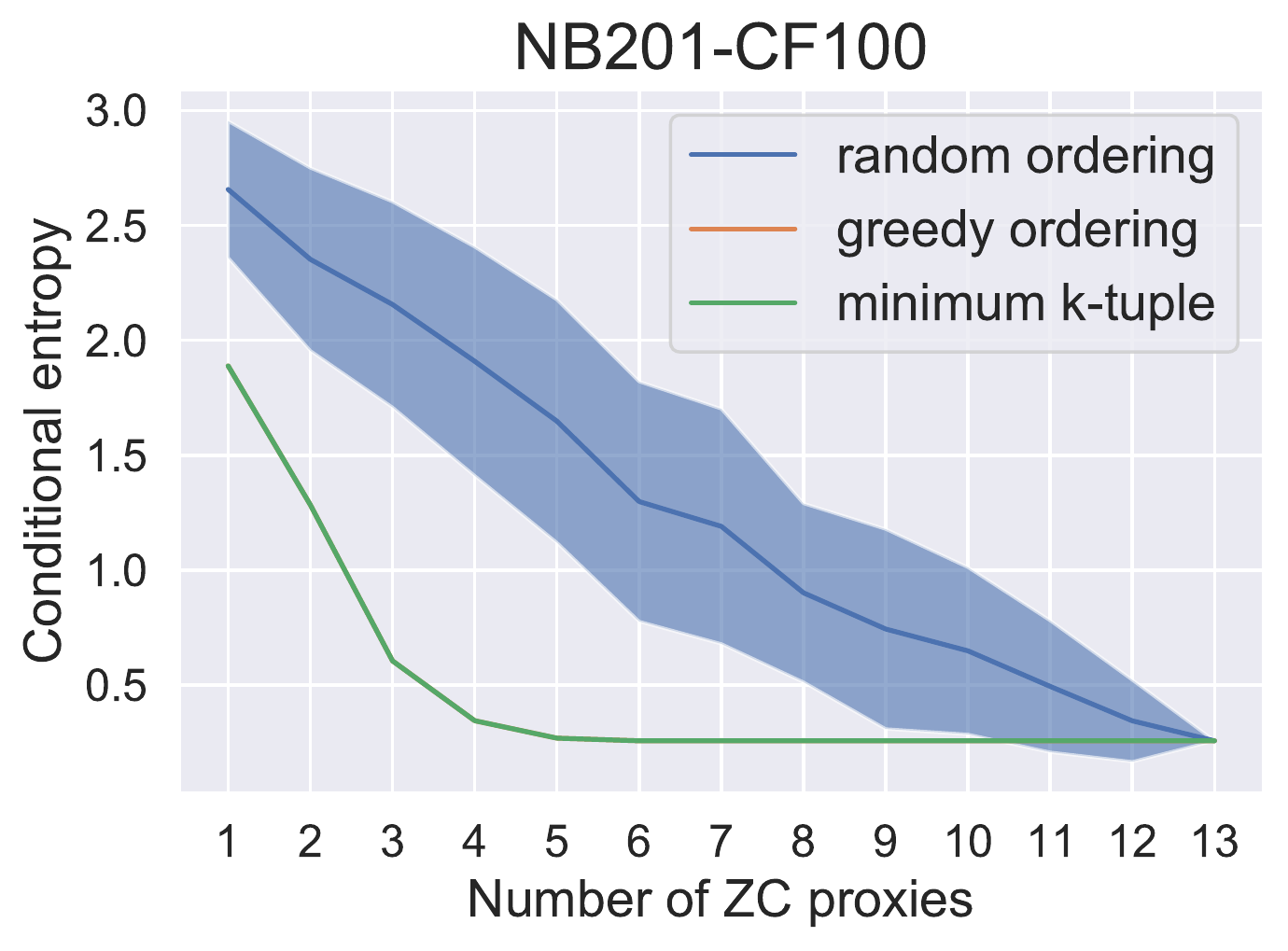}
    \includegraphics[width=.31\linewidth]{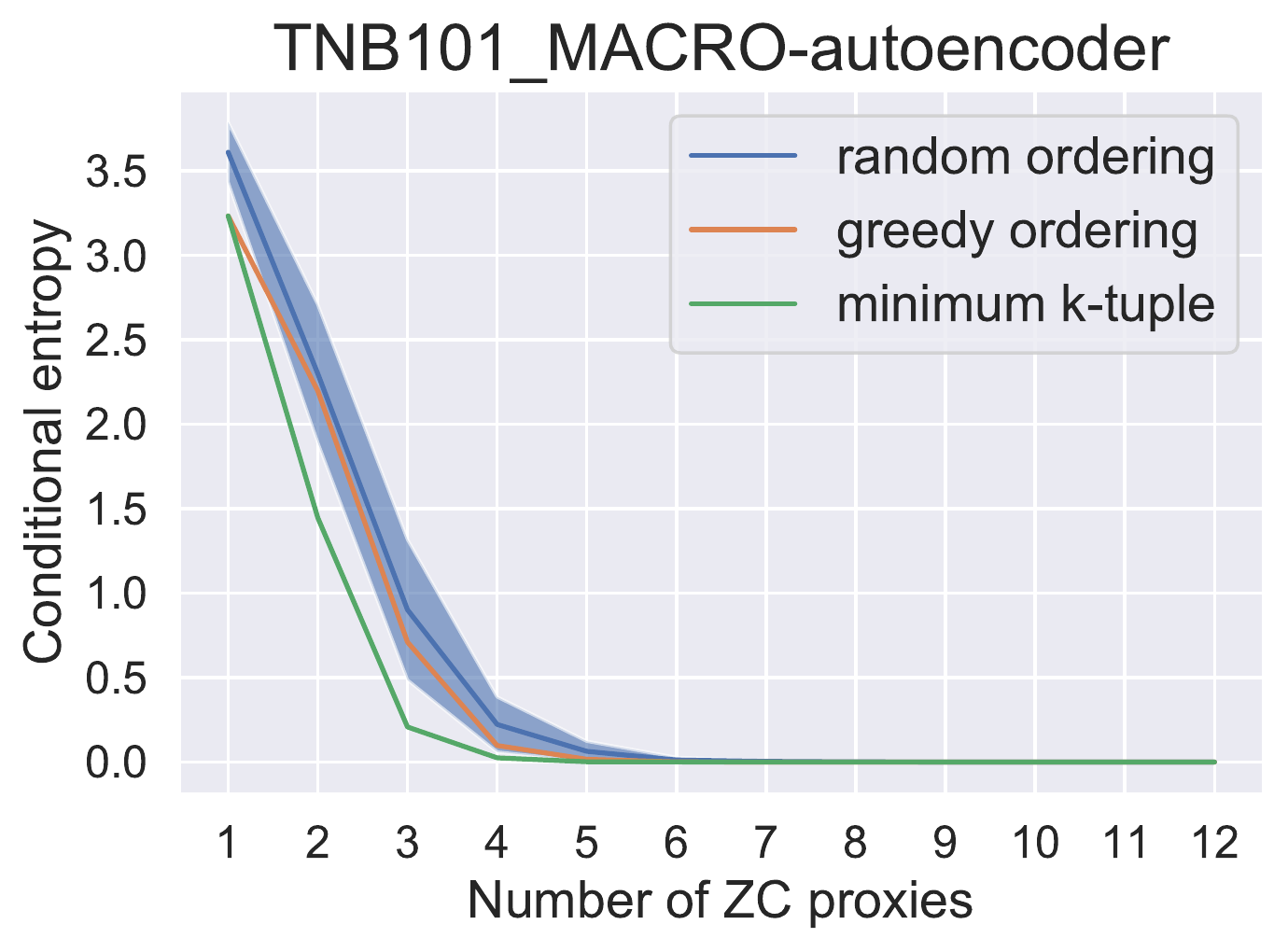}
    \\    
    \caption{
    Given a ZC proxy pair $(i,j)$, we compute the conditional entropy $H(y\mid z_i, z_j)$ (top left), and information gain $H(y\mid z_i) - H(y\mid z_i,z_j)$ (top right). 
    Conditional entropy $H(y\mid z_{i_1},\dots,z_{i_k})$ vs.\ $k$, where the ordering $z_{i_1},\dots,z_{i_k}$ is selected using three different strategies. The minimum $k$-tuple and greedy ordering significantly overlap in the first two figures (bottom).
    }
    \label{fig:cond_entropy}
\end{figure}

Formally, given a search space $S$, let $\Y$ denote the uniform distribution of validation accuracies over the search space, and let $y$ denote a random sample from $\Y$.
Similarly, for a ZC proxy $i$ from 1 to \nproxies, let $\Z_i$ denote the uniform distribution of the ZC proxy values, and let $z_i$ denote a random sample from $\Z_i$.
Let $H(\cdot)$ denote the entropy function. 
% First, we compute the information gain H - H.
% add ~ four random matrices 
For all pairs $z_i, z_j$ of ZC proxies, we compute the conditional entropy $H(y\mid z_i,z_j)$, as well as the information gain
$H(y\mid z_i) - H(y\mid z_i,z_j).$
See Figure \ref{fig:cond_entropy}. 
The entropy computations are based on 1000 randomly sampled architectures, using 24-bin histograms for density smoothing (see Appendix \ref{app:analysis} for more details).
We see that 
%the pair of ZC proxies with the maximum conditional entropy is \texttt{synflow, plain}. In other words, 
\texttt{synflow} and
\texttt{plain} together give the most information about the ground truth validation accuracies, due to their substantial complementary information.
% todo mention more trends, or cite the appendix for a guide

\begin{comment}

%%% separate figures

\begin{figure}[t]
    \centering
    \includegraphics[width=.48\linewidth]{02_neurips_dbt_2022/plots/section_4/information_theory_entropy/nasbench201-cifar10.pdf}
    \includegraphics[width=.48\linewidth]{02_neurips_dbt_2022/plots/section_4/information_theory_gain/nasbench201-cifar10.pdf}\\
    
    %\includegraphics[width=.48\linewidth]{02_neurips_dbt_2022/plots/section_4/information_theory_entropy/nasbench301-cifar10.pdf}
    %\includegraphics[width=.48\linewidth]{02_neurips_dbt_2022/plots/section_4/information_theory_gain/nasbench301-cifar10.pdf} \\
    \caption{
    Given a ZC proxy pair $(i,j)$, we compute the
    conditional entropy $H(y\mid z_i, z_j)$ (right), and information gain $H(y\mid z_i) - H(y\mid z_i,z_j)$ (left).
    }
    %from adding the column's proxy to the row's proxy on two benchmarks.
    \label{fig:cond_entropy}

\end{figure}

\begin{figure}[t]
    \centering
    \includegraphics[width=.31\linewidth]{02_neurips_dbt_2022/plots/section_4/entropy_vs_proxies/entropy_nb301_cf10.pdf}
    \includegraphics[width=.31\linewidth]{02_neurips_dbt_2022/plots/section_4/entropy_vs_proxies/entropy_nb201_cf100.pdf}
    \includegraphics[width=.31\linewidth]{02_neurips_dbt_2022/plots/section_4/entropy_vs_proxies/entropy_tnb101macro_autoencoder.pdf}
    \\
    \caption{Conditional entropy $H(y\mid z_{i_1},\dots,z_{i_k})$ vs.\ $k$, 
    where the ordering $z_{i_1},\dots,z_{i_k}$ is selected using three different strategies.}
    \label{fig:k_cond_entropy}
\end{figure}

\end{comment}

Now we can ask the same question for $k$ tuples of ZC proxies. Given an ordered list of $k$ ZC proxies $z_{i_1}, z_{i_2}, \dots z_{i_k}$, we define the information gain of $z_{i_k}$ conditioned on $y$ as follows:
\begin{equation}
%\vspace{-2mm}
\textbf{IG}(z_{i_k})\coloneqq H(y\mid z_{i_1},\dots,z_{i_{k-1}}) -  H(y\mid z_{i_1},\dots,z_{i_k}).
\end{equation}
Intuitively, \textbf{IG} computes the marginal information we learn about $y$ when $z_{i_k}$ is revealed, assuming we already knew the values of $z_{i_1},\dots,z_{i_{k-1}}$.
We compare the conditional entropy vs.\ number of ZC proxies for three different orderings of the ZC proxies.
The first is a random ordering (averaged over 100 random trials), which tells us the average information gain when iteratively adding more ZC proxies.
The second is a greedy ordering, computed by iteratively selecting the ZC proxy that maximizes $\textbf{IG}(z_{i_k})$, for $k$ from 1 to \nproxies.
The final plot exhaustively searches through ${\nproxies{} \choose k}$ sets to find the $k$ proxies which minimize $H(y\mid z_{i_1},\dots z_{i_k})$, for $k$ from 1 to \nproxies{} (note that this may not define a valid ordering).
See Figure \ref{fig:cond_entropy}, and Appendix \ref{app:analysis} for the complete results.
We see that there is very substantial information gain when iteratively adding ZC proxies, even if the ZC proxies are randomly chosen. Optimizing the order of adding ZC proxies yields much higher \textbf{IG} in certain benchmarks (e.g., NB201-CF100), and a greedy approach is shown to be not far from the optimum. 

% todo: summarize before giving the answer

\noindent{}\textbf{Answer to RQ 2:} \emph{In some benchmarks, we see substantial complementary information among ZC proxies. However, the degree of complementary information depends heavily on the NAS benchmark at hand.}
This suggests that we cannot always expect ZC proxies to yield complementary information, but a machine learning model might be able to identify useful combinations of ZC proxies.

\vspace{-1mm}
\subsection{RQ 3: Do ZC proxies contain biases, such as a bias toward certain operations or sizes, and can we mitigate these biases?} \label{subsec:bias}
\vspace{-1mm}

Identifying biases in ZC proxies can help explain weaknesses and facilitate the development of higher-performing ZC proxies. We  define bias metrics and study ZC proxy scores for thousands of architectures for their correlation with biases. This systematic approach yields generalizable conclusions and avoids the noise from assessing singular architectures. 
We consider the following biases: \textbf{conv:pool} (the numerical advantage of convolution to pooling operations in the cell), \textbf{cell size} (the number of non-zero operations in the cell), \textbf{num.\ skip connections}, and \textbf{num.\ parameters}.

For each search space, ZC proxy, and bias, we compute the Pearson correlation coefficient between the ZC proxy values and the bias values. We consider all 44K architectures referenced in Table \ref{tab:evaluations}. See Table \ref{tab:biases} and Appendix \ref{app:analysis} for the full results.
We find that many ZC proxies exhibit biases to various degrees.
Interestingly, some biases are consistent across search spaces, while others are not. For example, \texttt{l2-norm} has a conv:pool  bias on both NB201-C10 and NB301-C10, while \texttt{nwot} has a strong conv:pool bias on NB301-C10 and almost no bias on NB201-C10. While validation accuracy does not correlate with number of skip connections, most ZC proxies in the benchmark exhibit a negative bias towards this metric. 

Next, we present a procedure for removing these biases.
For this study, we use ZC proxies that had large biases in Table \ref{tab:biases}, and we attempt to answer the following questions: \emph{(1)} can we remove these biases, and \emph{(2)} if we can remove the biases, does the performance of ZC proxies improve?

Given a search space of architectures $A$, let $f:A\rightarrow \mathbb{R}$ denote a ZC proxy (a function that takes as input an architecture, and outputs a real number).
Furthermore, let $b:A\rightarrow\mathbb{R}$ denote a bias measure such as ``cell size''.
Recall that Table \ref{tab:biases} showed that the correlation between a ZC proxy $f$ and a bias measure $b$ may be high. For example, the correlation between \texttt{synflow} and ``cell size'' is high, which means using \texttt{synflow} would favor larger architectures.
To reduce bias, we use a simple heuristic:
\begin{equation} \label{eq:bias}
f'(a) = f(a) \cdot \frac{1}{b(a)+ C}.
\end{equation}

In this expression, $C$ is a constant that we can tune.
In deciding on a strategy to tune $C$, we make two observations.
First, for most bias measures, the bias of \texttt{val\_acc} is not zero, which means completely de-biasing ZC proxies could hurt performance.
Second, depending on the application, we may want to fully remove the bias of a ZC proxy, or else remove bias only insofar as it improves performance.

Therefore, we test three different strategies to tune $C$ by brute force:
\emph{(1)} ``minimize'', to minimize bias, \emph{(2)} ``equalize'', to match the bias with the bias of \texttt{val\_acc}, and \emph{(3)} ``performance'', to optimize the performance (Pearson correlation).
See Table \ref{tab:bias_mitigation} for the results.

We find that using the ``performance'' strategy, we are able to increase the performance of ZC proxies by reducing their bias. 
Furthermore, the ``equalize'' strategy sometimes provide good results on par with the ``performance'' strategy.
This suggests a good bias mitigation strategy when we do not know the ground truth but have information on how the ground truth correlations with bias. 
This may have important consequences for the future design of ZC proxies.

\noindent{}\textbf{Answer to RQ 3:} \emph{Many ZC proxies do exhibit different types of biases to various degrees, but the biases can be mitigated, thereby improving performance.}

\input{bias_tables}

%% file: bias_tables.tex
\begin{table}[t]
\caption{Pearson correlation coefficients between predictors and bias metrics (in bold) on different datasets. For example, for \textbf{Cell size} on NB201-CF100, \texttt{snip} has a correlation of -0.04 (indicating very little bias), while \texttt{synflow} has a correlation of 0.57 (meaning it favors larger architectures).
}
%\vspace{1mm}
%\vspace{3mm}
\resizebox{\linewidth}{!}{%
\centering
%\begin{adjustbox}{width=0.8\columnwidth}
\begin{tabular}{@{}l|c|c|c|c|c|c|c|c@{}}
\toprule
\multicolumn{1}{l|}{\textbf{Name}} & \multicolumn{2}{c|}{\textbf{Conv:pool}} & \multicolumn{2}{c|}{\textbf{Cell size}} & \multicolumn{2}{c|}{\textbf{Num.\ skip connections}} &  
\multicolumn{2}{c}{\textbf{Num.\ parameters}} \\
&NB201-CF10 & NB301-CF10 &NB201-CF100 & NB201-IM & NB301-CF10 & NB201-CF100 & NB101-CF10 &NB301-CF10\\
% &nb201(c10)&nb301(c10)&nb201(c100)&nb201(IM)&nb301(c10)&nb201(c100)&nb101(c10)&nb301(c10)\\
\midrule 
\texttt{epe-nas} & 0.05&-0.02&0.35& 0.35& 0.01 & 0.09 & -0.02 & -0.01 \\
\texttt{fisher}  & 0.05&0.01&-0.03&-0.05& -0.15 & -0.03 & 0.11 & 0.17\\
\texttt{flops} & 0.59&0.70&0.30&0.30& \textbf{-0.35} & -0.30 & \textbf{1.00} & 0.99 \\
\texttt{grad-norm} & 0.35& 0.27& -0.04& -0.05& -0.26 & -0.26 & 0.30 & 0.51 \\
\texttt{grasp} & 0.01& 0.28& -0.01& 0.01& 0.03 & 0.00 & -0.03 & 0.24\\
\texttt{l2-norm} & \textbf{0.87}& 0.76& 0.41& 0.41& -0.33 & \textbf{-0.41} & 0.62 & 0.99\\
\texttt{jacov}  & 0.05&-0.11& 0.35& 0.35& 0.08 & 0.09 & -0.18 & -0.10  \\
\texttt{nwot} & 0.06&\textbf{0.78}& 0.28& 0.28& -0.21 & 0.06 & 0.74 & 0.95\\
\texttt{params} & 0.61&\textbf{0.78}& 0.29& 0.29& -0.32 & -0.29 & \textbf{1.00} & \textbf{1.00}  \\
\texttt{plain} & -0.33&-0.45& 0.14& 0.14& 0.02 & 0.02 & 0.03 & -0.45 \\
\texttt{snip} &0.37& 0.27&-0.04& -0.04& -0.28 & -0.28 & 0.44 & 0.50  \\
\texttt{synflow}  &0.53& 0.41& \textbf{0.57}& \textbf{0.58}& -0.20 & -0.14 & 0.57 & 0.62  \\
\texttt{zen-score} &0.05& 0.75& 0.35& 0.35& -0.33 & 0.09 & 0.68 & 0.99  \\
\midrule
\texttt{val-acc} &0.36 & 0.45 & 0.35 & 0.43 & 0.13 & -0.06 & 0.09 & 0.47 \\
\bottomrule
\end{tabular}
}
\label{tab:biases}
\end{table}

\begin{table}
\caption{Bias mitigation strategies tested on the ZC proxies with the most biases.
We test three different strategies by tuning $C$ from Equation \ref{eq:bias} for different objectives:
minimize (tune $C$ to minimize bias), equalize (tune $C$ to match ground truth's correlation with bias metric), and performance (tune $C$ to maximize correlation with ground truth). 
Bias and performance are Pearson correlation coefficients of the proxy score with the bias metric and with the ground truth accuracy, respectively. 
$C$ is searched between -10 and 1000. 
%While optimizing, we consider using the original values as an option.
} \label{tab:bias_mitigation}
\centering
\begin{tabular}{llllllll} 
\hline
ZC proxy                  & dataset                      & \begin{tabular}[c]{@{}l@{}}bias \\metric~\end{tabular} & \begin{tabular}[c]{@{}l@{}}original \\bias\end{tabular} & \begin{tabular}[c]{@{}l@{}}original \\perf.\end{tabular} & \begin{tabular}[c]{@{}l@{}}new \\bias\end{tabular} & \begin{tabular}[c]{@{}l@{}}new \\perf.\end{tabular} & strategy     \\ 
\hline
\multirow{3}{*}{l2-norm~} & \multirow{3}{*}{NB201-CF10}  & \multirow{3}{*}{conv:pool}                             & \multirow{3}{*}{0.87}                                   & \multirow{3}{*}{0.42}                                          & 0.00                                               & 0.10                                                      & minimize     \\
                          &                              &                                                        &                                                         &                                                                & 0.37                                               & 0.11                                                      & equalize     \\
                          &                              &                                                        &                                                         &                                                                & 0.70                                               & 0.44                                                      & performance  \\ 
\hline
\multirow{3}{*}{nwot}     & \multirow{3}{*}{NB301-CF10}  & \multirow{3}{*}{conv:pool}                             & \multirow{3}{*}{0.78}                                   & \multirow{3}{*}{0.49}                                          & 0.00                                               & 0.03                                                      & minimize     \\
                          &                              &                                                        &                                                         &                                                                & 0.29                                               & 0.14                                                      & equalize     \\
                          &                              &                                                        &                                                         &                                                                & 0.78                                               & 0.49                                                      & performance  \\ 
\hline
\multirow{3}{*}{synflow}  & \multirow{3}{*}{NB201-CF100} & \multirow{3}{*}{cell size}                             & \multirow{3}{*}{0.57}                                   & \multirow{3}{*}{0.68}                                          & 0.01                                               & 0.64                                                      & minimize     \\
                          &                              &                                                        &                                                         &                                                                & 0.35                                               & 0.71                                                      & equalize     \\
                          &                              &                                                        &                                                         &                                                                & 0.35                                               & 0.71                                                      & performance  \\ 
\hline
\multirow{3}{*}{synflow}  & \multirow{3}{*}{NB201-IM}    & \multirow{3}{*}{cell size}                             & \multirow{3}{*}{0.58}                                   & \multirow{3}{*}{0.76}                                          & 0.01                                               & 0.62                                                      & minimize     \\
                          &                              &                                                        &                                                         &                                                                & 0.43                                               & 0.76                                                      & equalize     \\
                          &                              &                                                        &                                                         &                                                                & 0.46                                               & 0.76                                                      & performance  \\ 
\hline
\multirow{3}{*}{flops}    & \multirow{3}{*}{NB301-CF10}  & \multirow{3}{*}{num. skip}                             & \multirow{3}{*}{-0.35}                                  & \multirow{3}{*}{0.43}                                          & -0.01                                              & 0.06                                                      & minimize     \\
                          &                              &                                                        &                                                         &                                                                & 0.12                                               & -0.05                                                     & equalize     \\
                          &                              &                                                        &                                                         &                                                                & -0.35                                              & 0.43                                                      & performance    \\
\hline
\end{tabular}
\end{table}

%% file: 5_nas.tex
\vspace{-1mm}
\section{Integration into NAS} \label{sec:nas}
\vspace{-1mm}

The findings in Section \ref{subsec:info} showed that ZC proxies contain substantial complementary information, conditioned on the ground-truth validation accuracies. However, no prior work has combined more than four ZC proxies, or used a combination strategy other than a simple vote.
In this section, we combine and integrate all \nproxies{} ZC proxies into predictor-based NAS algorithms by adding the ZC proxies directly as features into the surrogate (predictor) models.

%\paragraph{Standalone model-based prediction.}

%First, we run experiments on a standalone model-based performance predictor.
We run experiments on two common predictor-based NAS algorithms: BANANAS, based on Bayesian optimization \citep{bananas}, and NPENAS, based on evolution \citep{npenas}.
Both algorithms use a model-based performance predictor: a model that takes in an architecture encoding as features (e.g., the adjacency matrix encoding \citep{white2020study}), and outputs a prediction of that architecture's validation accuracy. The model is retrained throughout the search algorithm, as more and more architectures are fully trained.
Recent work has shown that boosted trees such as XGBoost achieve strong performance in NAS \citep{nasbench301, white2021powerful}.

\vspace{-2mm}
\paragraph{Experimental setup.}
For both algorithms, we use the \texttt{NASLib} implementation \citep{ruchte2020naslib} and default parameters reported in prior work \citep{white2021powerful}. 
First, we assess the standalone performance of XGBoost when ZC proxies are added as features in addition to the architecture encoding, by randomly sampling 100 training architectures and 1000 disjoint test architectures, and computing the Spearman rank correlation coefficient between the set of predicted validation accuracies and the ground-truth accuracies. On NAS-Bench-201 CIFAR-100, averaged over 100 trials, the Spearman rank correlation ($\pm$ std.\ dev.) improves from $0.640\pm 0.0420$ to \textbf{0.908 $\pm$ 0.012} with the addition of ZC proxies, representing an \emph{improvement of 41.7\%}.
Even more surprisingly, using the ZC proxies alone as features without the architecture, results in a Spearman rank correlation of \textbf{0.907 $\pm$ 0.013}, implying that the ZC proxies subsume nearly all information contained in the architecture encoding itself. We present the full results in Appendix \ref{app:nas}.
These results show that an ensemble of ZC proxies can substantially increase the performance of model-based predictors.

Similar to the previous experiment, we run both NAS algorithms three different ways:
using only the encoding, only the ZC proxies, and both, as features of the predictor.
Each algorithm is given 200 architecture evaluations, and we plot performance over time, averaged over 400 trials. See Figure \ref{fig:bananas} for the results of BANANAS, and Appendix \ref{app:nas} for the full results.
We find that the ZC proxies give the NAS algorithms a boost in performance, especially in the early stages of the search.

%% file: 99_conclusions.tex
\vspace{-1mm}
\section{Conclusions, Limitations, and Broader Impact} \label{sec:conclusion}
\vspace{-1mm}
%In this work, we conducted a large-scale comparison of ZC proxies by implementing \nproxies{} ZC proxies and evaluating them across \ntasks{} tasks. We analyzed the rank correlation with validation accuracy for all architectures, as well as the correlation among all pairs of proxies, and mutual information. Motivated by the high level of complimentary information among ZC proxies, we ran NAS experiments using an ensemble of ZC proxies, showing that when used together, they substantially improve over the performance of any individual ZC proxy. Our findings make the case for ZC proxies as ``weak learners'' which grow stronger together and can be used to improve the performance of NAS methods at very little extra cost.

\begin{figure}[t]
    \centering
    \includegraphics[width=.32\linewidth]{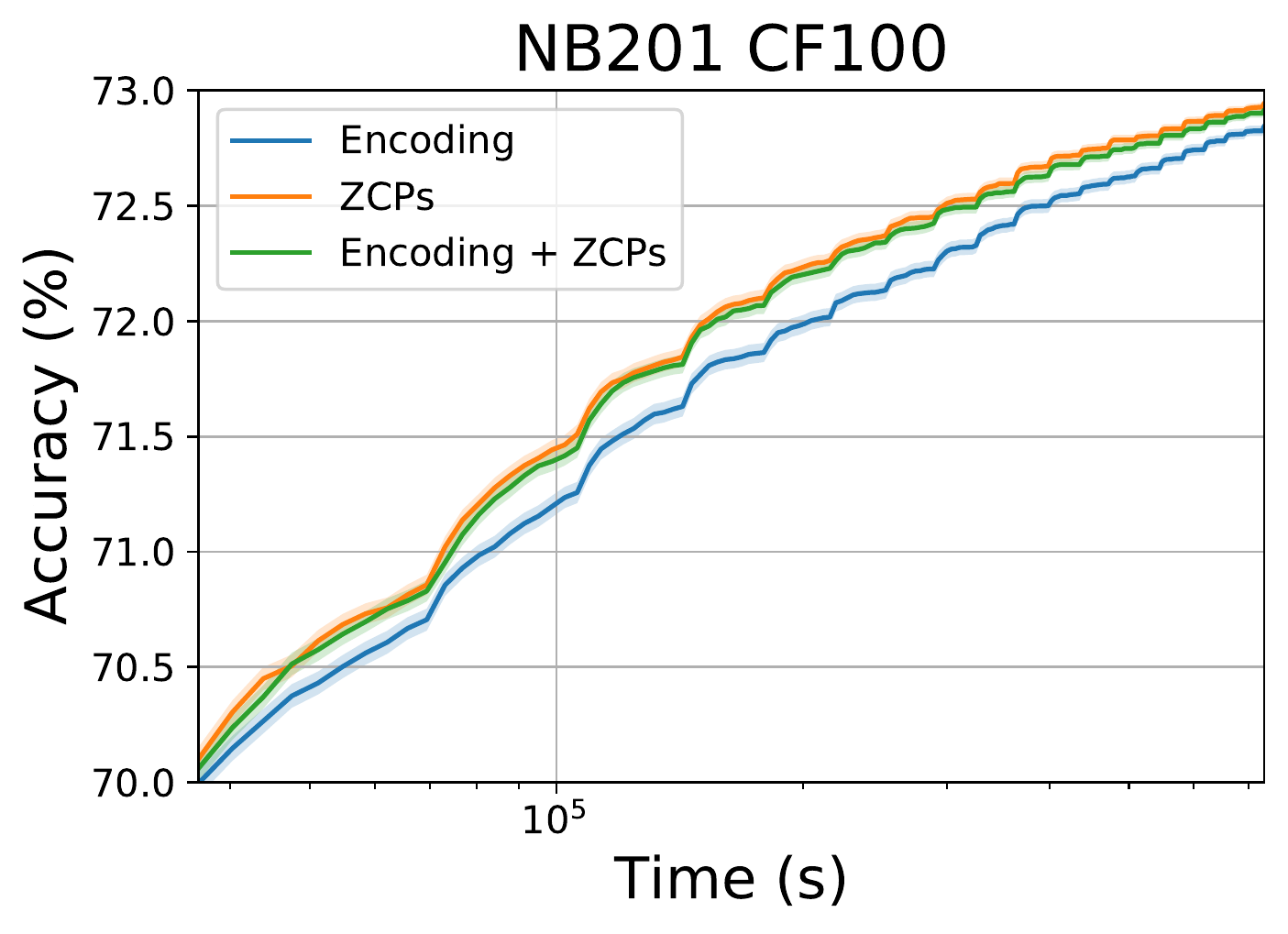}
    \includegraphics[width=.32\linewidth]{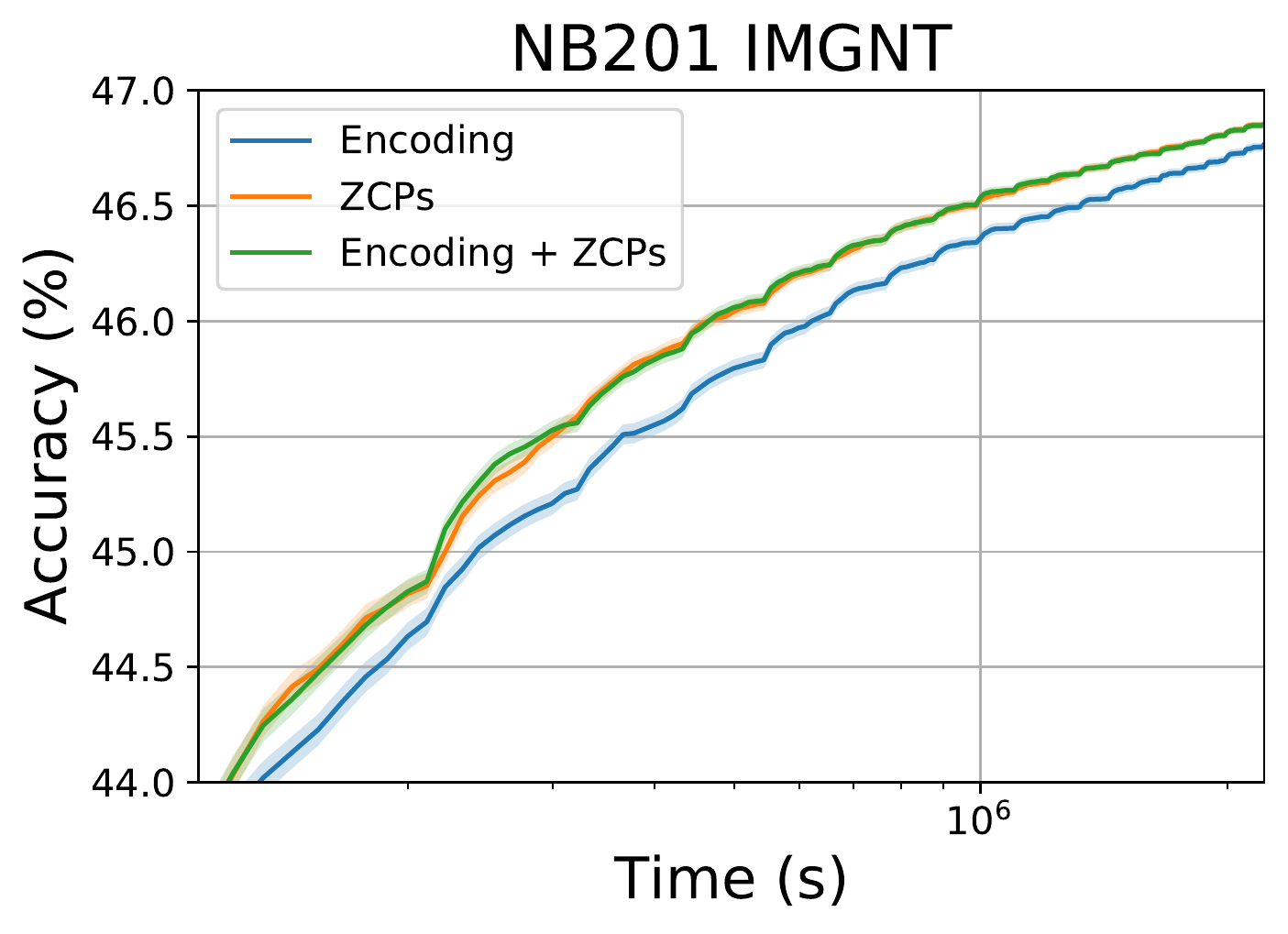}
    \includegraphics[width=.32\linewidth]{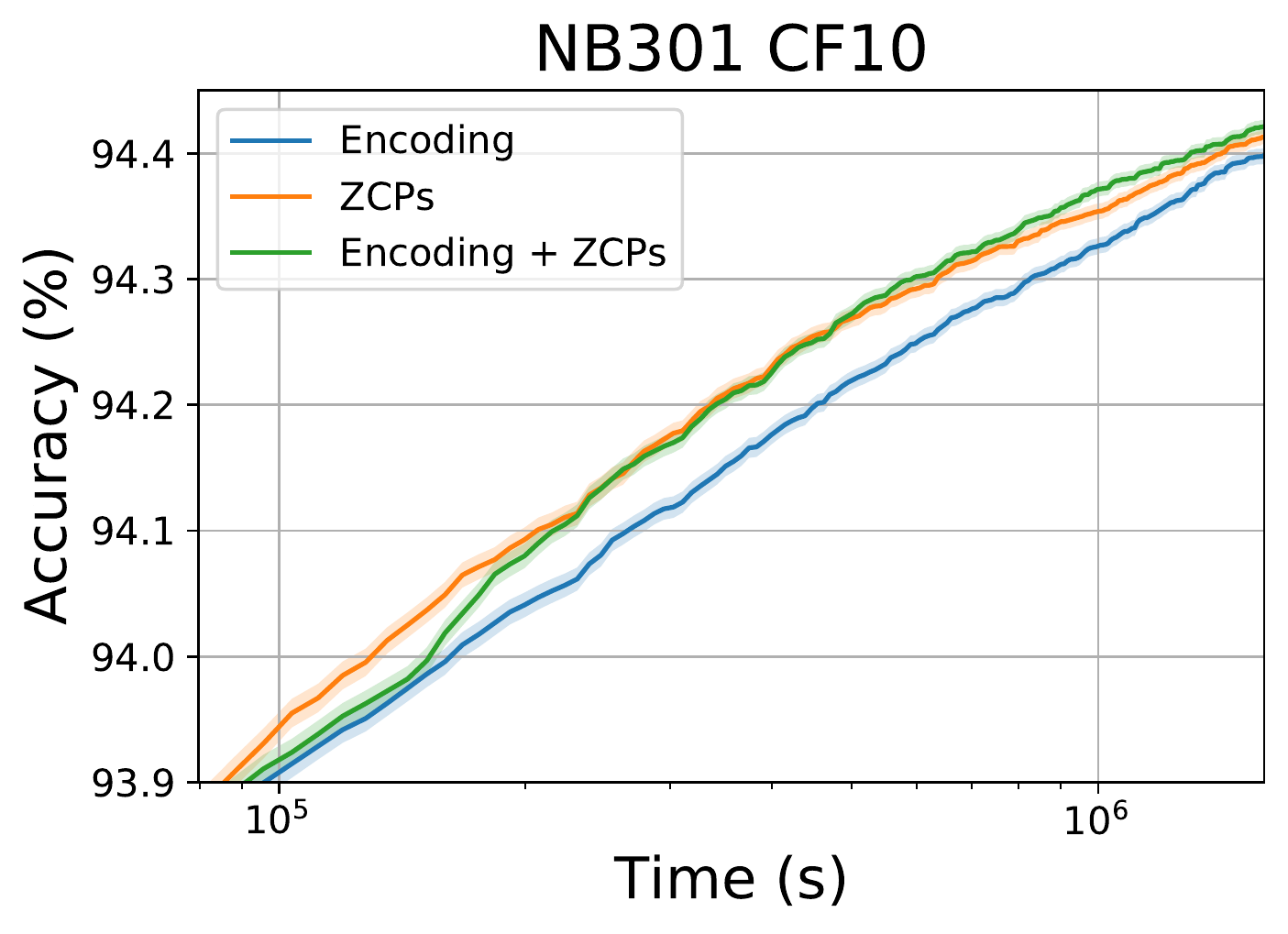} \\
    \caption{Performance of BANANAS with and without ZC proxies as additional features in the surrogate model. Each curve shows the mean and standard error across 400 trials.}
    \label{fig:bananas}
    \vspace{-3mm}
\end{figure}

In this work, we created \suite: an extensible collection of \nproxies{} ZC proxies (covering the majority that currently exist), accessible through a unified interface, which can be evaluated on a suite of \ntasks{} NAS benchmark tasks.
In addition to the codebase, we release precomputed ZC proxy scores across all \nproxies{} ZC proxies and \ntasks{} tasks, giving \nevals{} total ZC proxy evaluations. This dataset can be used to speed up ZC proxy-based NAS experiments, e.g., from 9 hours to 4 minutes (see Section \ref{sec:suite_overview}). Overall, \suite{} eliminates the overhead in ZC proxy research, with respect to comparing against different methods and across a diverse set of tasks.

To motivate the usefulness of \suite, we conducted a large-scale analysis of the generalizability, bias, and the first information-theoretic analysis of ZC proxies. Our empirical analysis showed substantial complementary information of ZC proxies conditioned on validation accuracy, motivating us to ensemble all \nproxies{} into predictor-based NAS algorithms. We show that using several ZC proxies together significantly improves the performance of the surrogate models used in NAS, as well as improving the NAS algorithms themselves. 
%We release our code, datasets, and all information needed to reproduce our results.

\vspace{-2mm}
\paragraph{Limitations and future work.}
Although our work makes substantial progress towards motivating and increasing the speed of ZC proxy research, there are still some limitations of our analysis. First, our work is limited to empirical analysis. However, we discuss existing theoretical results in Appendix \ref{app:zcp_theory}.
Furthermore, there are some benchmarks on which we did not give a comprehensive evaluation. 
For example, on NAS-Bench-301, we only computed ZC proxies on $11\,000$ architectures, since the full space of $10^{18}$ architectures is computationally infeasible. 
In the future, a surrogate model \citep{nasbench301, nasbenchx11} could be trained to predict the performance of ZC proxies on the remaining architectures.
Finally, there is very recent work on applying ZC proxies to one-shot NAS methods \citep{xiang2021zero}, which tested one ZC proxy at a time with one-shot models. Since our work motivates the ensembling of ZC proxies, an exciting problem for future work is to incorporate \nproxies{} ZC proxies into the one-shot framework.

\vspace{-2mm}
\paragraph{Broader impact.}
%We present a new benchmark suite, along with an analysis of ZC proxies.
The goal of our work is to make it faster and easier for researchers to run reproducible, generalizable ZC proxy experiments and to motivate further study on exploiting the complementary strengths of ZC proxies. By pre-computing ZC proxies across many benchmarks, researchers can run many trials of NAS experiments cheaply on a CPU, reducing the carbon footprint of the experiments \citep{patterson2021carbon, hao2019training}.
Due to the notoriously high GPU consumption of prior research in NAS \citep{zoph2017neural, real2019regularized}, this reduction in CO2 emissions is especially worthwhile.
Furthermore, our hope is that our work will have a positive impact in the NAS and automated machine learning communities by showing which ZC proxies are useful in which settings, and showing how to most effectively combine ZC proxies to achieve the best predictive performance. By open-sourcing all of our code and datasets, AutoML researchers can use our library to further test and develop ZC proxies for NAS.
%Our work is an abstraction away from real applications -- ZC proxy-based NAS algorithms may be used for both harmful applications (e.g.\ facial identification used in drone strikes) and beneficial applications (e.g.\ reducing CO2 emissions).

%% file: C_dataset_documentation.tex
\section{Dataset Documentation} \label{app:documentation}
%Author statement that they bear all responsibility in case of violation of rights, etc., and confirmation of the data license.

Here, we give an overview of our dataset documentation. For the full details, including links to the dataset, usage, and tutorials, see
\url{https://github.com/automl/NASLib/tree/zerocost}.

\subsection{Author responsibility and license}
We, the authors, bear all responsibility in case of violation of rights. The license of our dataset and repository is the \textbf{Apache License 2.0}. For more information, see \url{https://github.com/automl/NASLib/blob/Develop/LICENSE}.

In addition, we include the licenses of the datasets we used in Table \ref{tab:licenses}.

\begin{table}[t]
\caption{Licenses for the datasets that we use.}
%\vspace{1mm}
%\vspace{3mm}
%\resizebox{\linewidth}{!}{%
\centering
%\begin{adjustbox}{width=0.8\columnwidth}
\begin{tabular}{@{}l|c|c@{}}
\toprule
\multicolumn{1}{l}{\textbf{Dataset}} & \multicolumn{1}{c}{\textbf{License}} & \multicolumn{1}{c}{\textbf{URL}} \\
\midrule 
NAS-Bench-101 & Apache 2.0 & \url{https://github.com/google-research/nasbench} \\
NAS-Bench-201 & MIT & \url{https://github.com/D-X-Y/NAS-Bench-201} \\
NAS-Bench-301 & Apache 2.0 & \url{https://github.com/automl/nasbench301} \\
TransNAS-Bench-101 & MIT & \url{https://github.com/yawen-d/TransNASBench} \\
NAS-Bench-360 & MIT & \url{https://github.com/rtu715/NAS-Bench-360} \\
\bottomrule
\end{tabular}
%\end{adjustbox}
%}
\label{tab:licenses}
\end{table}

\subsection{Maintenance plan}
%Hosting, licensing, and maintenance plan. The choice of hosting platform is yours, as long as you ensure access to the data (possibly through a curated interface) and will provide the necessary maintenance.

% todo: where is the actual data located?
The data is available on GitHub at \url{https://github.com/automl/NASLib/tree/zerocost}.
We plan to actively maintain the repository, and we also welcome contributions from the community. For more information, see \url{https://github.com/automl/NASLib/tree/zerocost}.

\subsection{Code of conduct}
Our Code of Conduct is from the Contributor Covenant, version 2.0. See \\
\url{https://www.contributor-covenant.org/version/2/0/code_of_conduct.html}. %\\

\begin{comment}
The policy is copied below.

\begin{quote}
    ``We as members, contributors, and leaders pledge to make participation in our community a harassment-free experience for everyone, regardless of age, body size, visible or invisible disability, ethnicity, sex characteristics, gender identity and expression, level of experience, education, socio-economic status, nationality, personal appearance, race, caste, color, religion, or sexual identity and orientation.''
\end{quote}

\end{comment}

\subsection{Datasheet}

We include a datasheet \citep{gebru2021datasheets} for \suite.
% remove for CR version:
Please see \\ \url{https://github.com/automl/NASLib/blob/zerocost/docs/DATASHEET.md}.

% remove for arxiv version:
%\input{datasheet}

%% file: D_appendix_related_work.tex
\section{Related Work Continued} \label{app:related_work}
In this section, we give additional details on related work, continued from Section \ref{sec:background}.

Multiple recent works have investigated the performance of ZC proxies in ranking architectures over different NAS benchmarks.
\citep{ning2021evaluating} provides rank correlations and pairwise correlations of 10 ZC proxies across 7 tasks, and concludes that the relative performance of different ZC proxies highly depends on the search space. They further analyze how ZC proxies have improper biases.
\citep{white2021powerful} compares 6 ZC proxies across four tasks, and further shows how
\texttt{jacov} can be used to accelerate the search in predictor-based NAS. In particular, OMNI \citep{white2021powerful} combines \texttt{jacov} with \emph{sum of training losses} \citep{ru2021speedy} in the surrogate models of BANANAS and predictor-guided evolution. However, the predictor-based NAS experiments are restricted to NAS-Bench-201 and a single ZC proxy.
Similar to \citep{white2021powerful}, ProxyBO \citep{shen2021proxybo} introduces a NAS framework based on BO which uses ZC proxies to speed up NAS. It dynamically chooses whether to use a Gaussian process, \texttt{snip}, \texttt{jacov}, or \texttt{synflow} as the surrogate model in BO. Experiments were done on five tasks. Note that although the NAS method makes use of three different ZC proxies, each are used \emph{separately} to make predictions on the performance of architectures.

%Given the design similarity of the widely used benchmarks NAS-Bench-101/201/301 and motivated by the need to mimic manually designed networks, 
Recently, NAS-Bench-Zero was introduced \citep{chen2021bench}, a new benchmark based on popular computer vision models ResNet \citep{resnet} and MobileNetV2 \citep{sandler2018mobilenetv2}, and examined different characteristics of 10 ZC proxies across these search space as well as three existing search spaces.
% todo: mention that they precomputed zc proxies on their three search spaces.
The study shows in particular that individual ZC proxies do not transfer across NAS benchmarks. They also show that voting among \texttt{synflow}, \texttt{zen}, \texttt{snip} and \texttt{synflow} is the optimal voting ZC proxy strategy.
% todo: expand a bit?
%
A recent overview of ZC proxies \citep{colin2022adeeperlook} computes rank correlation, pairwise correlation, and performance plots for 8 ZC proxies across 12 tasks.

Only two prior works combine the information of multiple ZC proxies together in architecture predictions \citep{abdelfattah2021zerocost, chen2021bench} and both only use the \emph{voting} strategy to combine three or four ZC proxies. Our work is the first to combine ZC proxies in a nontrivial way, and the first to combine \nproxies{} ZC proxies. We also conduct analysis on the largest set of ZC proxies and benchmarks to date.

\subsection{Theoretical results for ZC proxies} \label{app:zcp_theory}

While ZC proxies are starting to be used more widely today \citep{abdelfattah2021zerocost, colin2022adeeperlook, zhou2022training, javaheripi2022litetransformersearch}, still relatively little is known about them from a theoretical standpoint.
However, there have been a few works that do give theoretical results.
In this section, we survey the existing theoretical results for ZC proxies.

Ning et al.\ gave a theoretical preference analysis for \texttt{synflow}, proving that it favors larger architectures (Section B.3 in \citep{ning2021evaluating}).
Specifically, they prove that given an architecture, introducing a new fully-connected layer into an MLP architecture causes the \texttt{synflow} value to increase. The core of their argument is to prove the following statement: ``when introducing a new fully-connected layer, the expected loss gradients with respect to the existing parameters increases.''
The authors also claim that the intuition for this argument should extend to convolutional neural networks.
Finally, we note that our empirical results from Table \ref{tab:biases} confirm their theoretical finding.

Shu et al.\ \citep{Shu2022Unifying} attempted to give a unified, general theory for multiple ZC proxies.
First, the authors prove that ZC proxy values are asymptotically similar. Specifically, they show that assuming the loss function of the neural network is $\beta$-Lipschitz continuous, and $\gamma$-Lipschitz smooth, then with high-priority, then the values of 
\texttt{grad\_norm}, \texttt{snip}, and \texttt{grasp} are all asymptotically similar up to constants (i.e., the same under big-Oh notation) to the trace norm of the NTK matrix at initialization. 
This result implies that the values of these ZC proxies are highly correlated.

Next, Shu et al.\ establish generalization bounds for DNNs in terms of the ZC proxies.
Specifically, they show that the generalization error of a DNN is at most the sum of the training error of the DNN and $O\left(\kappa/\mathcal{M}\right)$, where $\mathcal{M}$ can be set to \texttt{grad\_norm}, \texttt{snip}, or \texttt{grasp}, and $\kappa$ is the condition number of the NTK matrix at initialization, i.e., given the NTK matrix $\Theta_0$, $\kappa=\lambda_{\max}(\Theta_0)/ \lambda_{\min}(\Theta_0)$.

As a corollary, they also bound the generalization error in terms of the ZC proxy value and other fixed constants of the neural network, without the training error term.

% n theoretically justify their empirically observed transferability in a similar way (Sec. 4.4)

Other than these results, a few works have derived new ZC proxies via a theoretical analysis or inspired by existing theories of deep learning.
Shu et al.\ \citep{shu2022nasi} introduce NASI by giving a theoretical analysis that shows the trace norm of the NTK has a similar form to gradient flow.
Other theory-inspired ZC proxies include TE-NAS \citep{chen2021neural}, which uses the spectrum of the NTK and the number of linear regions in the input space, and NNGP-NAS \citep{park2020towards}, which approximates the Neural Network Gaussian Process using Monte-Carlo methods.

% TE-NAS: https://openreview.net/forum?id=Cnon5ezMHtu
% unifying and boosting: https://arxiv.org/pdf/2201.09785.pdf

As ZC proxies gain in popularity, a further theoretical analysis is an important step in understanding their robustness on different datasets, and in designing higher-performing ZC proxies.

%% file: E_appendix.tex
\section{Details from Section \ref{sec:analysis}} \label{app:analysis}
In this section, we give additional details from Section \ref{sec:analysis}.

\subsection{Details from Section \ref{subsec:generalization}: generalization}
\begin{figure}[t]
    \centering
    \includegraphics[width=.5\linewidth]{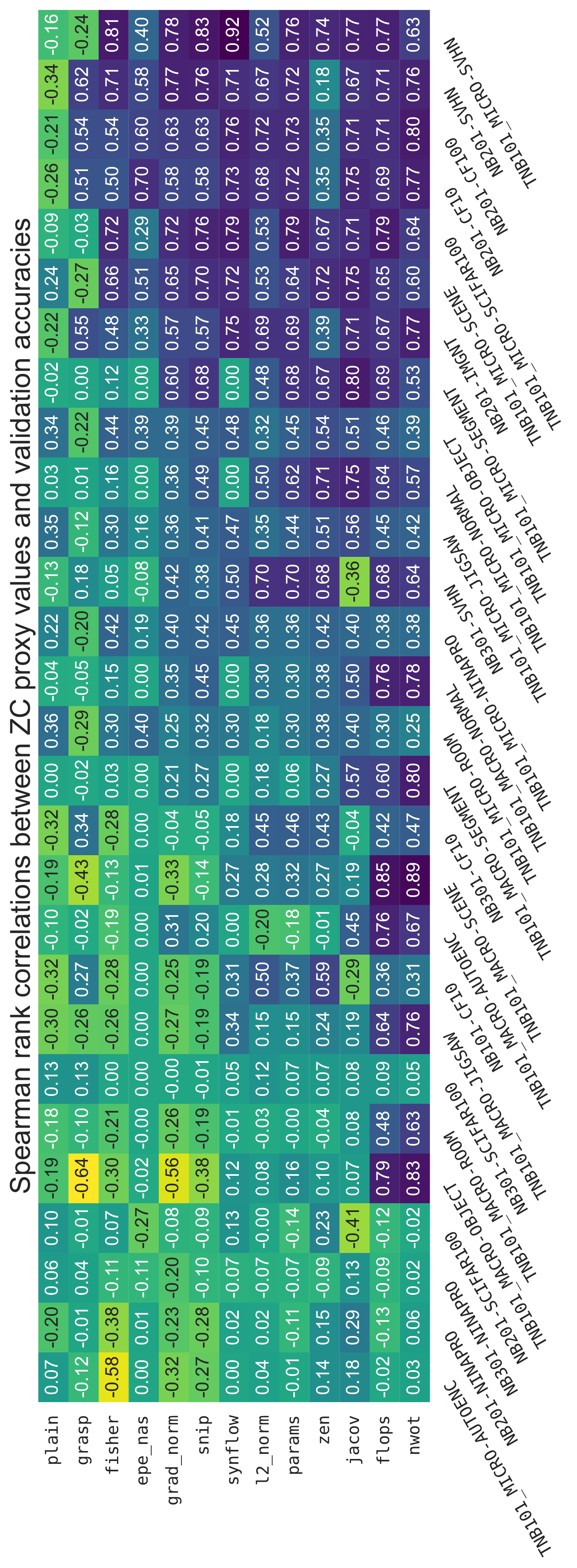}
    \caption{Spearman rank correlation coefficient between ZC proxy values and validation accuracies, for each ZC proxy and benchmark. 
    The rows and columns are ordered based on the mean scores across columns and rows, respectively. This is the full version of Figure \ref{fig:corr_zcp}.
    %Note that some ZC proxies only work on image classification tasks. In such cases we provide an value of 0 in the plot.
    }
    \label{fig:corr_extended}
\end{figure}

We give the full extensions of the experiments from Section \ref{subsec:generalization}.
In Figure \ref{fig:corr_extended}, for each ZC proxy and each benchmark, we compute the Spearman rank correlation (see Section \ref{sec:analysis}). This is the full version of Figure \ref{fig:corr_zcp}.

\begin{figure}[t]
    \centering
    \includegraphics[width=.95\linewidth]{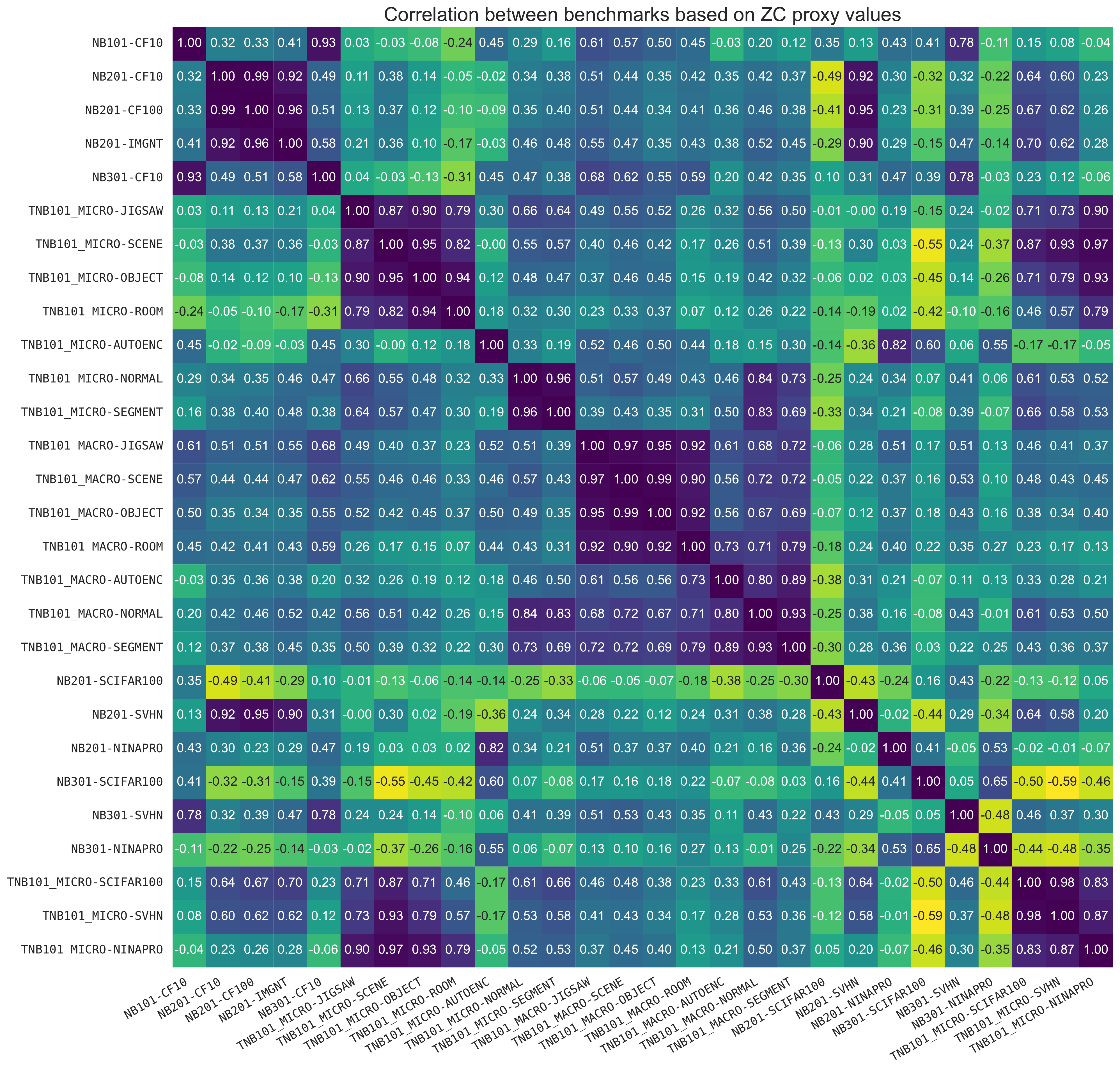}
    \caption{Pearson correlation coefficient between ZC proxy scores on pairs of benchmarks. The entries in the plot are ordered based on the mean score across each row and column. This is the full version of Figure \ref{fig:xcorr_zcp_bench}.}		\label{fig:xcorr_zcp_bench_appendix}
\end{figure}

In Figure \ref{fig:xcorr_zcp_bench_appendix}, we compute the Pearson correlation coefficient between ZC proxy scores on pairs of benchmarks. This is the full version of Figure \ref{fig:xcorr_zcp_bench}.

Next, we recompute Figure \ref{fig:corr_zcp} using different metrics: Precision@K and BestRanking@K \citep{ning2021evaluating, chen2021bench}. Let $M$ denote the number of architectures, and for each architecture $a_i$ from $i\in [1,M]$, denote the rankings of the ground truth and ZC proxy-estimated scores are $r_i$ and $n_i$, respectively. 
Given $K$, define $A_K=\{a_i\mid n_i < KM\}$. 
The definitions are as follows:

\begin{align*}
\text{Precision}@K &= \frac{\#\{i\mid r_i < K \wedge n_i < K\}}{K} \\
\text{BestRanking}@K &= \text{argmin}_{\alpha_i\in A_K} r_i/M
\end{align*}

\begin{figure}[t]
    \centering
    \includegraphics[width=\linewidth]{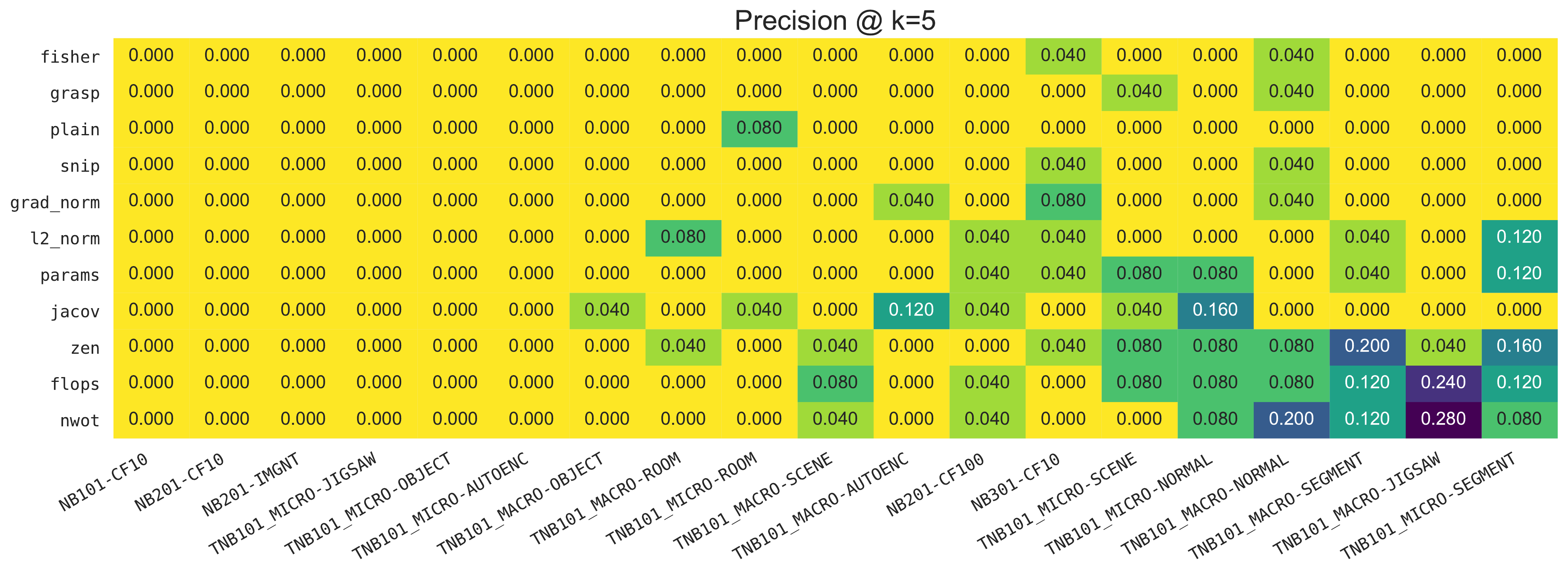}
    \includegraphics[width=\linewidth]{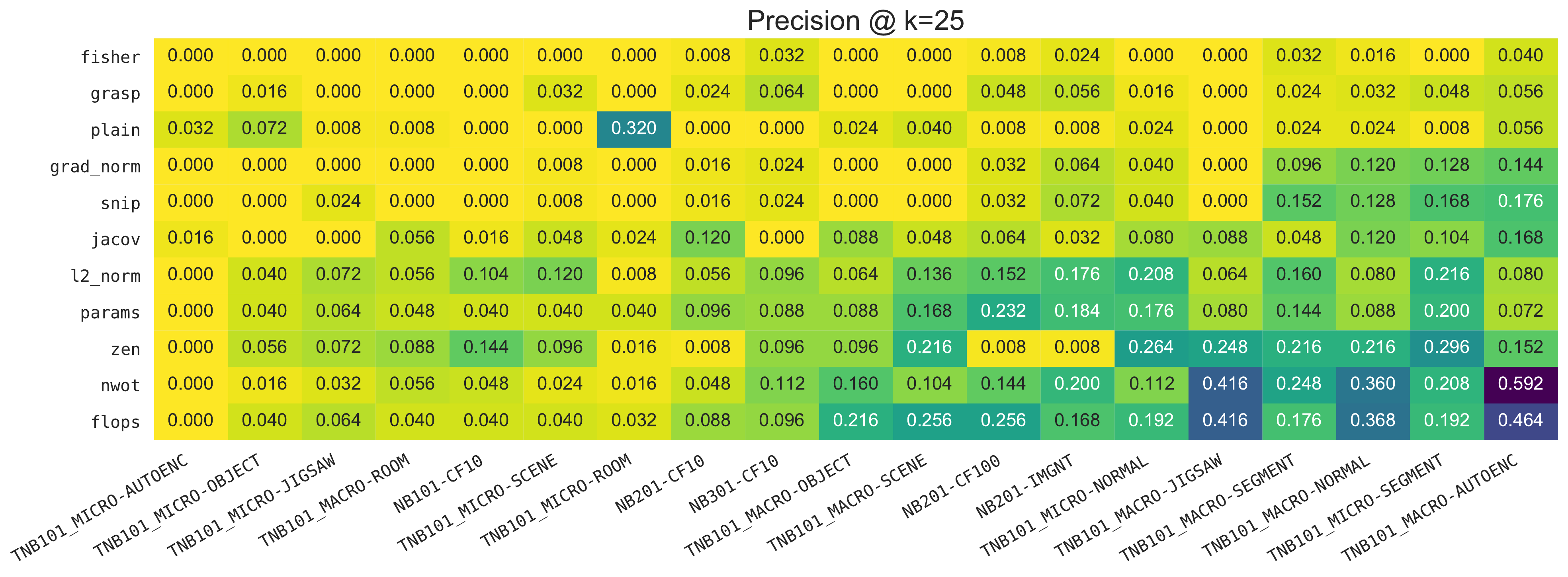}
    \includegraphics[width=\linewidth]{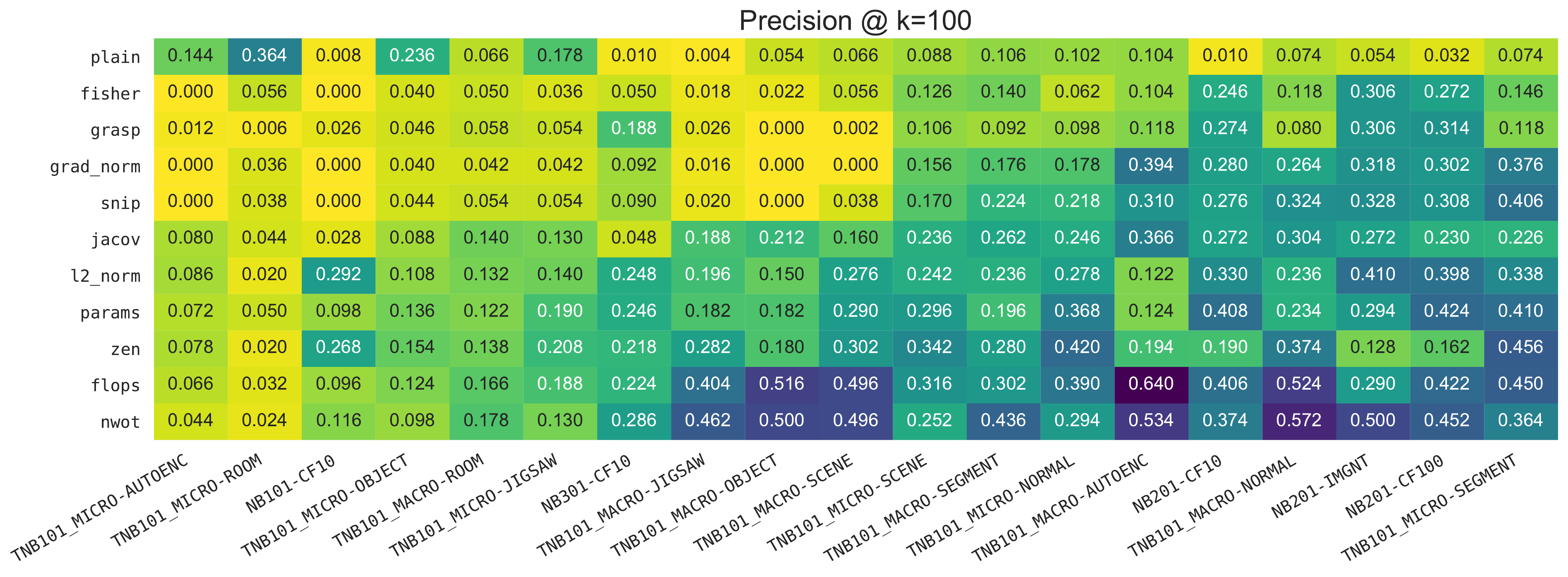}
    \caption{Precision@K between ZC proxy values and validation accuracies, for each ZC proxy and benchmark. 
    The rows and columns are ordered based on the mean scores across columns and rows, respectively. 
    }
    \label{fig:prec}
\end{figure}

\begin{figure}[t]
    \centering
    \includegraphics[width=\linewidth]{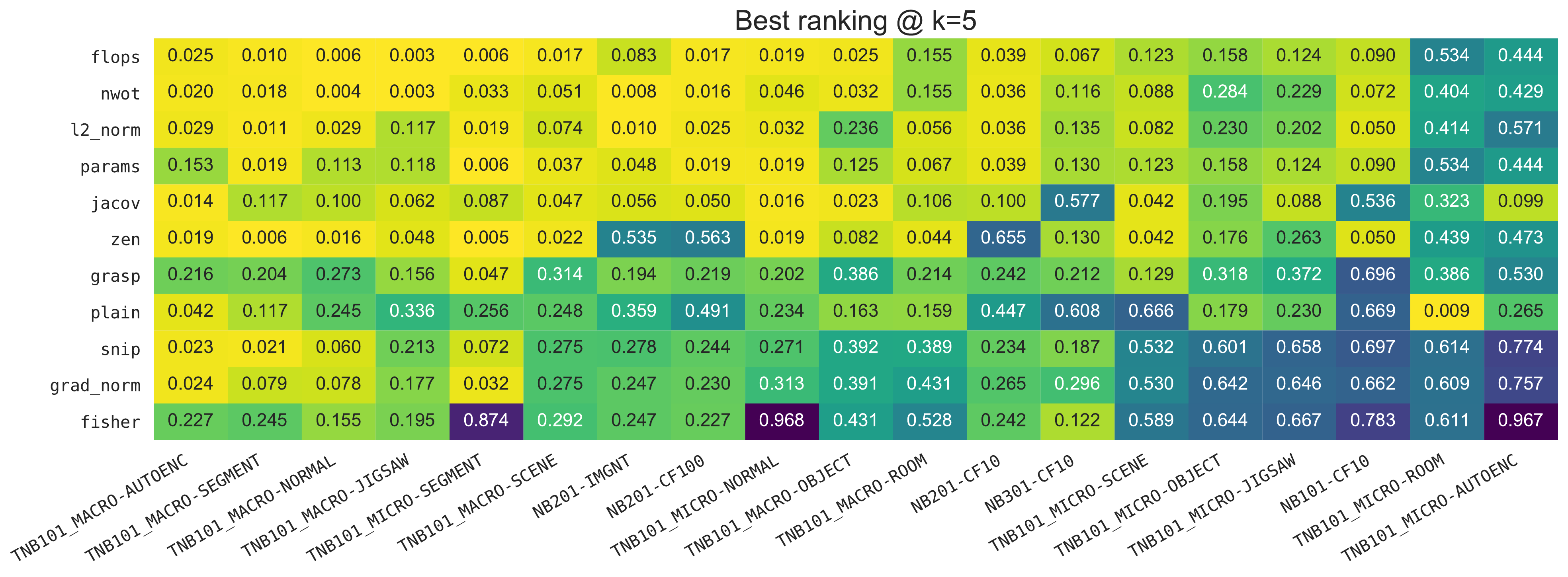}
    \includegraphics[width=\linewidth]{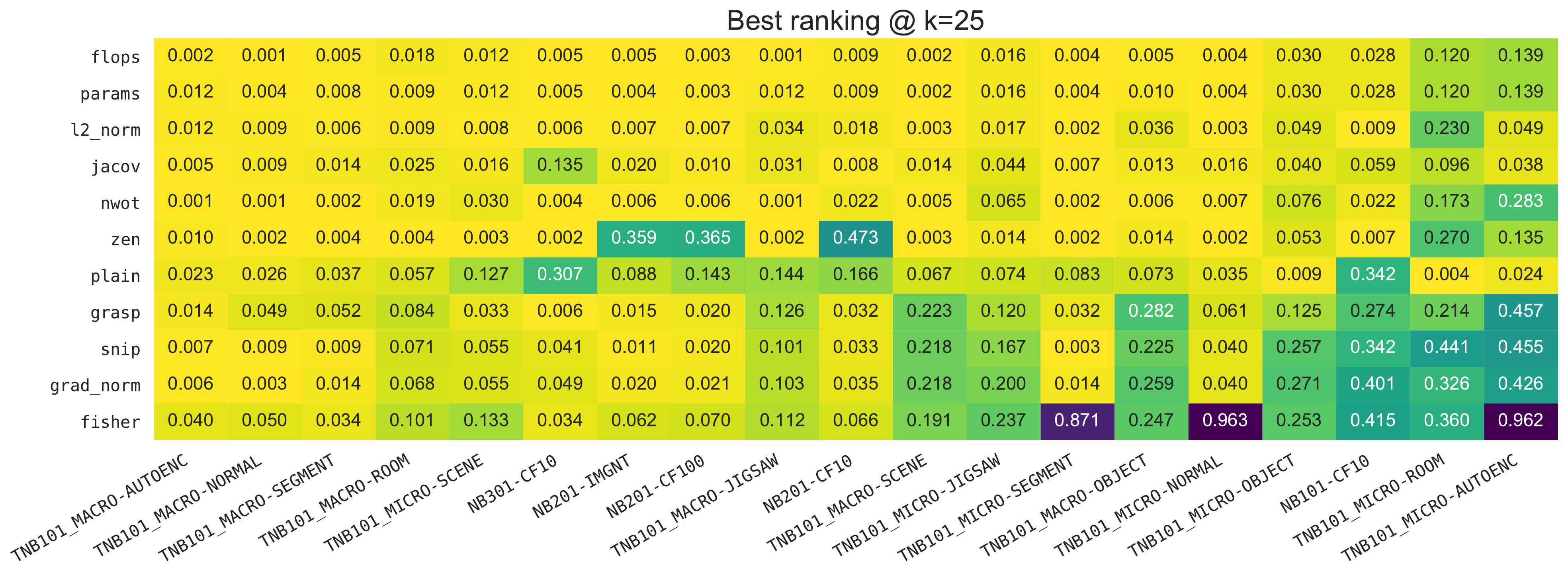}
    \includegraphics[width=\linewidth]{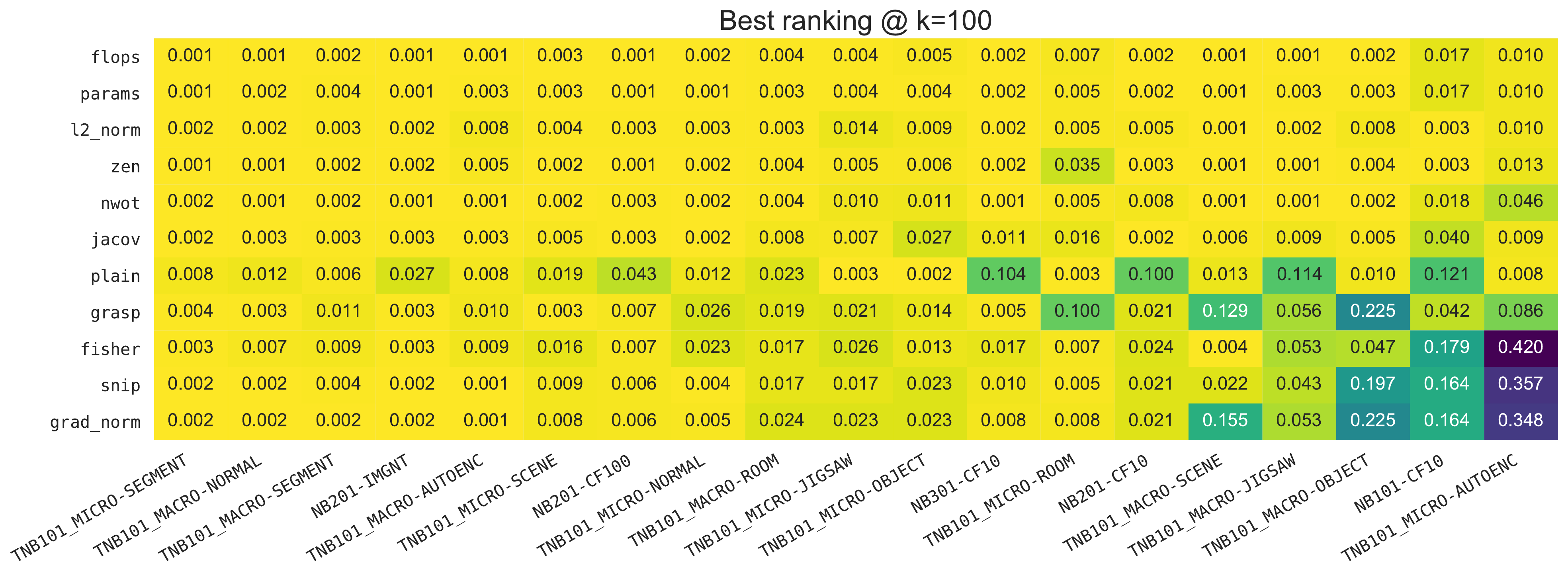}
    \caption{BestRanking@K between ZC proxy values and validation accuracies, for each ZC proxy and benchmark. 
    The rows and columns are ordered based on the mean scores across columns and rows, respectively. 
    }
    \label{fig:bestrank}
\end{figure}

In Figure \ref{fig:prec}, we recompute Figure \ref{fig:corr_zcp} using Precision@K, for $K=5, 25, 100$.
In Figure \ref{fig:bestrank}, we recompute Figure \ref{fig:corr_zcp} using BestRanking@K, for $K=5, 25, 100$.
Overall, we see similar trends to Figure \ref{fig:corr_zcp}, but we note that Precision@K and BestRanking@K may be more useful than Spearman in terms of NAS, since the goal of NAS is to find the very best architectures.

\subsubsection{Initial results with FBNet}
While \suite{} contains \ntasks{} tasks, the majority of search spaces used were designed for research.
Now, in contrast, we give initial results for FBNet \citep{fbnet} as a search space that has been used to achieve state-of-the-art results.

The FBNet search space consists of 22 searchable layers, with 9 operation choices each (3 filters and 3 kernel sizes), for a total of $9^{22}=10^{21}$ architectures in the search space.
The block structure is inspired by MobileNetV2 \citep{sandler2018mobilenetv2} and ShiftNet \citep{wu2018shift}.

See Table \ref{tab:fbnet} for the Spearman rank correlation values of the validation accuracy of 100 randomly drawn architectures compared to ZC proxies.
Even though the FBNet search space is size $10^{21}$, some of the ZC proxies perform surprisingly well, such as \texttt{snip}, \texttt{synflow}, and \texttt{flops}. The highest-performing ZC proxy is \texttt{flops}.

\begin{table}[t]
\caption{Spearman rank correlation for 100 architectures randomly drawn from the FBNet search space on various ZC proxies.
}
\resizebox{\linewidth}{!}{%
\centering
\begin{tabular}{@{}c|c|c|c|c|c|c|c|c@{}}
\toprule
\multicolumn{1}{c}{ZC Proxy} & 
\multicolumn{1}{c}{\texttt{fisher}} & 
\multicolumn{1}{c}{\texttt{flops}} &  
\multicolumn{1}{c}{\texttt{grad\_norm}} & 
\multicolumn{1}{c}{\texttt{grasp}} & 
\multicolumn{1}{c}{\texttt{jacov}} & 
\multicolumn{1}{c}{\texttt{params}} & 
\multicolumn{1}{c}{\texttt{snip}} &  
\multicolumn{1}{c}{\texttt{synflow}} \\
\midrule 
Spearman & 0.2574 & \textbf{0.6484} & 0.4278 & -0.262 & -0.0895 & 0.3762 & 0.5102 & 0.4954 \\
\bottomrule
\end{tabular}
%\end{adjustbox}
}
\label{tab:fbnet}
\end{table}

\subsection{Details from Section \ref{subsec:info}: information theory}
In this section, we give details from Section \ref{subsec:info}. We start with more details on the conditionial entropy, including why we chose this metric, how it is computed, and how to interpret the results.

\begin{itemize}[topsep=0pt, itemsep=2pt, parsep=0pt, leftmargin=5mm]
    \item \textit{Why do we choose conditional entropy as the metric?} \\
    The conditional entropy of a random variable Y given another random variable X is 
    \begin{equation}
    H(Y|X) = \mathbb{E}[-\log(p(y|x))] = -\sum_{x\in \mathcal{X}, y\in \mathcal{Y}} p(x,y) \log\frac{p(x,y)}{p(x)}, 
    \end{equation}
    for two support sets $\mathcal{X}, \mathcal{Y}.$ If we assume entropy to be a measure of information, in other words uncertainty within a random variable, conditional entropy essentially captures what is left of the uncertainty after conditioning. $H(Y|X)$ also has certain desirable properties: (1). $H(Y|X)=0$ if and only if $X$ completely determines the value of $Y$; (2). $H(Y|X)=H(Y)$ if and only if $X$ and $Y$ are completely independent; and (3). $H(Y|X_1, X_2) = H(Y, X_1, X_2) - H(X_1, X_2).$ We can then easily calculate conditional entropy when conditioning on multiple random variables, and use it as a metric for uncertain information. 
    
    \item \textit{Discretization of ZC proxy scores and ground-truth accuracies.} \\
    Calculating conditional entropy as prescribed above requires that all random variables be discrete, which is not the case for raw validation accuracies and ZC proxy scores. Implementation wise, we discretize all the float values and use Sturge's rule \citep{scott2009sturges} as a heuristic to choose the number of bins for discretization:
    \begin{equation} \label{eq:sturges}
    n_{\text{bins}} = \text{round}(1+3.322*\log(N))), \text{where $N$ is the sample size.}  
    \end{equation}
    Therefore, information about $Y$ does not reveal the exact validation accuracy but rather the interval in which the value falls.  
    
    \item \textit{Interpreting the information gain heatmap.} \\ 
    The information gain heatmap shows how much the conditional entropy of $y|z_{i_1}$ decreases to $y|z_{i_1}, z_{i_2}$ as the scores of ZC proxy on each column ($z_{i_2}$) is revealed, given that we already know the scores of ZC proxy on each row ($z_{i_1}$). For instance, on Figure \ref{fig:cond_entropy} (top right), the value $1.42$ on the second row, first column shows that $H(y|scores(\text{synflow}) - H(y|scores(\text{synflow}), scores(\text{epe\_nas})) = 1.42$. Note that (1). all values on the diagonal are $0.0$ because no information is gained when we add a copy of the existing ZC proxy scores; (2). The heatmap is \textbf{not} symmetric like pairwise conditional entropy. The order in which conditioning is applied affects the amount of information gain, i.e. $\textbf{IG}(y|z_{i_1}, z_{i_2}) \neq \textbf{IG}(y|z_{i_2}, z_{i_1})$; (3). \textbf{IG} measures how much one ZC proxy's information complements that of another for determining the ground-truth accuracy. It does \textbf{not} serve as a direct indicator of the quality of individual ZC proxy themselves. 
    
    \item \textit{Interpreting the entropy vs. number of ZC proxies plot.} \\
    Conditional entropy monotonically decreases as we condition the validation accuracy, $y$, on an increasing amount of ZC proxy scores, $z_{i_1}, \ldots z_{i_k}$, which always brings in additional information. In most cases, marginal \textbf{IG} drastically decreases as the amount of ZC proxies $k$ reaches 4, but this is only true if the proxies are chosen strategically, using either a greedy or a brute-force minimization approach. For the majority of benchmarks, the less computationally intensive greedy strategy matches up to the brute-force strategy. On the other hand, randomly choosing the ZC proxies does not have stable performance and could be suboptimal, such as on NAS-Bench-201 + CIFAR-100 in Figure \ref{fig:cond_entropy} (bottom middle). 
    
\end{itemize}

\begin{figure}[t]
    \centering
    \includegraphics[width=.95\linewidth]{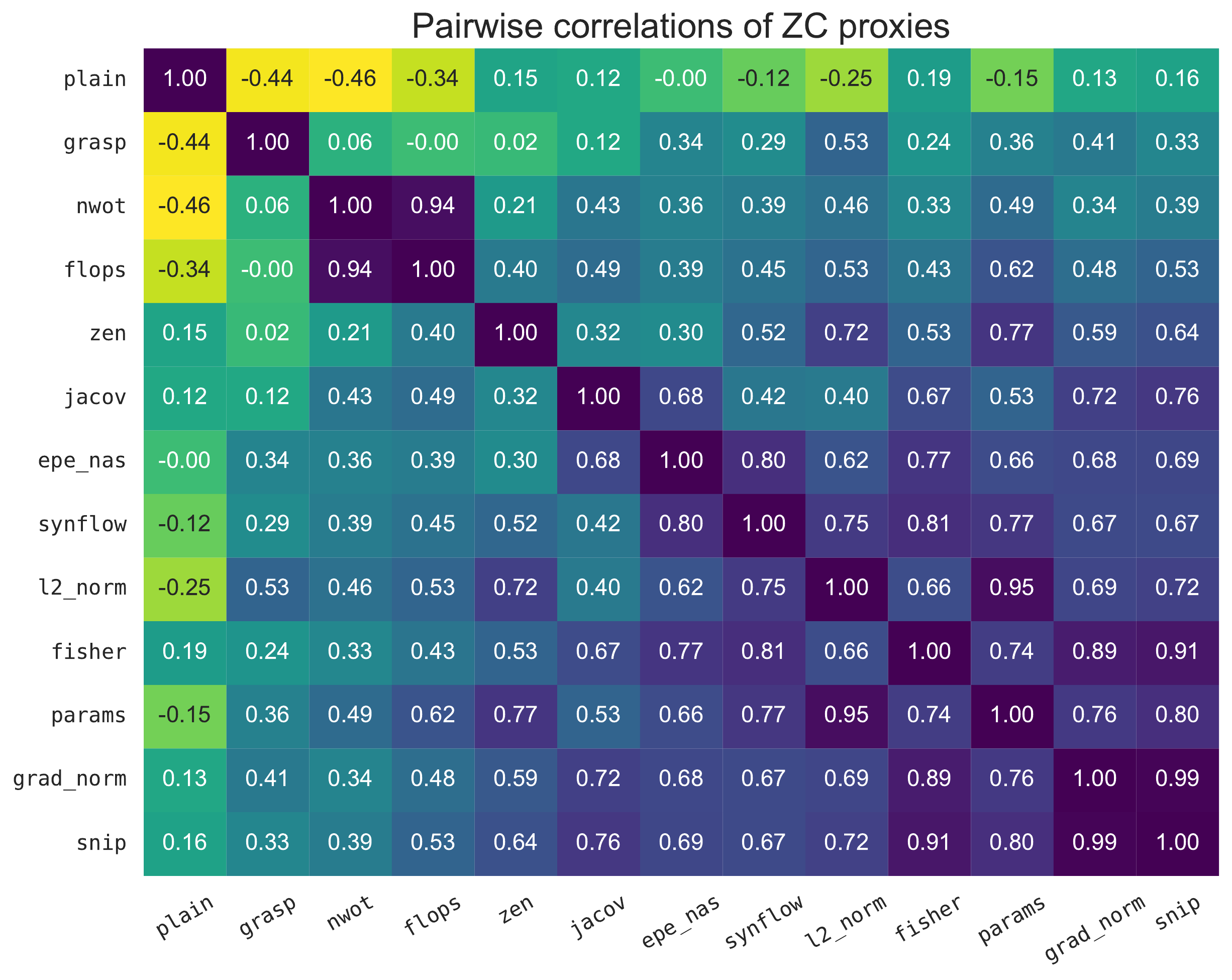}
    \caption{Pearson correlation coefficient for each pair of ZC proxies, averaged over all benchmarks. The entries in the plot are ordered based on the mean score across each row and column.}		\label{fig:pairwise_corr_avg}
\end{figure}

For completion, in Figure \ref{fig:pairwise_corr_avg}, we plot the average pairwise correlation for all pairs of ZC proxies.

\begin{figure}[ht]
    \centering
    \includegraphics[width=.32\linewidth]{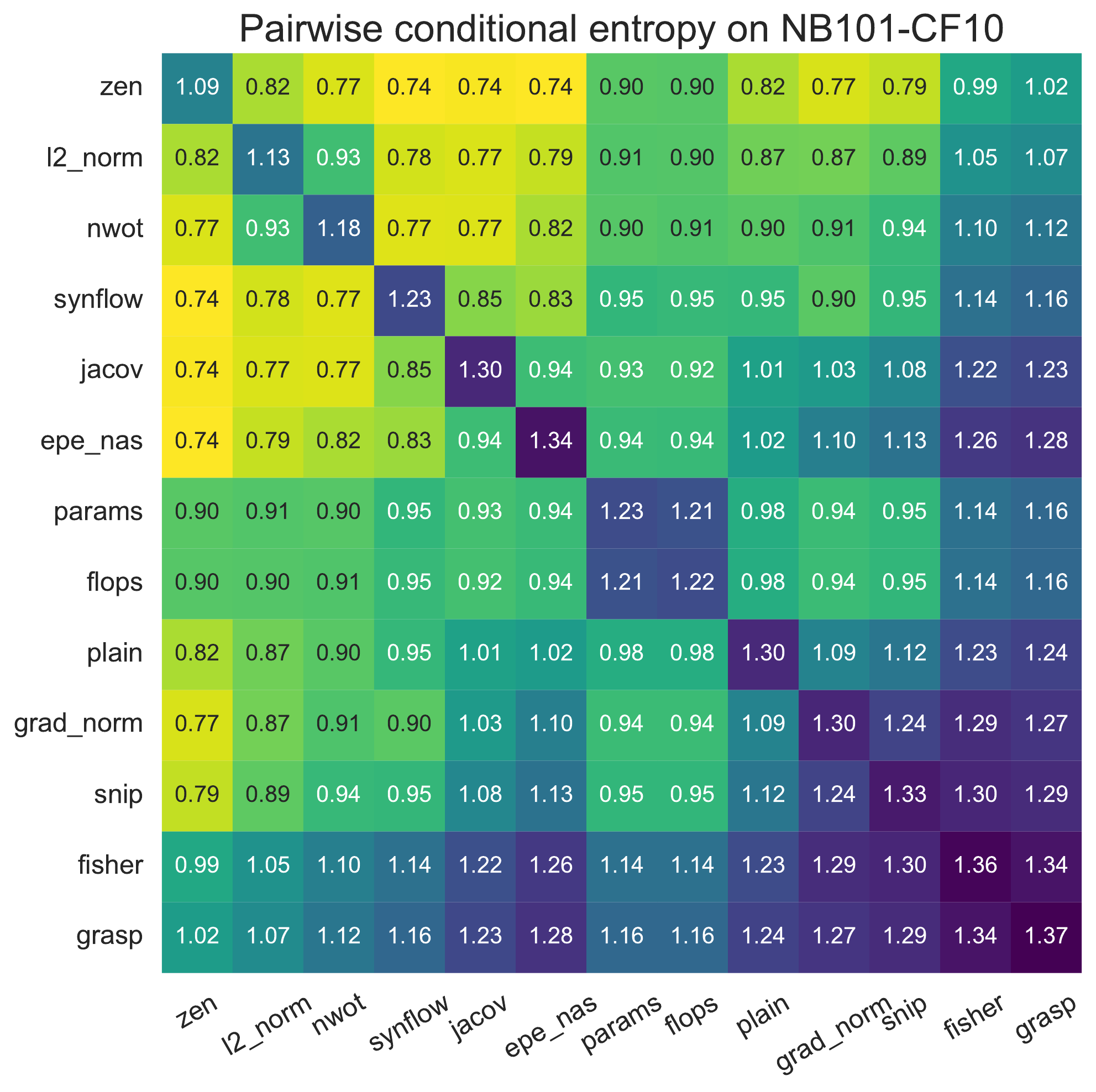}
    \includegraphics[width=.32\linewidth]{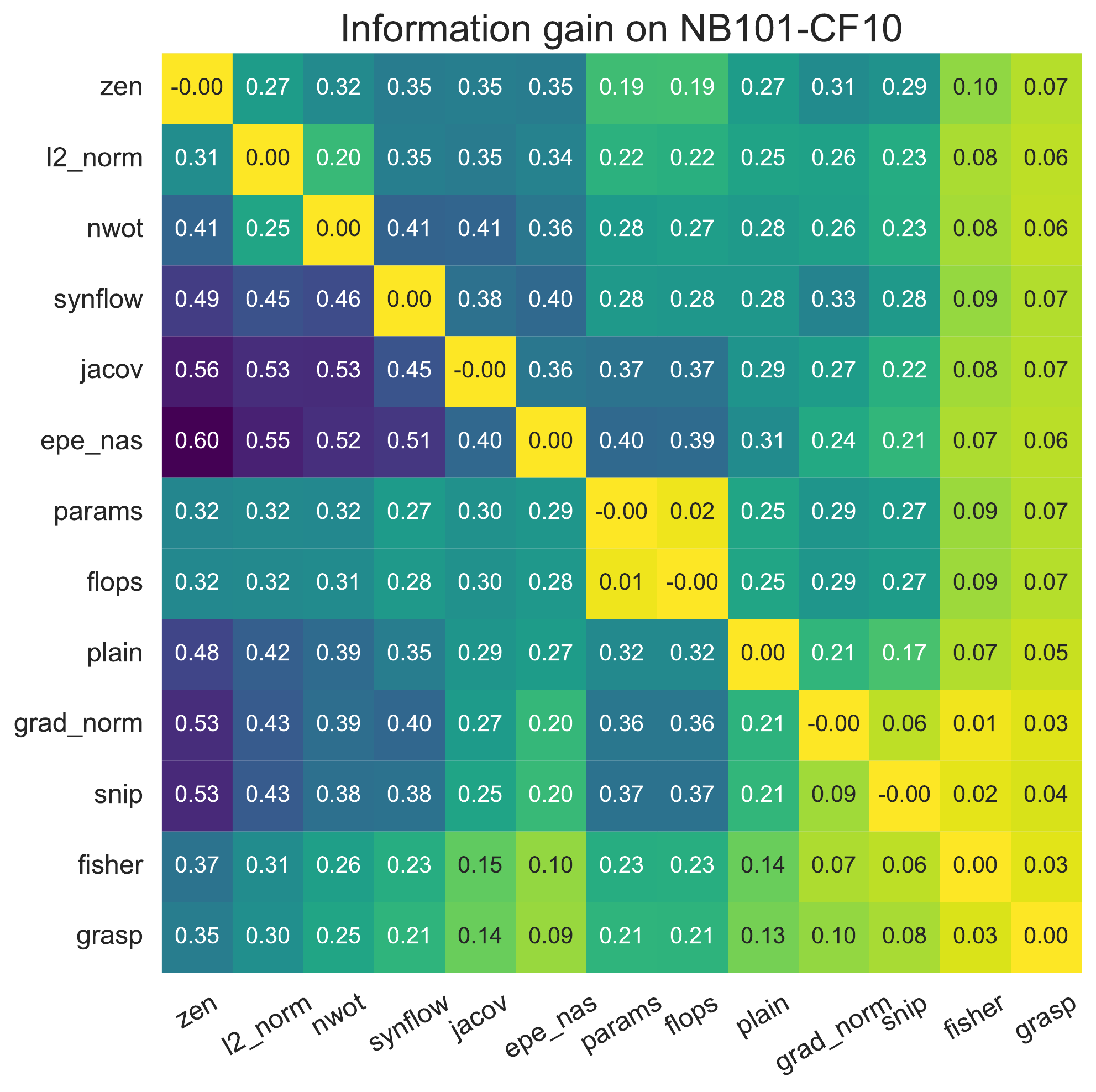}
    \includegraphics[width=.32\linewidth]{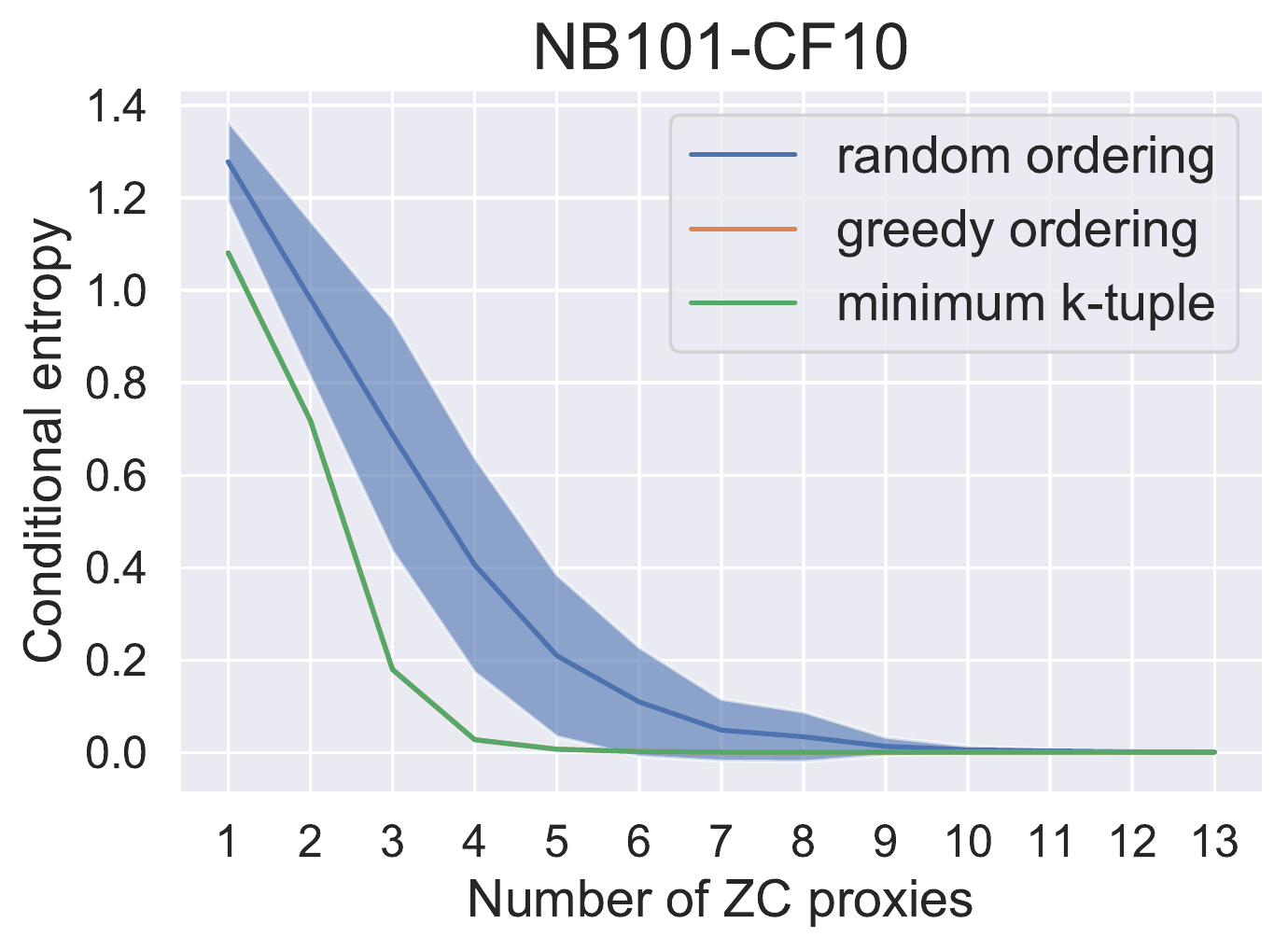} \\

    \includegraphics[width=.32\linewidth]{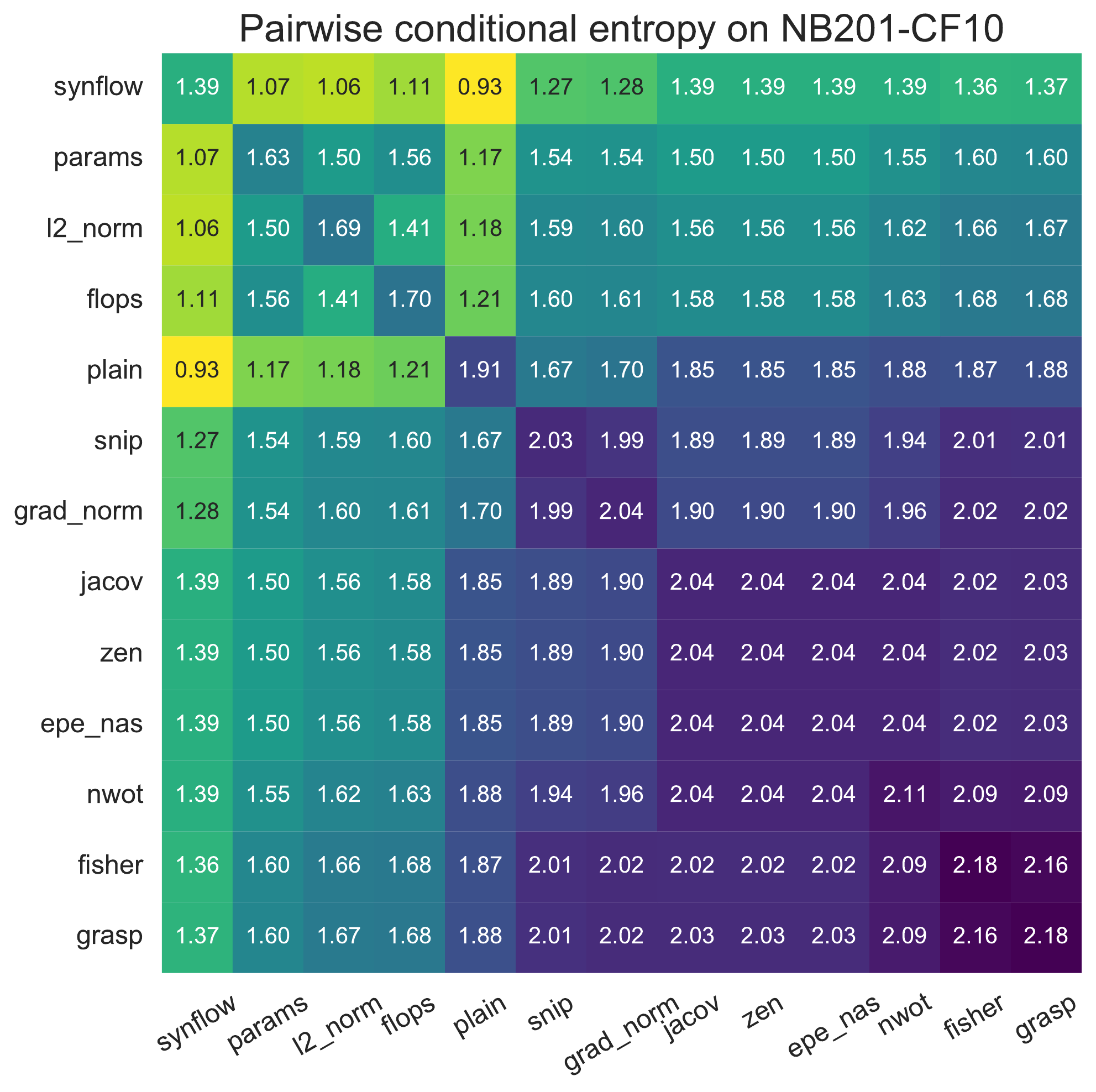}   
    \includegraphics[width=.32\linewidth]{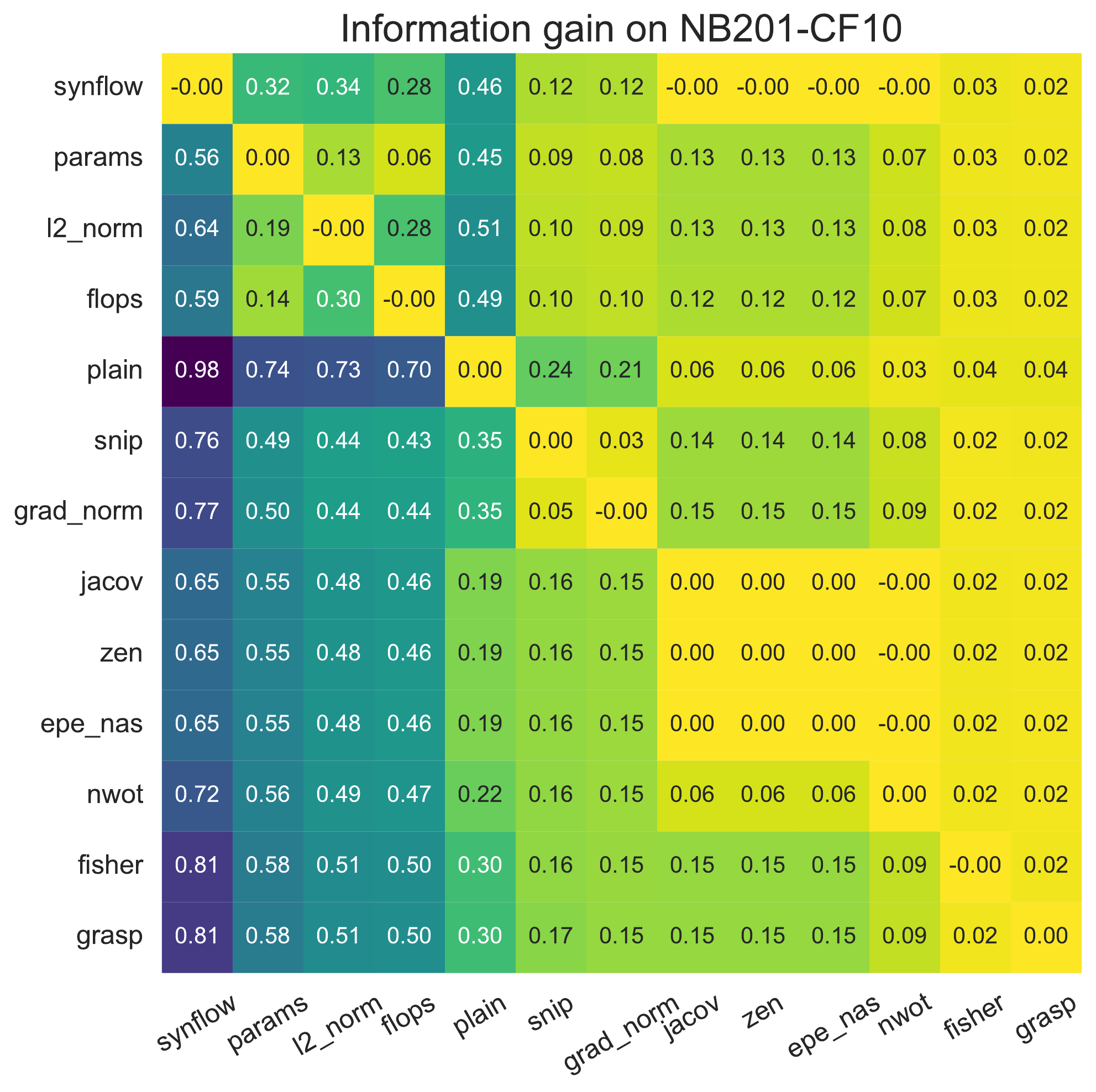}
    \includegraphics[width=.32\linewidth]{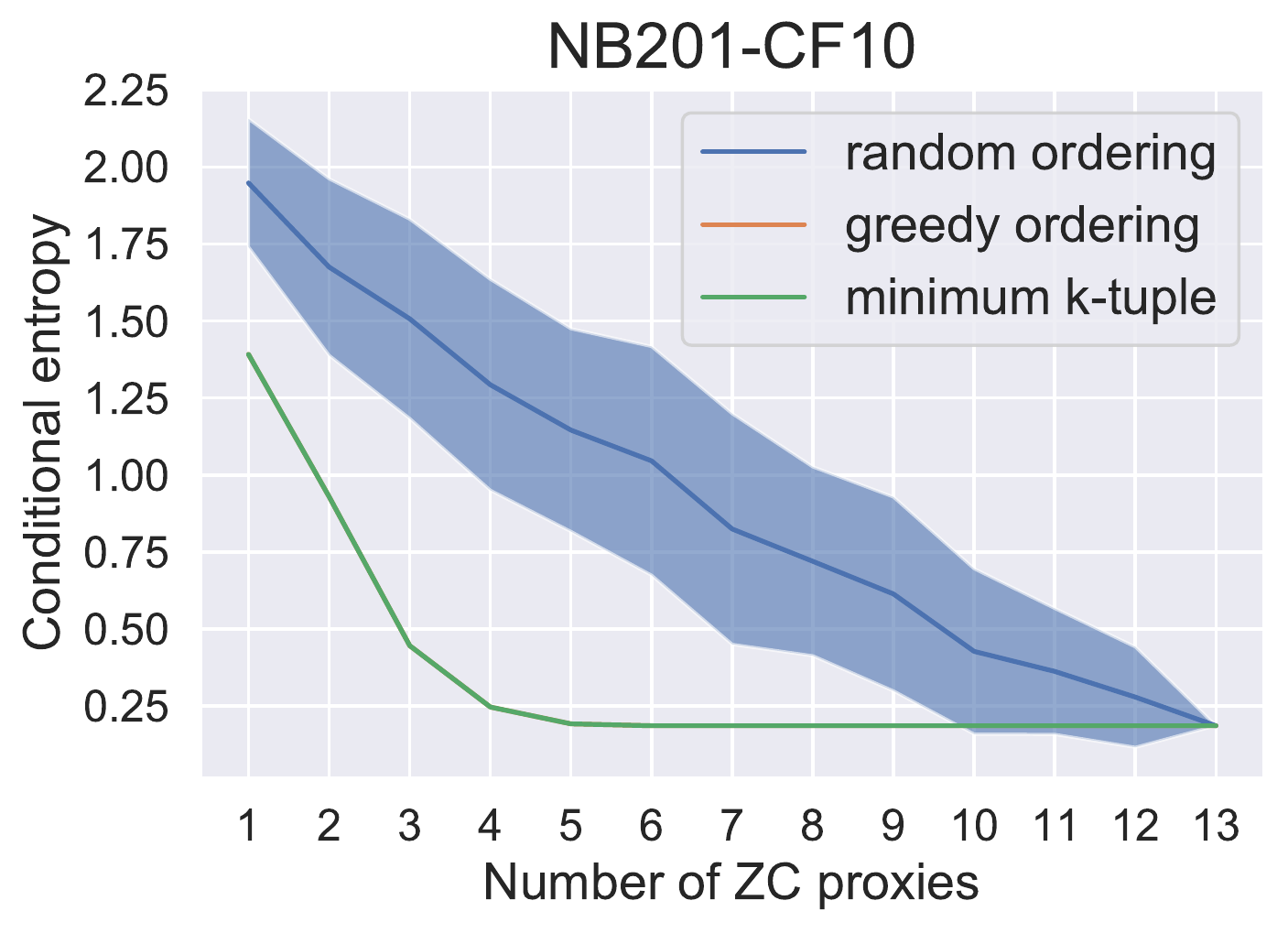}\\
    \includegraphics[width=.32\linewidth]{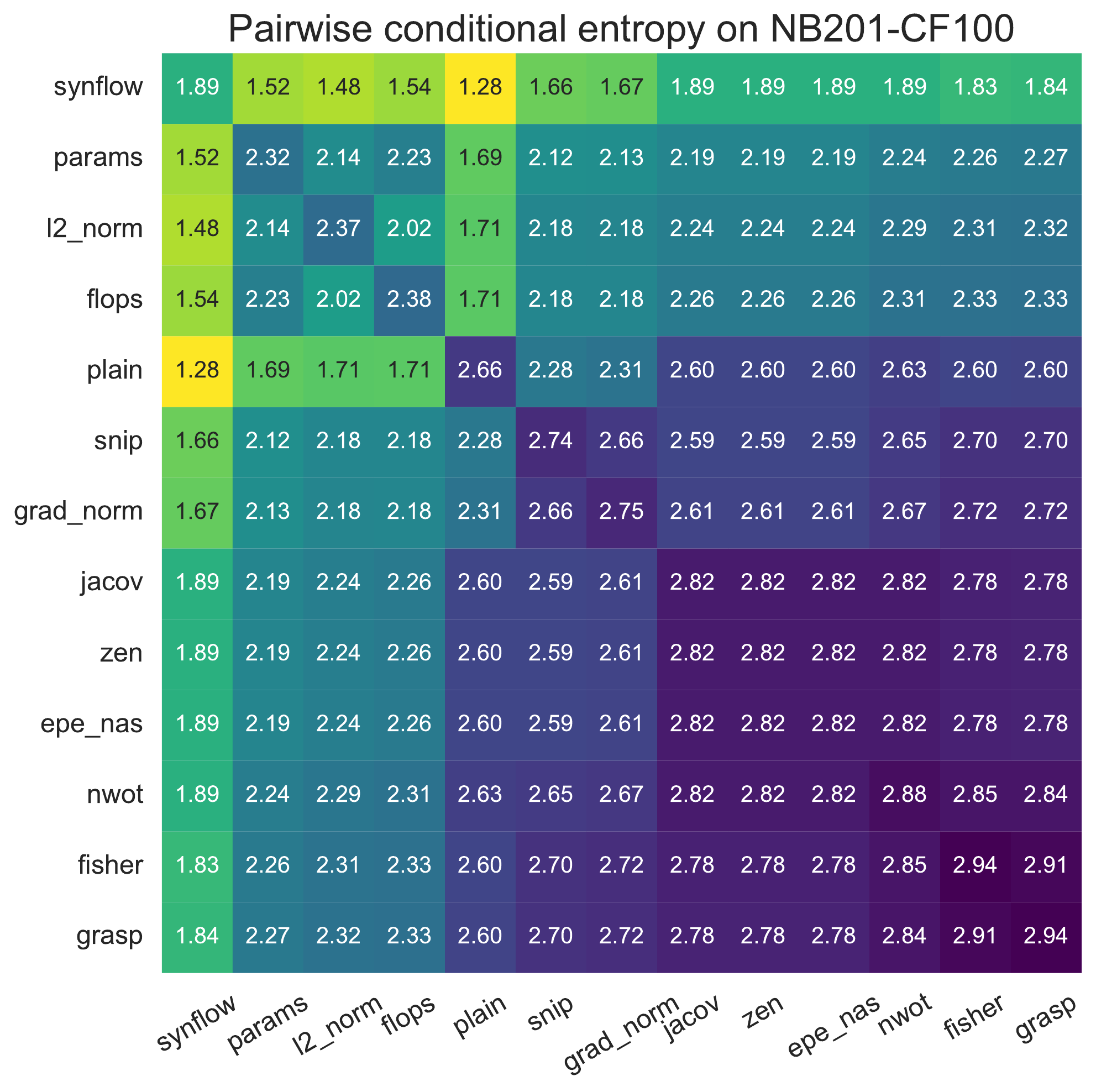}   
    \includegraphics[width=.32\linewidth]{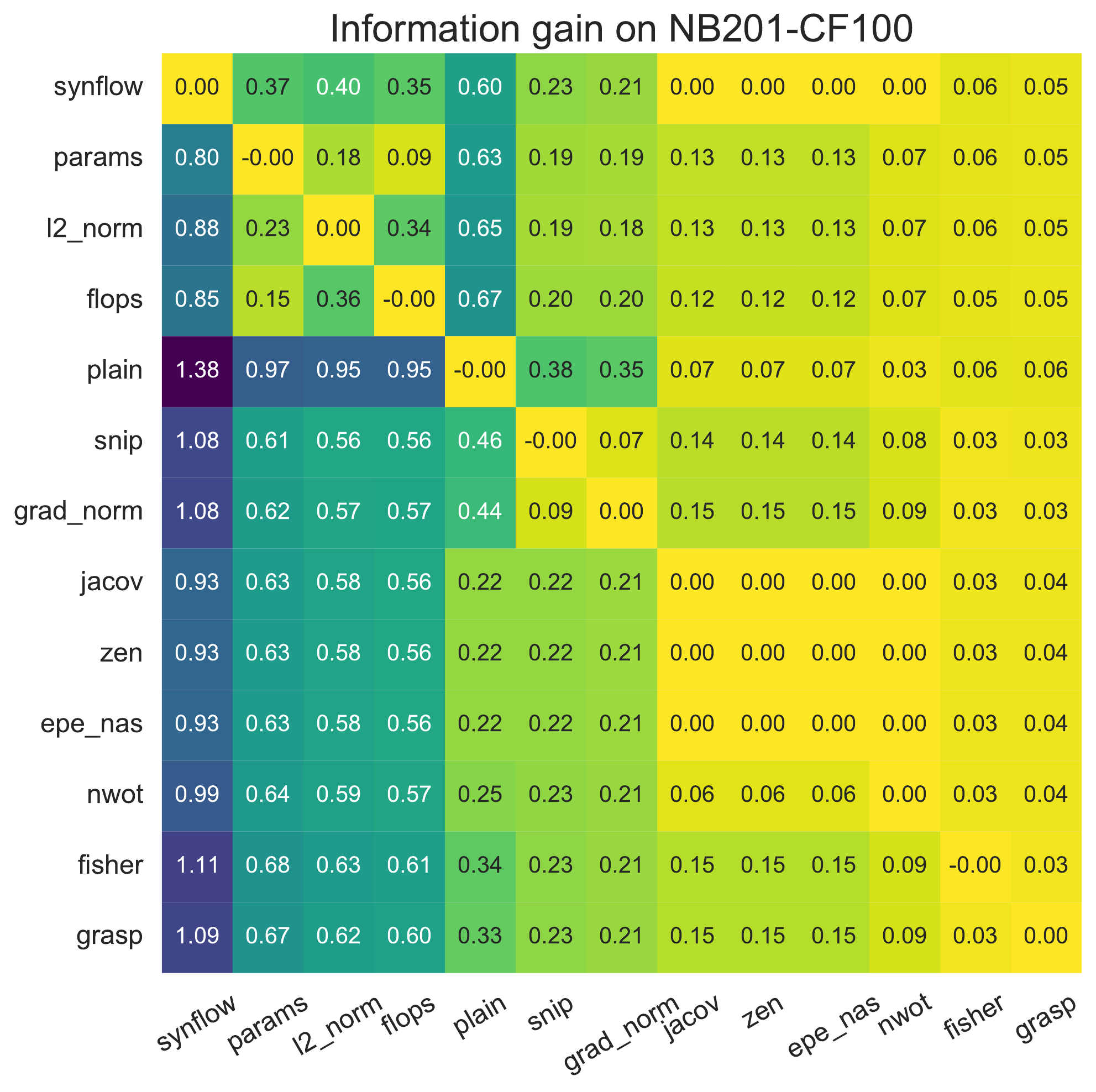}
    \includegraphics[width=.32\linewidth]{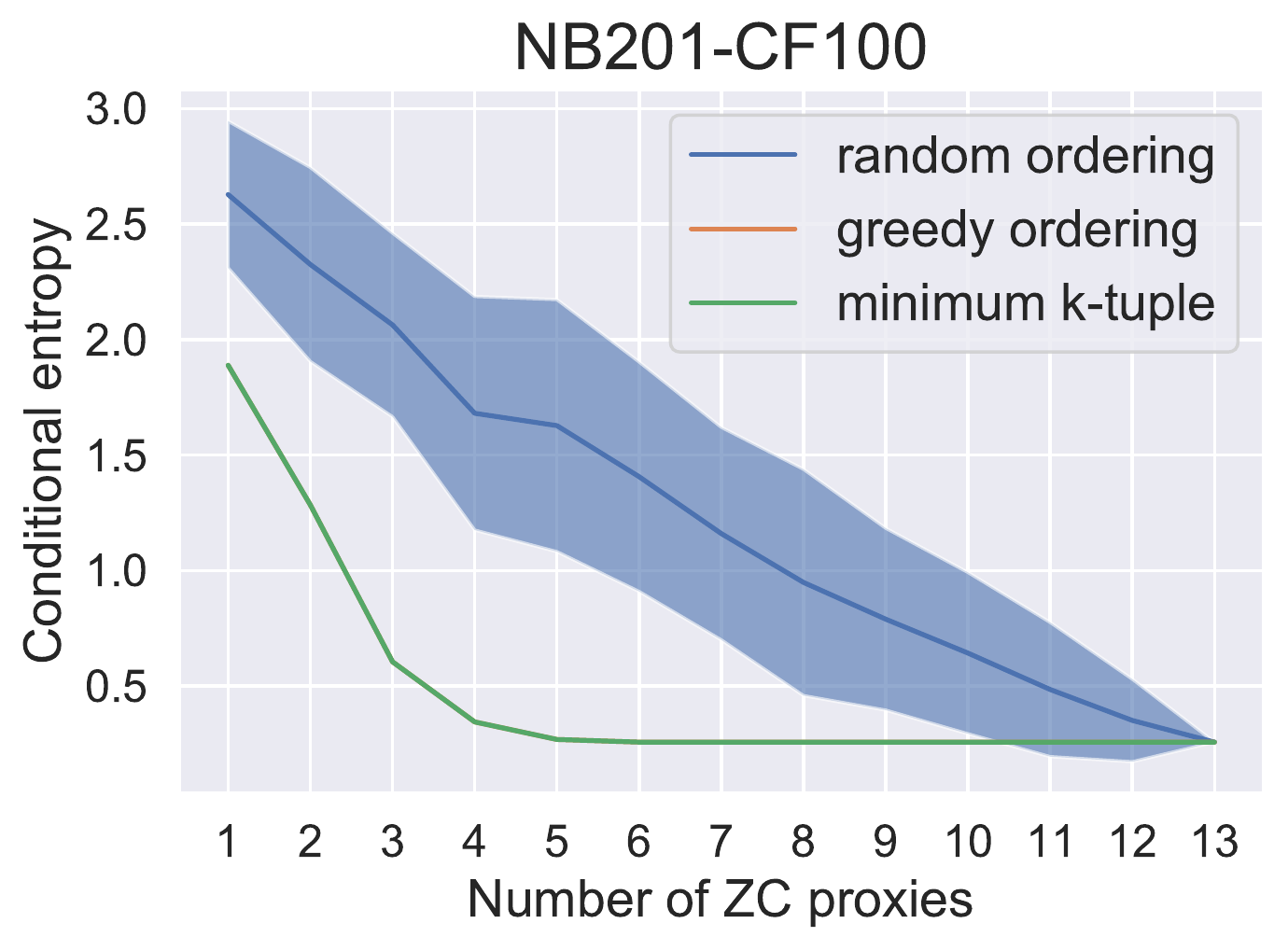}
    \\
    \includegraphics[width=.32\linewidth]{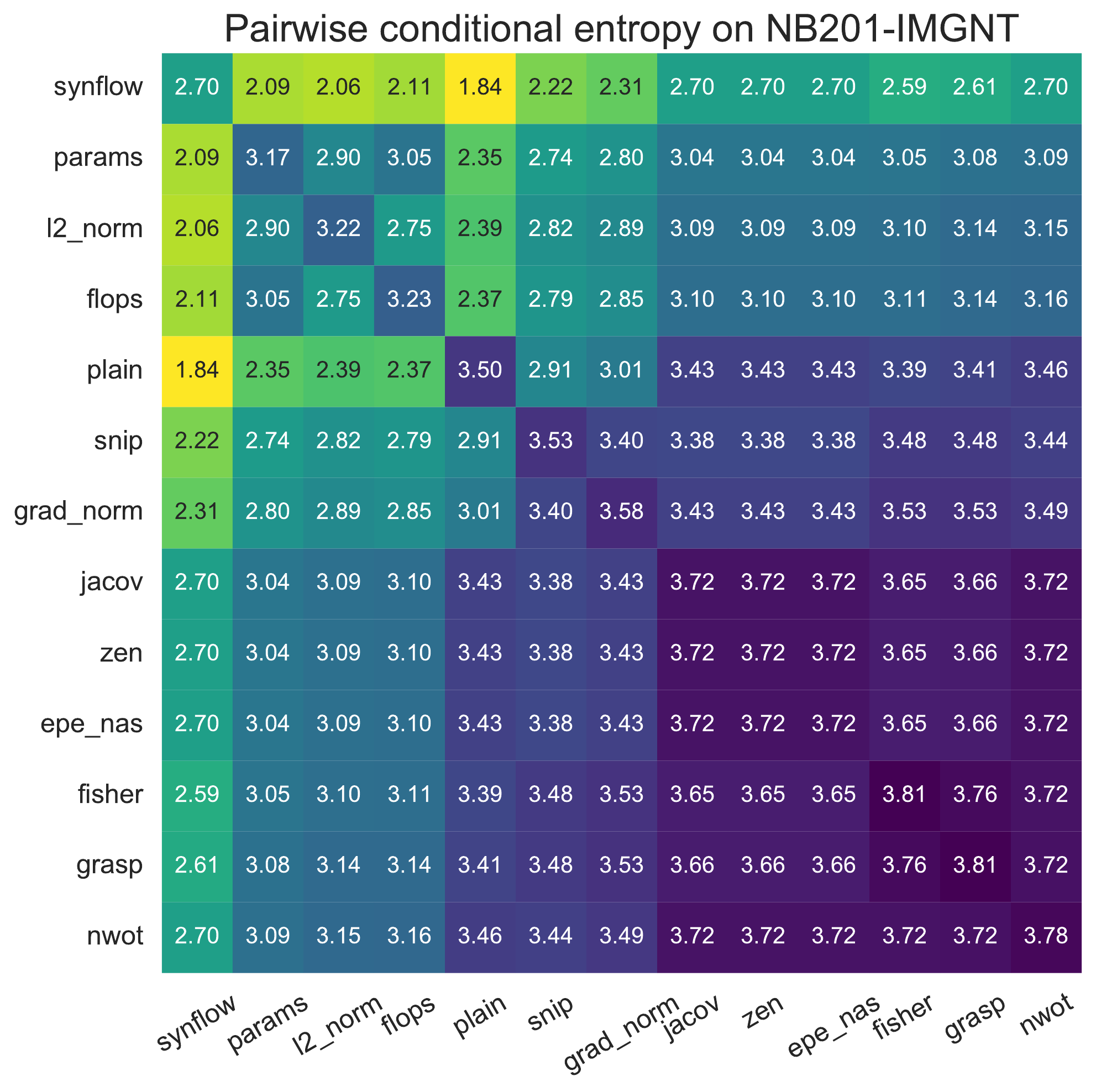}
    \includegraphics[width=.32\linewidth]{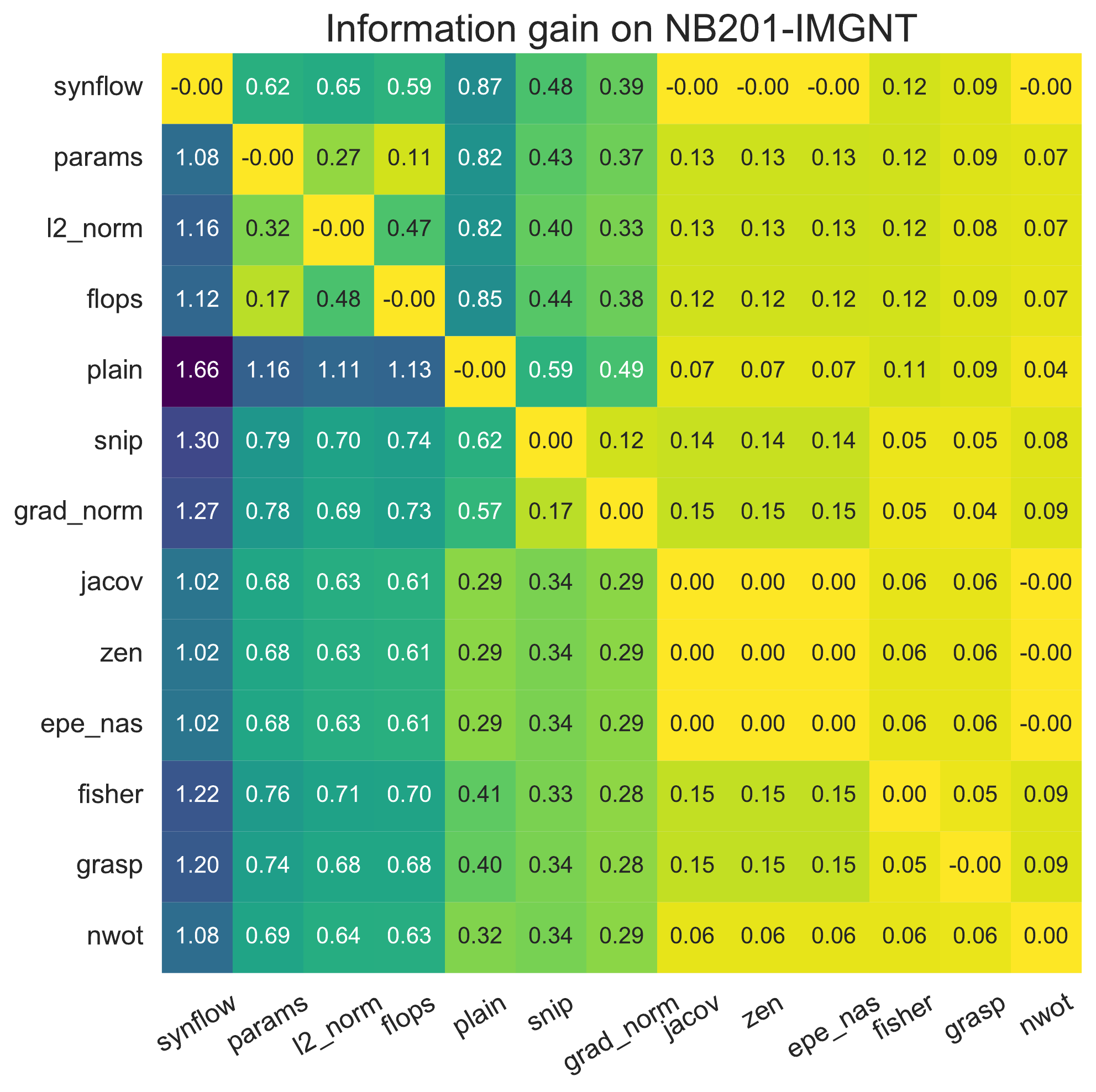}
    \includegraphics[width=.32\linewidth]{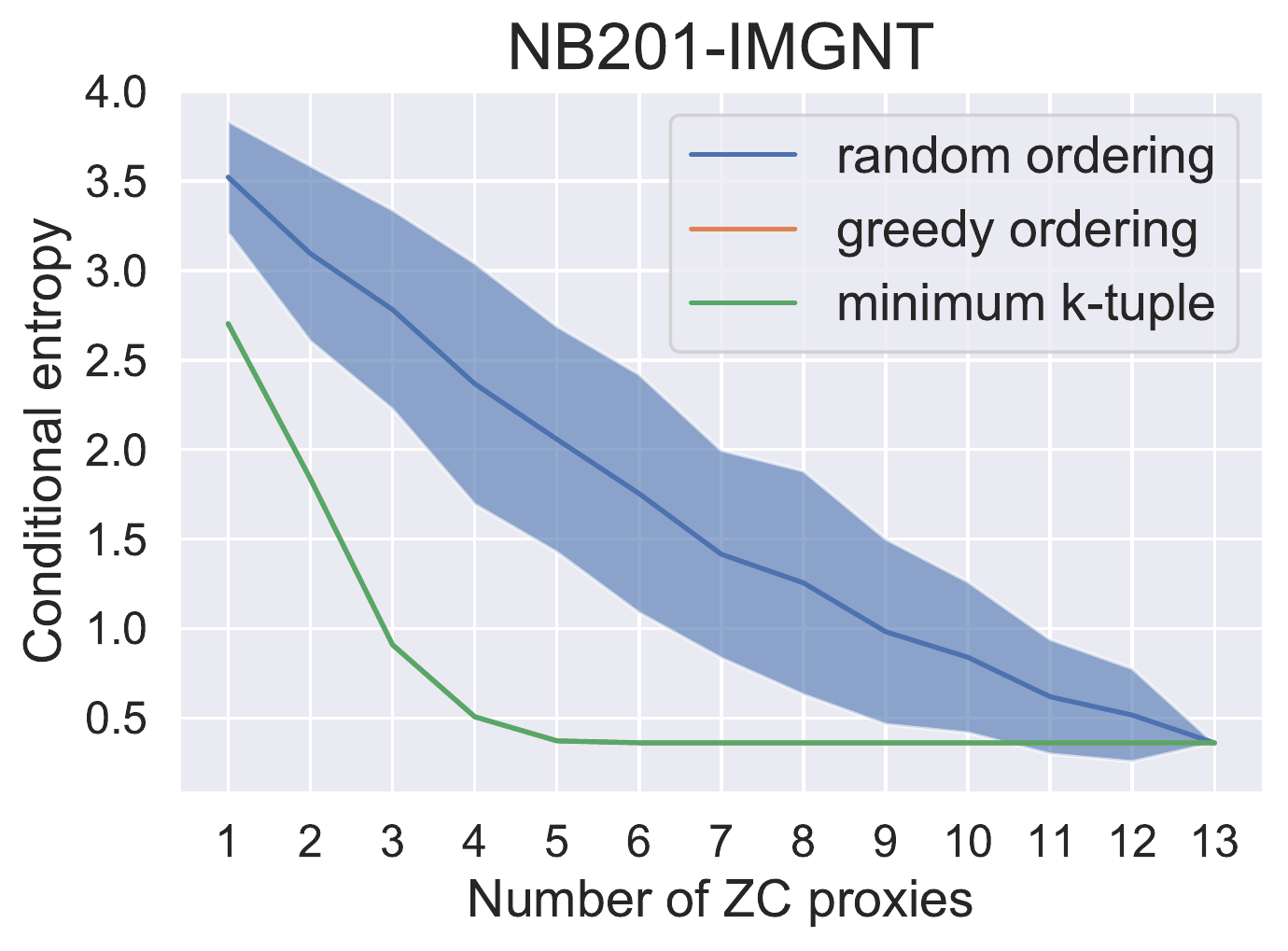}\\

    \caption{Conditional entropy and information gain (\textbf{IG}) for each ZC proxy pair across all search spaces and datasets (Left and Middle). Conditional entropy $H(y\mid z_{i_1},\dots,z_{i_k})$ vs.\ $k$, 
    where the ordering $z_{i_1},\dots,z_{i_k}$ is selected using three different strategies (Right). (1/5)}
    \label{fig:info_theory_appendix_1}
\end{figure}

\begin{figure}[ht]
    \centering

    \includegraphics[width=.32\linewidth]{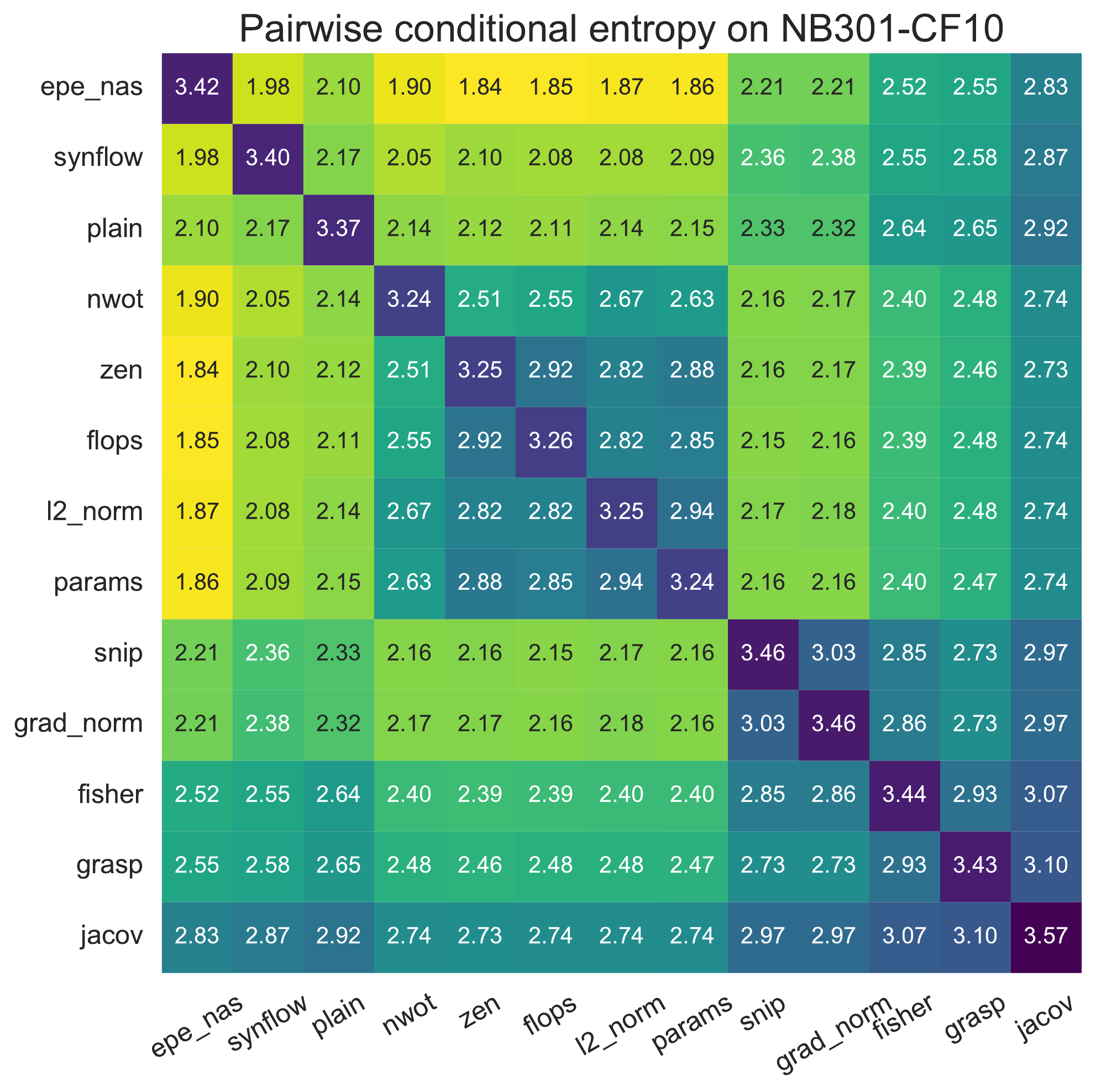}    
    \includegraphics[width=.32\linewidth]{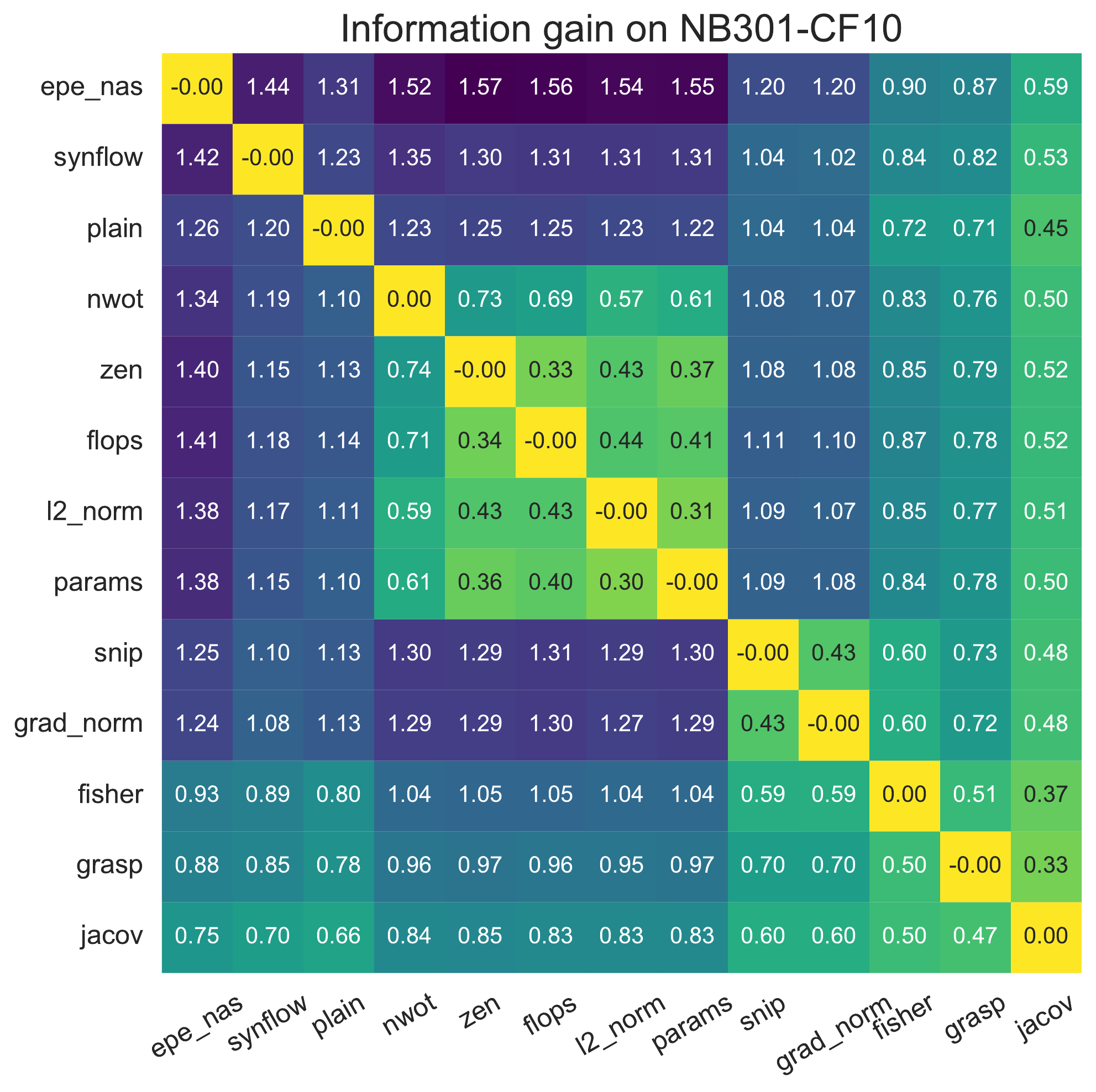}
    \includegraphics[width=.32\linewidth]{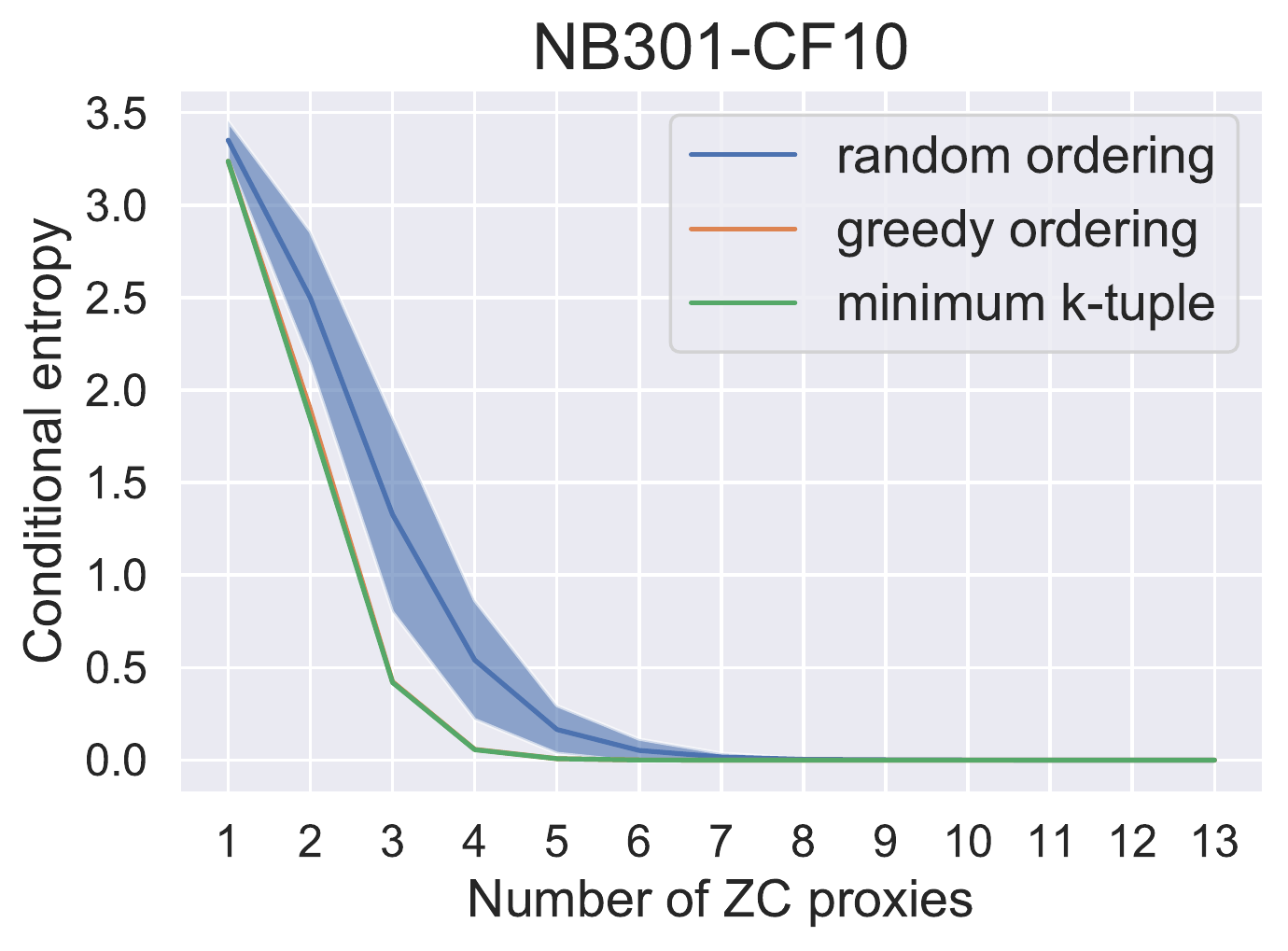}\\    
    \includegraphics[width=.32\linewidth]{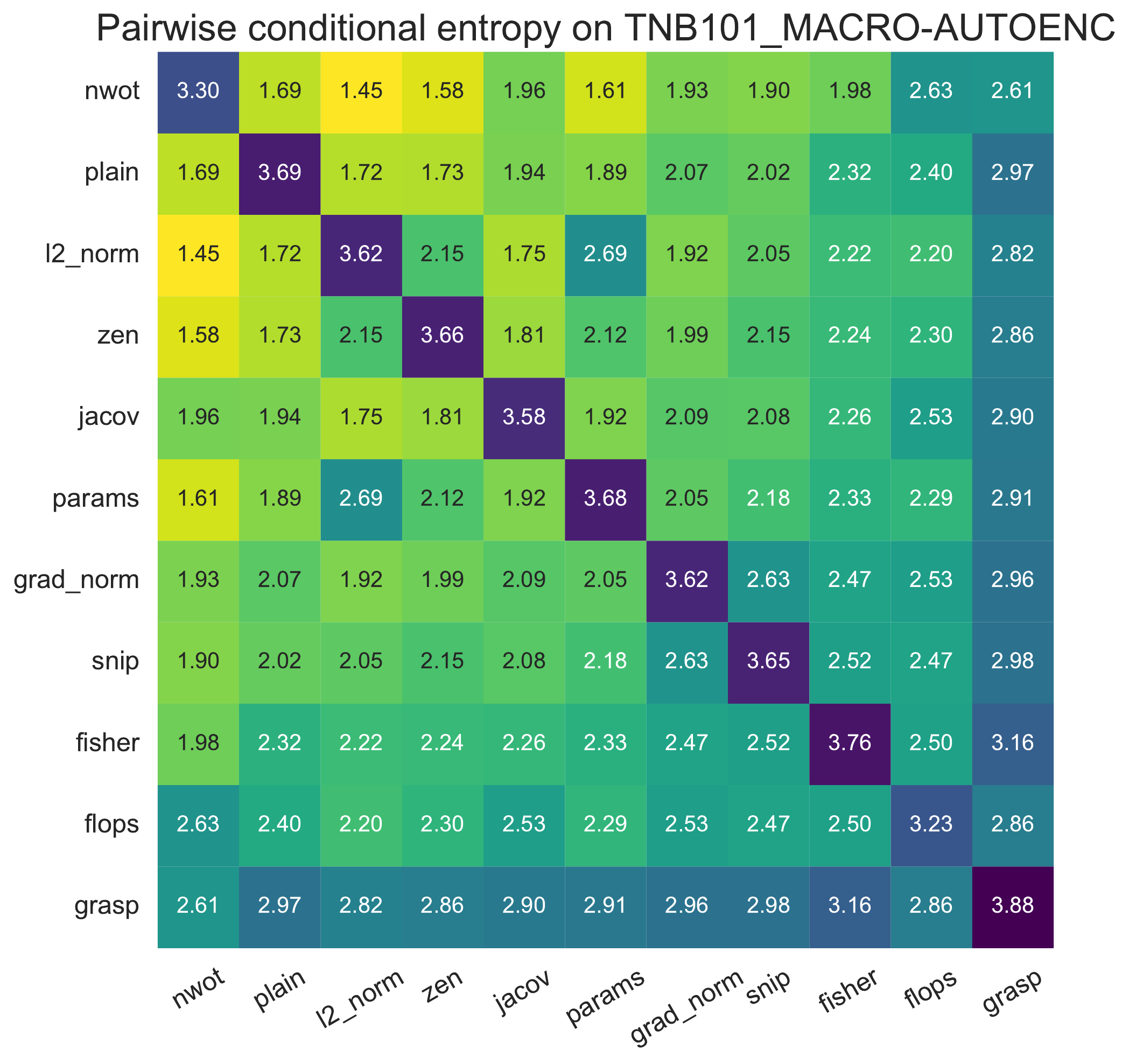}    
    \includegraphics[width=.32\linewidth]{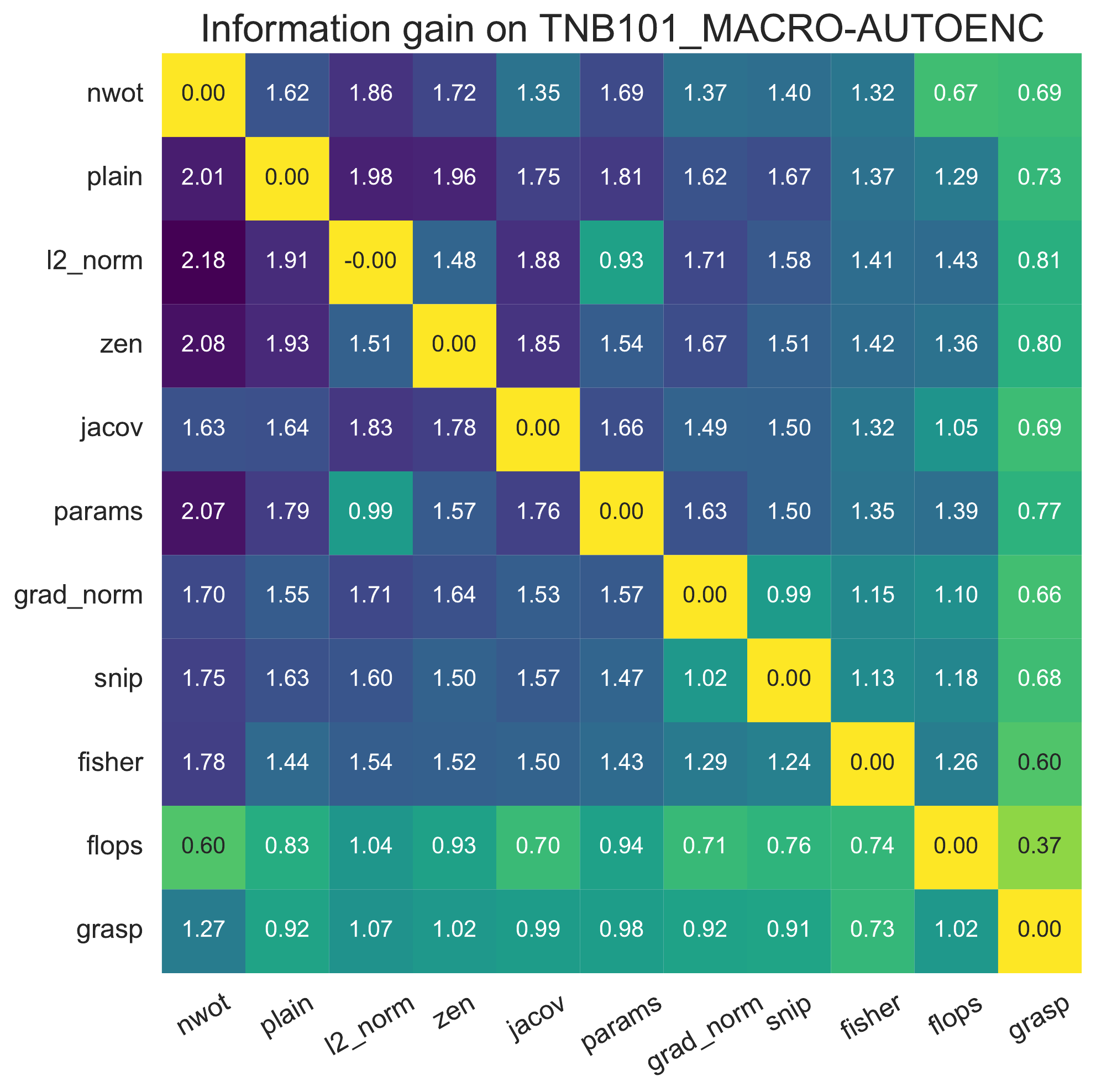}
    \includegraphics[width=.32\linewidth]{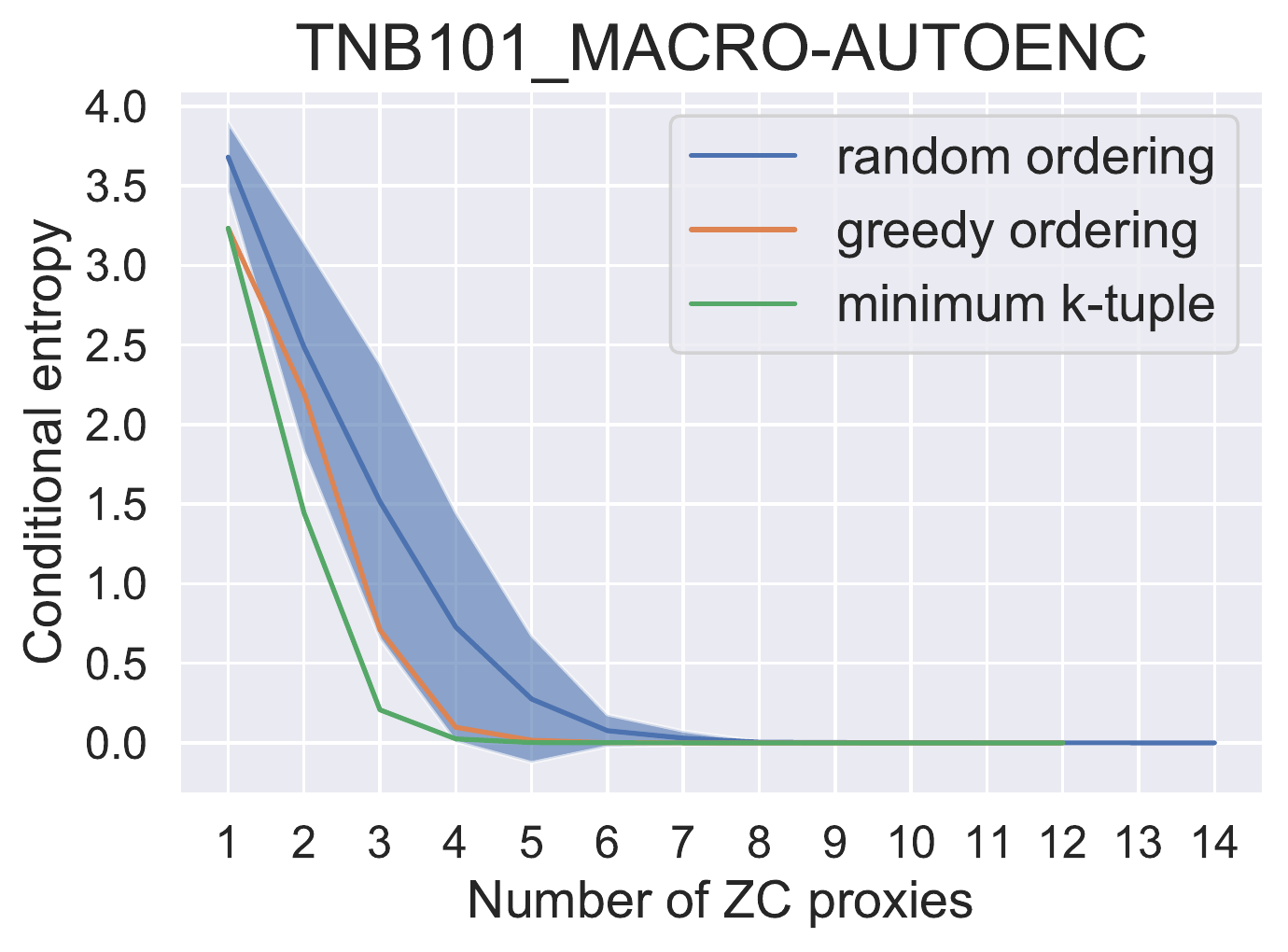}\\ 
    \includegraphics[width=.32\linewidth]{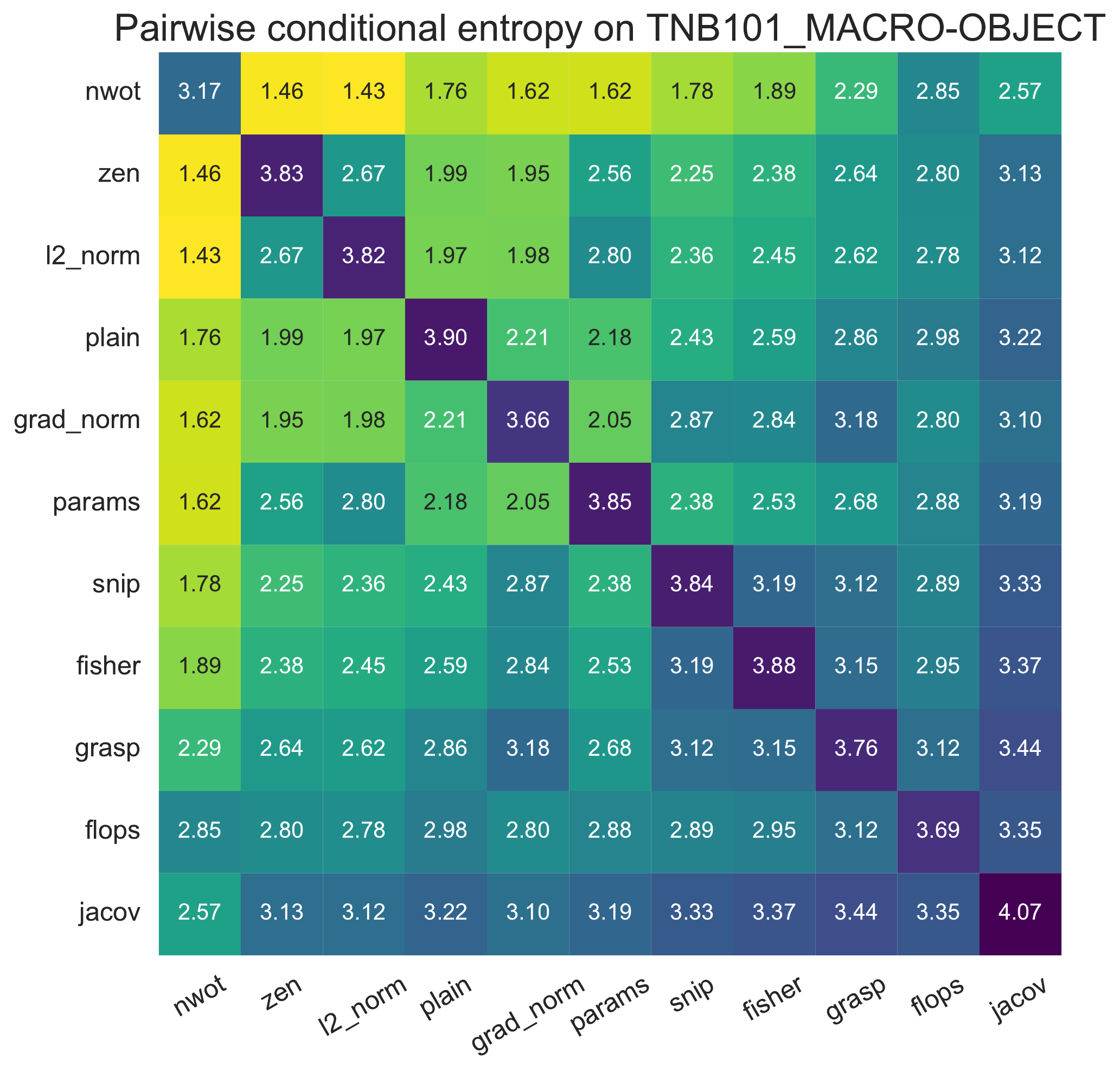}    
    \includegraphics[width=.32\linewidth]{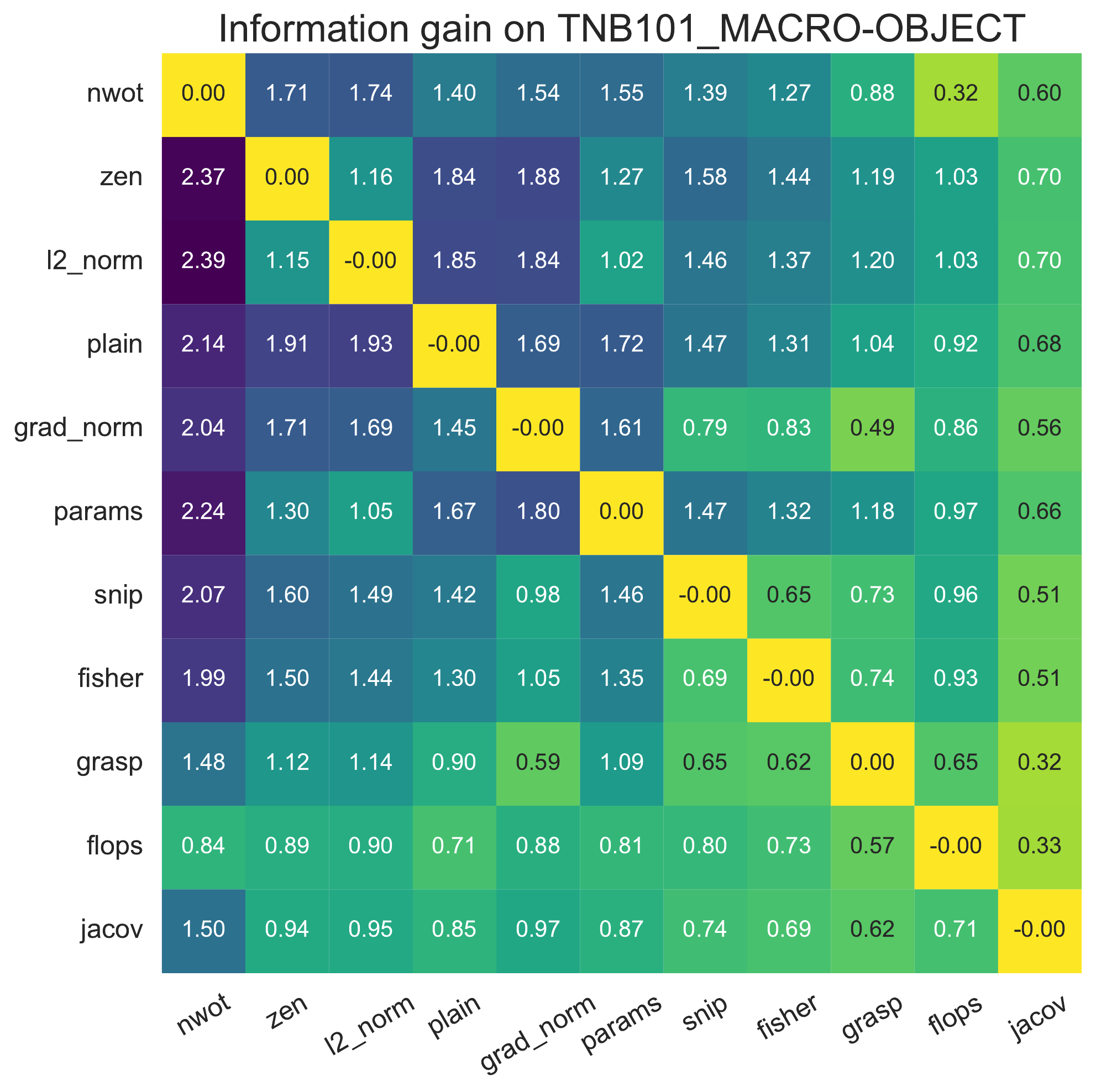}
    \includegraphics[width=.32\linewidth]{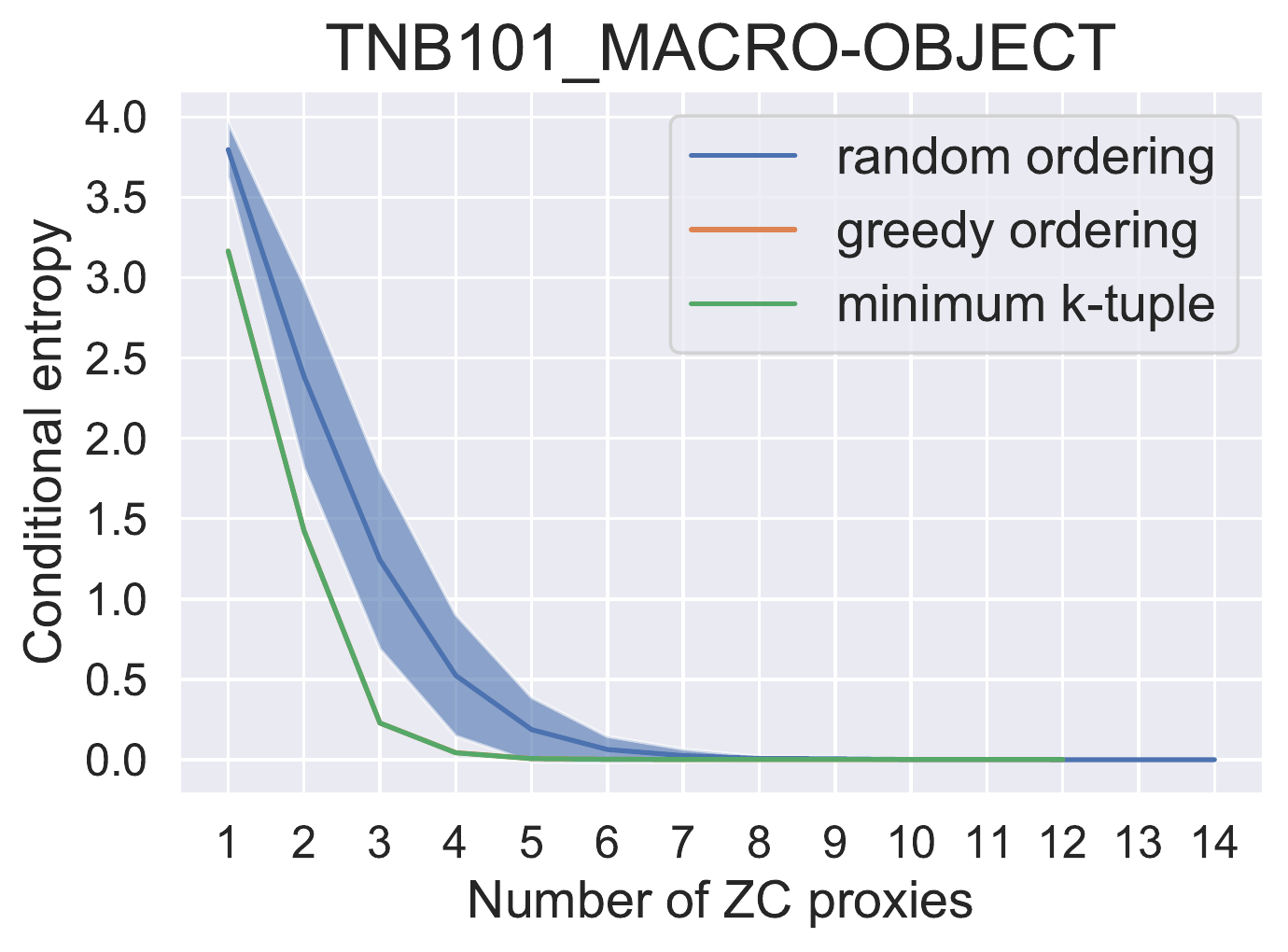}\\   
    \includegraphics[width=.32\linewidth]{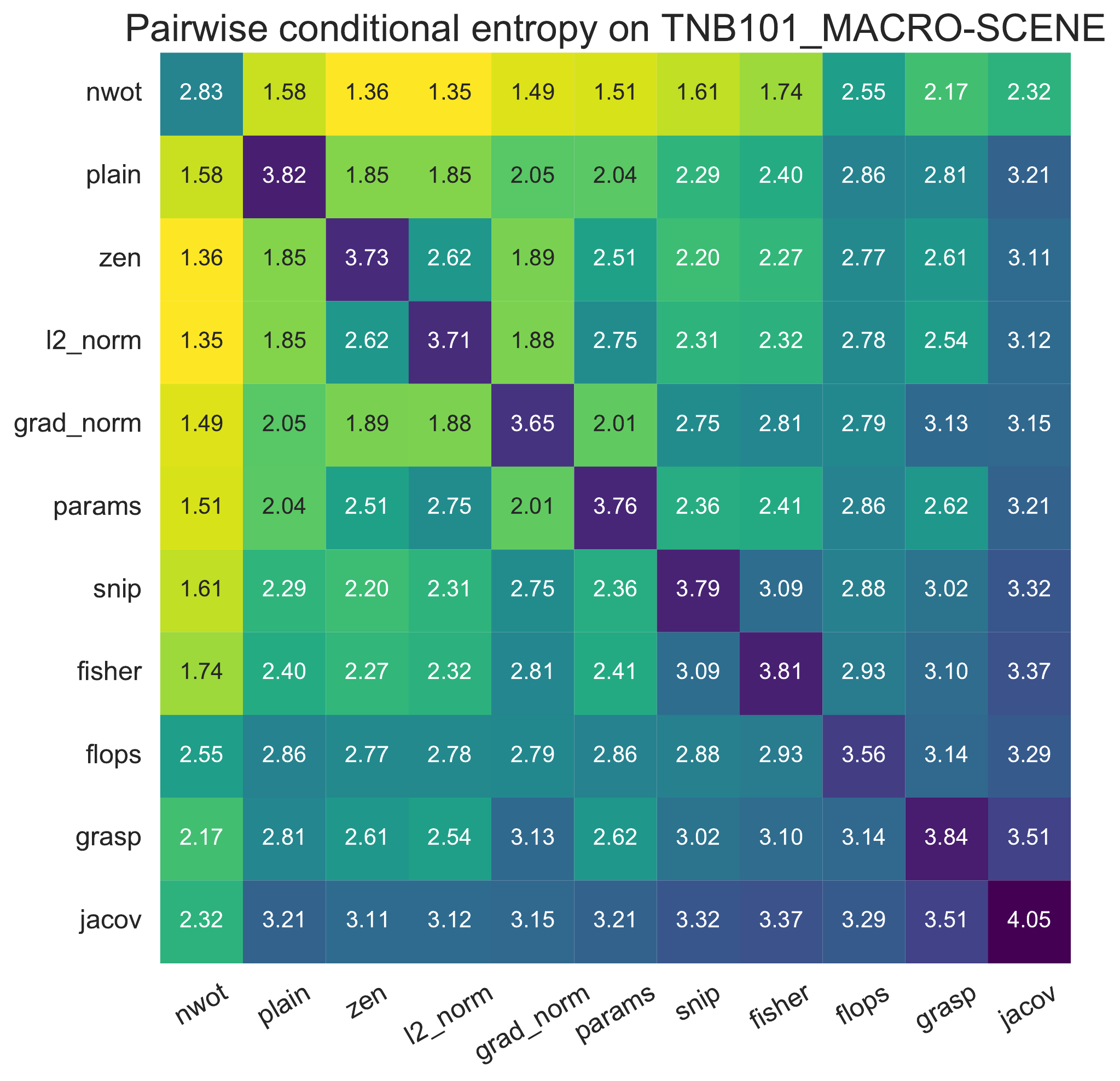}    
    \includegraphics[width=.32\linewidth]{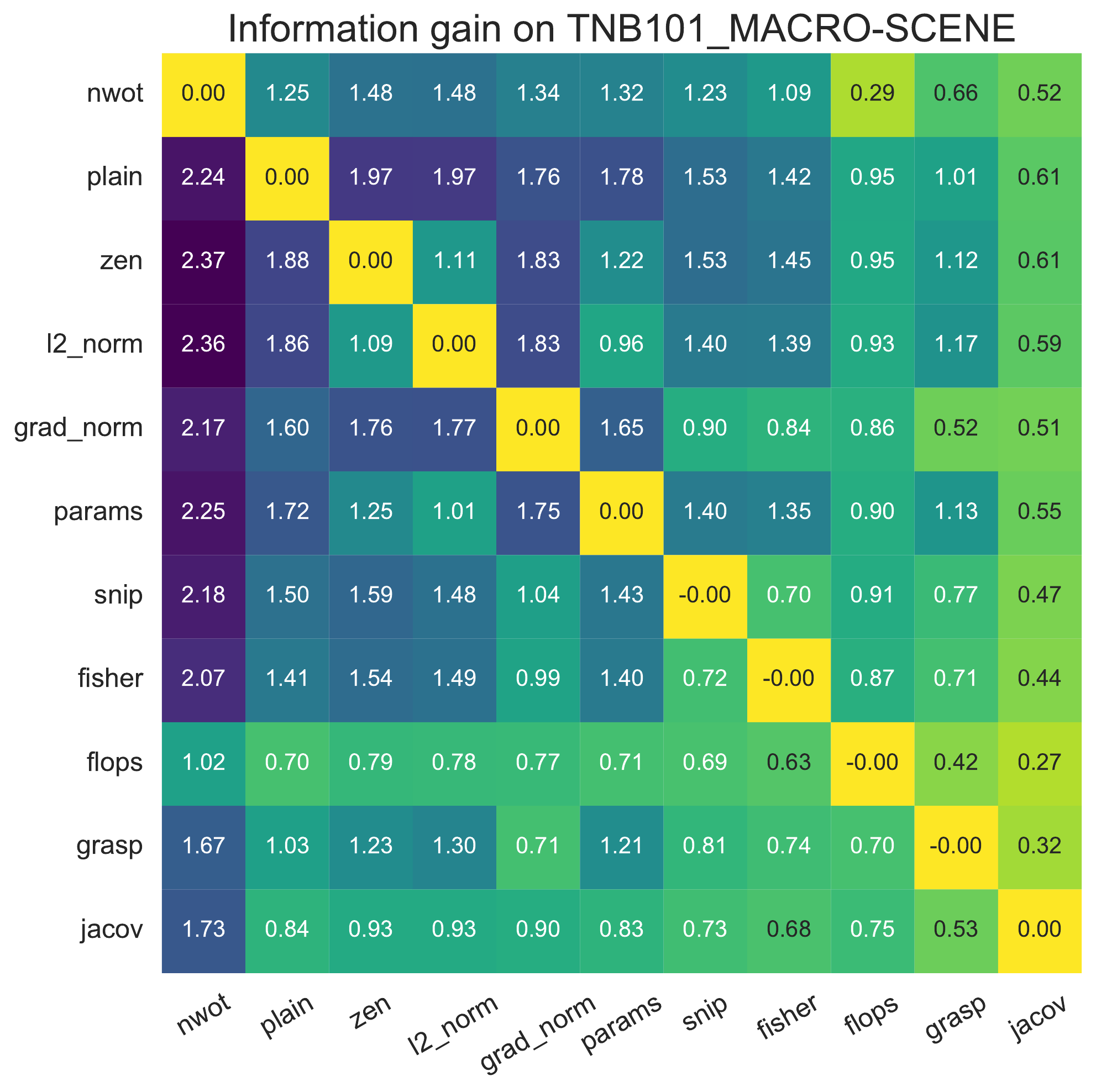}
    \includegraphics[width=.32\linewidth]{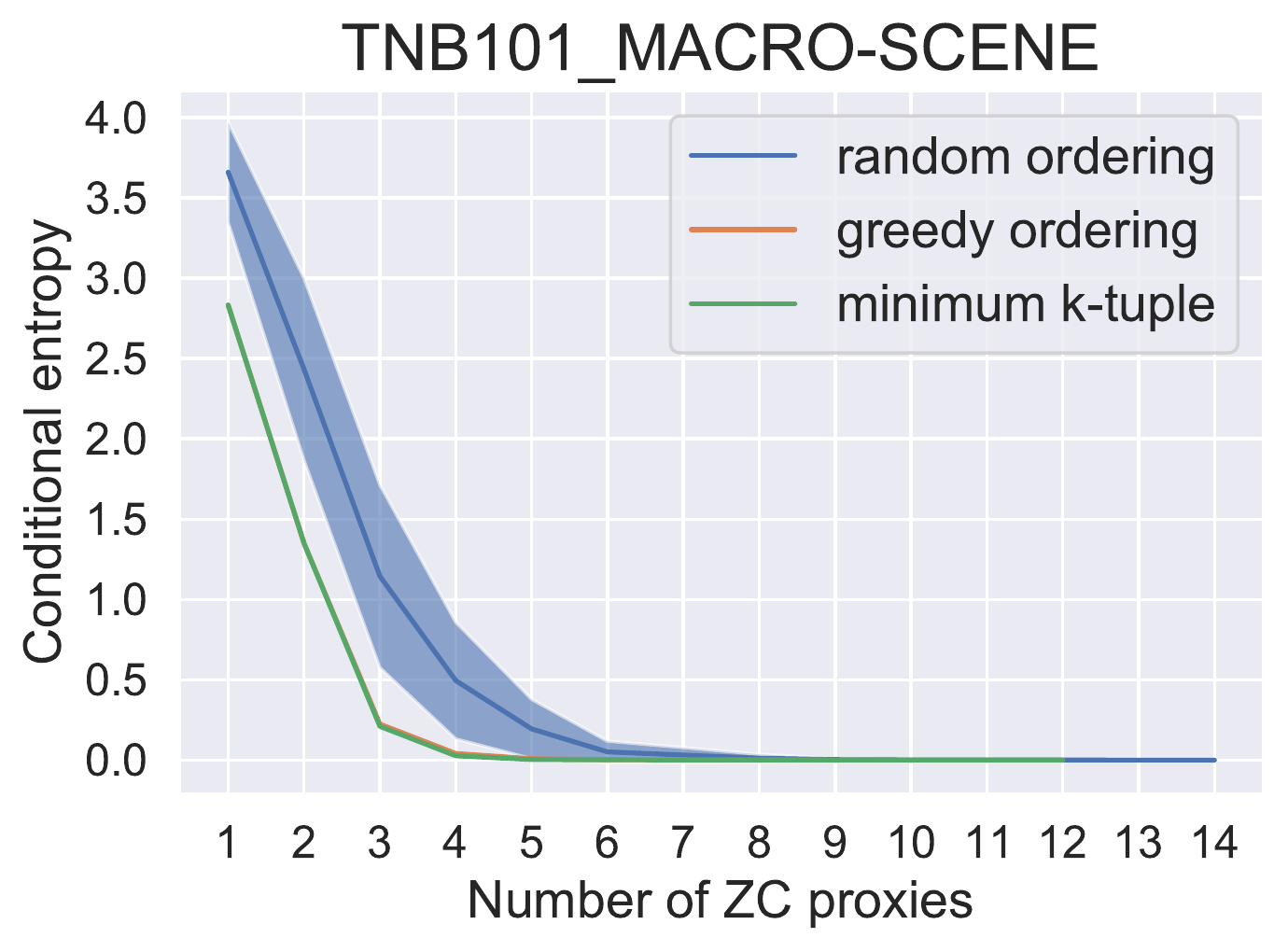}\\ 

    \caption{Conditional entropy and information gain (\textbf{IG}) for each ZC proxy pair across all search spaces and datasets (Left and Middle). Conditional entropy $H(y\mid z_{i_1},\dots,z_{i_k})$ vs.\ $k$, 
    where the ordering $z_{i_1},\dots,z_{i_k}$ is selected using three different strategies (Right). (2/5)}

    \label{fig:info_theory_appendix_2}
\end{figure}

\begin{figure}[ht]
    \centering
   
    \includegraphics[width=.32\linewidth]{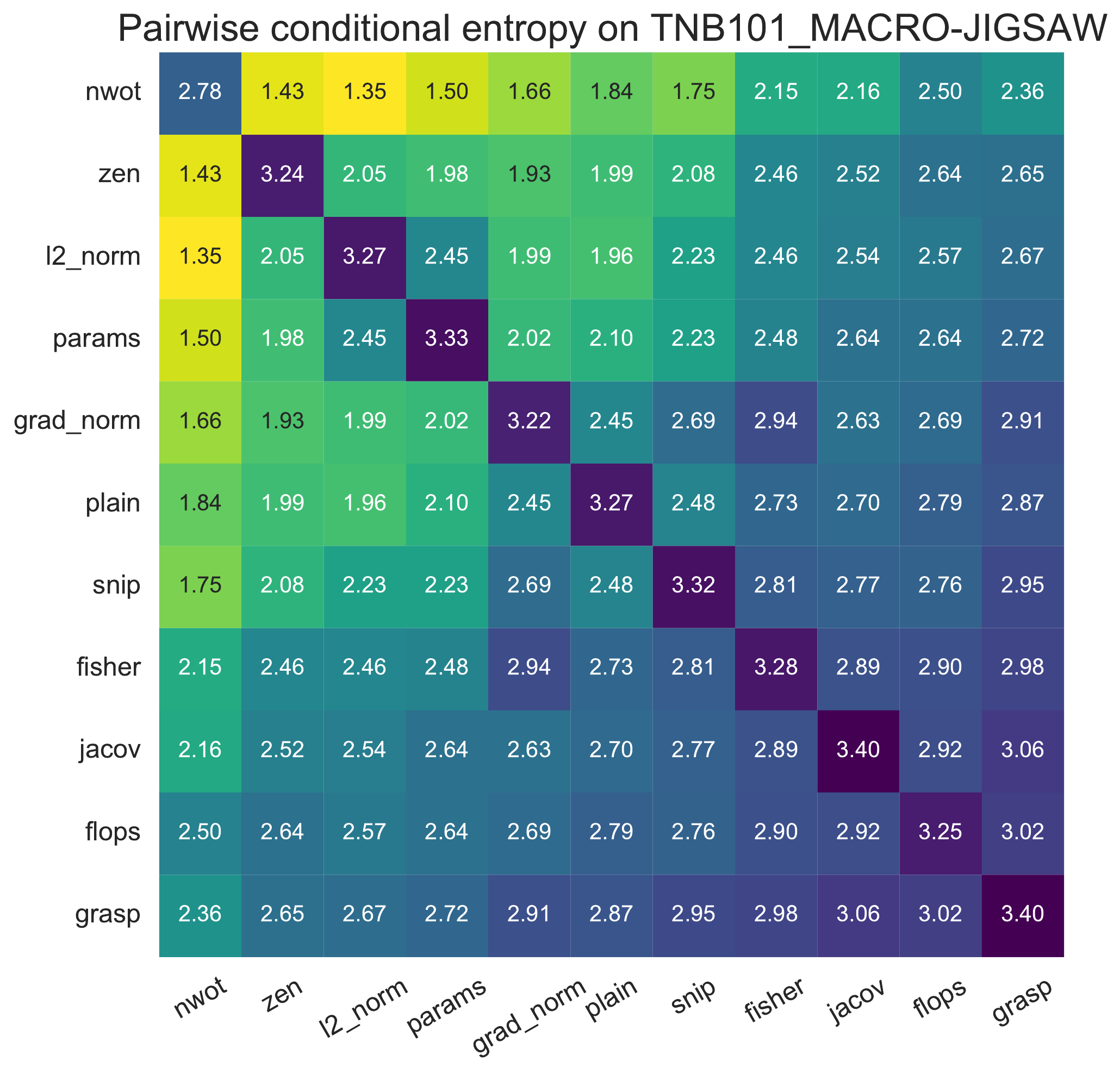}    
    \includegraphics[width=.32\linewidth]{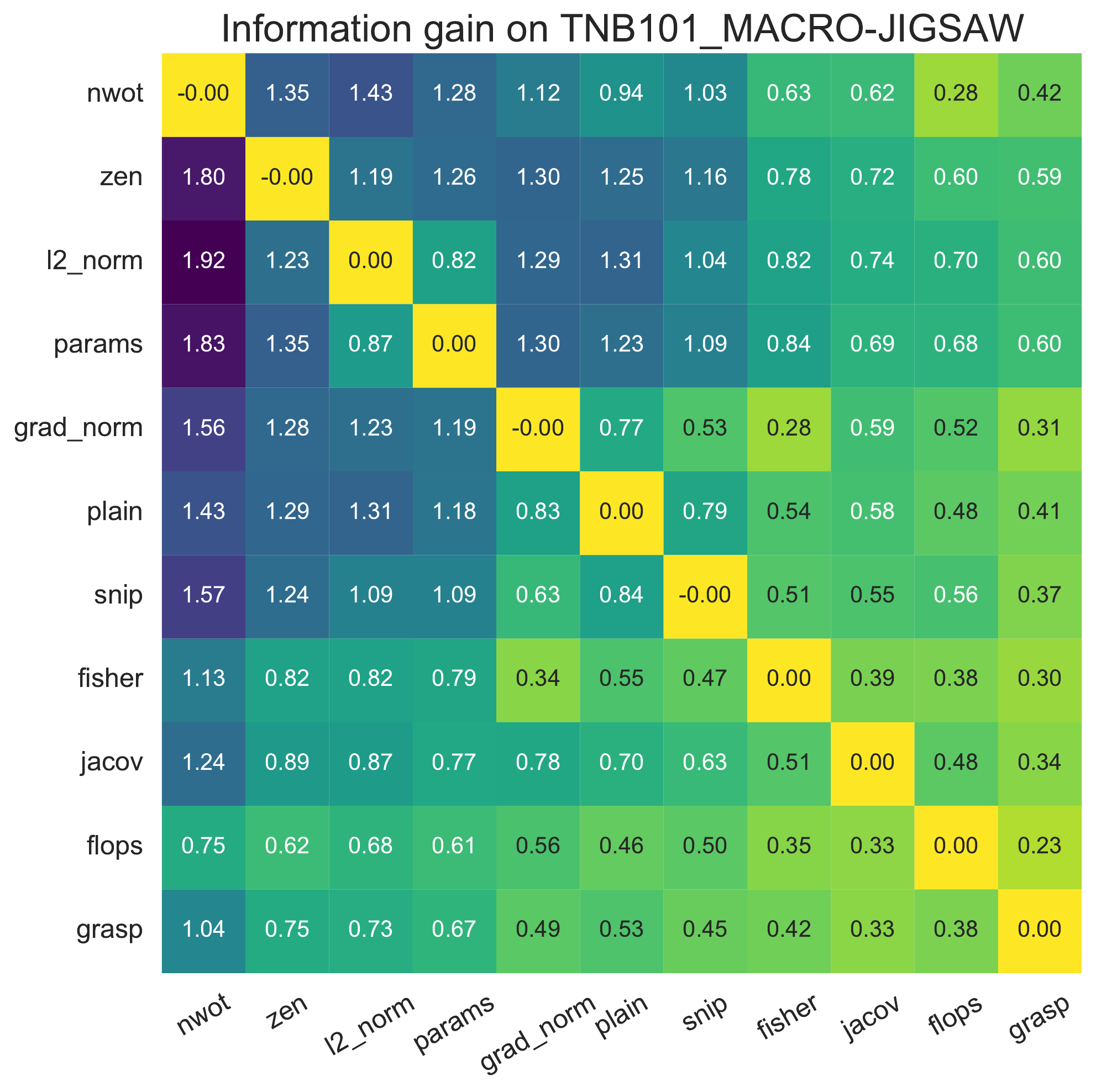}
    \includegraphics[width=.32\linewidth]{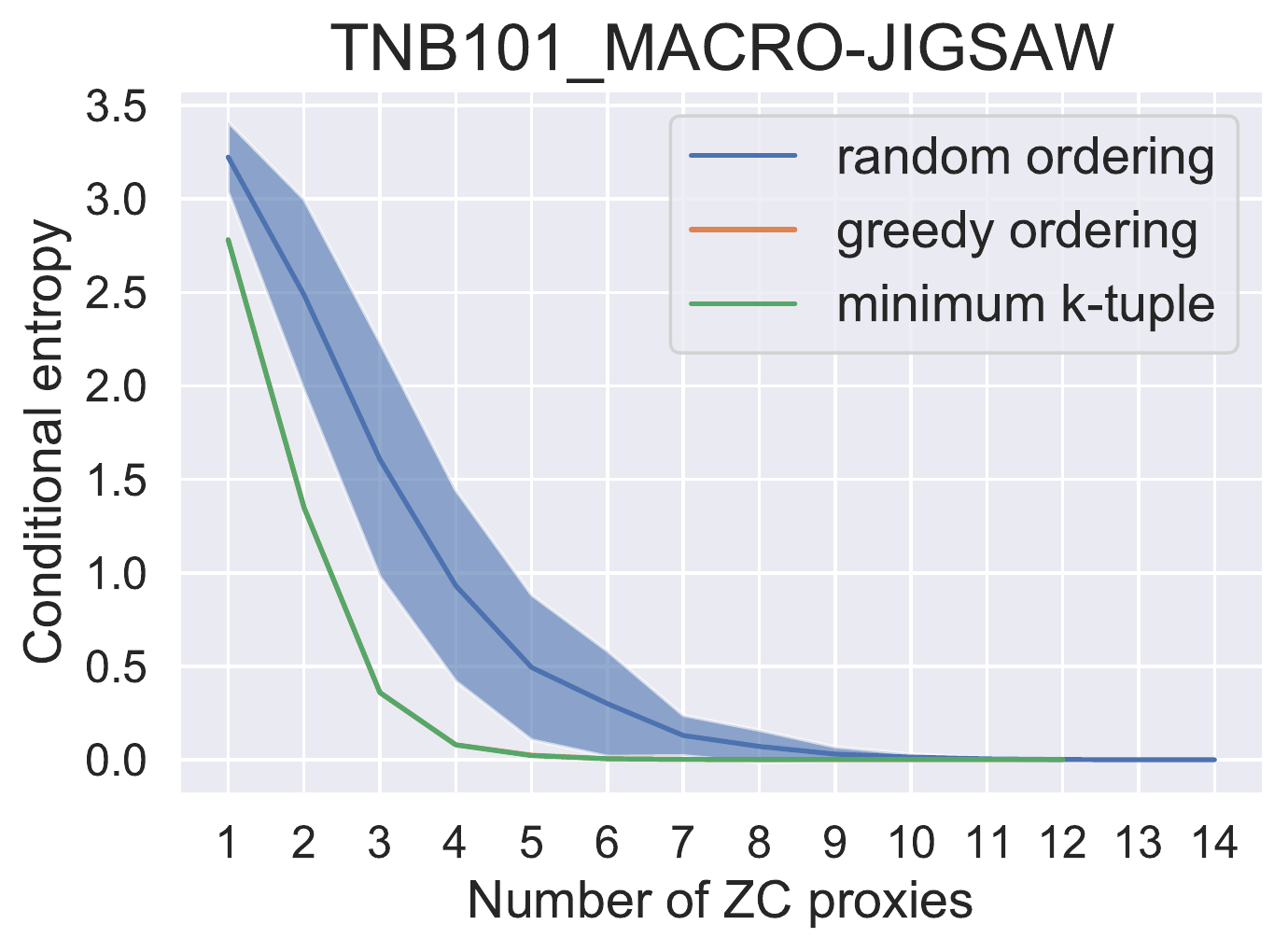}\\
    \includegraphics[width=.32\linewidth]{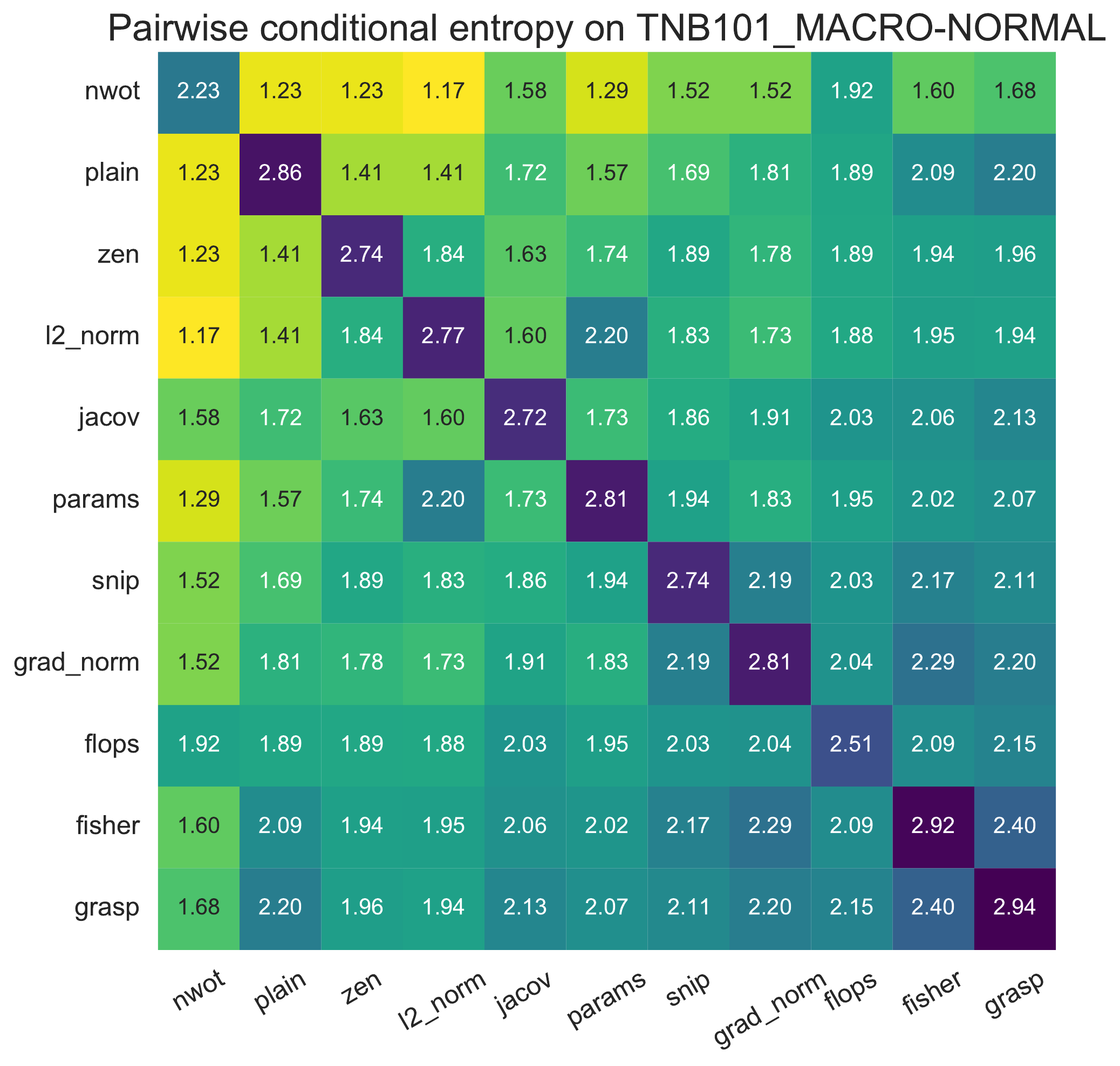}    
    \includegraphics[width=.32\linewidth]{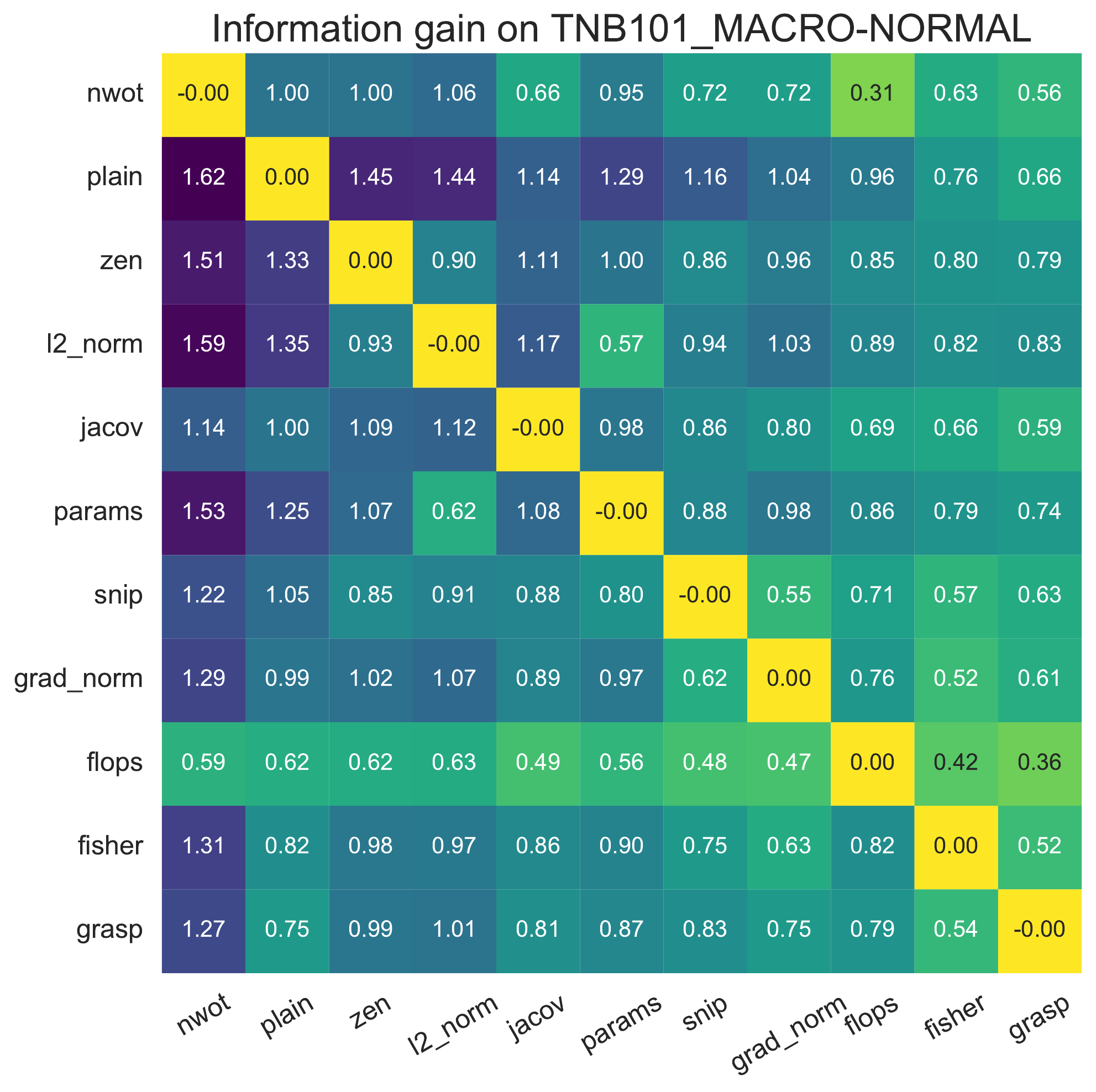}
    \includegraphics[width=.32\linewidth]{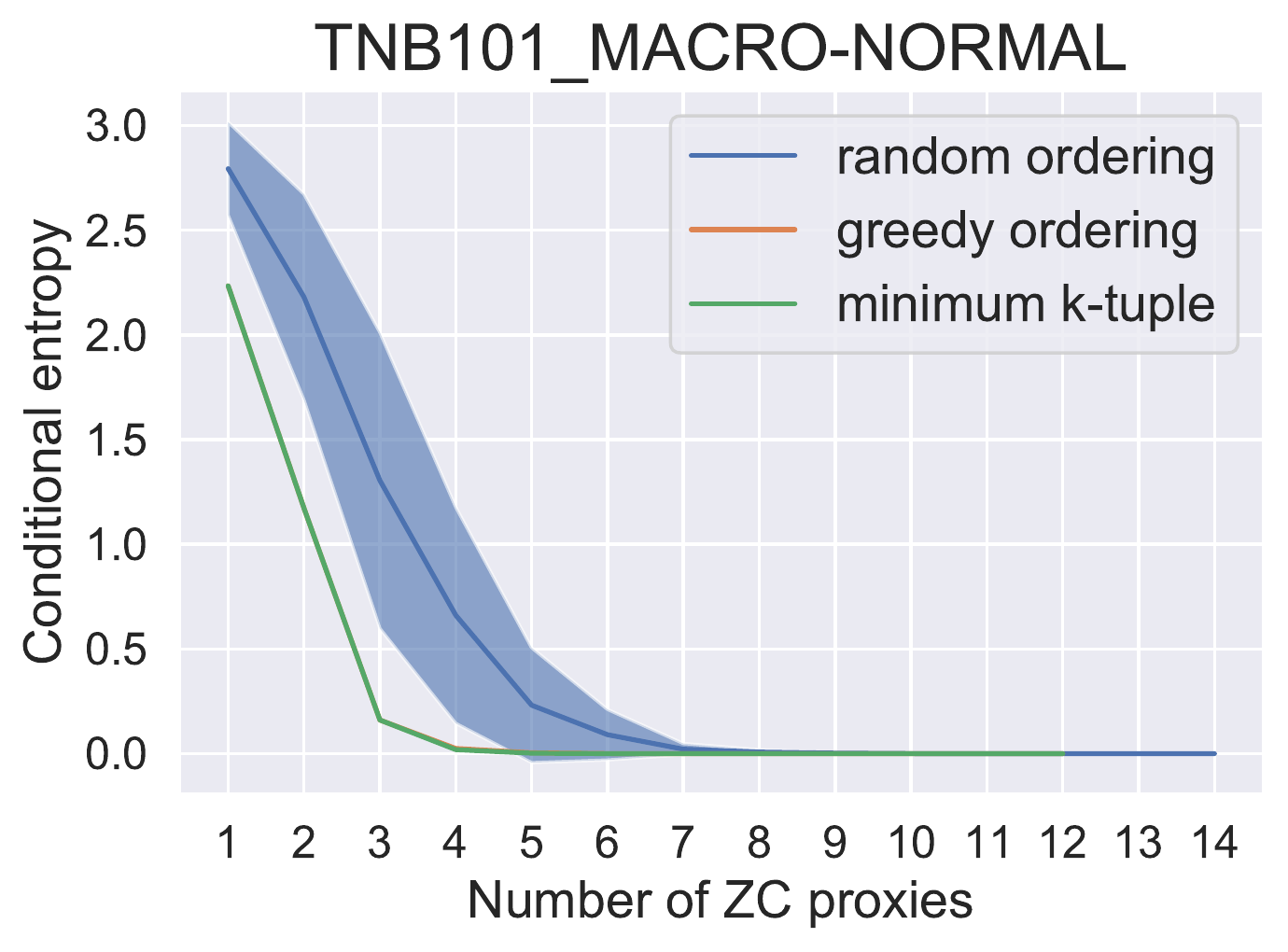}\\    
    \includegraphics[width=.32\linewidth]{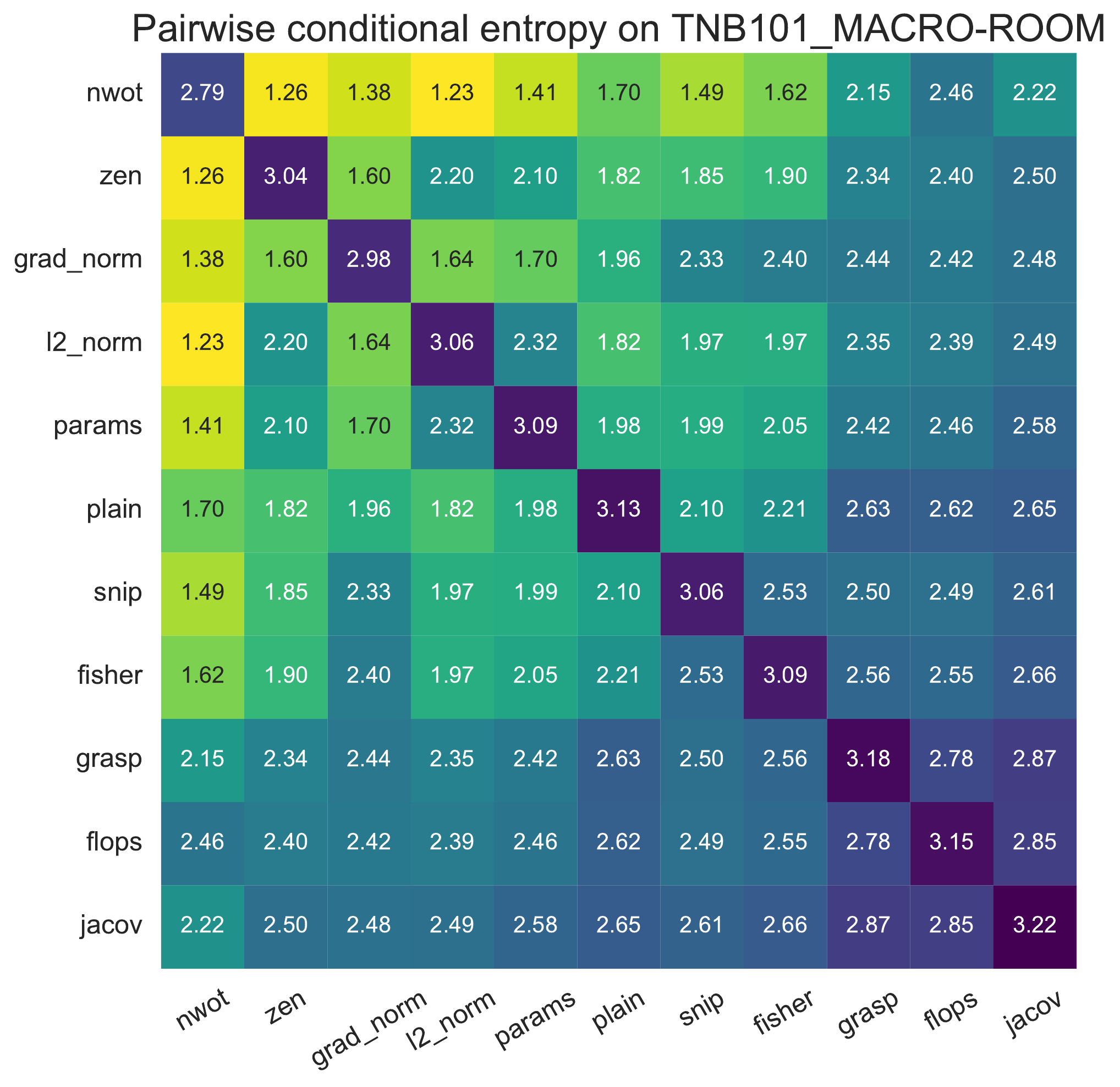}    
    \includegraphics[width=.32\linewidth]{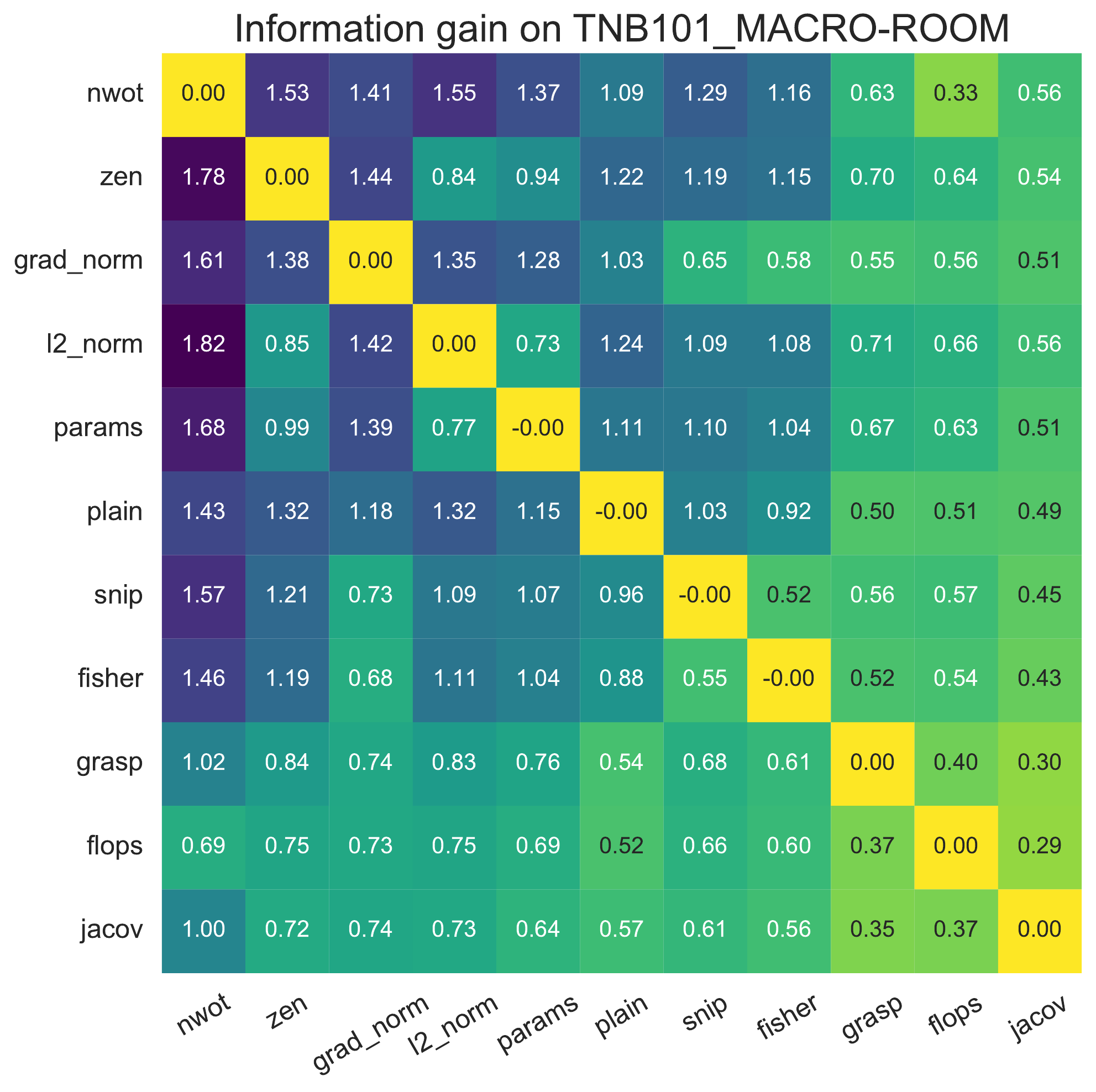}
    \includegraphics[width=.32\linewidth]{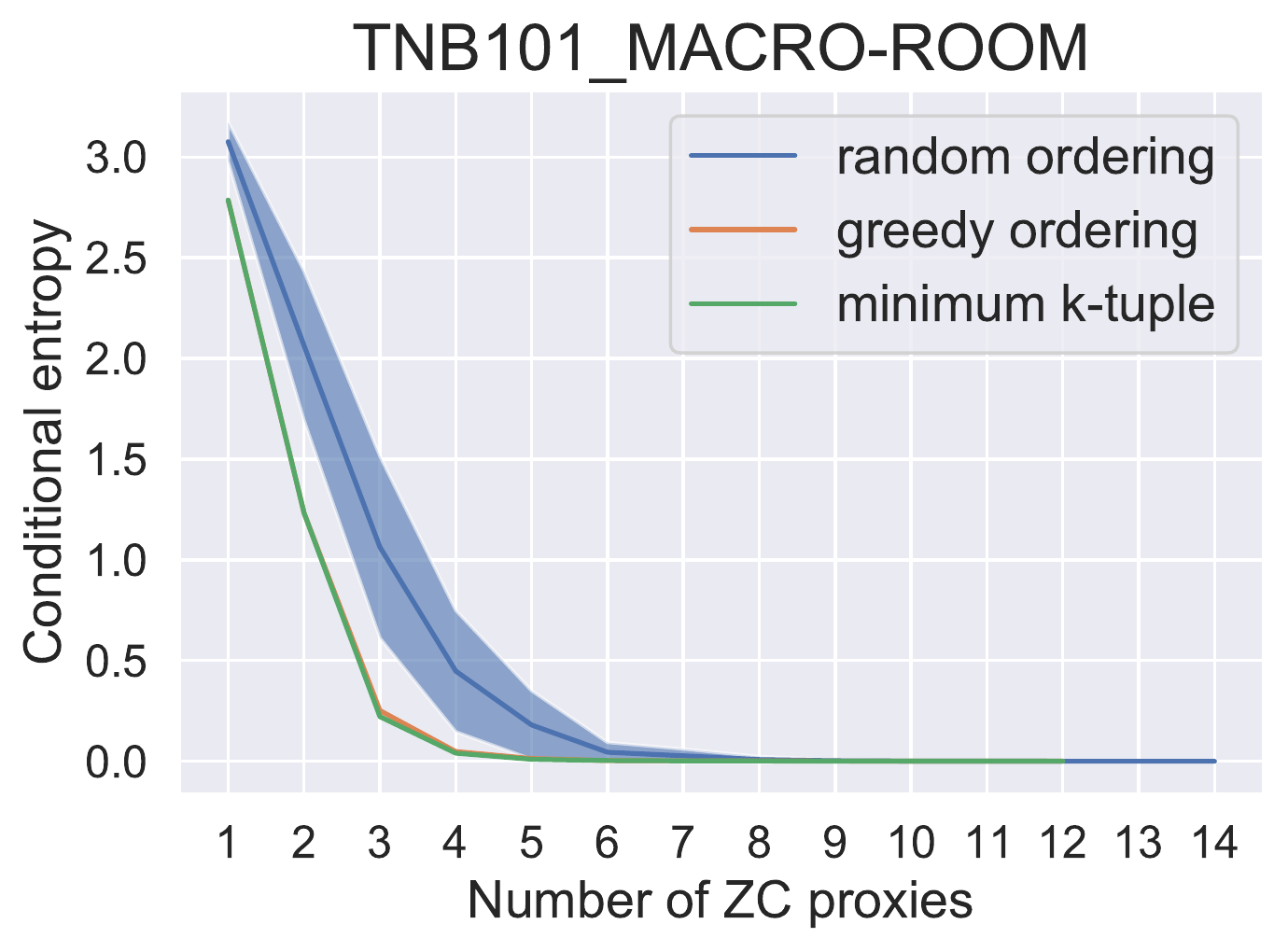}\\  
    \includegraphics[width=.32\linewidth]{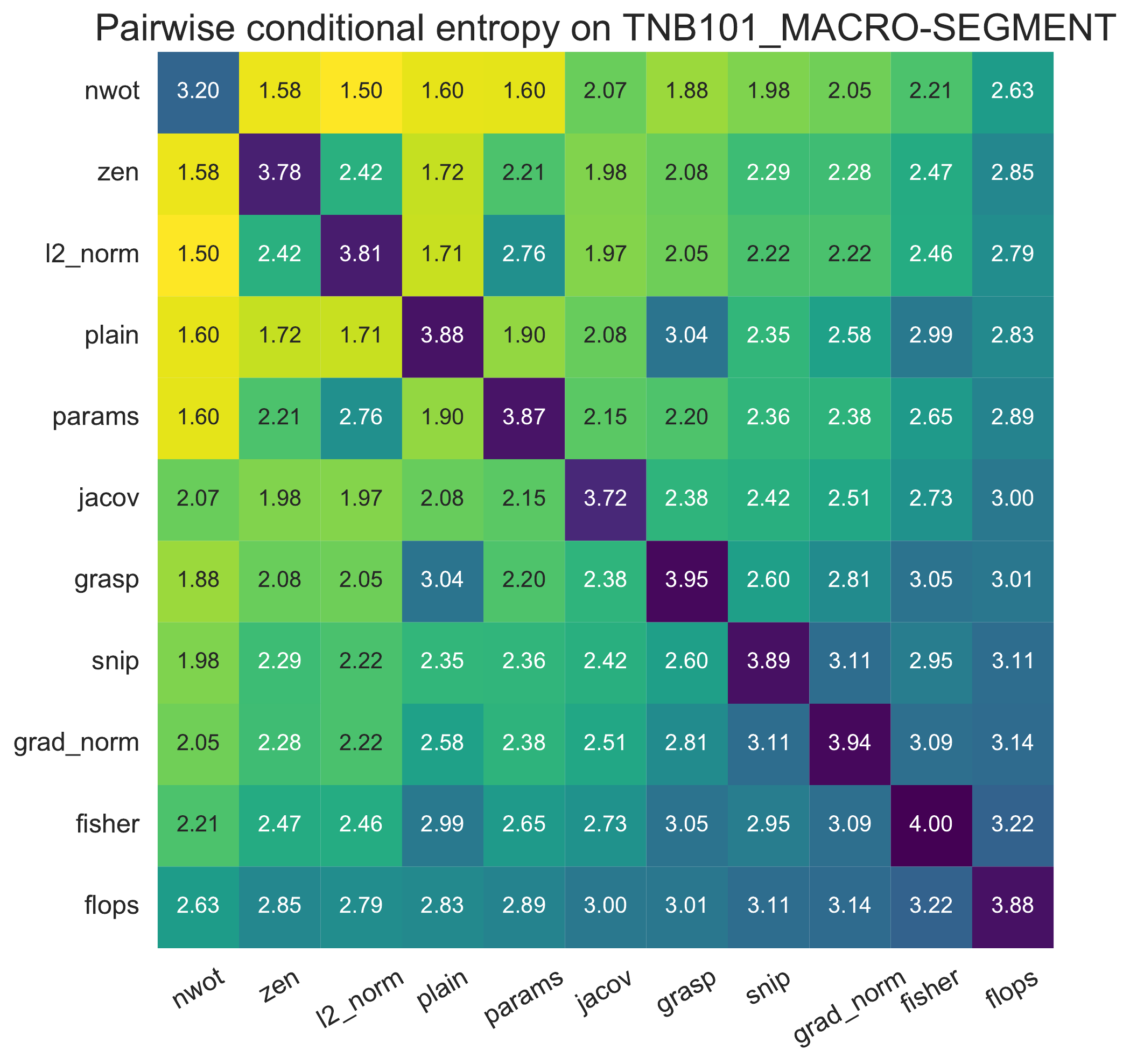}    
    \includegraphics[width=.32\linewidth]{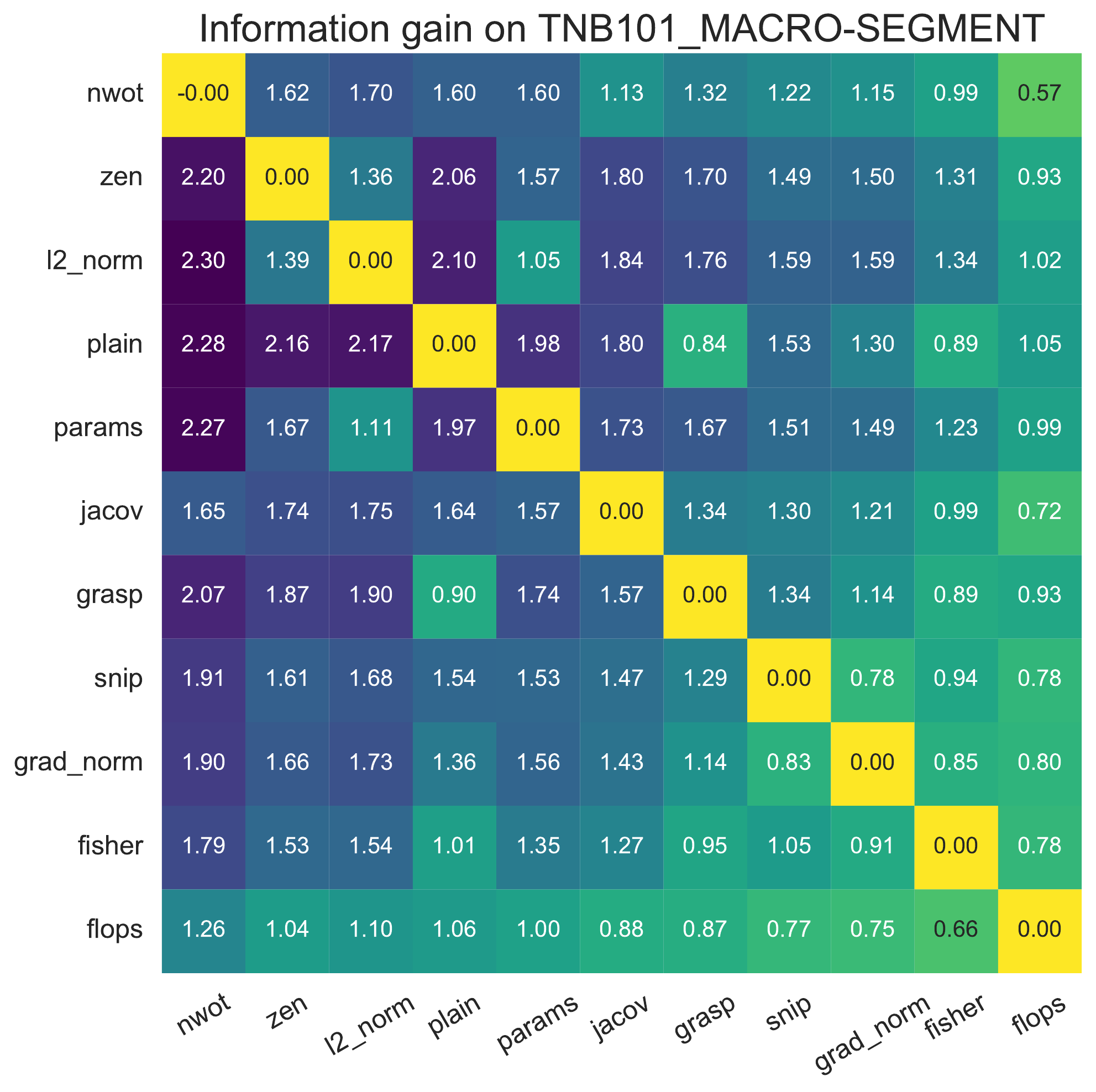}
    \includegraphics[width=.32\linewidth]{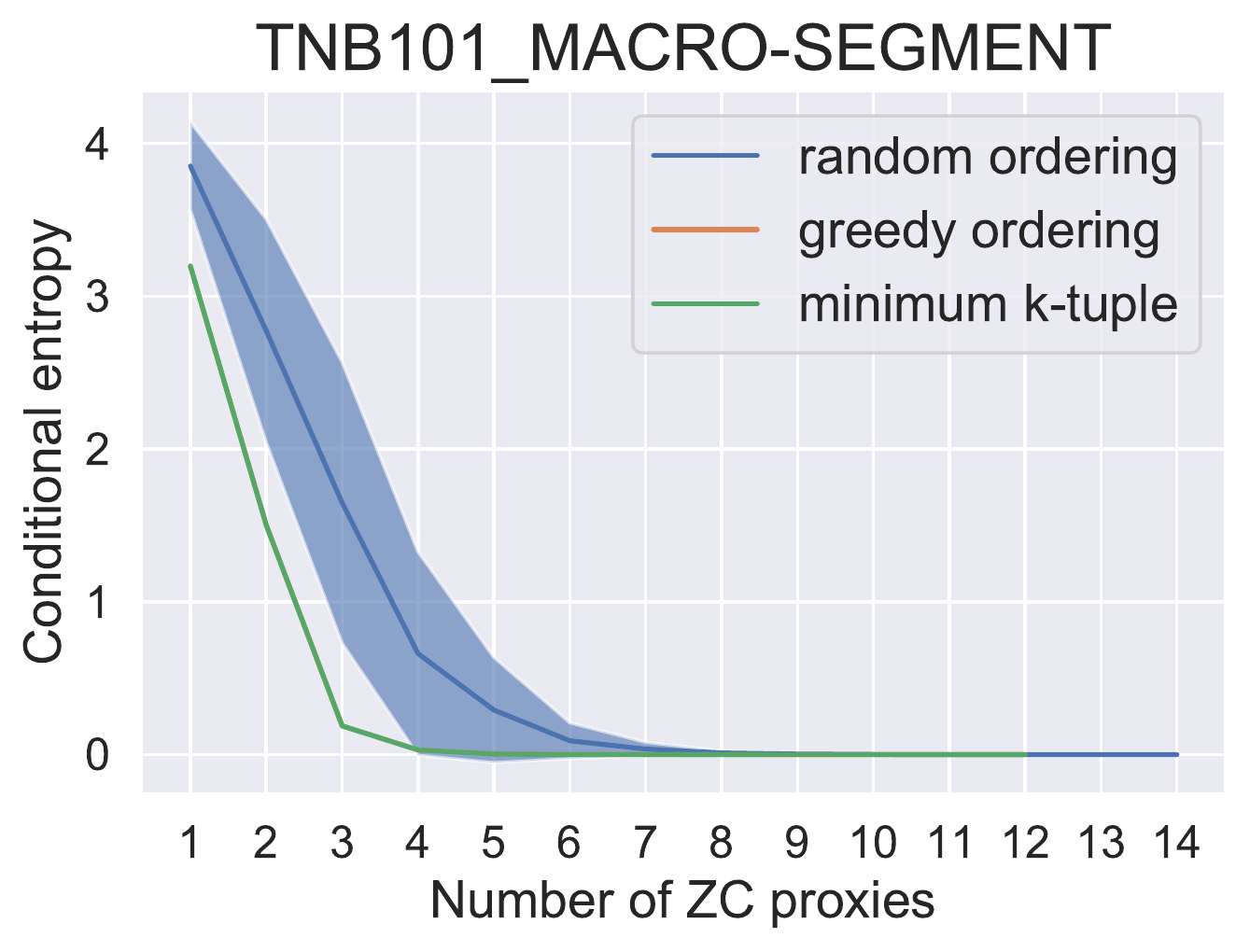}\\ 

    \caption{Conditional entropy and information gain (\textbf{IG}) for each ZC proxy pair across all search spaces and datasets (Left and Middle). Conditional entropy $H(y\mid z_{i_1},\dots,z_{i_k})$ vs.\ $k$, 
    where the ordering $z_{i_1},\dots,z_{i_k}$ is selected using three different strategies (Right). (3/5)}
    \label{fig:info_theory_appendix_3}
\end{figure}

\begin{figure}[ht]
    \centering
    \includegraphics[width=.32\linewidth]{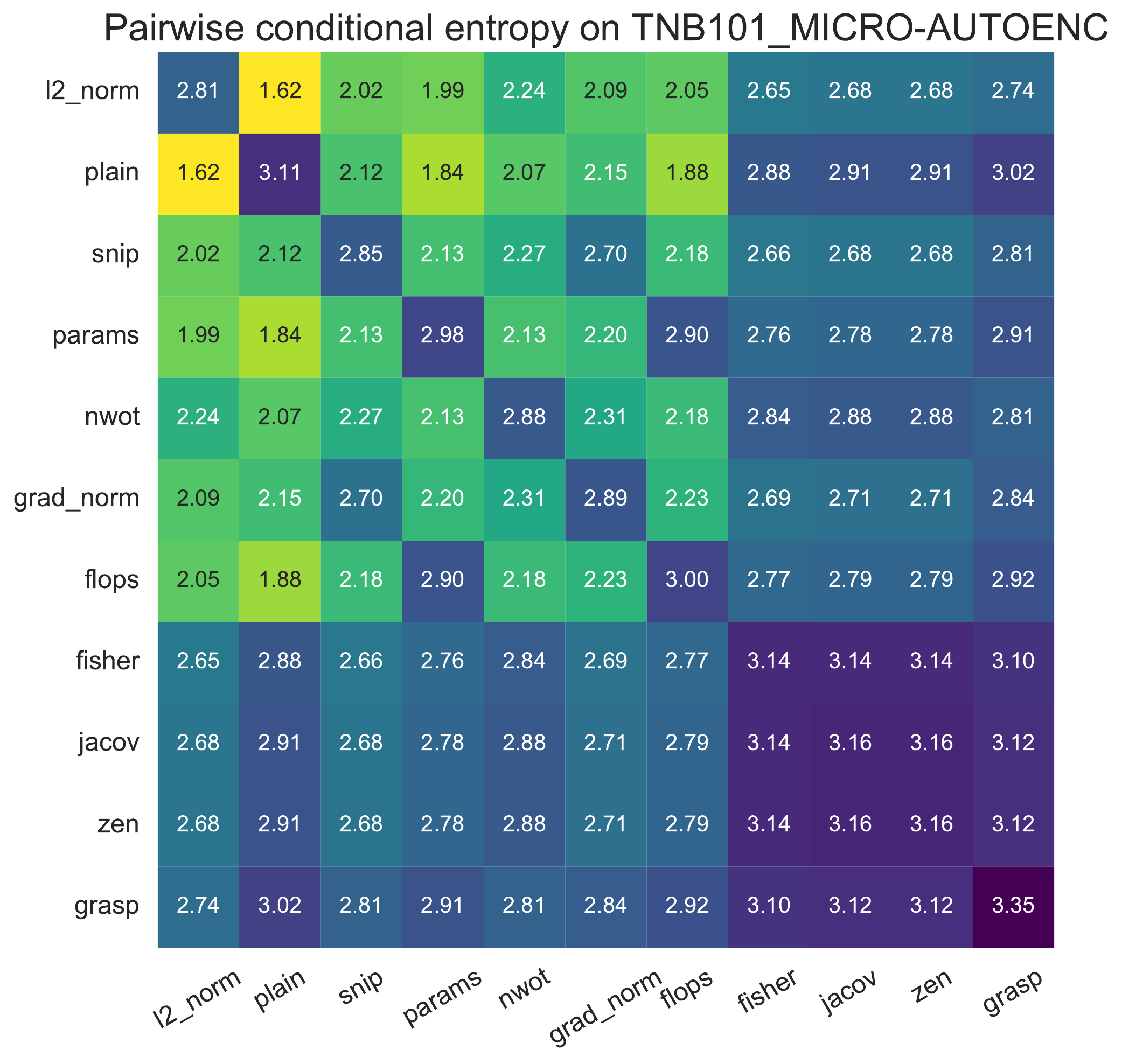}    
    \includegraphics[width=.32\linewidth]{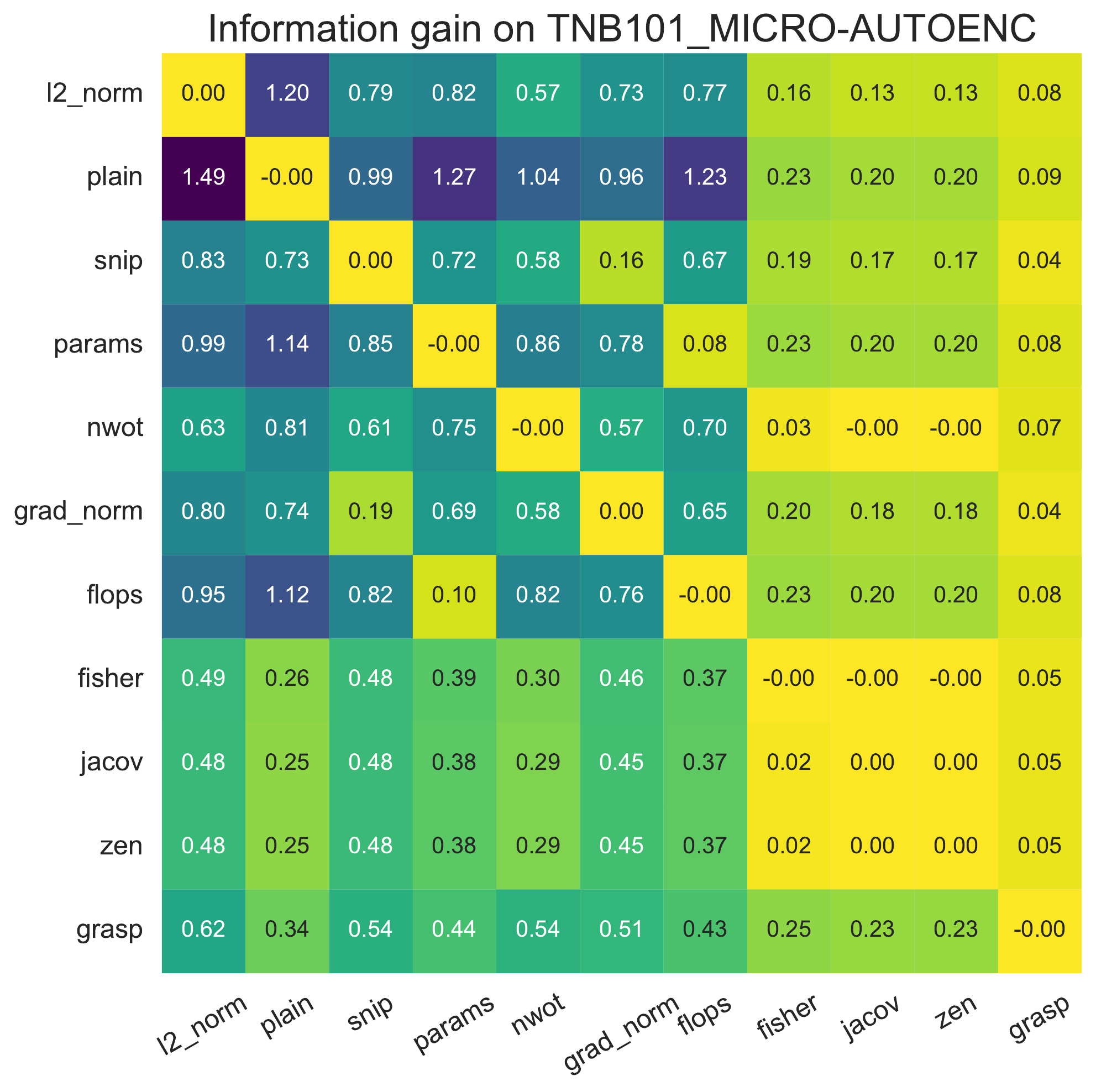}
    \includegraphics[width=.32\linewidth]{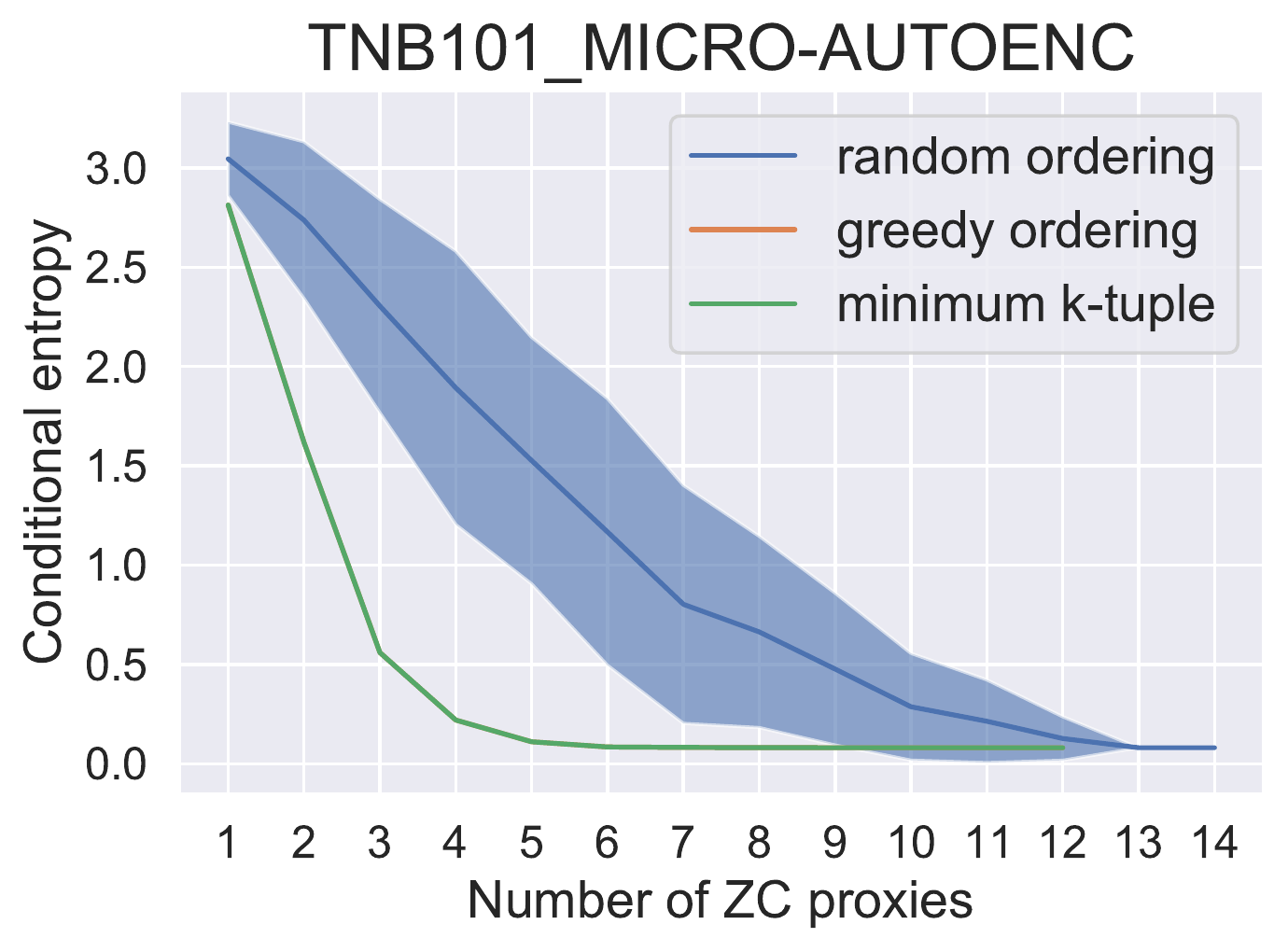}\\    
    \includegraphics[width=.32\linewidth]{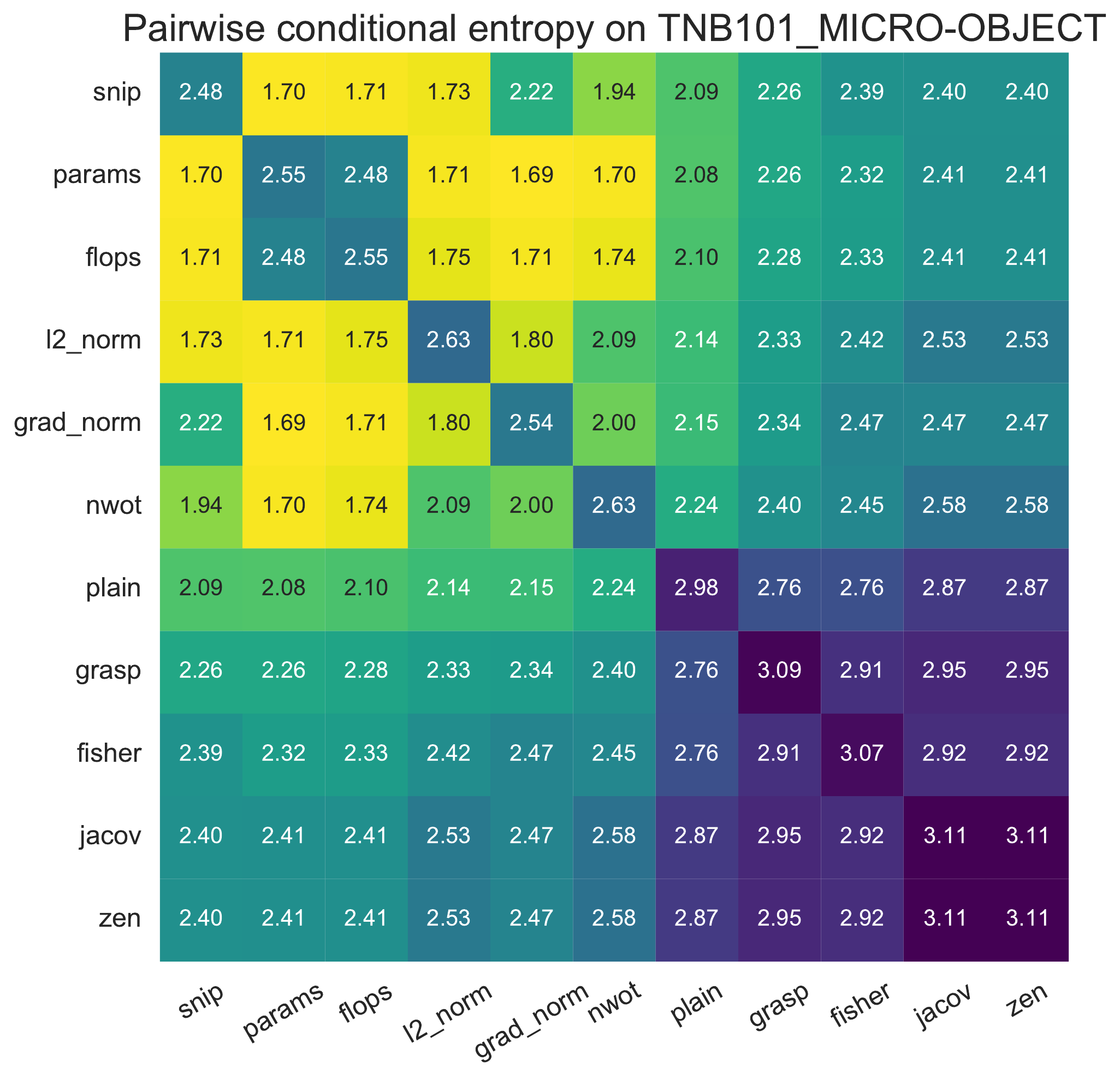}    
    \includegraphics[width=.32\linewidth]{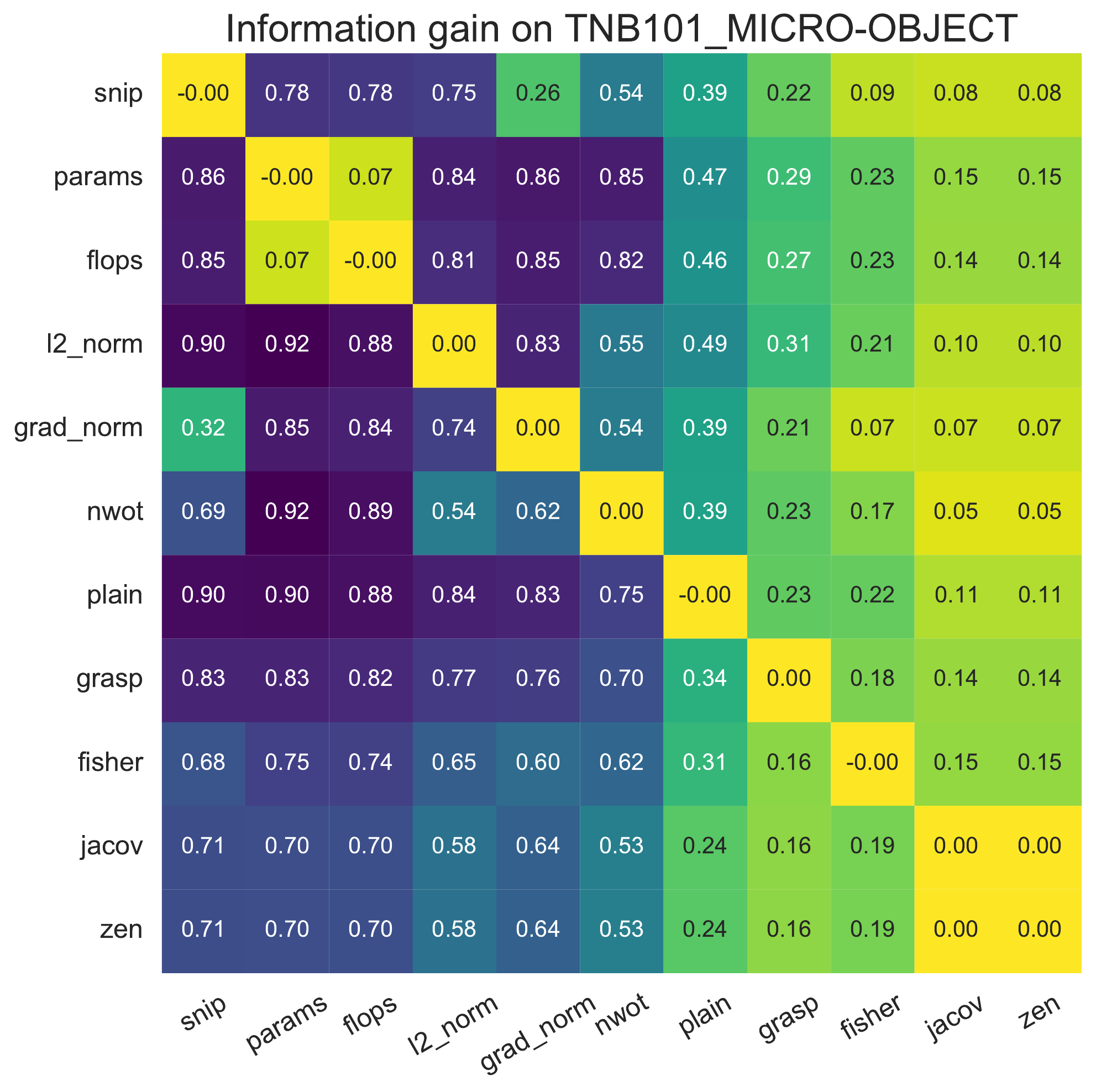}
    \includegraphics[width=.32\linewidth]{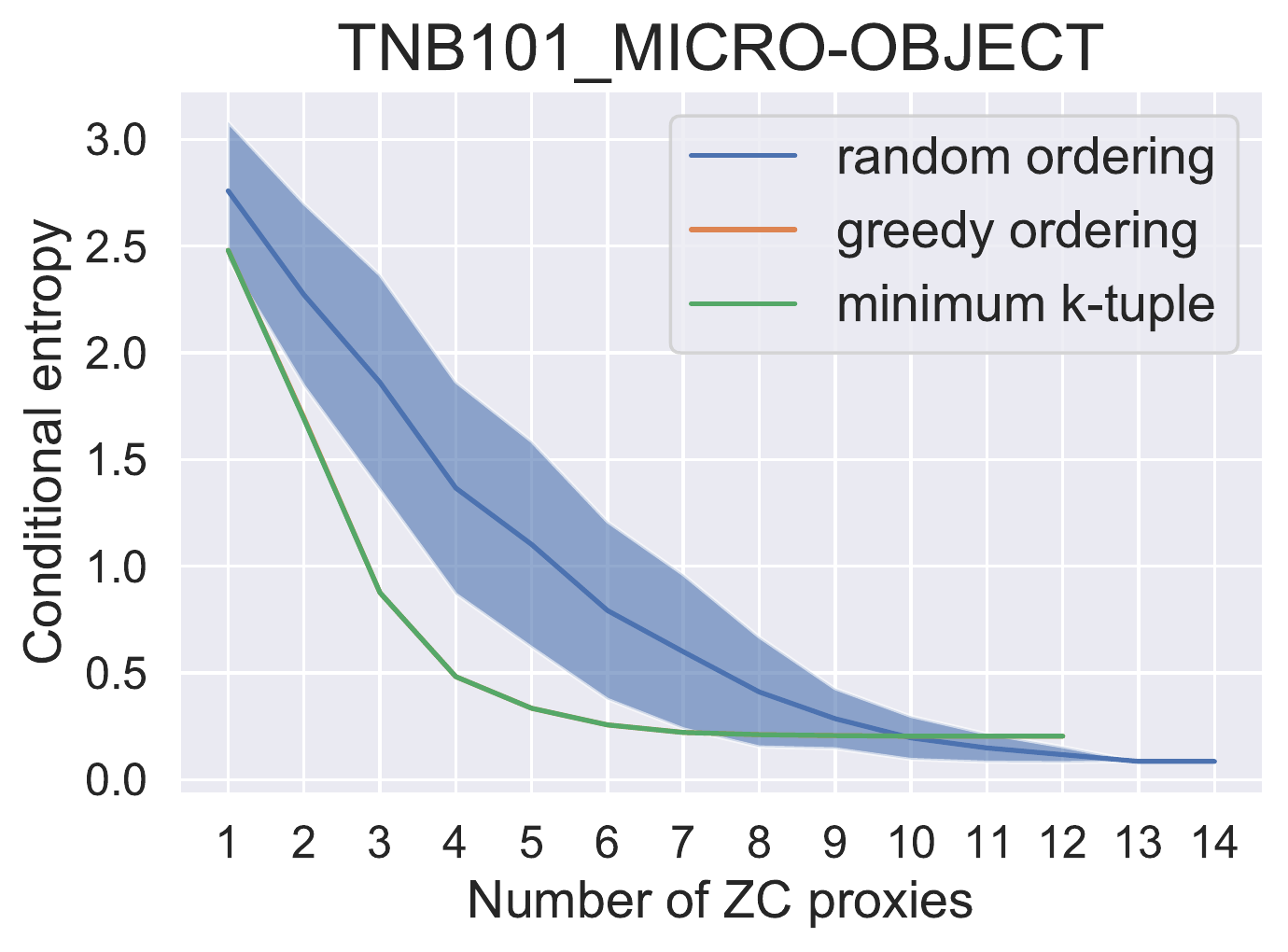}\\    
    \includegraphics[width=.32\linewidth]{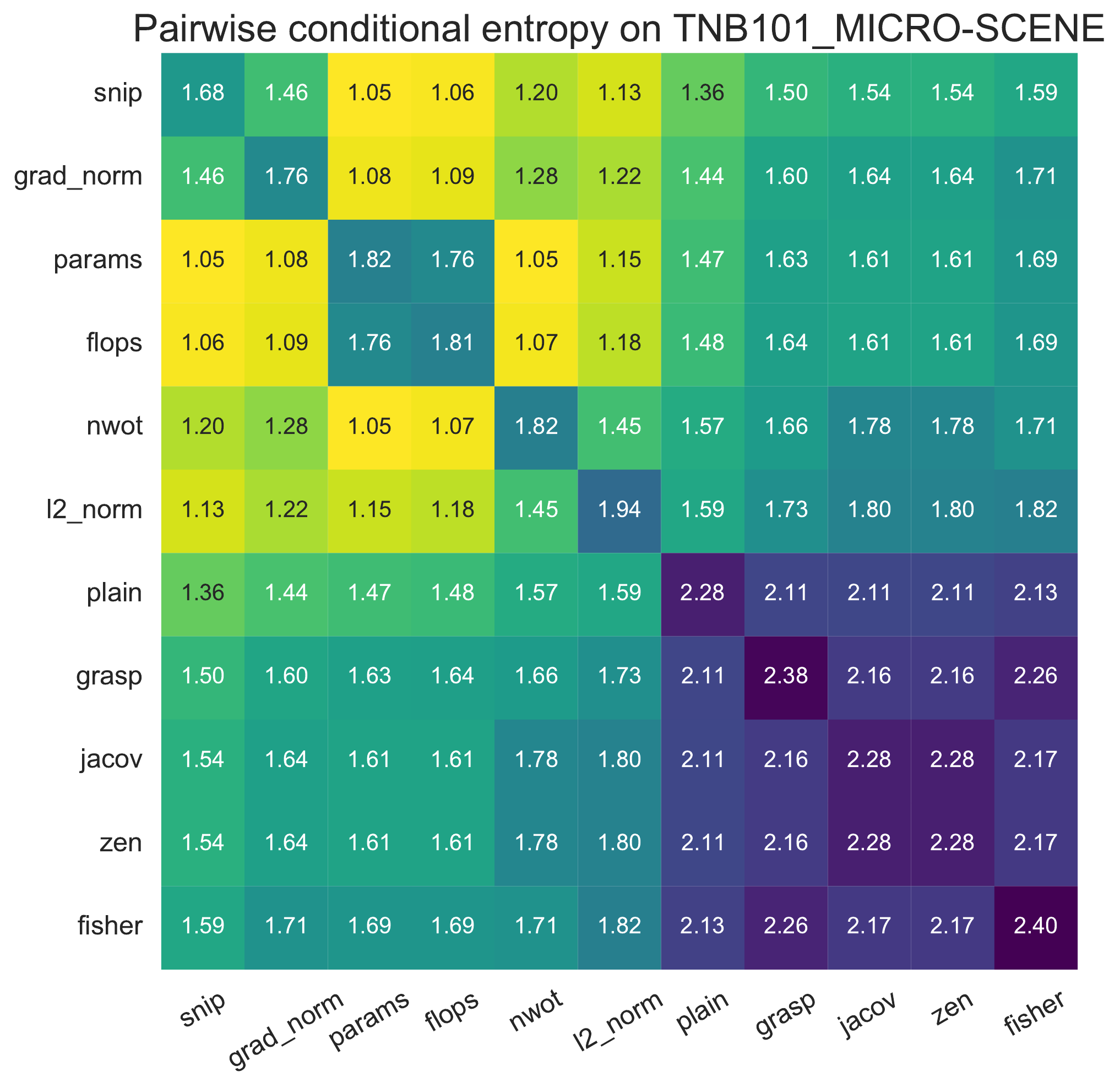}    
    \includegraphics[width=.32\linewidth]{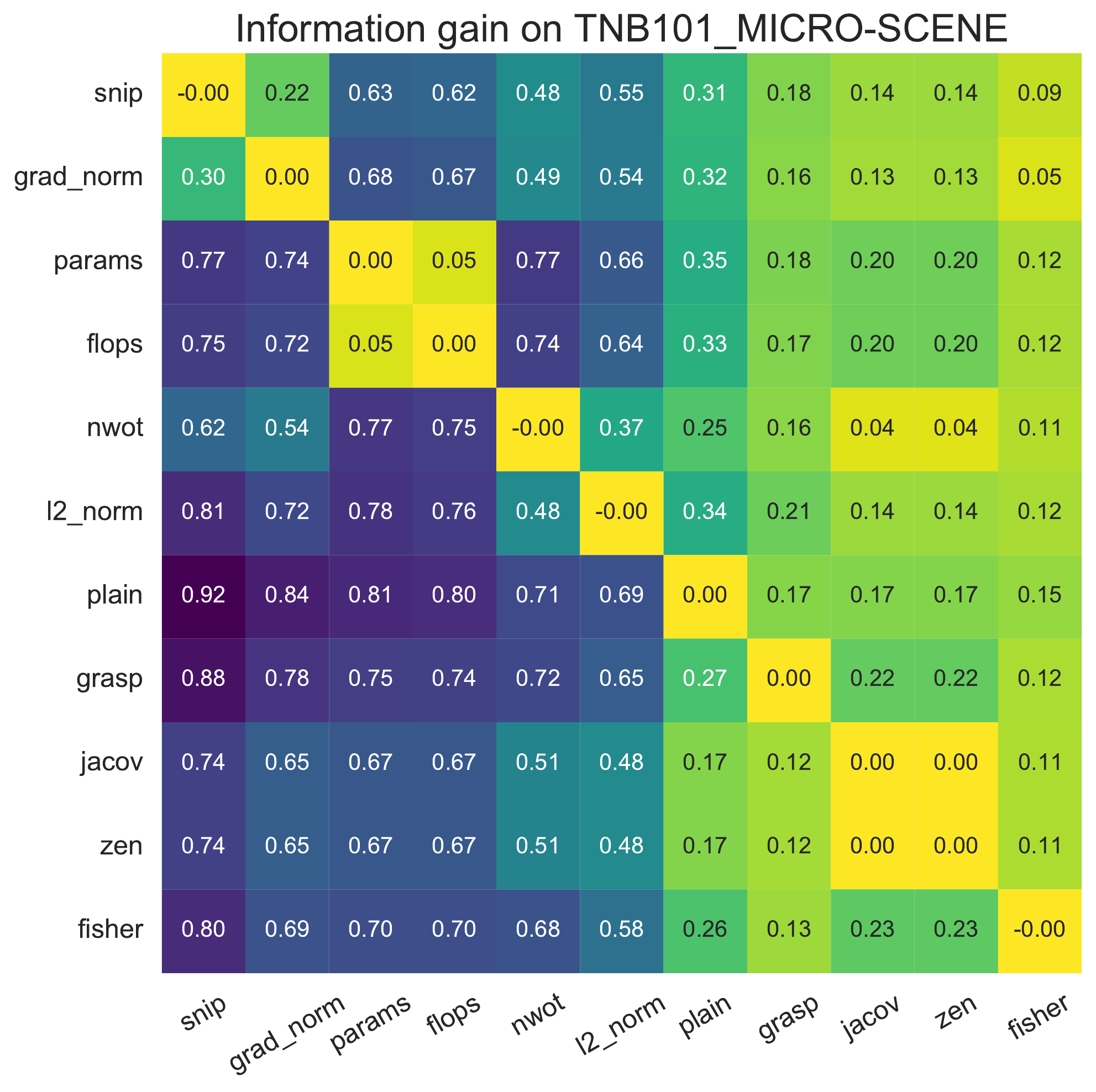}
    \includegraphics[width=.32\linewidth]{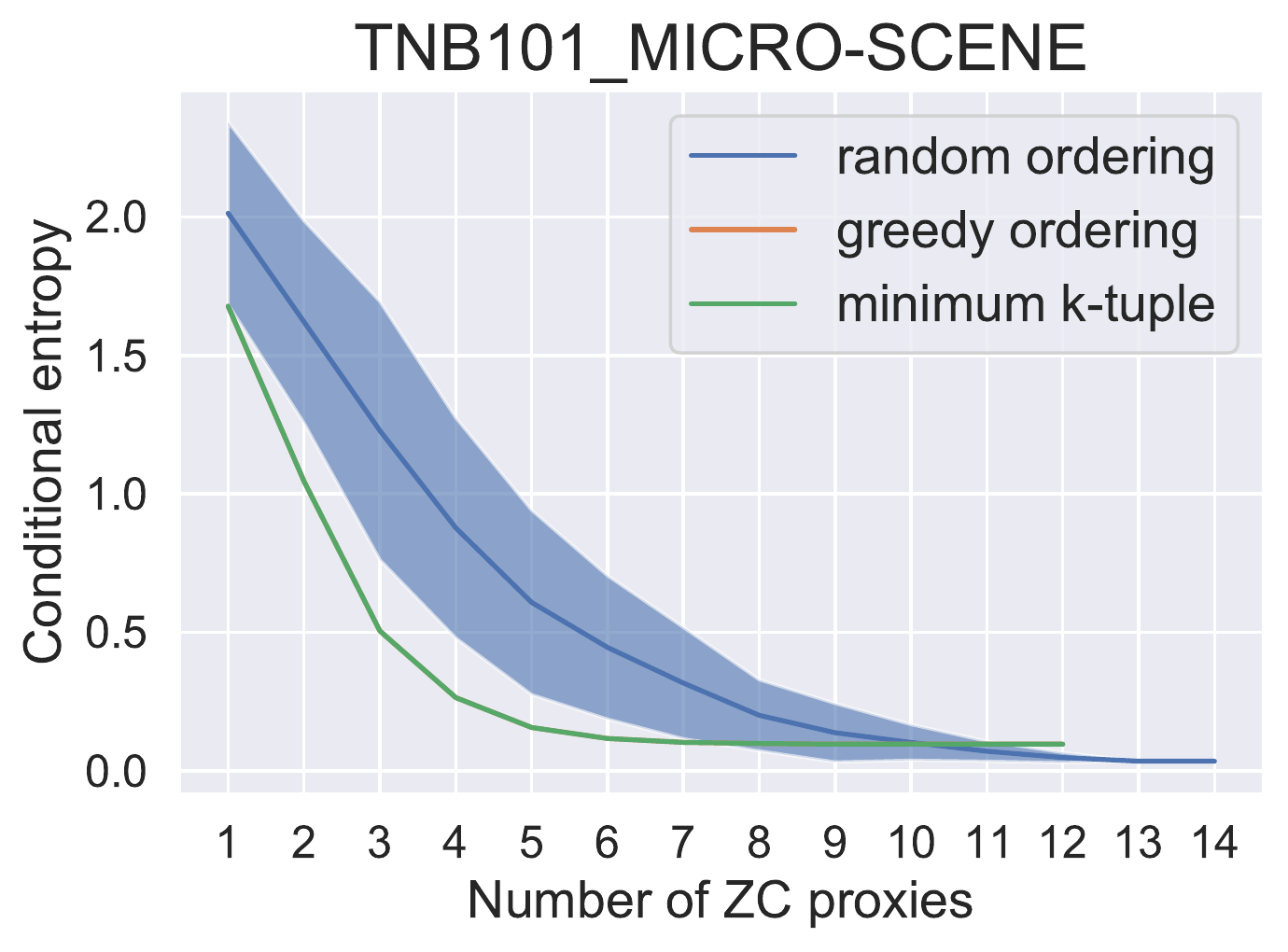}\\
    \includegraphics[width=.32\linewidth]{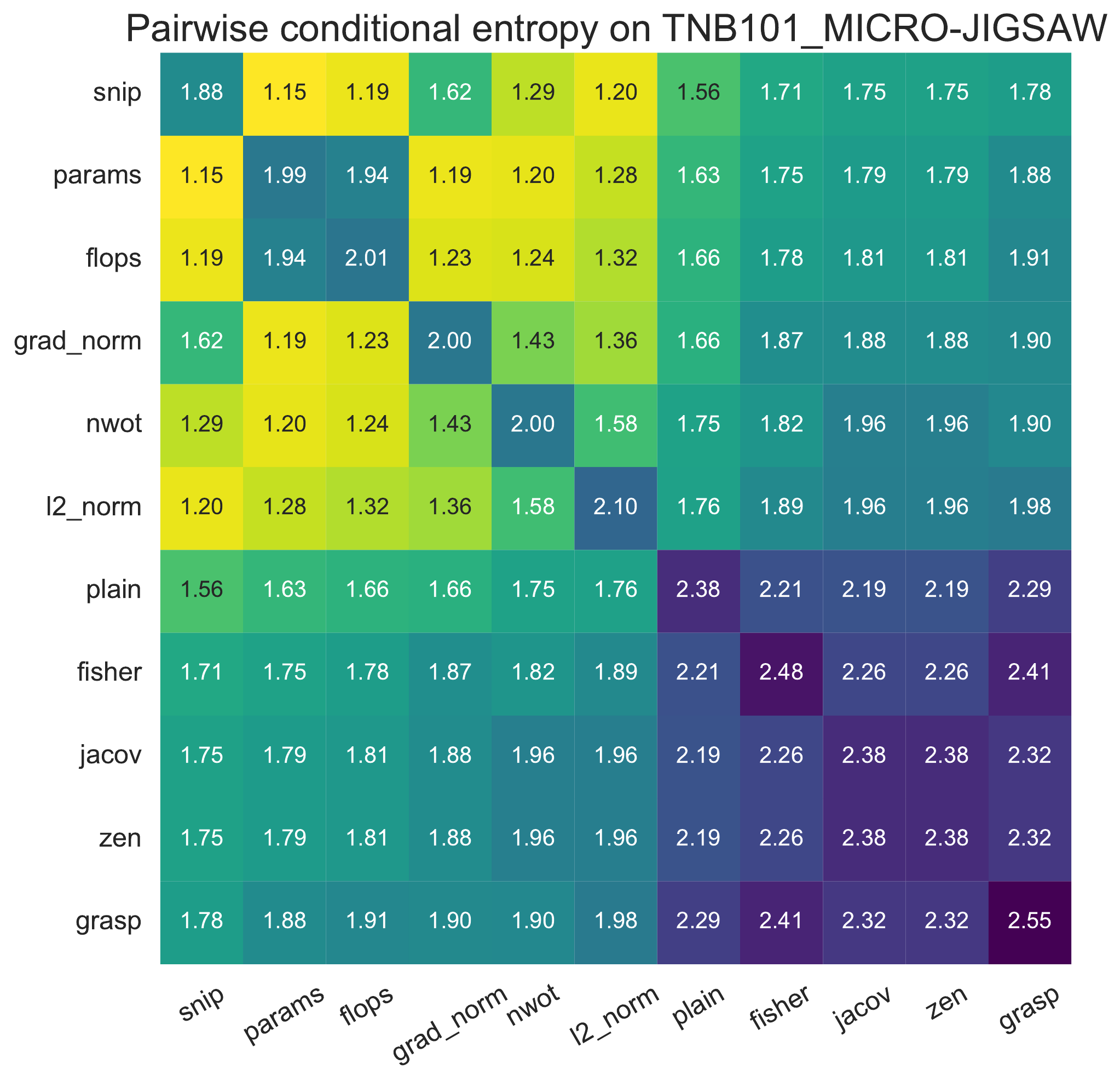}    
    \includegraphics[width=.32\linewidth]{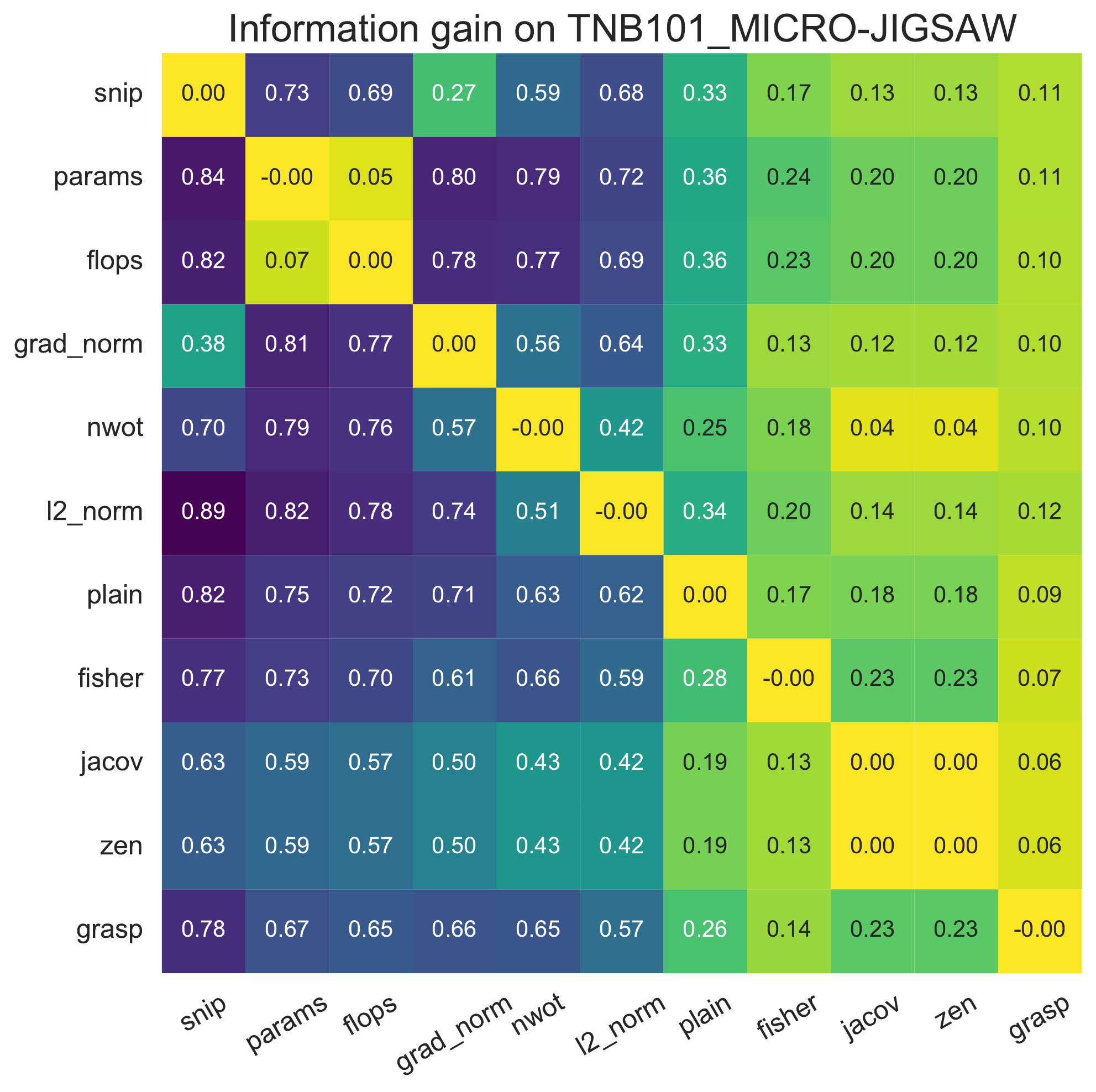}
    \includegraphics[width=.32\linewidth]{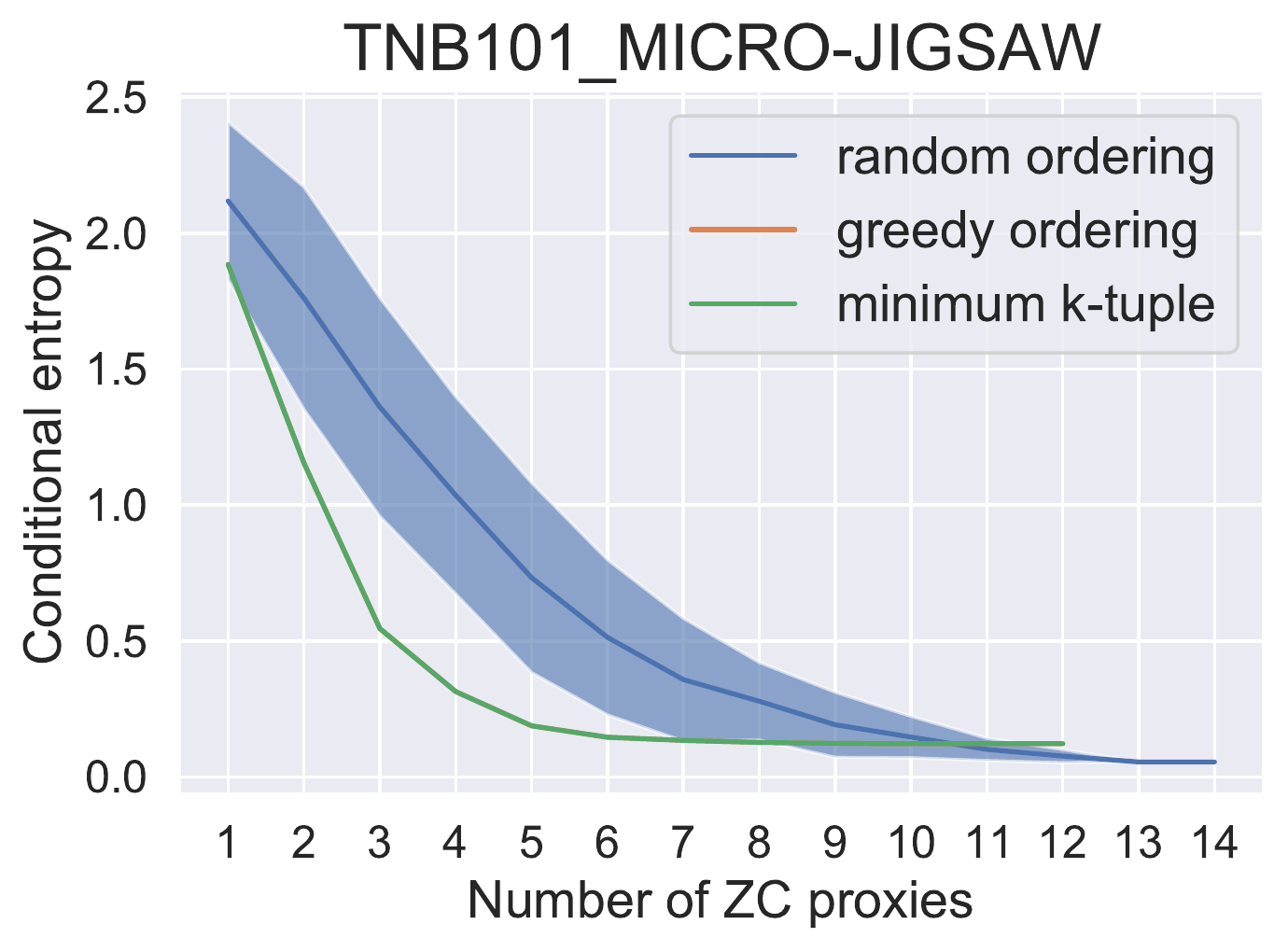}\\ 

    \caption{Conditional entropy and information gain (\textbf{IG}) for each ZC proxy pair across all search spaces and datasets (Left and Middle). Conditional entropy $H(y\mid z_{i_1},\dots,z_{i_k})$ vs.\ $k$, 
    where the ordering $z_{i_1},\dots,z_{i_k}$ is selected using three different strategies (Right). (4/5)}
    \label{fig:info_theory_appendix_4}

\end{figure}

\begin{figure}[ht]
    \centering
    \includegraphics[width=.32\linewidth]{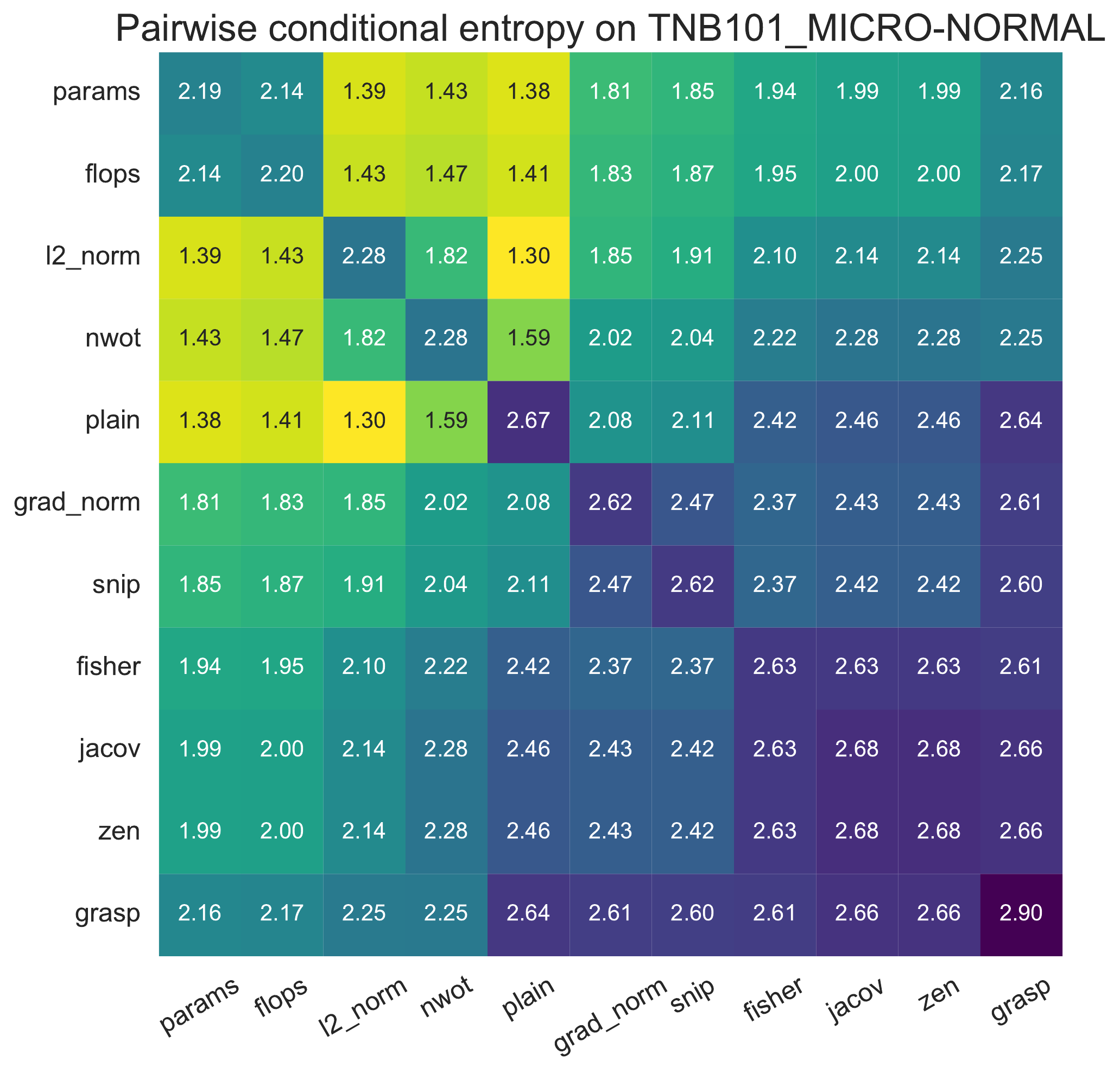}    
    \includegraphics[width=.32\linewidth]{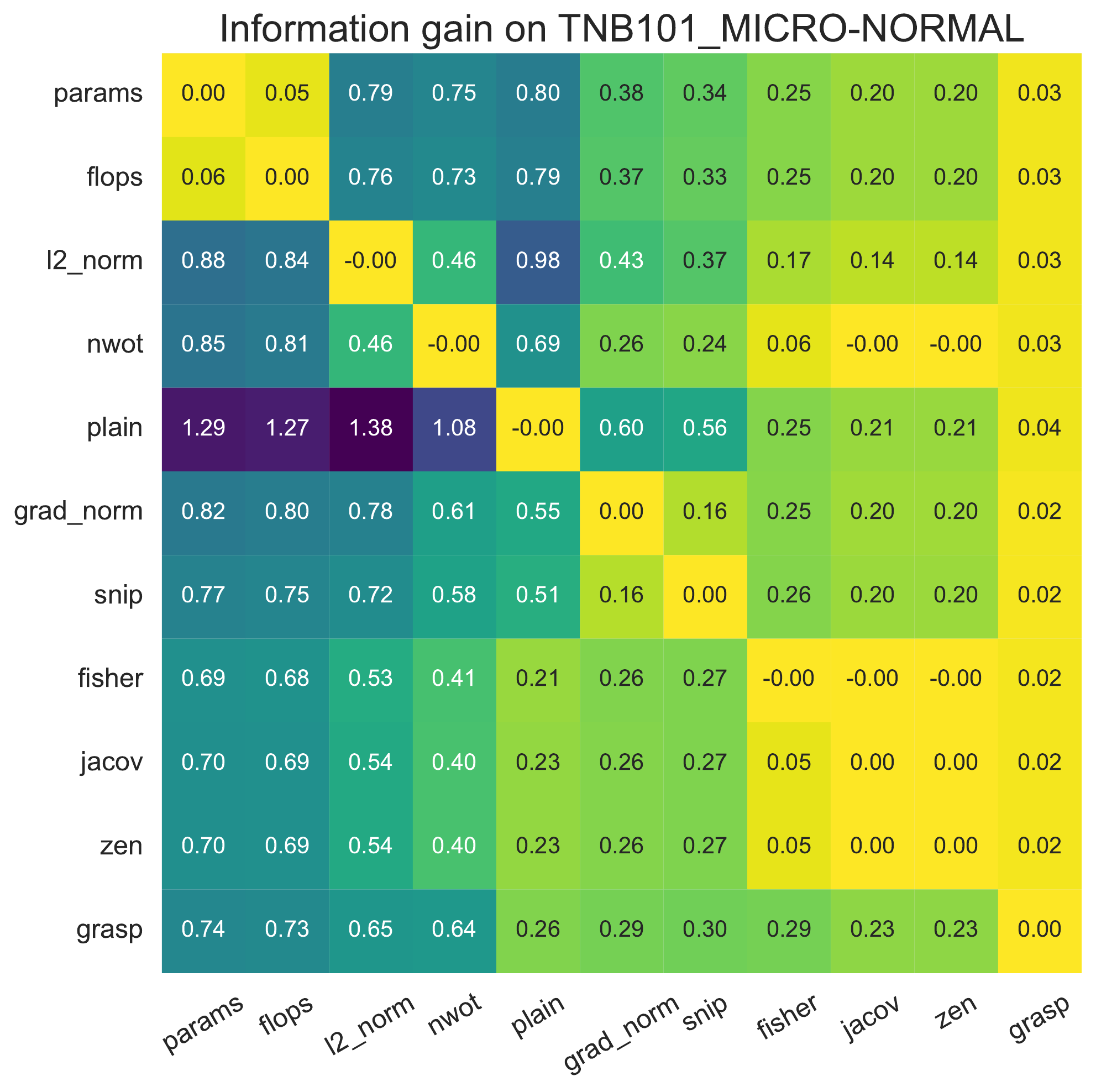}
    \includegraphics[width=.32\linewidth]{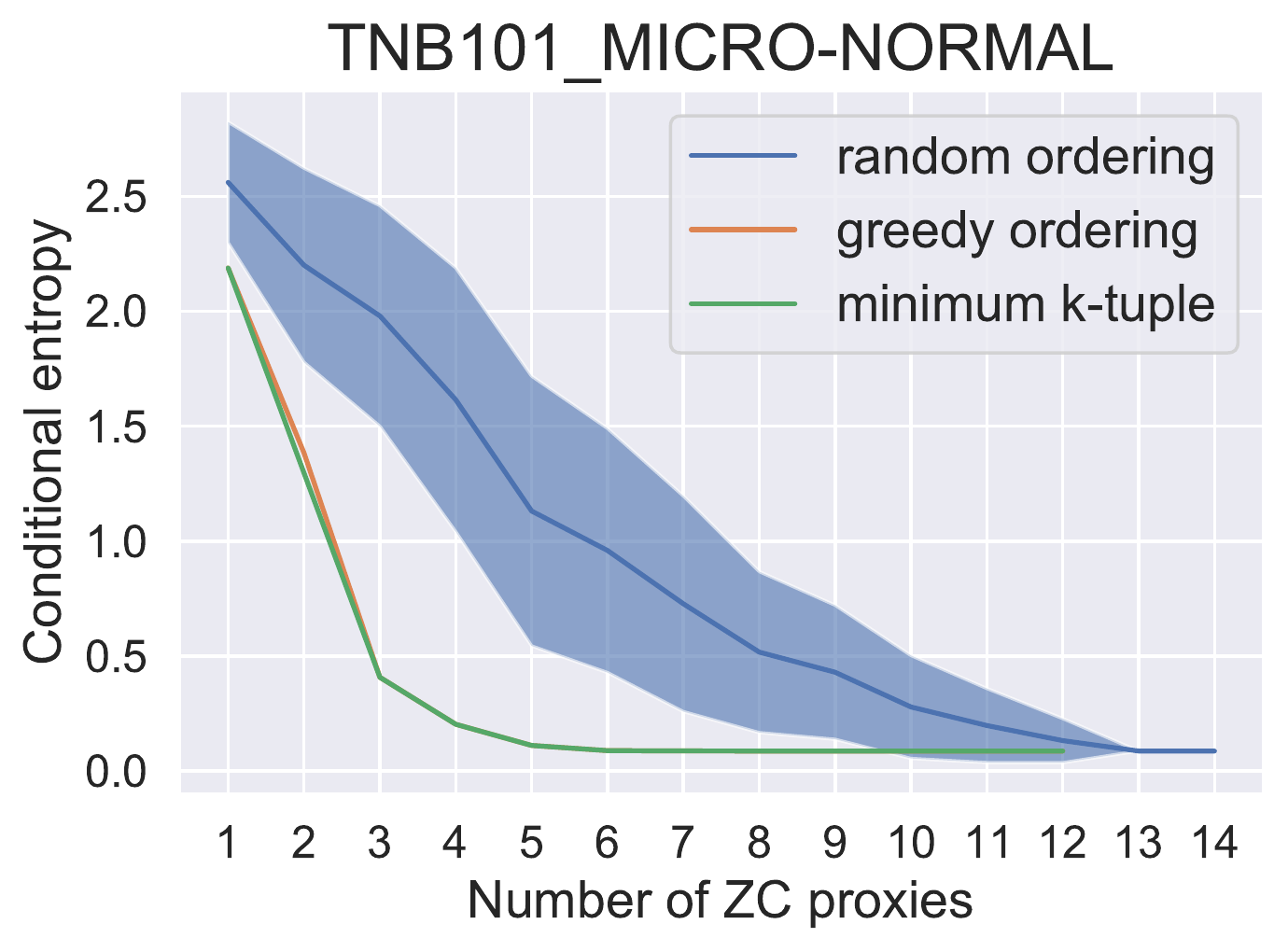}\\    
    \includegraphics[width=.32\linewidth]{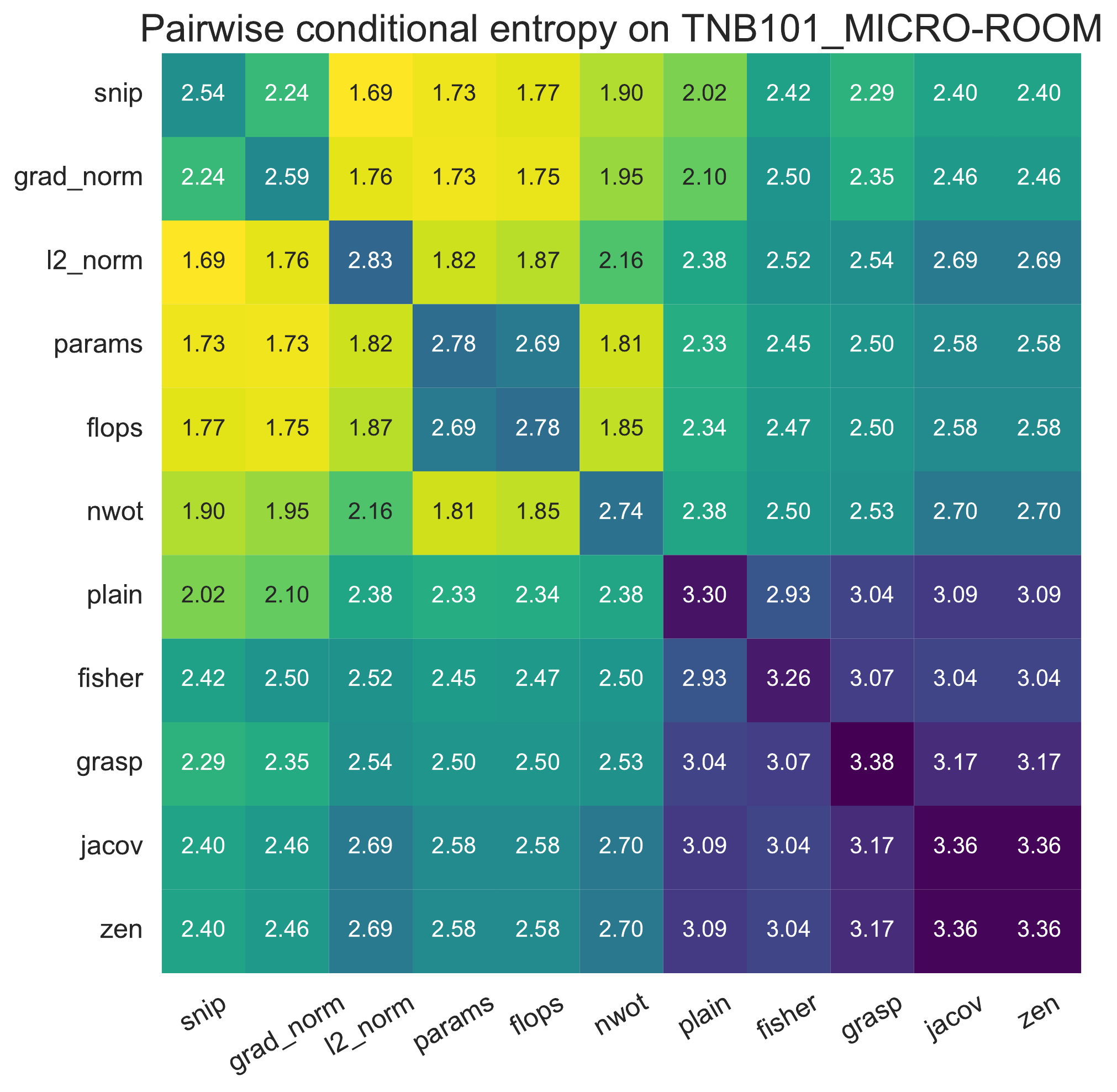}    
    \includegraphics[width=.32\linewidth]{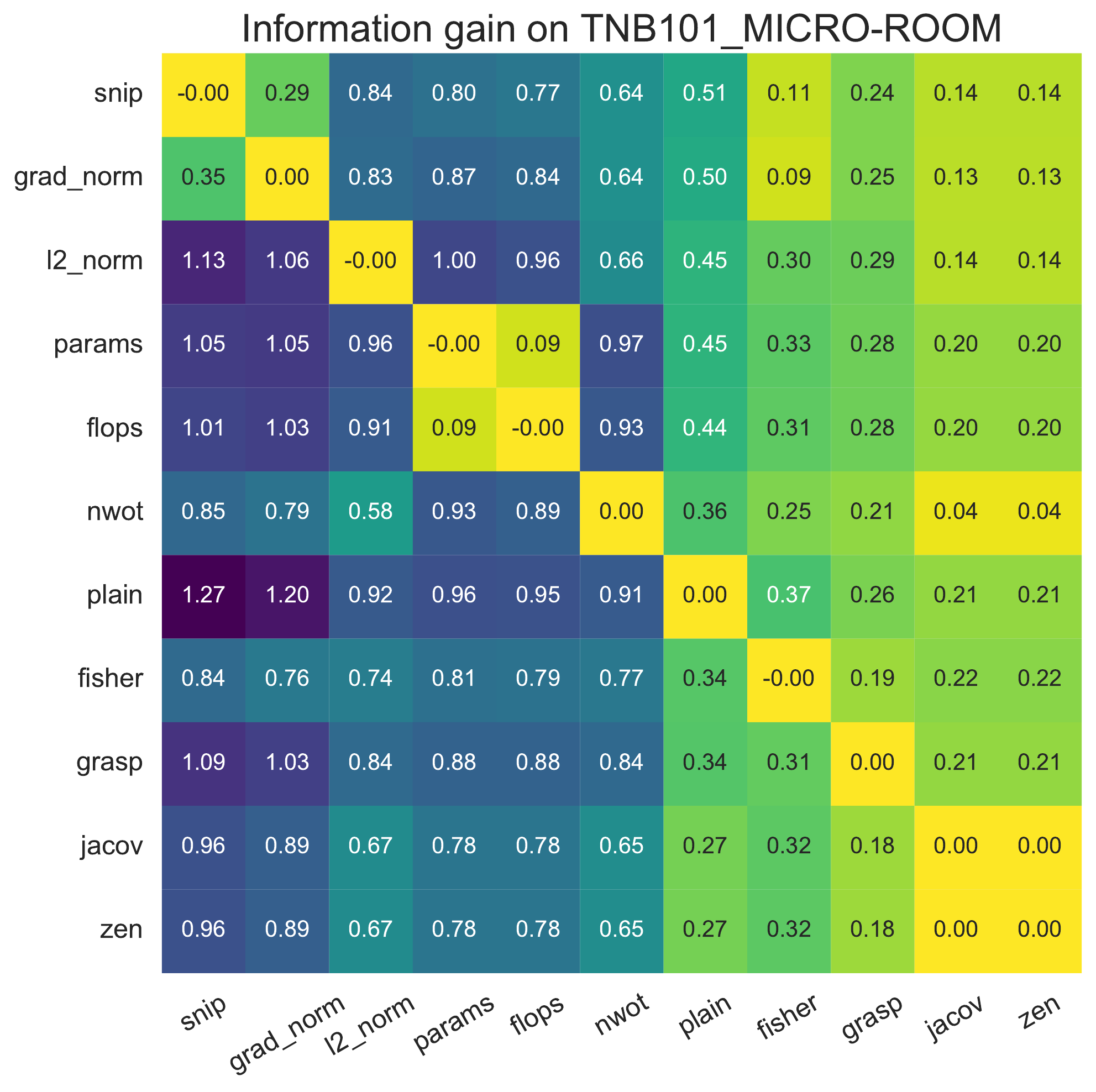}
    \includegraphics[width=.32\linewidth]{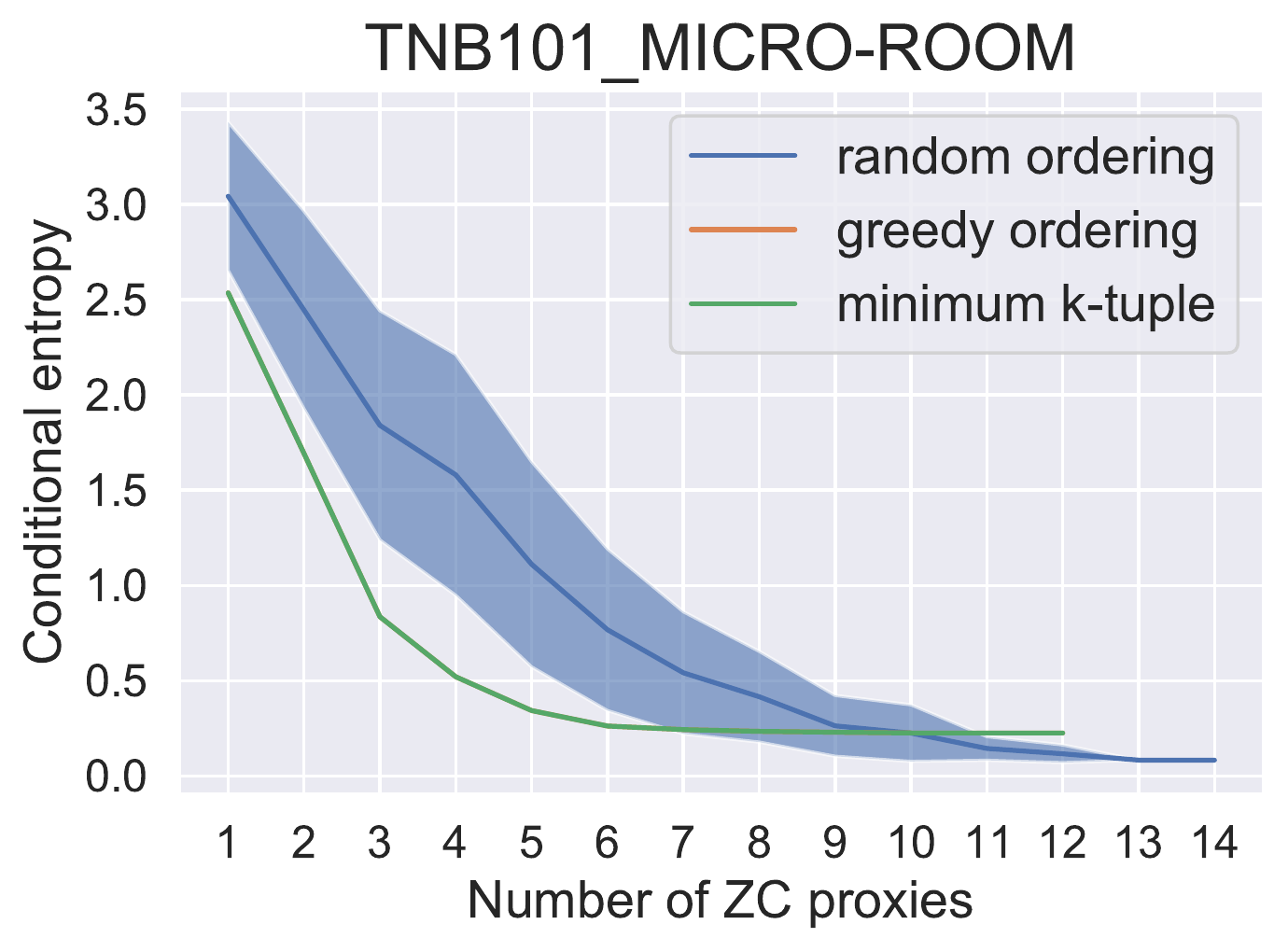}\\  
    \includegraphics[width=.32\linewidth]{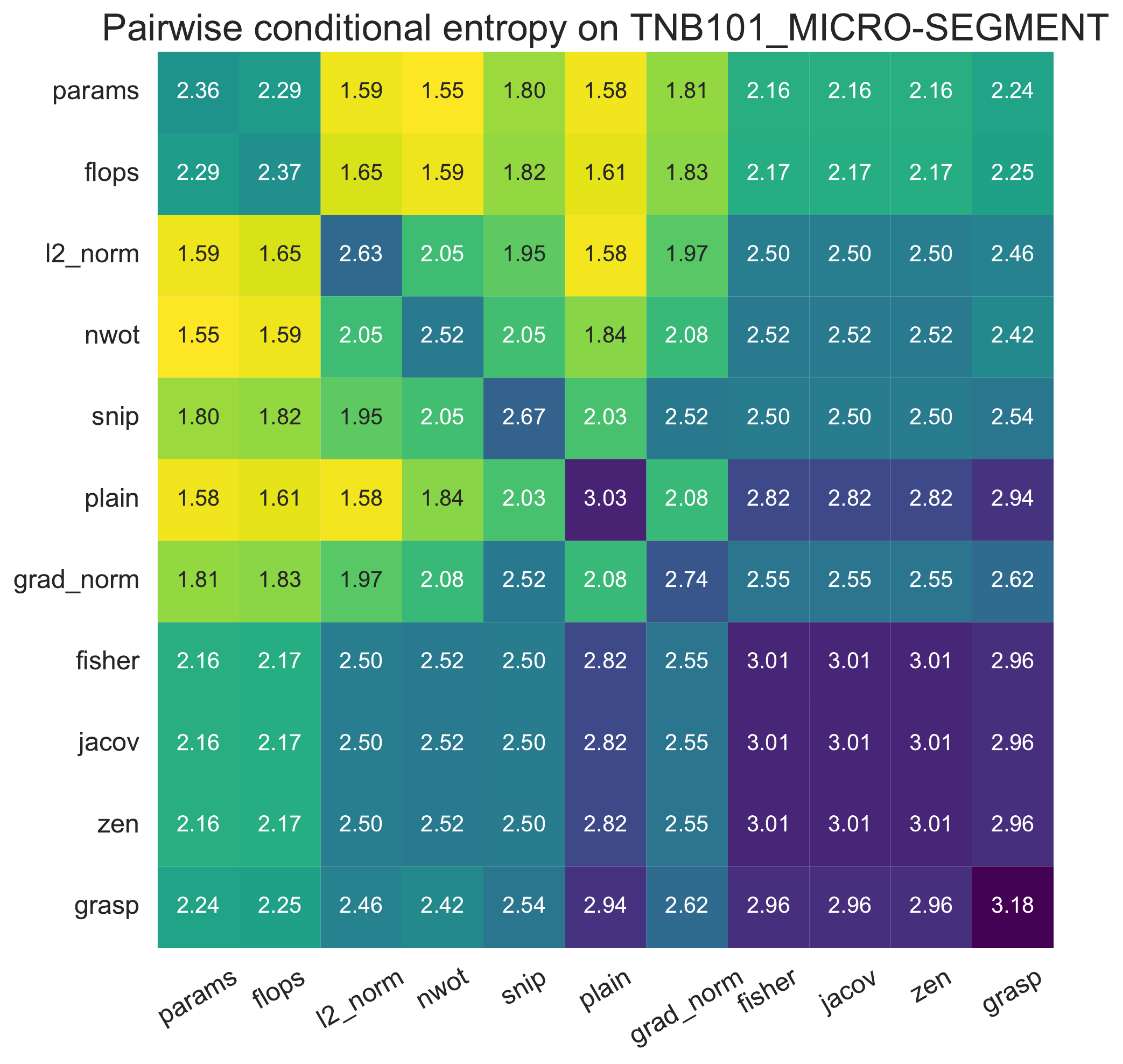}
    \includegraphics[width=.32\linewidth]{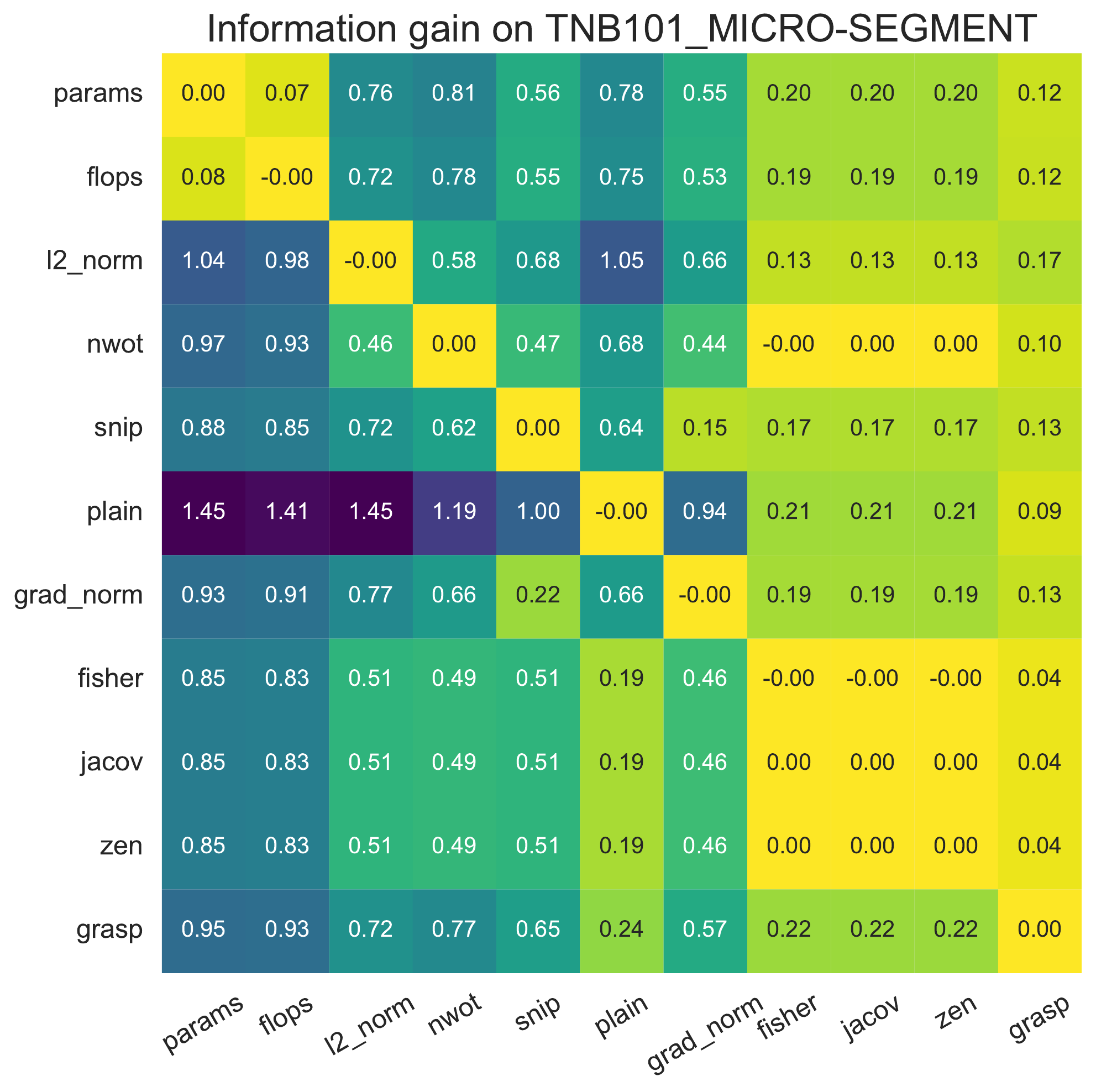}
    \includegraphics[width=.32\linewidth]{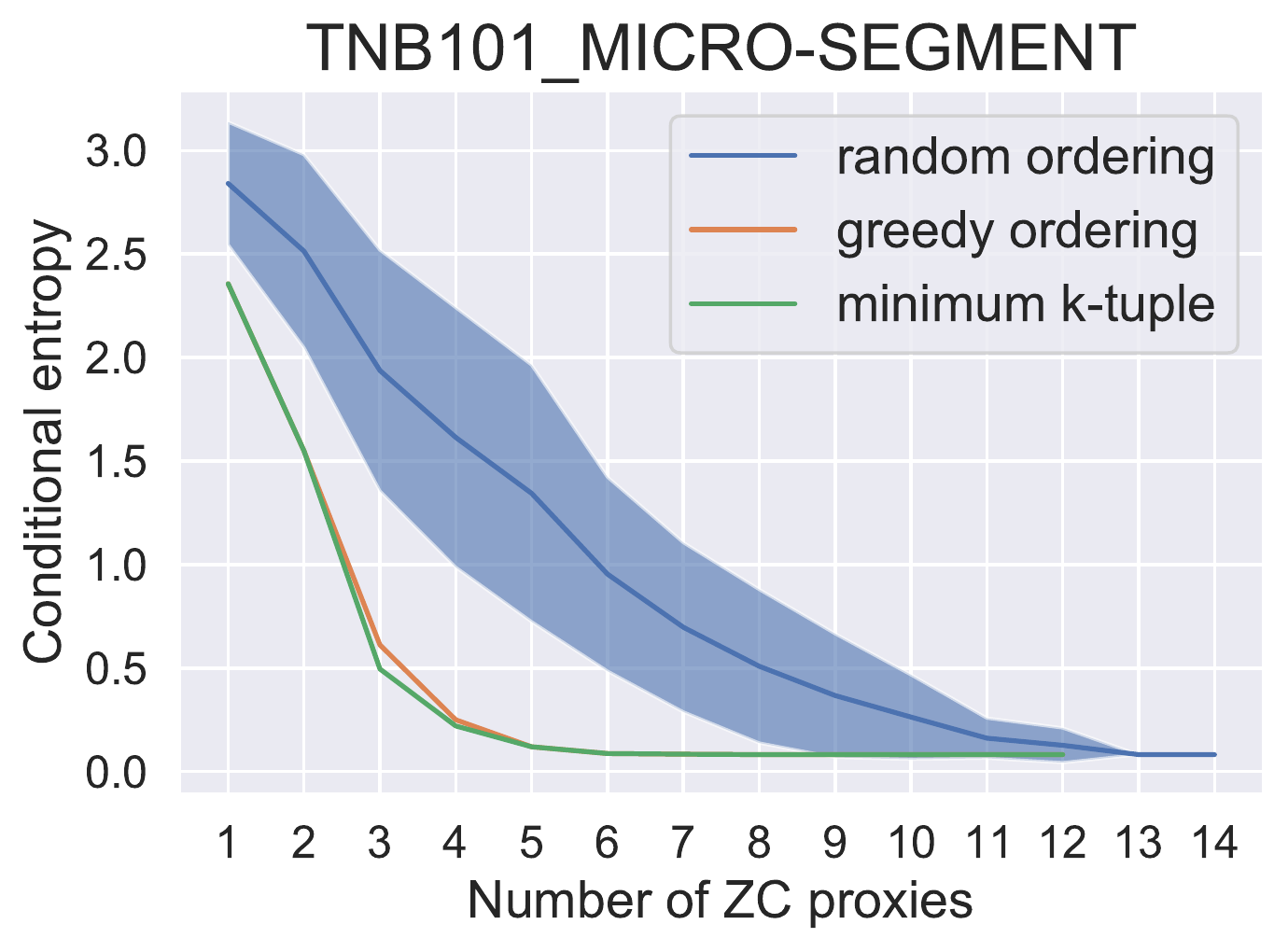}\\
    \caption{Conditional entropy and information gain (\textbf{IG}) for each ZC proxy pair across all search spaces and datasets (Left and Middle). Conditional entropy $H(y\mid z_{i_1},\dots,z_{i_k})$ vs.\ $k$, 
    where the ordering $z_{i_1},\dots,z_{i_k}$ is selected using three different strategies (Right). (5/5)}
    \label{fig:info_theory_appendix_5}

\end{figure}

In Figures \ref{fig:info_theory_appendix_1}, \ref{fig:info_theory_appendix_2}, \ref{fig:info_theory_appendix_3}, \ref{fig:info_theory_appendix_4}, \ref{fig:info_theory_appendix_5}, we show all the conditional entropy and information gain heatmaps, in addition to the entropy vs. number of ZC proxies plots for all benchmark, dataset pairs. Note that for TransNAS-Bench-101, there are no results for \texttt{epe\_nas} because it is not defined on non-classification tasks. Similarly, \texttt{synflow} returns 0.0 for certain non-classification tasks such as the ones in TransNAS-Bench-101, so we also removed \texttt{synflow} from the TransNAS-Bench-101 plots. 

\begin{figure}[t]
    \centering
    \includegraphics[width=.48\linewidth]{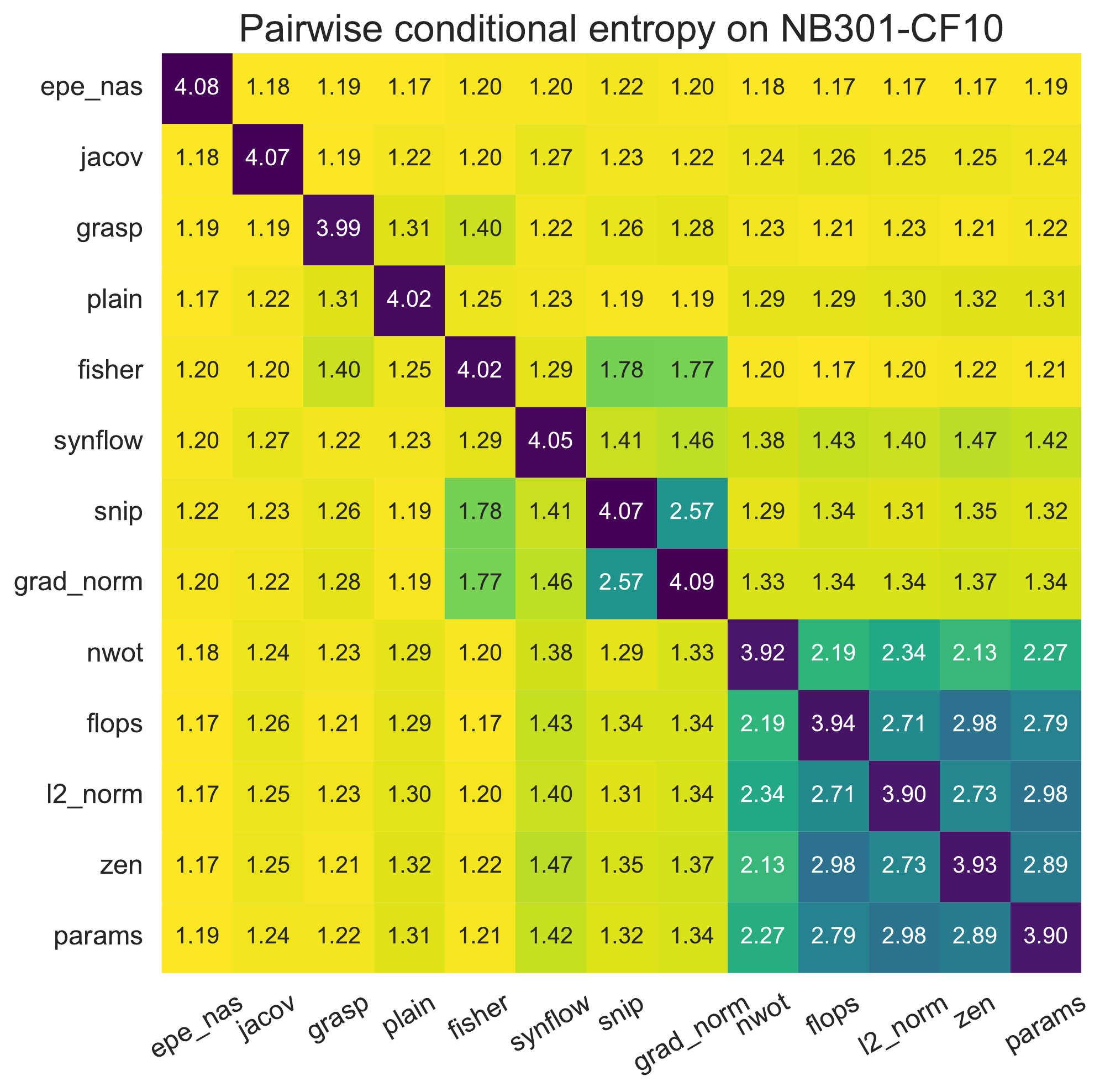}
    \includegraphics[width=.48\linewidth]{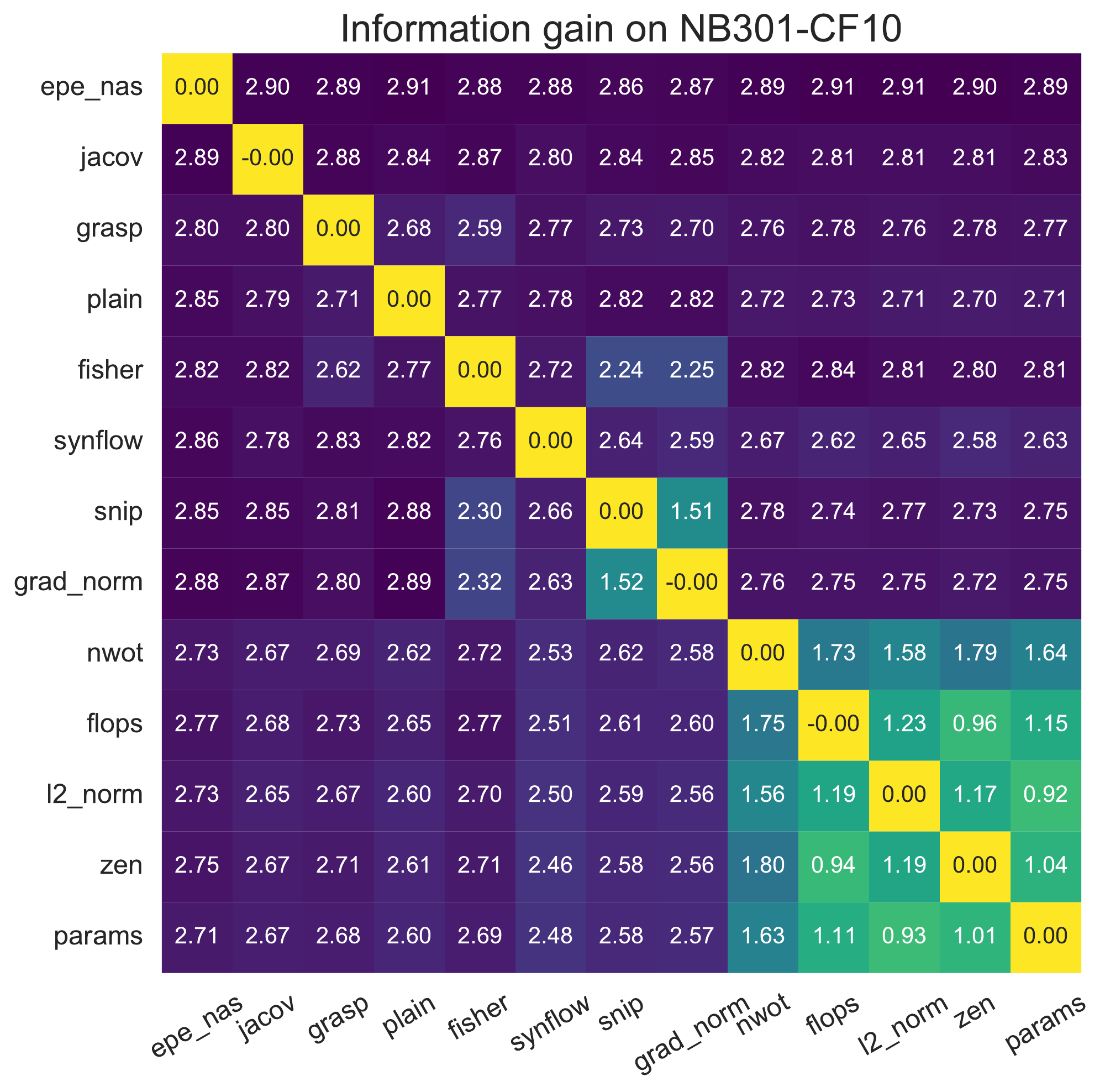} \\
    \includegraphics[width=.31\linewidth]{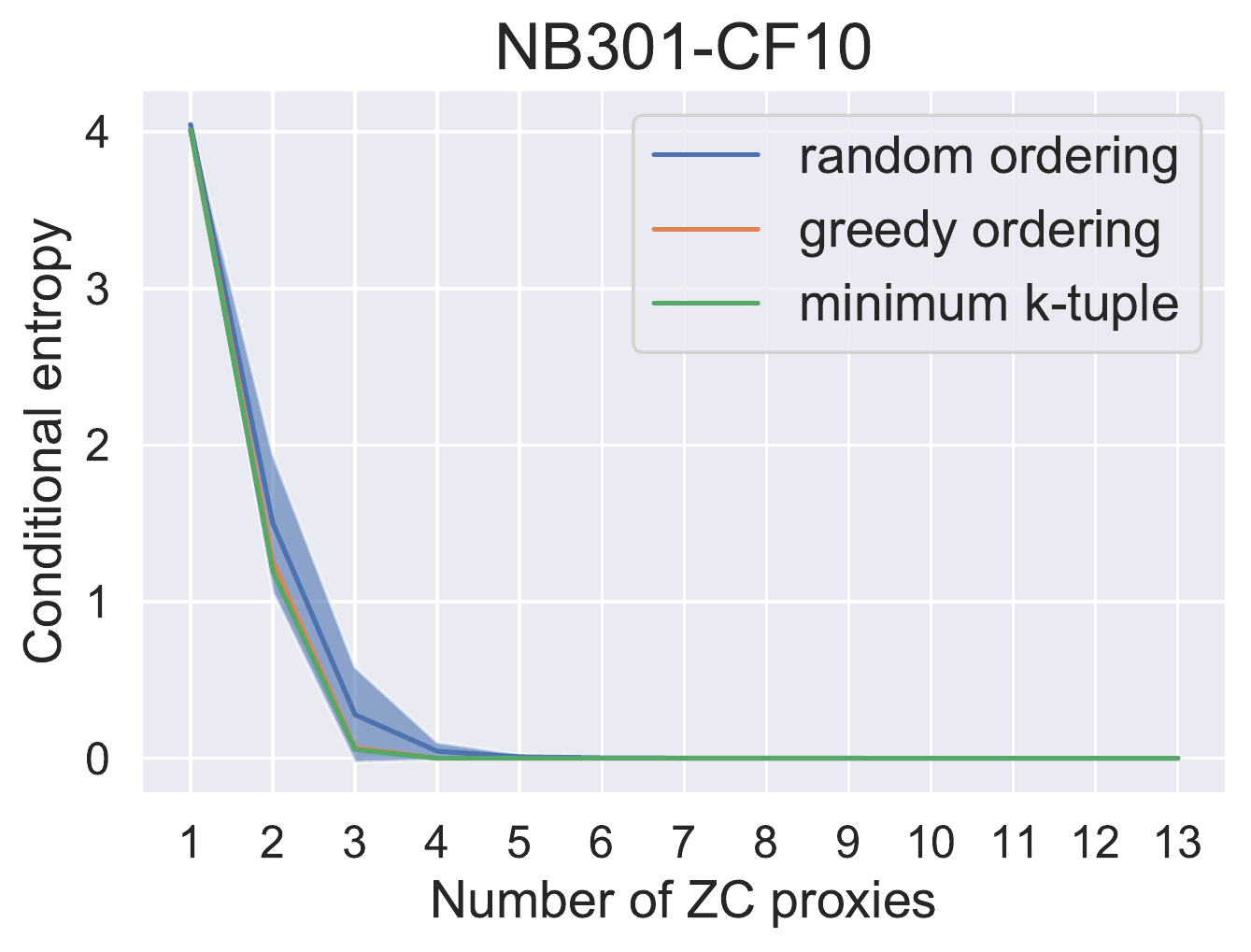}
    \includegraphics[width=.31\linewidth]{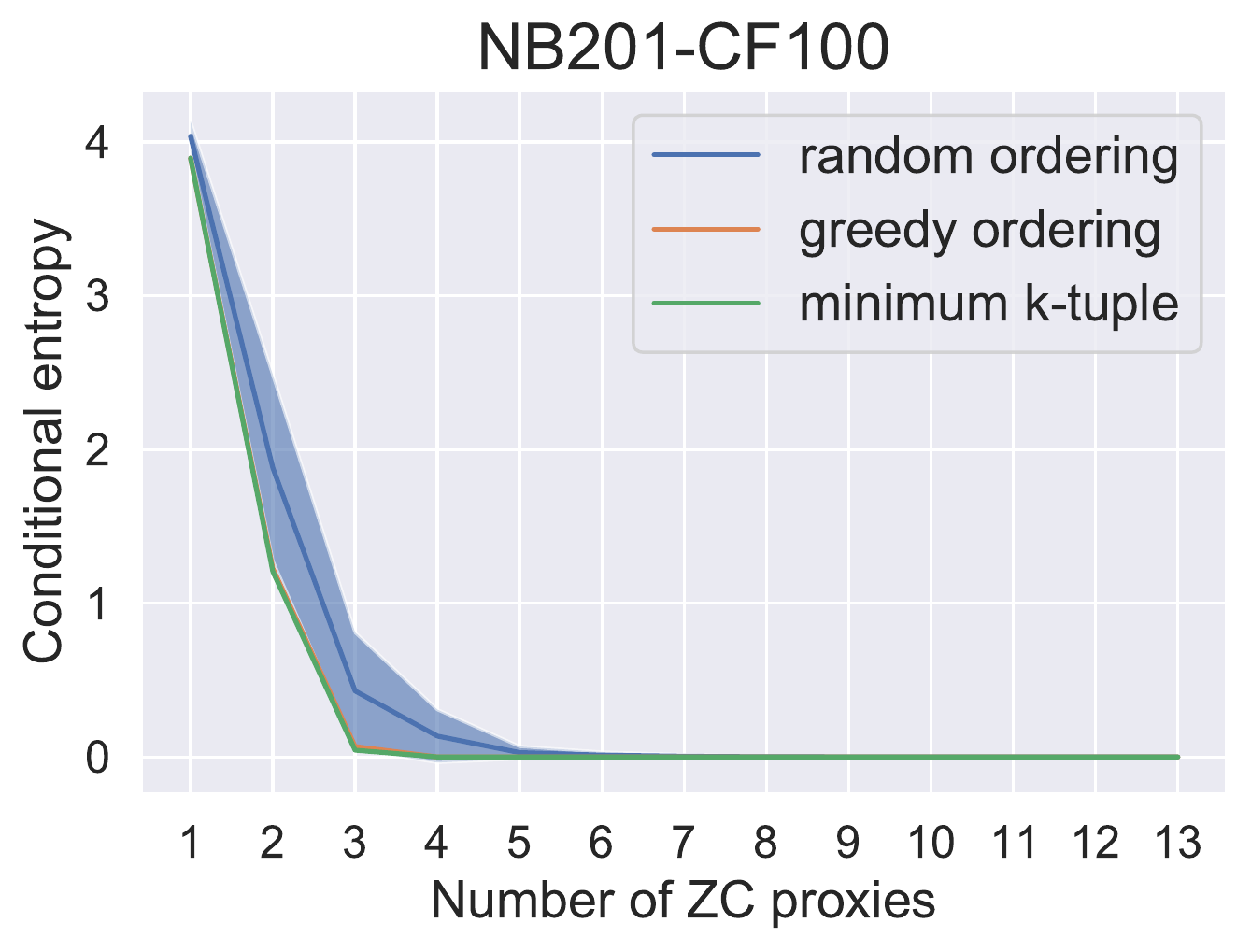}
    \includegraphics[width=.31\linewidth]{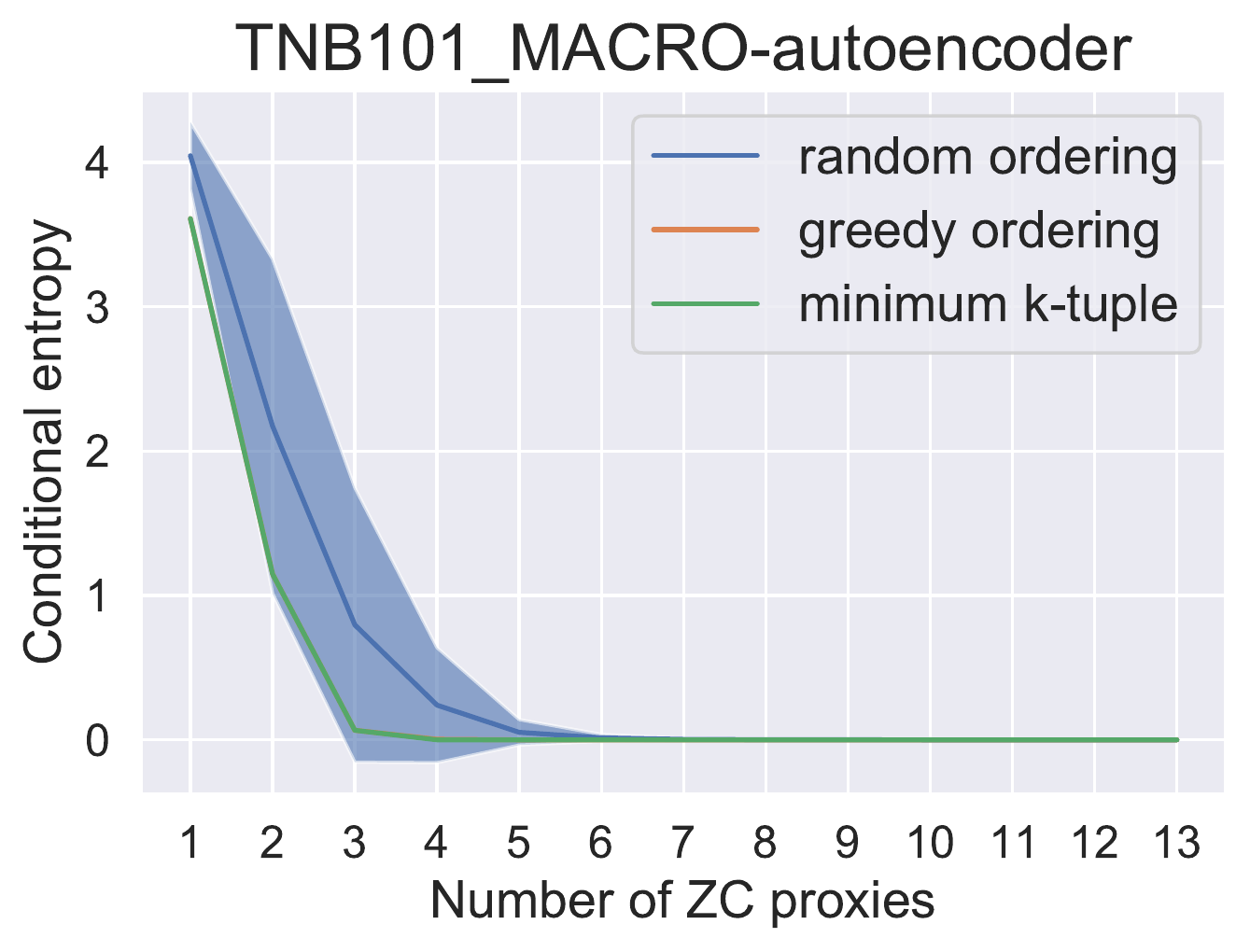}
    \\    
    \caption{
    Given a ZC proxy pair $(i,j)$, we compute the
    conditional entropy $H(y\mid z_i, z_j)$ (top left), and information gain $H(y\mid z_i) - H(y\mid z_i,z_j)$ (top right). 
    Conditional entropy $H(y\mid z_{i_1},\dots,z_{i_k})$ vs.\ $k$, 
    where the ordering $z_{i_1},\dots,z_{i_k}$ is selected using three different strategies. The minimum $k$-tuple and greedy ordering significantly overlap in the first two figures (bottom).
    Similar to Figure \ref{fig:cond_entropy}, but using a different bin discretization strategy.
    }
    %from adding the column's proxy to the row's proxy on two benchmarks.
    \label{fig:bin_ablation}
\end{figure}

While the conditional entropy and information gain plots from Figure \ref{fig:cond_entropy}
was computed using Equation \ref{eq:sturges} to compute the number of bins, 
% what was the original bin discretization strategy?
we also run the same experiment using a different discretization strategy: the bin dividers are computed based on percentages of the data.
See Figure \ref{fig:bin_ablation} (top). 
While the scales differ, we see largely the same trends. 
For example, there is still a cluster among \texttt{nwot}, \texttt{flops}, \texttt{l2\_norm}, \texttt{zen}, and \texttt{params}.
This suggests that this analysis is robust to the two different discretization strategies.
Next, we also re-run the experiment on conditional entropy vs.\ $k$ from Figure \ref{fig:cond_entropy} using the top 1000 architectures only, which may be important in the context of NAS, since NAS is concerned with finding the \emph{best} architectures. 
See Figure \ref{fig:bin_ablation} (bottom). 
We find that the random ordering performs comparatively better, predictably implying that it is harder to distinguish architectures that are in the top 1000 vs.\ randomly drawn architectures.

\subsection{Details from Section \ref{subsec:bias}: biases}
In this section, we give details from Section \ref{subsec:bias}.
In Table \ref{tab:bias_full}, for each bias metric we assess, we show the ZC proxies with the highest and lowest absolute correlation for each search space and dataset, if applicable. For the number of parameters bias, we do not consider the ZC proxies of \texttt{params} and \texttt{flops} since they trivially have 1.00 correlation. Note that operation biases are not available in TransNASBench101-Macro because the search space is architecture-level. 
This is an extension of Table \ref{tab:biases}.

\begin{table}[t]
\caption{
Pearson correlation coefficients between predictors and bias metrics (in bold) on different datasets, for the most and least biased ZC proxies
on each search space and task.
For example, for the \textbf{Conv:pool} bias on NB201-CF10, \texttt{synflow} is most biased, with a correlation of 0.76, while \texttt{grasp} is least biased (in terms of absolute value), with a correlation of -0.01. 
%Note that some biases are constant on some search spaces and are therefore non applicable. For example, the cell size for NB301 
} \label{tab:bias_full}
\resizebox{\linewidth}{!}{%
\centering
\begin{tabular}{@{}l|c|c|c|c|c|c|c|c@{}}
\toprule
\multicolumn{1}{l|}{\textbf{Name}} & \multicolumn{2}{c|}{\textbf{Conv:pool}} & \multicolumn{2}{c|}{\textbf{Cell size}} & \multicolumn{2}{c|}{\textbf{Num.\ skip connections}} &  
\multicolumn{2}{c}{\textbf{Num.\ parameters}} \\

&Most biased & Least biased & Most biased & Least biased & Most biased & Least Biased & Most biased & Least biased \\

\midrule 
\multirow{2}{*}{\texttt{NB101-CF10}} & synflow & grasp & \multirow{2}{*}{n/a}& \multirow{2}{*}{n/a}& \multirow{2}{*}{n/a} & \multirow{2}{*}{n/a} & nwot & epe\_nas \\ 
& 0.76 & -0.01 &&&&&0.74& -0.02 \\ [2ex]
\multirow{2}{*}{\texttt{NB201-CF10}}  & l2\_norm &grasp& synflow &grasp& l2\_norm &grasp & l2\_norm & grasp \\
& 0.87 & 0.01 & 0.57 & -0.02 & -0.41& -0.01 & 0.70 & 0.00\\ [2ex]

\multirow{2}{*}{\texttt{NB201-CF100}}  & l2\_norm &grasp& synflow &grasp& l2\_norm &grasp & l2\_norm & fisher \\
& 0.87 & 0.01 & 0.57 & -0.01 & -0.41& -0.01 & 0.70 & 0.01\\ [2ex]

\multirow{2}{*}{\texttt{NB201-IM}}  & l2\_norm &grasp& synflow &grasp& l2\_norm &grasp & l2\_norm & grasp \\
& 0.87 & 0.01 & 0.58 & 0.01 & -0.41& -0.01 & 0.70 & 0.01\\ [2ex]

\multirow{2}{*}{\texttt{NB301-CF10}}  & params &fisher& \multirow{2}{*}{n/a} &\multirow{2}{*}{n/a}& flops &epe\_nas & zen & epe\_nas \\
& 0.78 & 0.01 &  &  & -0.35& 0.01 & 0.99 & -0.01\\ [2ex]

\multirow{2}{*}{\texttt{TNB101\_MICRO-JIGSAW}} & \multirow{2}{*}{n/a}& \multirow{2}{*}{n/a}& l2\_norm& grasp& plain & grasp & l2\_norm & grasp\\
&  & & 0.70 & -0.02 & 0.50 & -0.01 & 0.64  & 0.02\\ [2ex] 

\texttt{TNB101\_MICRO-SCENE}  & \multirow{2}{*}{n/a}& \multirow{2}{*}{n/a}& l2\_norm & fisher& plain & grasp & snip & grasp  \\
&  & & 0.70 & 0.07 & 0.49 & -0.10 & 0.64  & -0.04\\ [2ex] 

\texttt{TNB101\_MICRO-OBJECT} & \multirow{2}{*}{n/a}& \multirow{2}{*}{n/a}& l2\_norm & fisher& plain & grasp & l2\_norm & grasp  \\
&  & & 0.70 & -0.08 & 0.49 & -0.06 & 0.64  & -0.02\\ [2ex]  

\texttt{TNB101\_MICRO-AUTOENC} & \multirow{2}{*}{n/a}& \multirow{2}{*}{n/a}& l2\_norm & grasp& grad\_norm & grasp & l2\_norm & grasp  \\
&  & & 0.70 & -0.02 & -0.46 & -0.03 & 0.64  & 0.02\\ [2ex] 

\texttt{TNB101\_MICRO-NORMAL} & \multirow{2}{*}{n/a}& \multirow{2}{*}{n/a}& l2\_norm & plain& snip & grasp & l2\_norm & plain  \\
&  & & 0.70 & 0.01 & -0.45 & -0.01 & 0.64  & 0.00\\ [2ex] 

\texttt{TNB101\_MICRO-ROOM}  & \multirow{2}{*}{n/a}& \multirow{2}{*}{n/a}& l2\_norm & fisher& plain & jacov & l2\_norm & grasp  \\
&  & & 0.70 & 0.10 & 0.45 & 0.14 & 0.64  & -0.01\\ [2ex]  

\texttt{TNB101\_MICRO-SEGMENT}  & \multirow{2}{*}{n/a}& \multirow{2}{*}{n/a}& l2\_norm & grasp& grad\_norm & grasp & l2\_norm & grasp  \\
&  & & 0.70 & 0.00 & -0.43 & 0.01 & 0.64  & -0.01\\ [2ex]  

\texttt{TNB101\_MACRO-JIGSAW} & \multirow{2}{*}{n/a}& \multirow{2}{*}{n/a}& \multirow{2}{*}{n/a} & \multirow{2}{*}{n/a}& \multirow{2}{*}{n/a} & \multirow{2}{*}{n/a} & l2\_norm & plain  \\
& & & &  & &  & 0.89 & 0.04 \\ [2ex] 

\texttt{TNB101\_MACRO-SCENE}  & \multirow{2}{*}{n/a}& \multirow{2}{*}{n/a}& \multirow{2}{*}{n/a} & \multirow{2}{*}{n/a}& \multirow{2}{*}{n/a} & \multirow{2}{*}{n/a} & l2\_norm & plain  \\
& & & &  & &  & 0.90 & 0.05 \\ [2ex] 

\texttt{TNB101\_MACRO-OBJECT} & \multirow{2}{*}{n/a}& \multirow{2}{*}{n/a}& \multirow{2}{*}{n/a} & \multirow{2}{*}{n/a}& \multirow{2}{*}{n/a} & \multirow{2}{*}{n/a} & l2\_norm & plain  \\
& & & &  & &  & 0.89 & 0.05 \\ [2ex] 

\texttt{TNB101\_MACRO-AUTOENC}  & \multirow{2}{*}{n/a}& \multirow{2}{*}{n/a}& \multirow{2}{*}{n/a} & \multirow{2}{*}{n/a}& \multirow{2}{*}{n/a} & \multirow{2}{*}{n/a} & l2\_norm & plain  \\
& & & &  & &  & 0.89 & 0.01 \\ [2ex] 

\texttt{TNB101\_MACRO-NORMAL} & \multirow{2}{*}{n/a}& \multirow{2}{*}{n/a}& \multirow{2}{*}{n/a} & \multirow{2}{*}{n/a}& \multirow{2}{*}{n/a} & \multirow{2}{*}{n/a} & l2\_norm & grasp  \\
& & & &  & &  & 0.89 & -0.02 \\ [2ex] 

\texttt{TNB101\_MACRO-ROOM} & \multirow{2}{*}{n/a}& \multirow{2}{*}{n/a}& \multirow{2}{*}{n/a} & \multirow{2}{*}{n/a}& \multirow{2}{*}{n/a} & \multirow{2}{*}{n/a} & l2\_norm & grasp  \\
& & & &  & &  & 0.89 & 0.00 \\ [2ex] 

\texttt{TNB10\_MACRO-SEGMENT} & \multirow{2}{*}{n/a}& \multirow{2}{*}{n/a}& \multirow{2}{*}{n/a} & \multirow{2}{*}{n/a}& \multirow{2}{*}{n/a} & \multirow{2}{*}{n/a} & l2\_norm & plain  \\
& & & &  & &  & 0.89 & 0.00 \\ [2ex] 

\bottomrule
\end{tabular}
}
\end{table}

%%%%%%%%%%%%

%%%%%%%%%%%%

\subsection{NAS-Bench-Suite-Zero Speedup Details}\label{app:speedup}
Here we show statistics on how our benchmark speeds up NAS experiments previously done with NAS-Bench-Suite by orders of magnitude. 
See Table \ref{tab:speedups}.

\begin{table}[t]
\caption{Runtimes (on an Intel Xeon Gold 6242 CPU) for all types of experiments done in this paper, with and without \suite.
The runtimes of the experiments with NBSuite are computed by using the average training times for randomly drawn architectures from each search space in NBSuite.
} \label{tab:speedups}
\resizebox{\linewidth}{!}{%
\centering
\begin{tabular}{llll}
\toprule
                           Experiment & With NBSuite (approx.) & With NBSuite + NBSuite-Zero & Speedup \\
\midrule
                    Mutual information study &          158.2 hours &       124.1 seconds &     4592$\times$ \\
                    Architecture bias study        &          6956 hours &         14.8 seconds &     1776003$\times$  \\
%               Rank correlation study &          NaN &                         NaN &     NaN \\
  Standalone XGBoost+ZC, 100 trials &          1033 hours&            100 seconds &     37180$\times$ \\
               BANANAS+ZC, 100 trials &          4694 hours &           4260 seconds &     3967$\times$ \\
                NPENAS+ZC, 100 trials &          1033 hours &          3470 seconds &     1071$\times$ \\
\bottomrule
\end{tabular}
}
\end{table}

\section{Details from Section \ref{sec:nas}} \label{app:nas}
In this section, we give the full details from Section \ref{sec:nas}.

We start by presenting the complete standalone predictor experiments.
In Section \ref{sec:nas}, we mentioned that on NAS-Bench-201 CIFAR-100, the Spearman rank correlation of XGBoost predictions trained on 100 randomly sampled architectures and averaged over 100 trials, improves from 0.640 to 0.908 when 13 ZC proxies are added.
Now, we present the results of this same experiment for all benchmarks. See Table \ref{tab:xgb_train_100}.
We see that the large improvement is consistent across the board.
We also run the same experiment when XGBoost is trained on 1000 randomly sampled architectures. 
See Table \ref{tab:xgb_train_1000}.
Even though the predictions with the original XGBoost already have high rank correlation, we show that ZC proxies improve the performance even more.

\subsection{Feature importances of ZC proxies}
\begin{figure}[t]
    \centering
    \includegraphics[width=.96\linewidth]{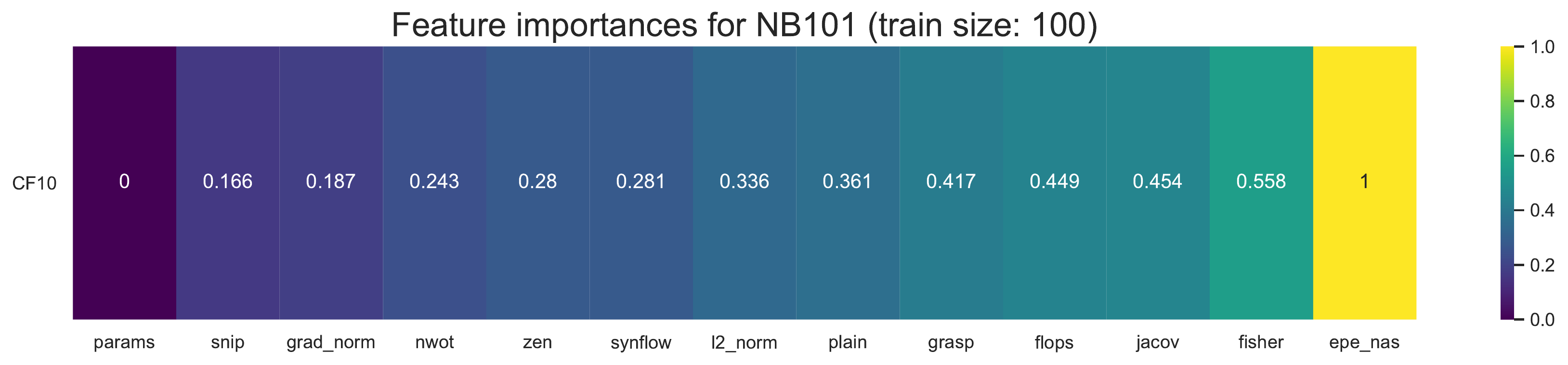}
    \includegraphics[width=.96\linewidth]{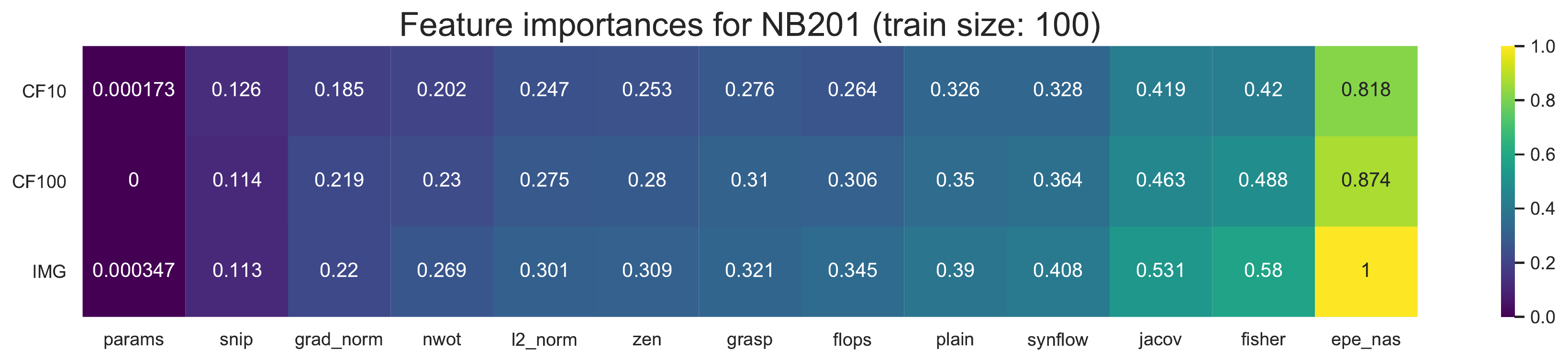}
    \includegraphics[width=.96\linewidth]{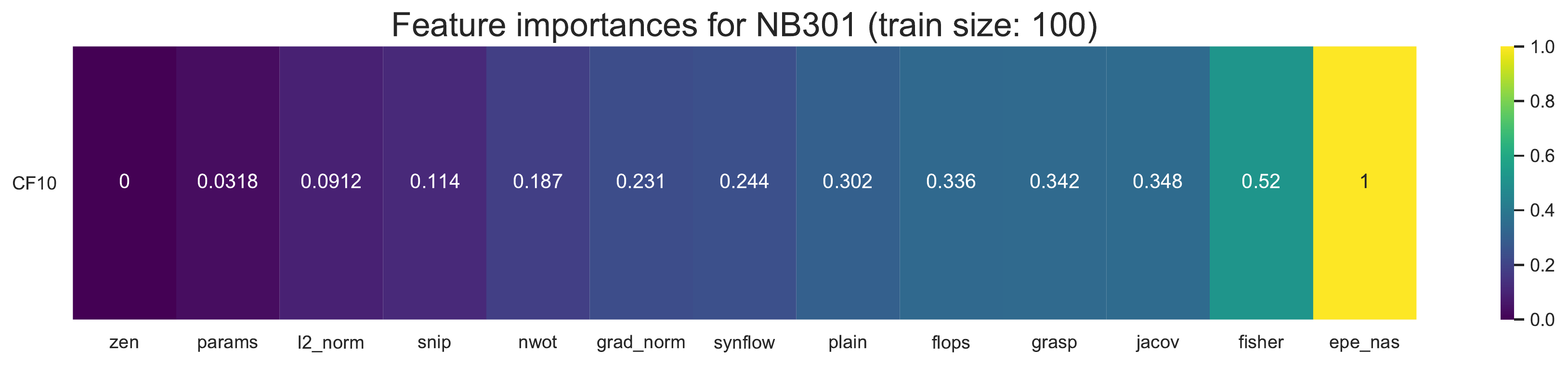}
    \includegraphics[width=\linewidth]{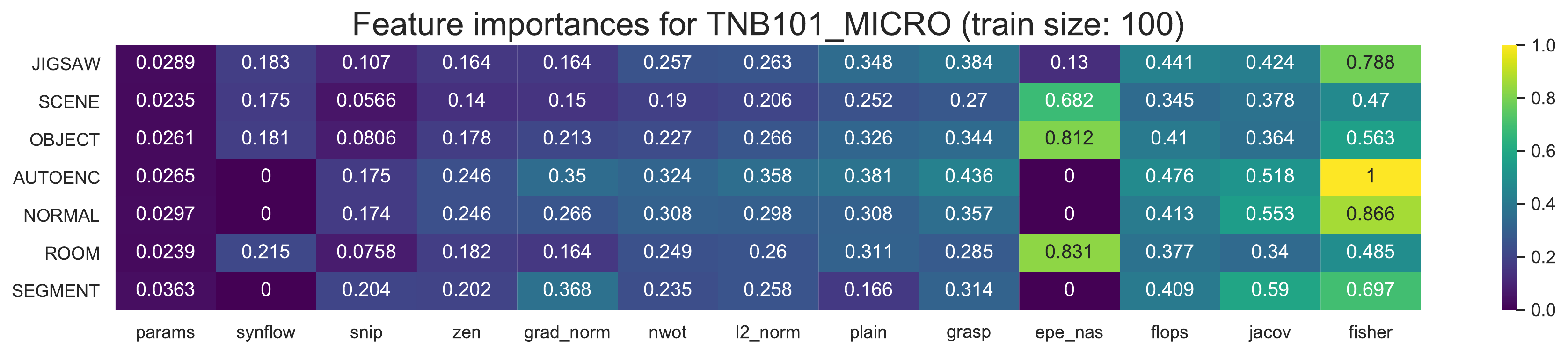}
    \includegraphics[width=\linewidth]{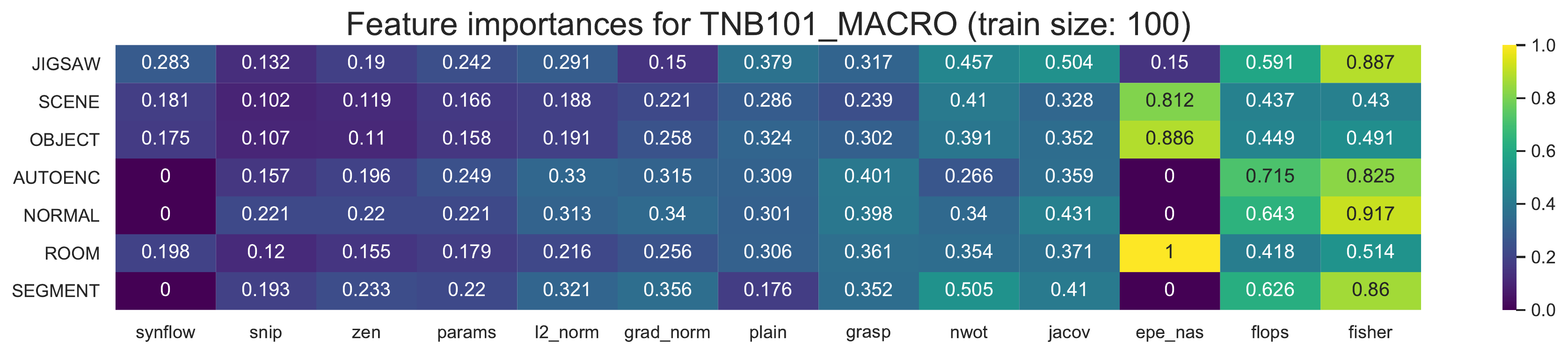}
    \caption{Feature importance values for XGBoost trained on a set of 100 architectures using ZC proxies as features.    
    }
    \label{fig:feat_imp_100}
\end{figure}

\begin{figure}[t]
    \centering
    \includegraphics[width=.96\linewidth]{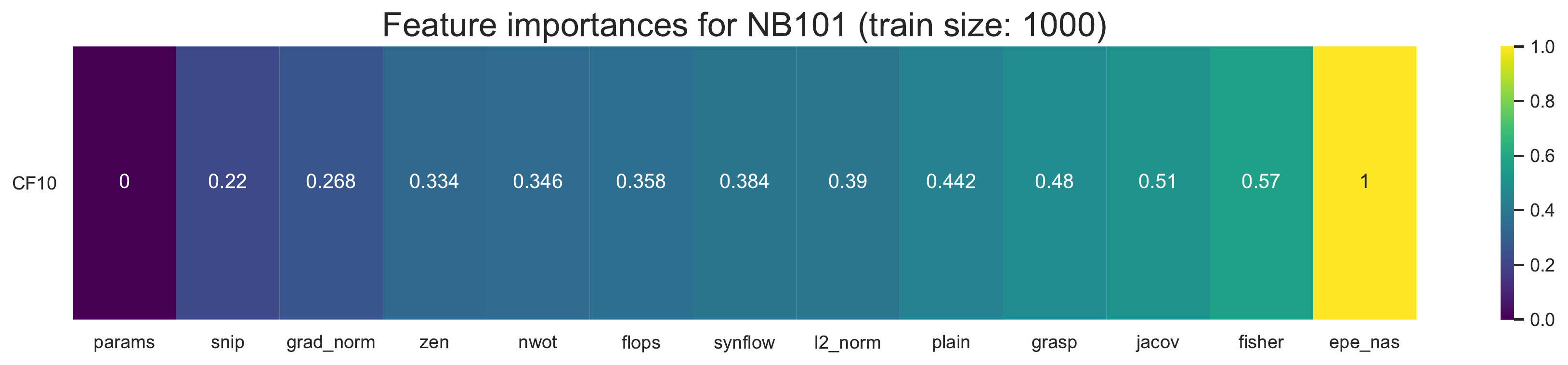}
    \includegraphics[width=.96\linewidth]{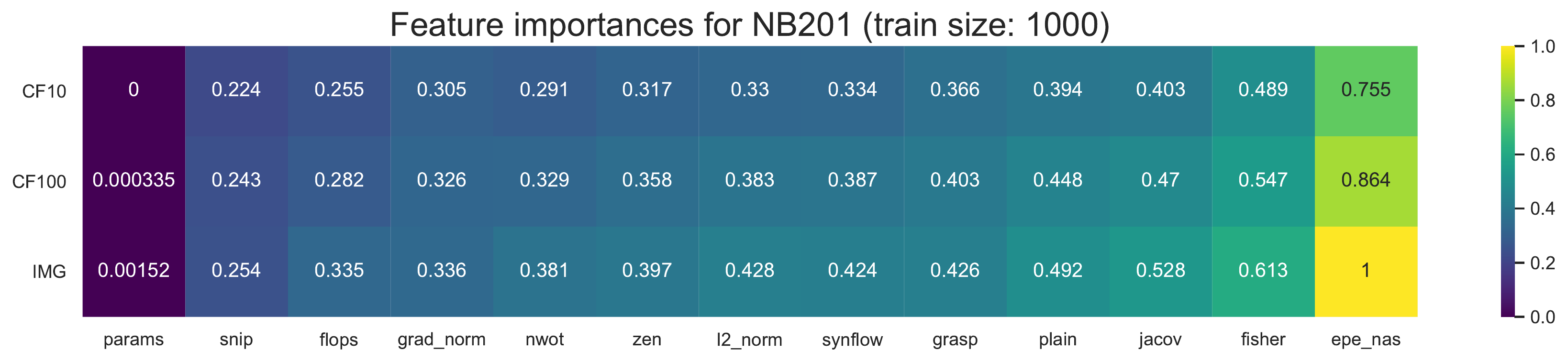}
    \includegraphics[width=.96\linewidth]{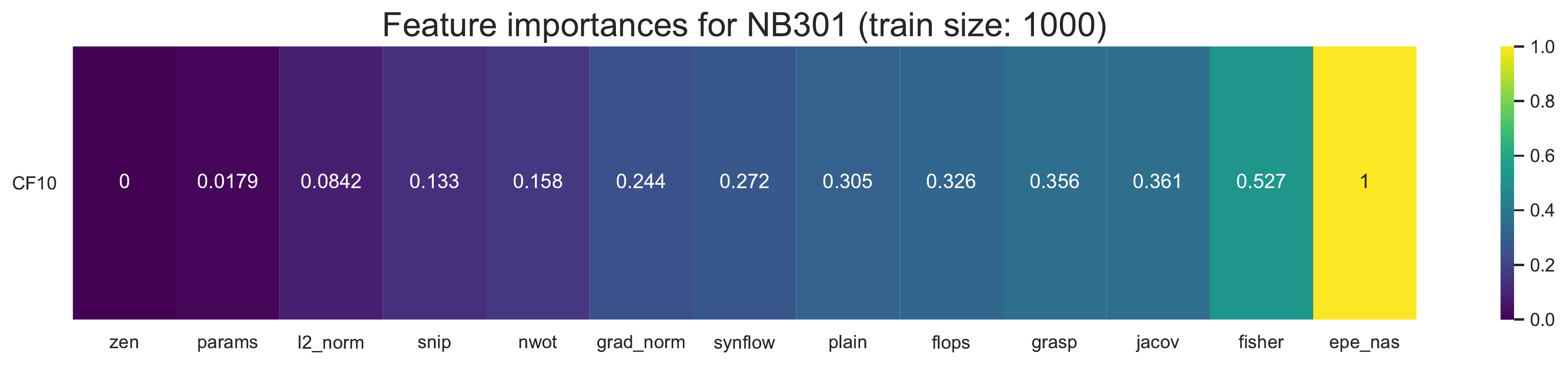}
    \includegraphics[width=\linewidth]{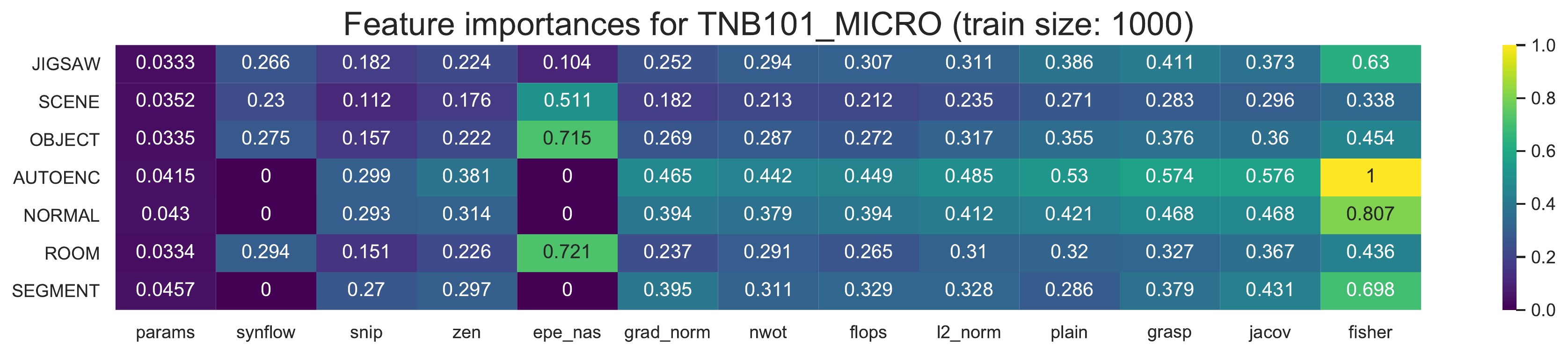}
    \includegraphics[width=\linewidth]{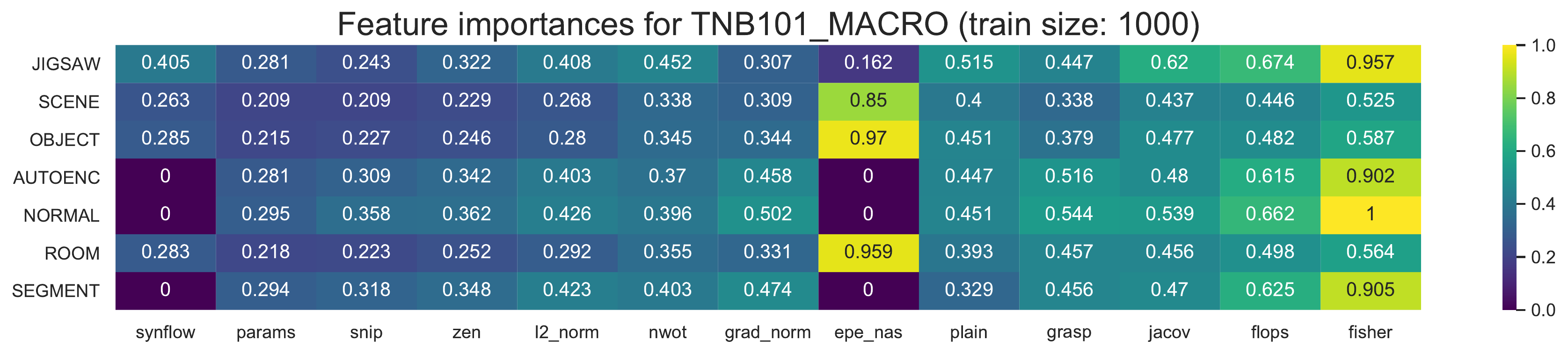}
    \caption{Feature importance values for XGBoost trained on a set of 1000 architectures using ZC proxies as features.    
    }
    \label{fig:feat_imp_1000}
\end{figure}

In this section, we train an XGBoost surrogate model on 100 and 1000 randomly drawn architectures using the ZC proxies as features, and then we plot feature importances for each feature.
The feature importance is calculated by the the number of times a feature is used to split the data across all trees (the default feature importance method in the XGBoost library \citep{chen2016xgboost}).
See Figures \ref{fig:feat_imp_100} and \ref{fig:feat_imp_1000} for the results with a training set size of 100 and 1000, respectively.

\begin{table}[t]
\centering
\caption{Average Spearman rank correlations between XGBoost predictions and validation accuracies, for each benchmark, across three different experiments: \textit{Encoding} uses only the encoding of the model, \textit{ZC} uses only the ZC features, and \textit{Both} concatenates ZC features to the encoding of the model. 100 models were used to train XGBoost.} \label{tab:xgb_train_100}
\resizebox{\linewidth}{!}{%
\begin{tabular}{lrrrrr}
\toprule
Features &  Encoding &     ZC &   Both &  \% Improvement (ZC) &  \% Improvement (Both) \\
Benchmark            &           &        &        &                     &                       \\
\midrule
NB101-CF10           &     0.546 &  0.708 &  0.718 &               29.67 &                 31.50 \\
NB201-CF10           &     0.622 &  0.905 &  0.906 &               45.50 &                 45.66 \\
NB201-CF100          &     0.640 &  0.907 &  0.908 &               41.71 &                 41.87 \\
NB201-IMGNT          &     0.683 &  0.879 &  0.883 &               28.70 &                 29.28 \\
NB301-CF10           &     0.314 &  0.405 &  0.465 &               28.98 &                 48.09 \\
TNB101\_MACRO-AUTOENC &     0.673 &  0.831 &  0.837 &               23.48 &                 24.37 \\
TNB101\_MACRO-JIGSAW  &     0.809 &  0.706 &  0.809 &              -12.73 &                  0.00 \\
TNB101\_MACRO-NORMAL  &     0.617 &  0.710 &  0.716 &               15.07 &                 16.05 \\
TNB101\_MACRO-OBJECT  &     0.736 &  0.840 &  0.843 &               14.13 &                 14.54 \\
TNB101\_MACRO-ROOM    &     0.683 &  0.589 &  0.707 &              -13.76 &                  3.51 \\
TNB101\_MACRO-SCENE   &     0.832 &  0.891 &  0.899 &                7.09 &                  8.05 \\
TNB101\_MACRO-SEGMENT &     0.900 &  0.807 &  0.876 &              -10.33 &                 -2.67 \\
TNB101\_MICRO-AUTOENC &     0.714 &  0.754 &  0.803 &                5.60 &                 12.46 \\
TNB101\_MICRO-JIGSAW  &     0.585 &  0.730 &  0.743 &               24.79 &                 27.01 \\
TNB101\_MICRO-NORMAL  &     0.657 &  0.801 &  0.809 &               21.92 &                 23.14 \\
TNB101\_MICRO-OBJECT  &     0.637 &  0.733 &  0.752 &               15.07 &                 18.05 \\
TNB101\_MICRO-ROOM    &     0.582 &  0.843 &  0.844 &               44.85 &                 45.02 \\
TNB101\_MICRO-SCENE   &     0.710 &  0.849 &  0.866 &               19.58 &                 21.97 \\
TNB101\_MICRO-SEGMENT &     0.767 &  0.886 &  0.897 &               15.51 &                 16.95 \\
\bottomrule
\end{tabular}
}
\end{table}

\begin{table}[t]
\centering
\caption{Average Spearman rank correlations between XGBoost predictions and validation accuracies, for each benchmark, across three different experiments: \textit{Encoding} uses only the encoding of the model, \textit{ZC} uses only the ZC features, and \textit{Both} concatenates ZC features to the encoding of the model. 1000 models were used to train XGBoost.} \label{tab:xgb_train_1000}
\resizebox{\linewidth}{!}{%
\begin{tabular}{lrrrrr}
\toprule
Features &  Encoding &     ZC &   Both &  \% Improvement (ZC) &  \% Improvement (Both) \\
Benchmark            &           &        &        &                     &                       \\
\midrule
NB101-CF10           &     0.748 &  0.811 &  0.851 &                8.42 &                 13.77 \\
NB201-CF10           &     0.890 &  0.954 &  0.961 &                7.19 &                  7.98 \\
NB201-CF100          &     0.906 &  0.953 &  0.959 &                5.19 &                  5.85 \\
NB201-IMGNT          &     0.922 &  0.948 &  0.957 &                2.82 &                  3.80 \\
NB301-CF10           &     0.678 &  0.496 &  0.705 &              -26.84 &                  3.98 \\
TNB101\_MACRO-AUTOENC &     0.890 &  0.903 &  0.917 &                1.46 &                  3.03 \\
TNB101\_MACRO-JIGSAW  &     0.812 &  0.801 &  0.856 &               -1.35 &                  5.42 \\
TNB101\_MACRO-NORMAL  &     0.692 &  0.759 &  0.764 &                9.68 &                 10.40 \\
TNB101\_MACRO-OBJECT  &     0.846 &  0.880 &  0.888 &                4.02 &                  4.96 \\
TNB101\_MACRO-ROOM    &     0.741 &  0.731 &  0.793 &               -1.35 &                  7.02 \\
TNB101\_MACRO-SCENE   &     0.936 &  0.936 &  0.953 &                0.00 &                  1.82 \\
TNB101\_MACRO-SEGMENT &     0.951 &  0.920 &  0.952 &               -3.26 &                  0.11 \\
TNB101\_MICRO-AUTOENC &     0.838 &  0.815 &  0.861 &               -2.74 &                  2.74 \\
TNB101\_MICRO-JIGSAW  &     0.768 &  0.827 &  0.833 &                7.68 &                  8.46 \\
TNB101\_MICRO-NORMAL  &     0.816 &  0.850 &  0.864 &                4.17 &                  5.88 \\
TNB101\_MICRO-OBJECT  &     0.806 &  0.841 &  0.858 &                4.34 &                  6.45 \\
TNB101\_MICRO-ROOM    &     0.874 &  0.943 &  0.947 &                7.89 &                  8.35 \\
TNB101\_MICRO-SCENE   &     0.862 &  0.929 &  0.943 &                7.77 &                  9.40 \\
TNB101\_MICRO-SEGMENT &     0.921 &  0.934 &  0.948 &                1.41 &                  2.93 \\
\bottomrule
\end{tabular}
}
\end{table}

\subsection{Ablation study on the number of ZC proxies}
\begin{figure}[t]
    \centering
    \includegraphics[width=.34\linewidth]{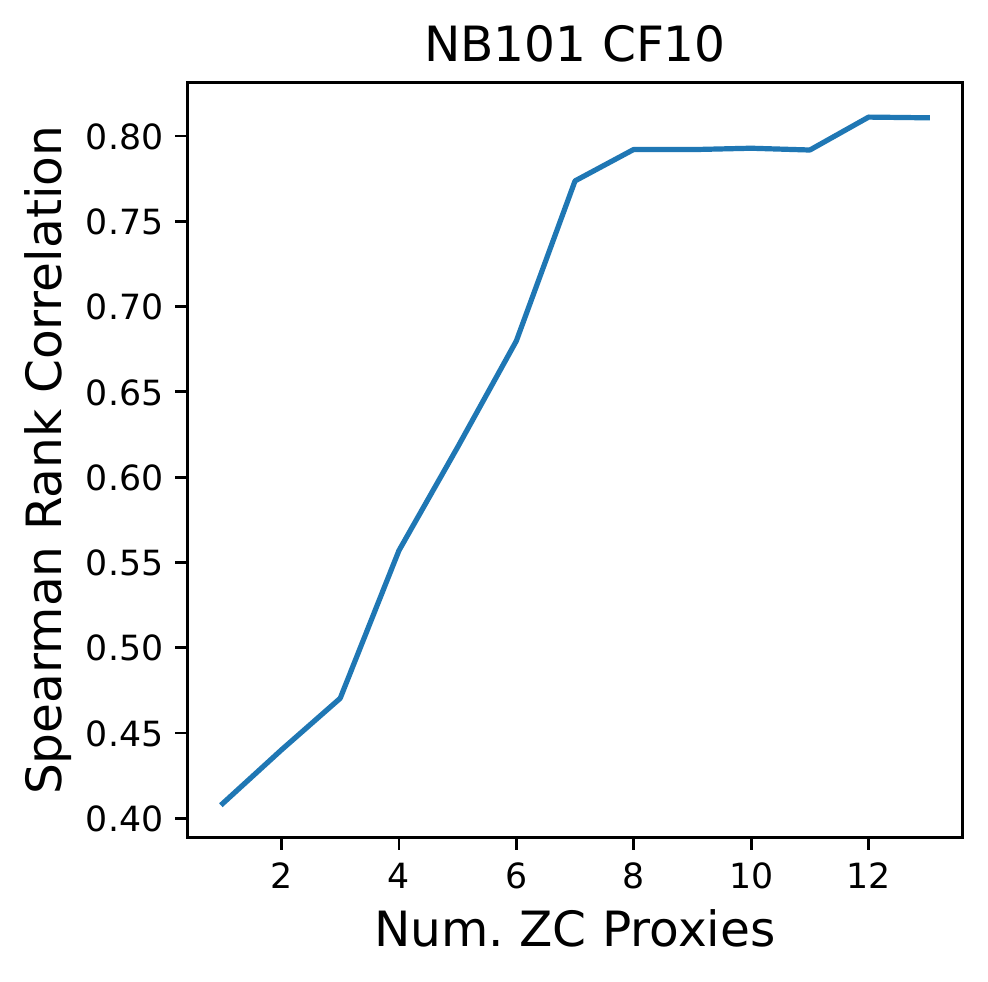}
    \includegraphics[width=.34\linewidth]{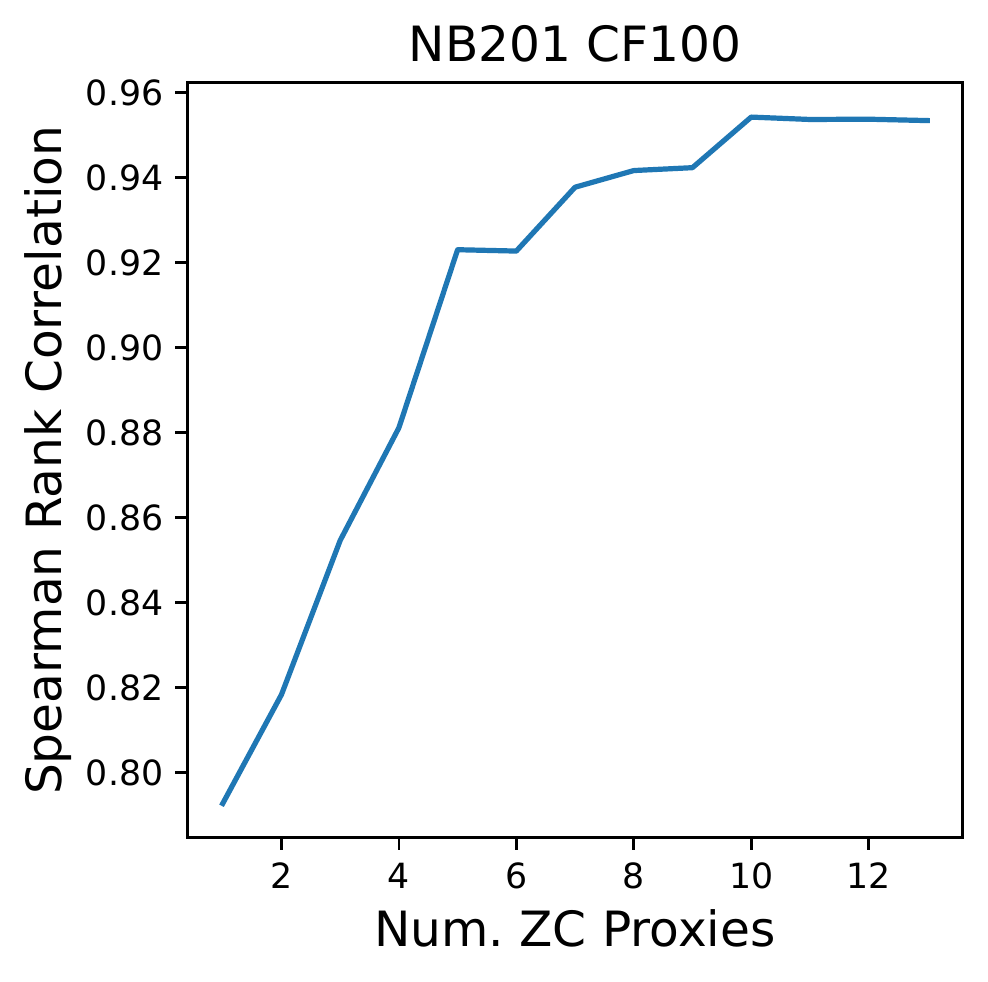}
    \includegraphics[width=.34\linewidth]{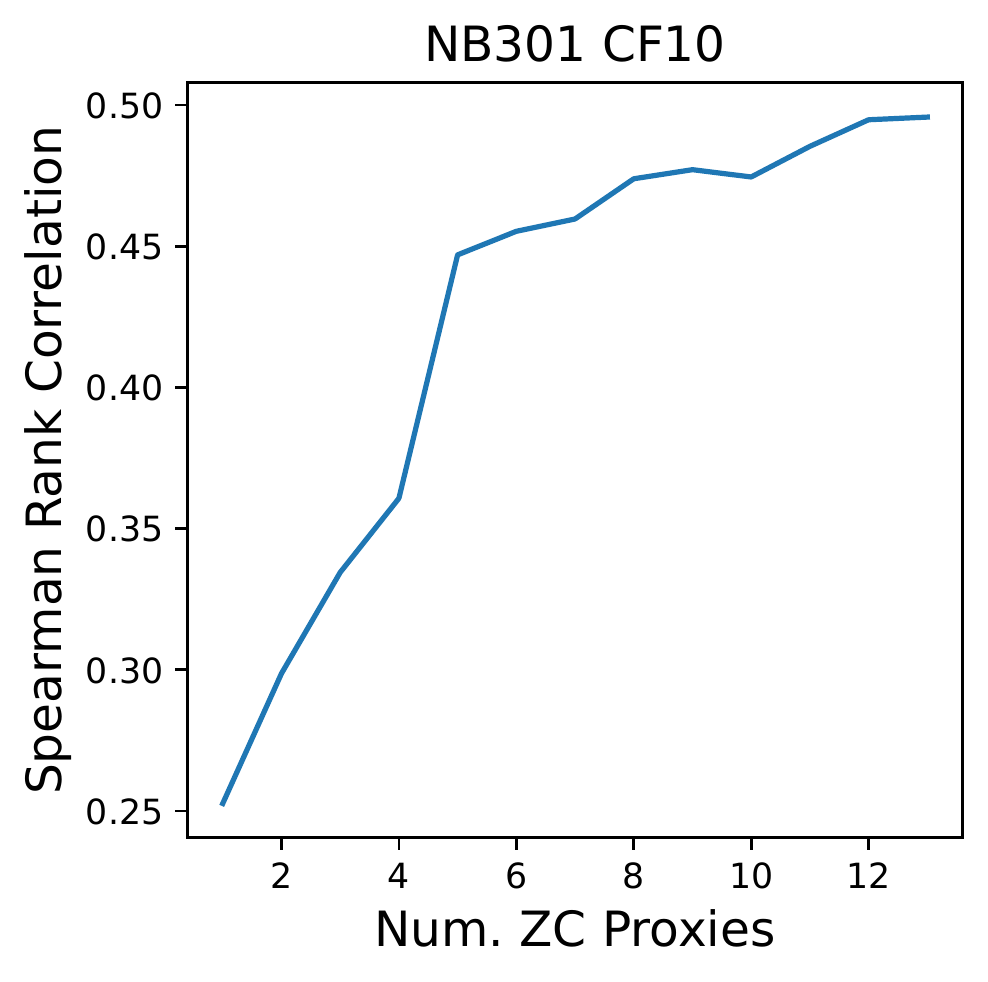}
    \includegraphics[width=.34\linewidth]{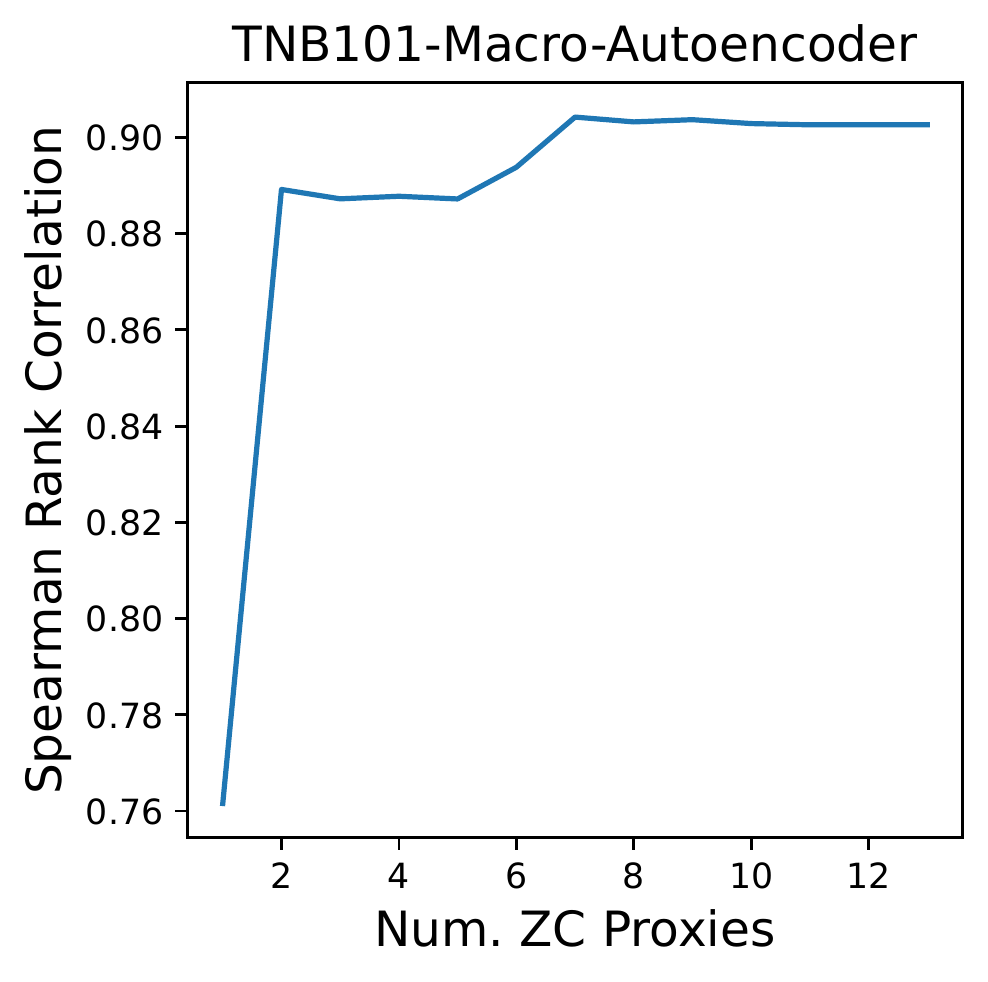}
    %'k_ablation_n101_c10.pdf', 'k_ablation_n201_c100.pdf', 'k_ablation_n301_c10.pdf', 'k_ablation_t_mac_auto.pdf']
    \\    
    \caption{
    Ablation study on the number of ZC proxies as features vs.\ rank correlation performance, for an XGBoost surrogate model trained on 1000 randomly drawn architectures. The ordering of ZC proxies is computed via the greedy method from Section \ref{subsec:bias}.
    }
    \label{fig:num_proxies_ablation}
\end{figure}

Next, we give an ablation study on the number of ZC proxies as features, for an XGBoost surrogate model trained on 1000 randomly drawn architectures. The ordering of ZC proxies is computed via the greedy method from Section \ref{subsec:bias}.
See Figure \ref{fig:num_proxies_ablation}.
We find that on all tasks, the best performance is achieved with all 13 ZC proxies (in some cases, there are ties). 
However, after 6-8 ZC proxies, there is only a small improvement up to the full 13 ZC proxies. This is consistent with our mutual information study from Section \ref{subsec:bias}.

%after 3-6 ZC proxies, the 

\subsection{Additional NAS results}
\begin{figure}[t]
    \centering
    \includegraphics[width=.32\linewidth]{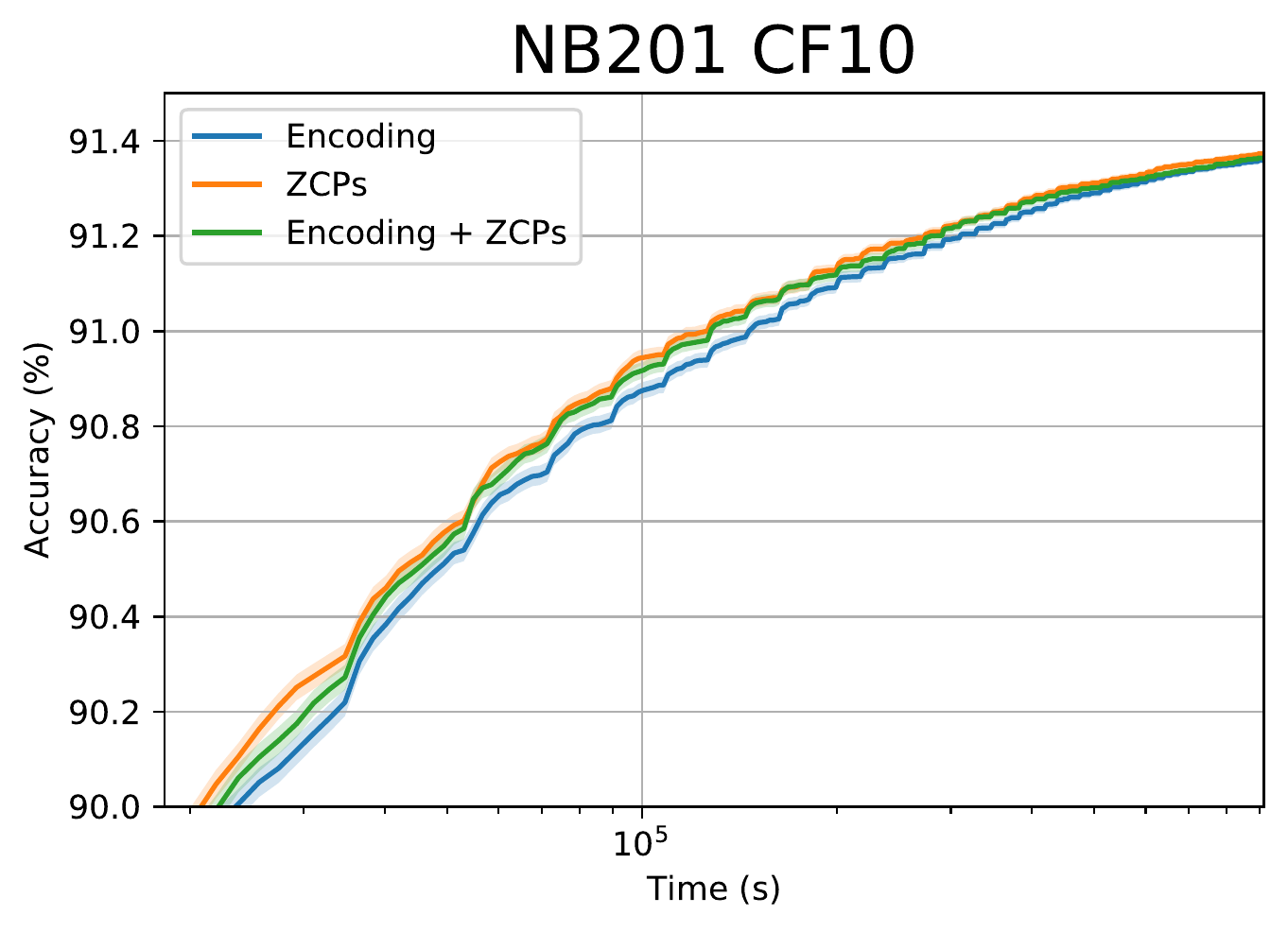}
    \includegraphics[width=.32\linewidth]{plots/section_4/bananas/NB201_CF100.pdf}
    \includegraphics[width=.32\linewidth]{plots/section_4/bananas/NB201_IMGNT.pdf} \\
    \includegraphics[width=.32\linewidth]{plots/section_4/bananas/NB301_CF10.pdf}
    \includegraphics[width=.32\linewidth]{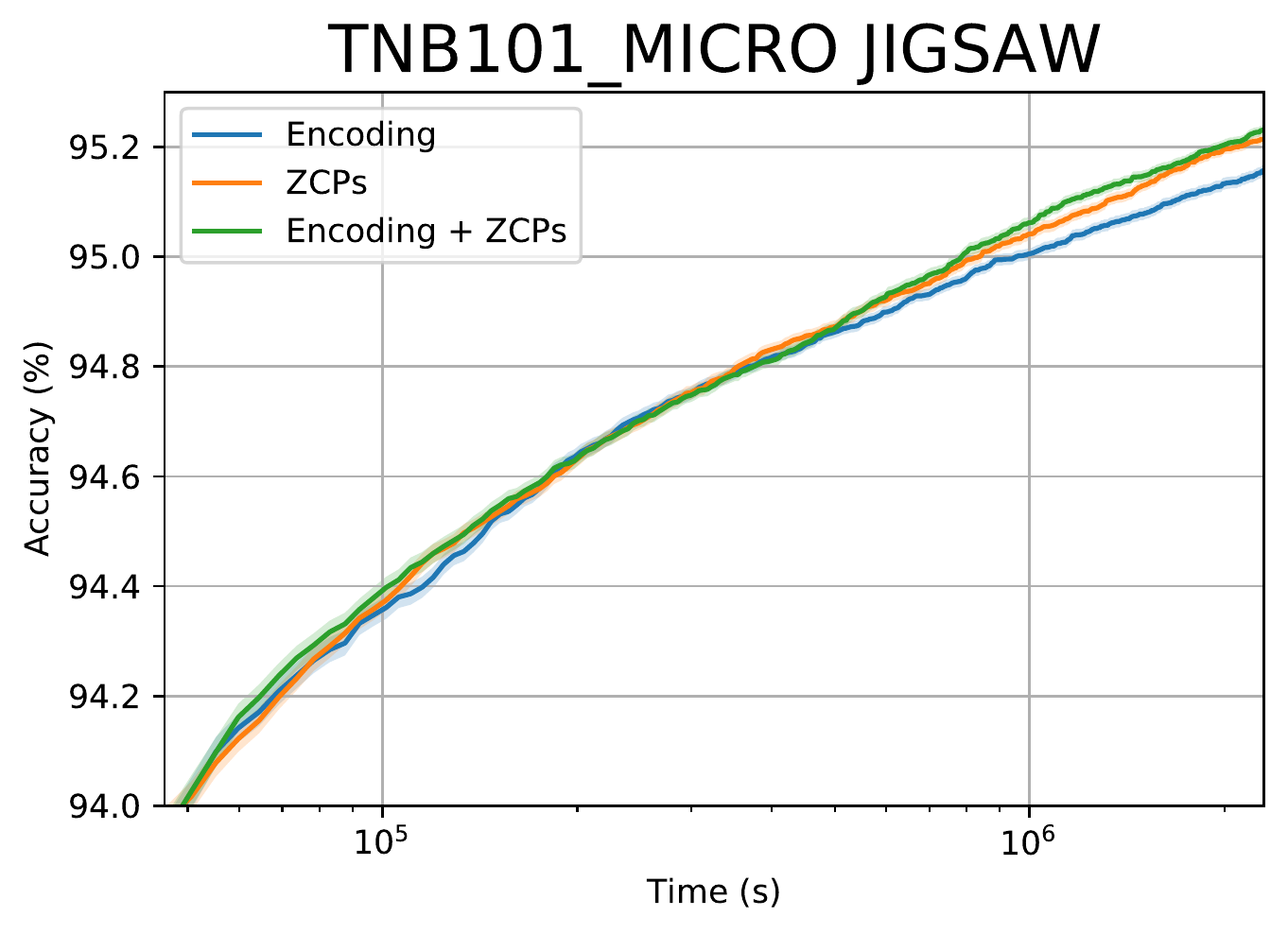}
    \includegraphics[width=.32\linewidth]{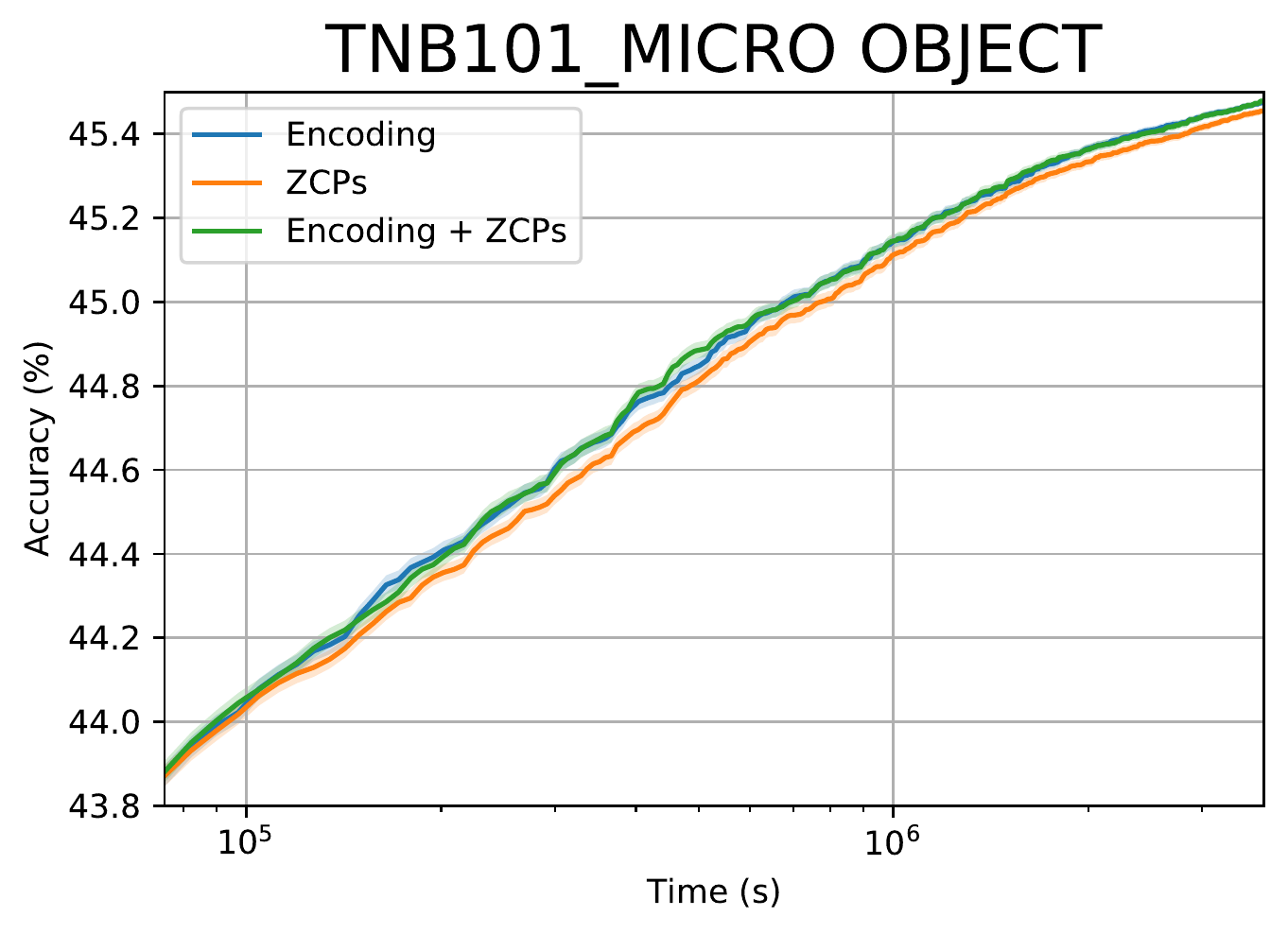}\\
    \includegraphics[width=.32\linewidth]{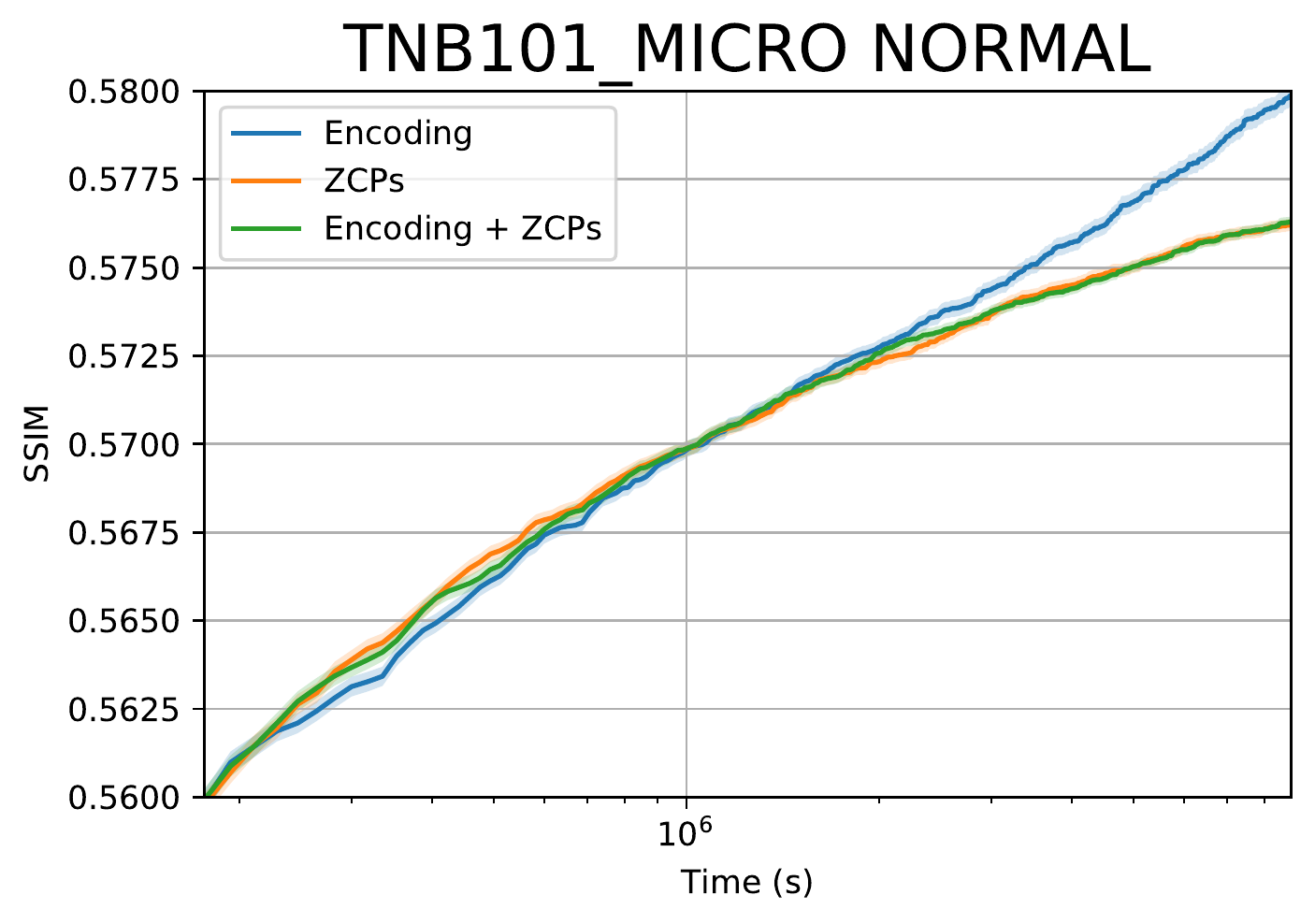}
    \includegraphics[width=.32\linewidth]{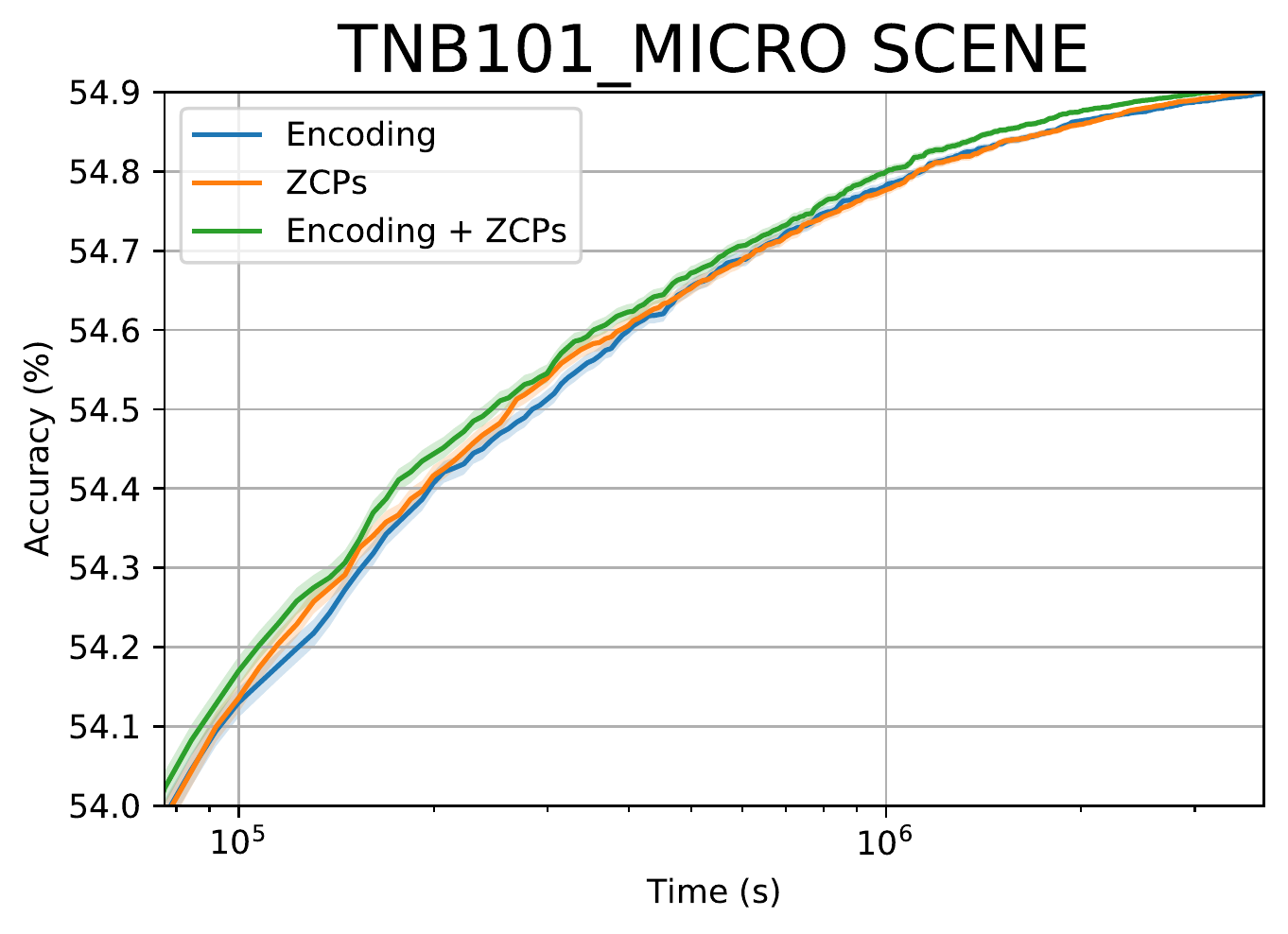}
    \includegraphics[width=.32\linewidth]{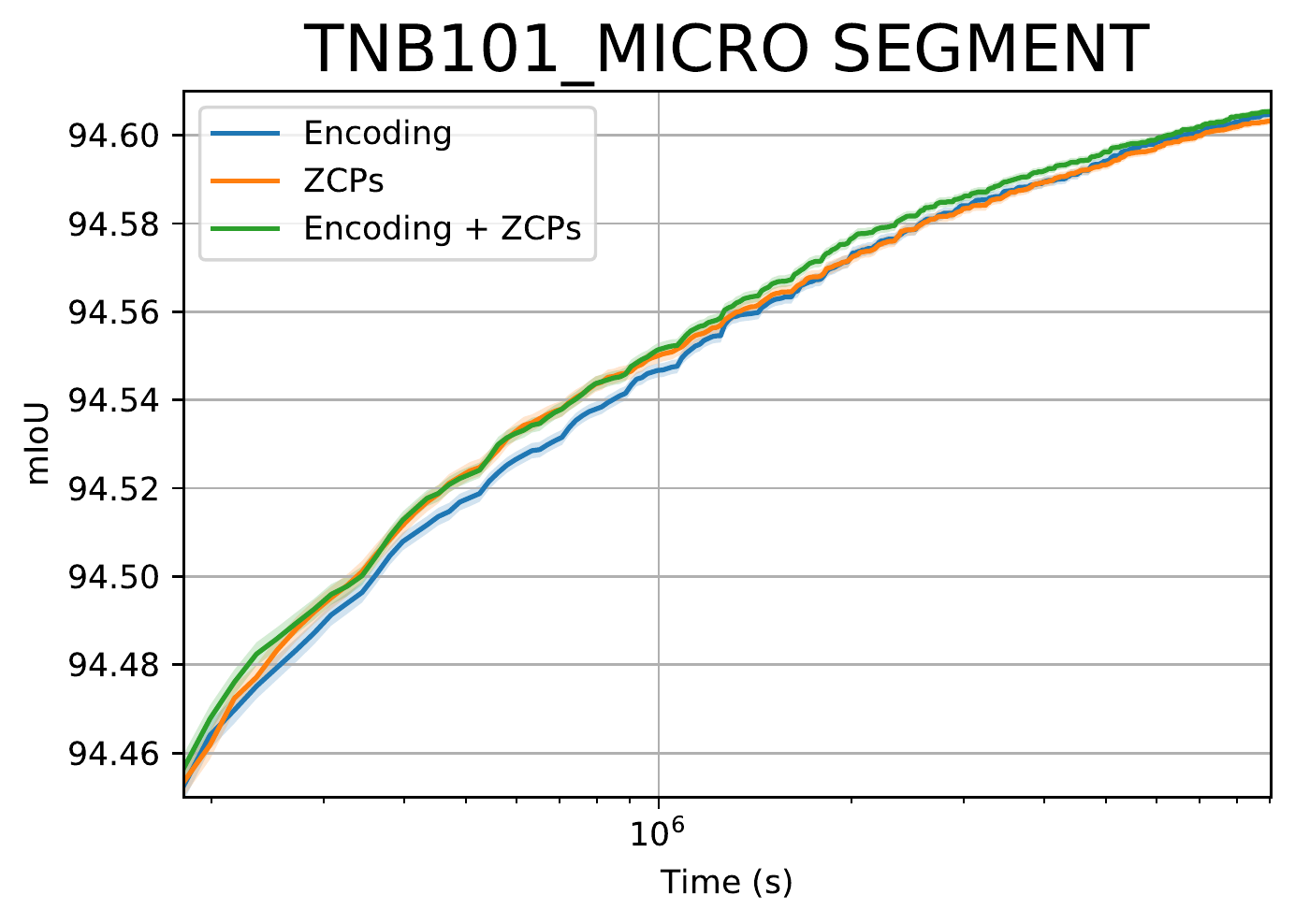}\\
    \includegraphics[width=.32\linewidth]{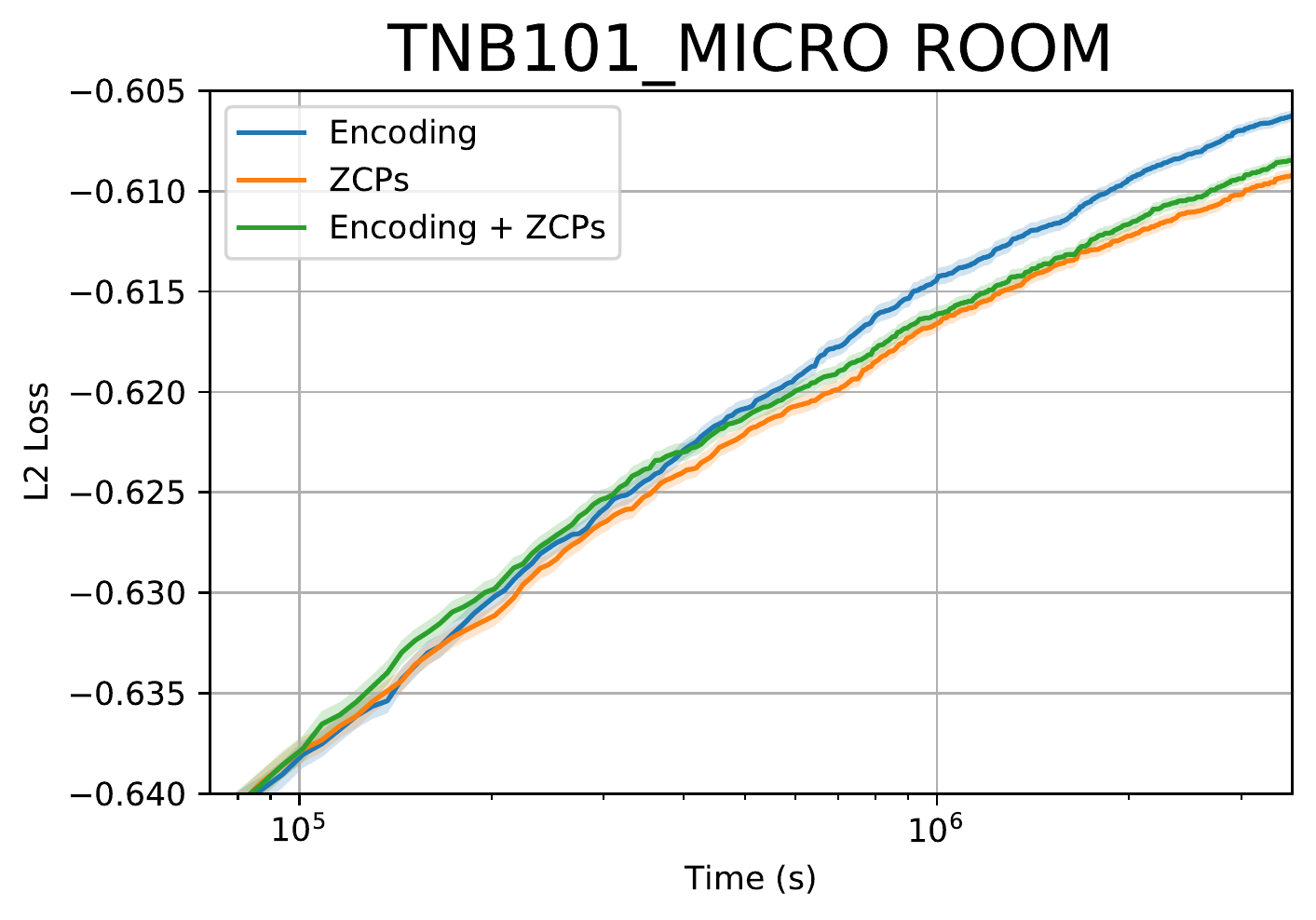}
    \includegraphics[width=.32\linewidth]{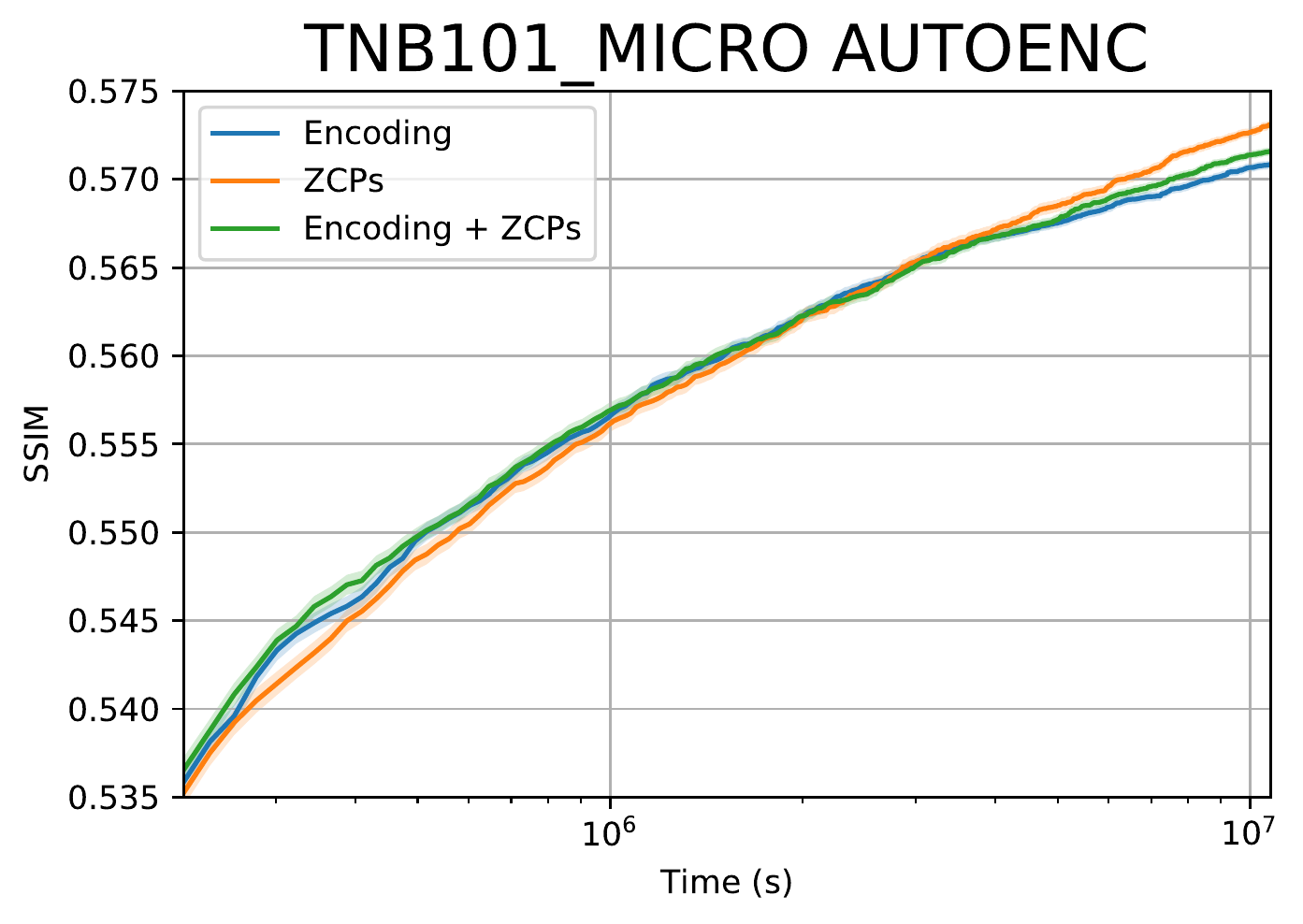}
    \caption{Performance of BANANAS with the vanilla XGBoost surrogate model vs. XGBoost using the additional ZC proxy scores (concatenated to the architecture encoding) as input.}
    \label{fig:bananas_all}
\end{figure}

\begin{figure}[ht]
    \centering
    \includegraphics[width=.32\linewidth]{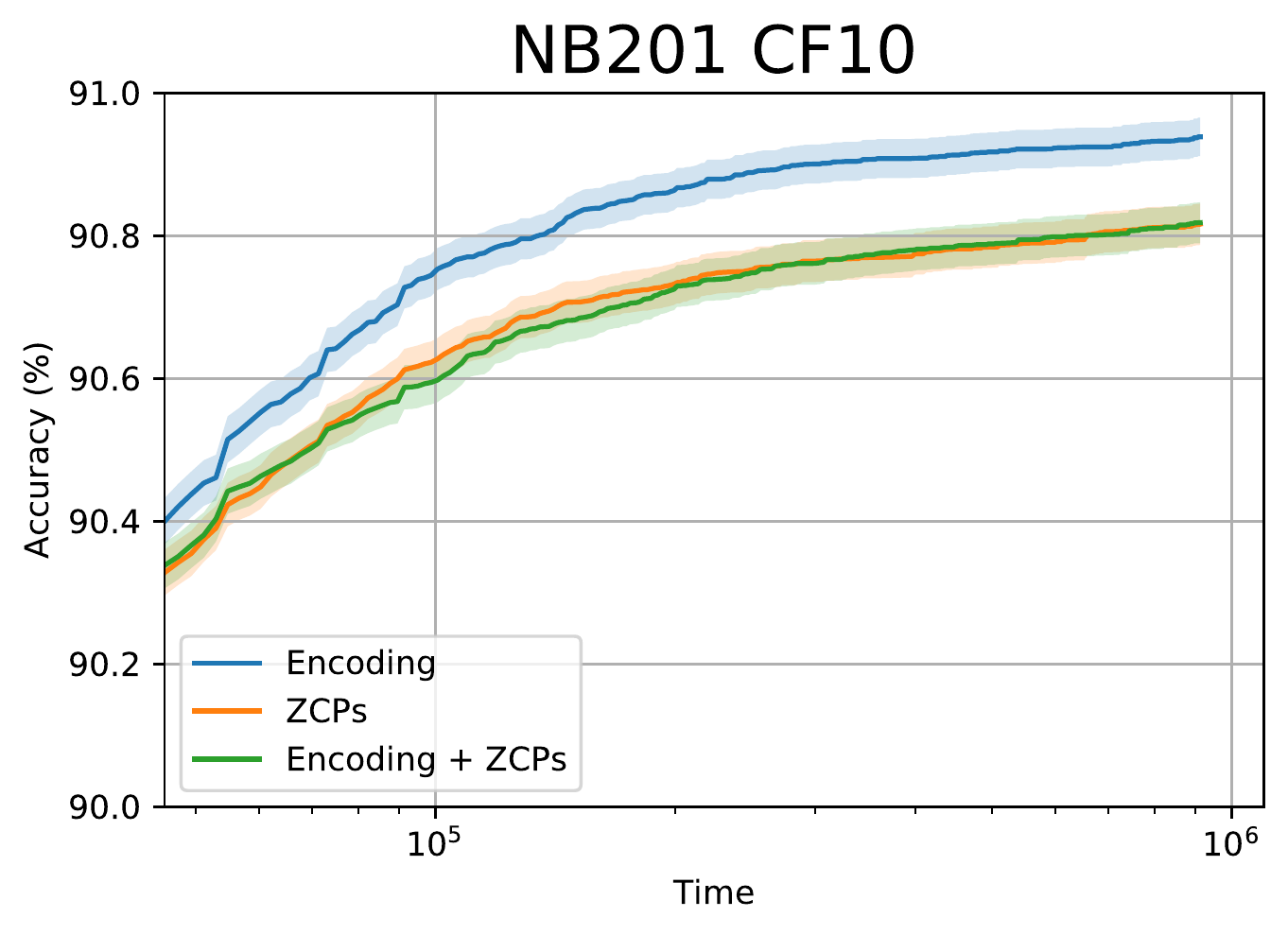}
    \includegraphics[width=.32\linewidth]{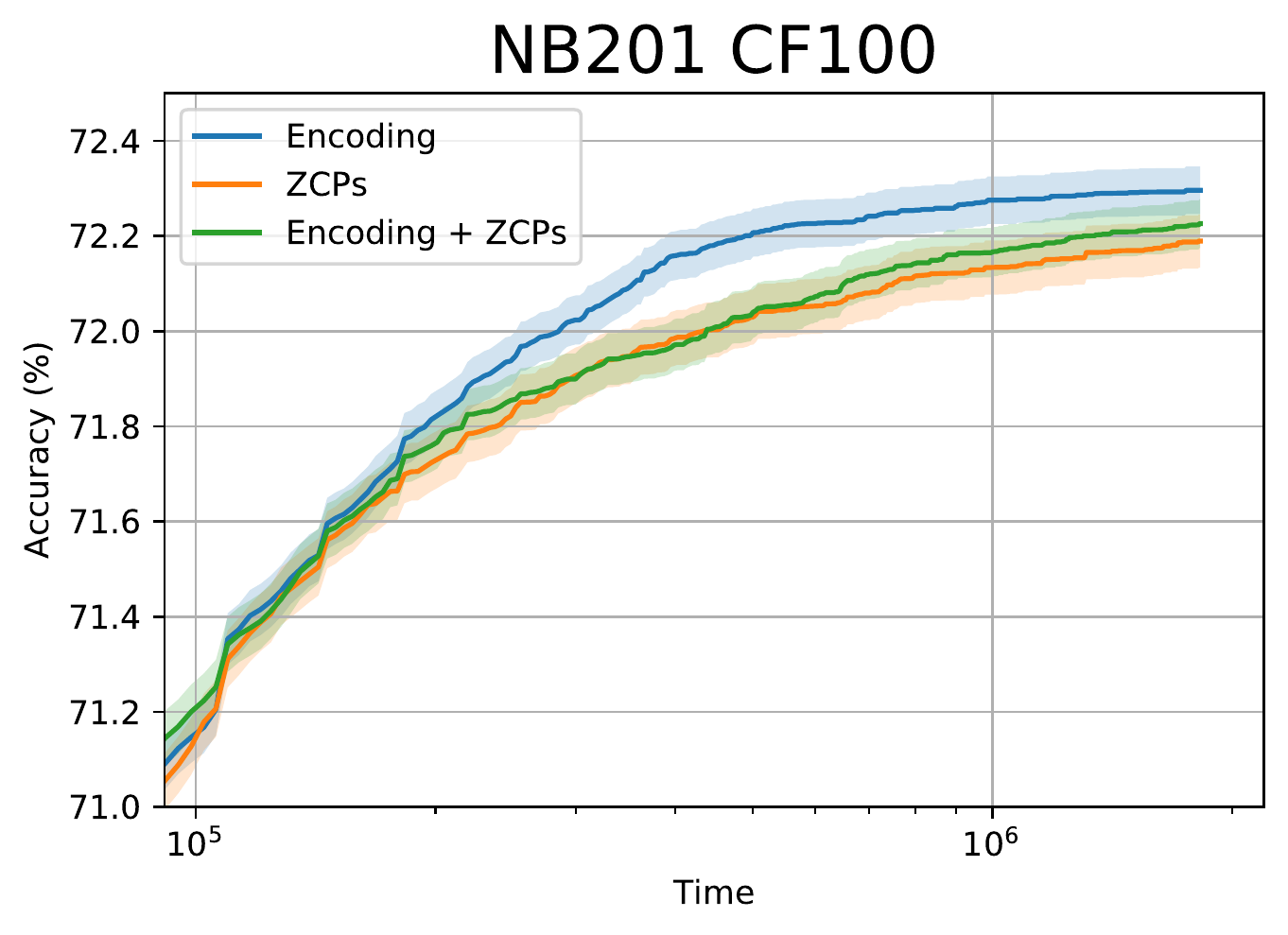}
    \includegraphics[width=.32\linewidth]{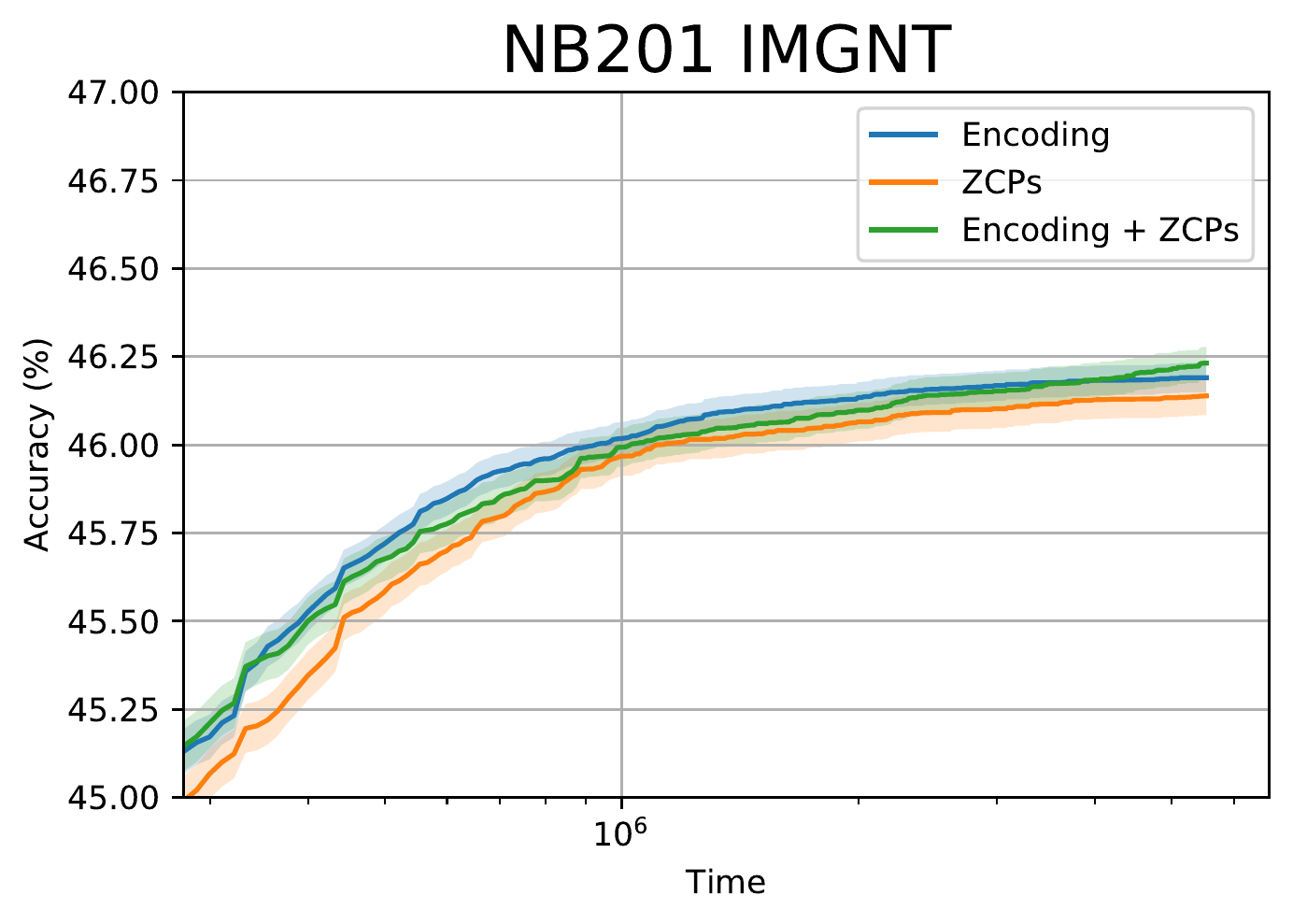} \\
    \includegraphics[width=.32\linewidth]{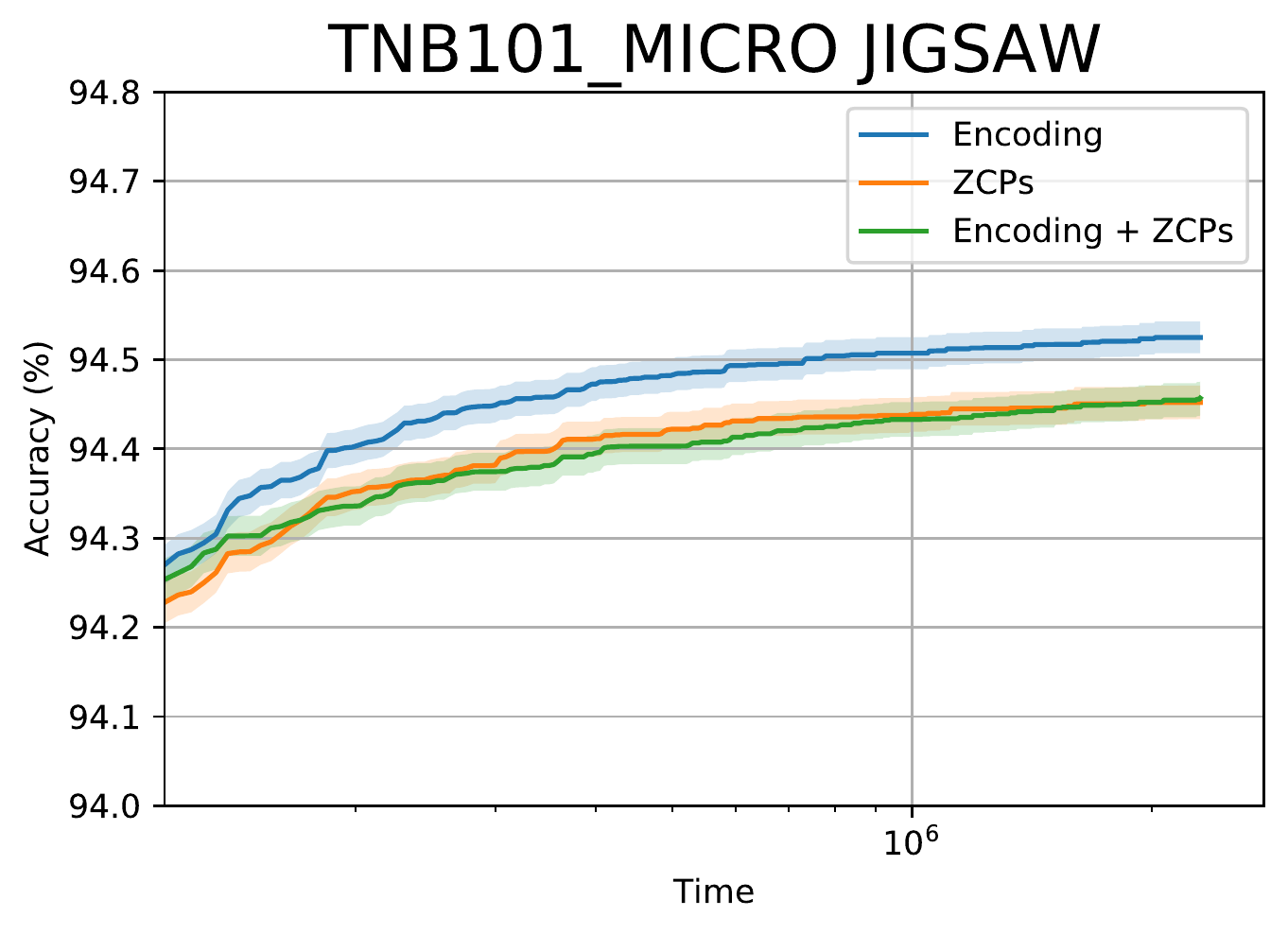}
    \includegraphics[width=.32\linewidth]{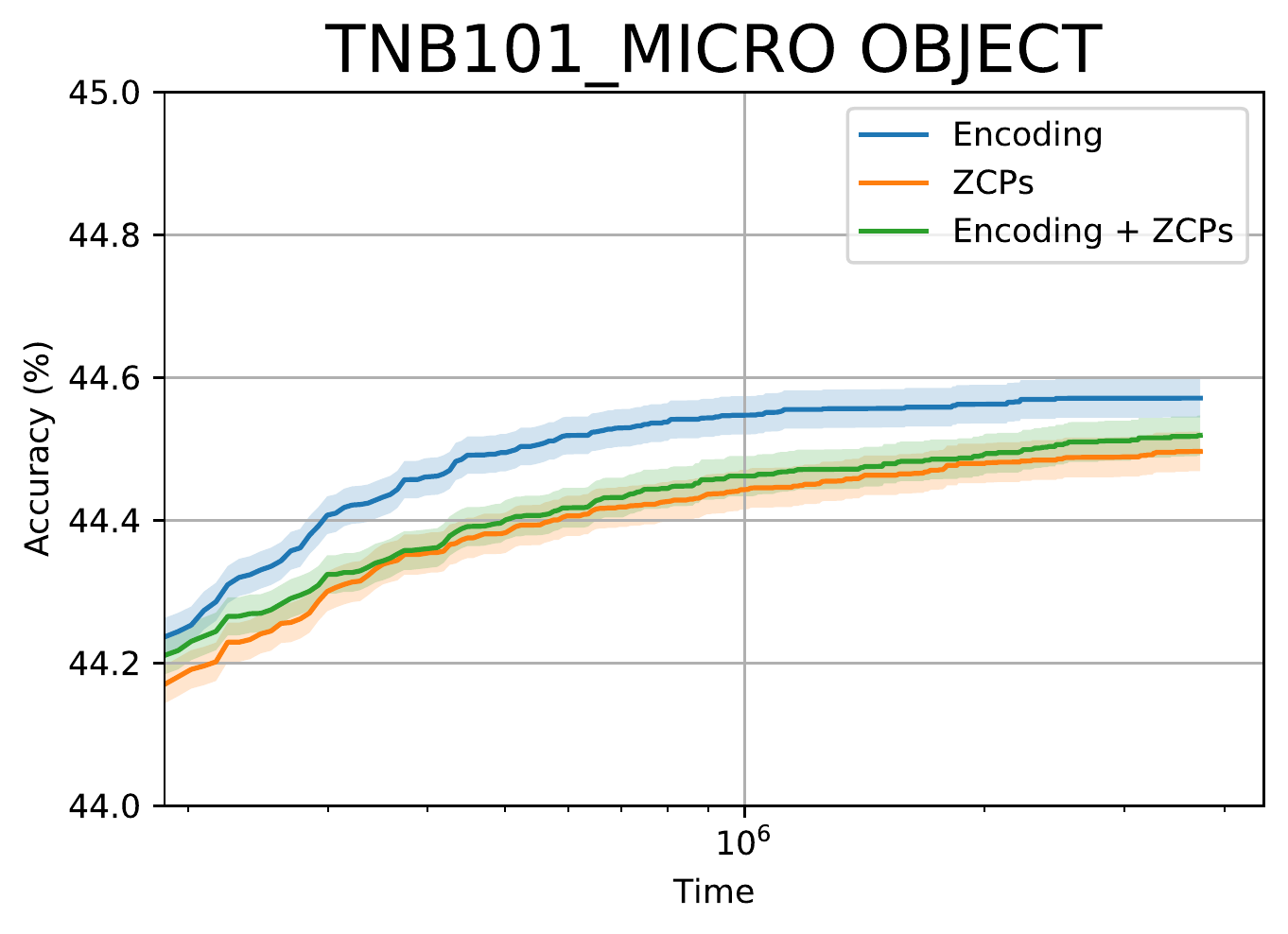}
    \includegraphics[width=.32\linewidth]{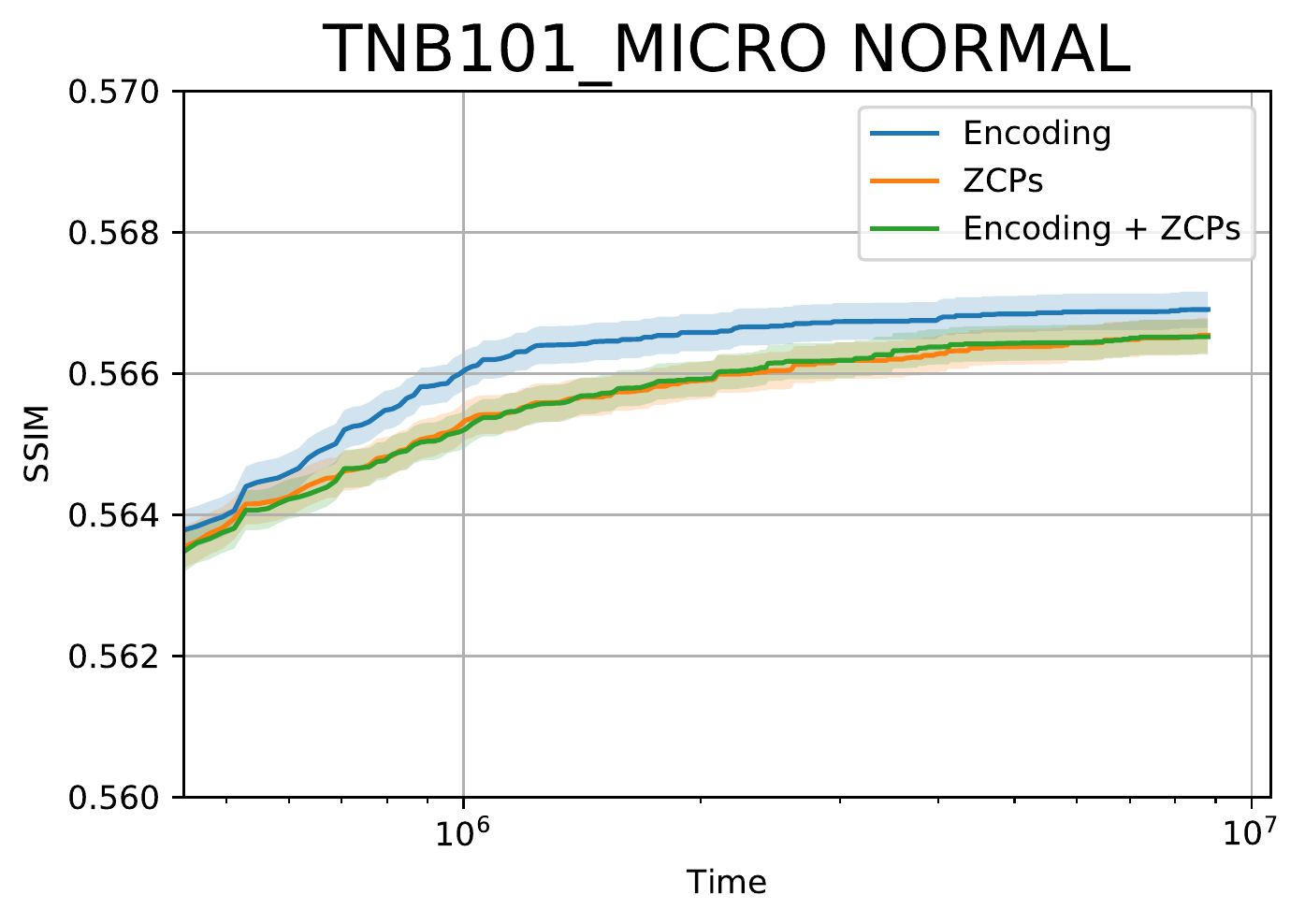}\\
    \includegraphics[width=.32\linewidth]{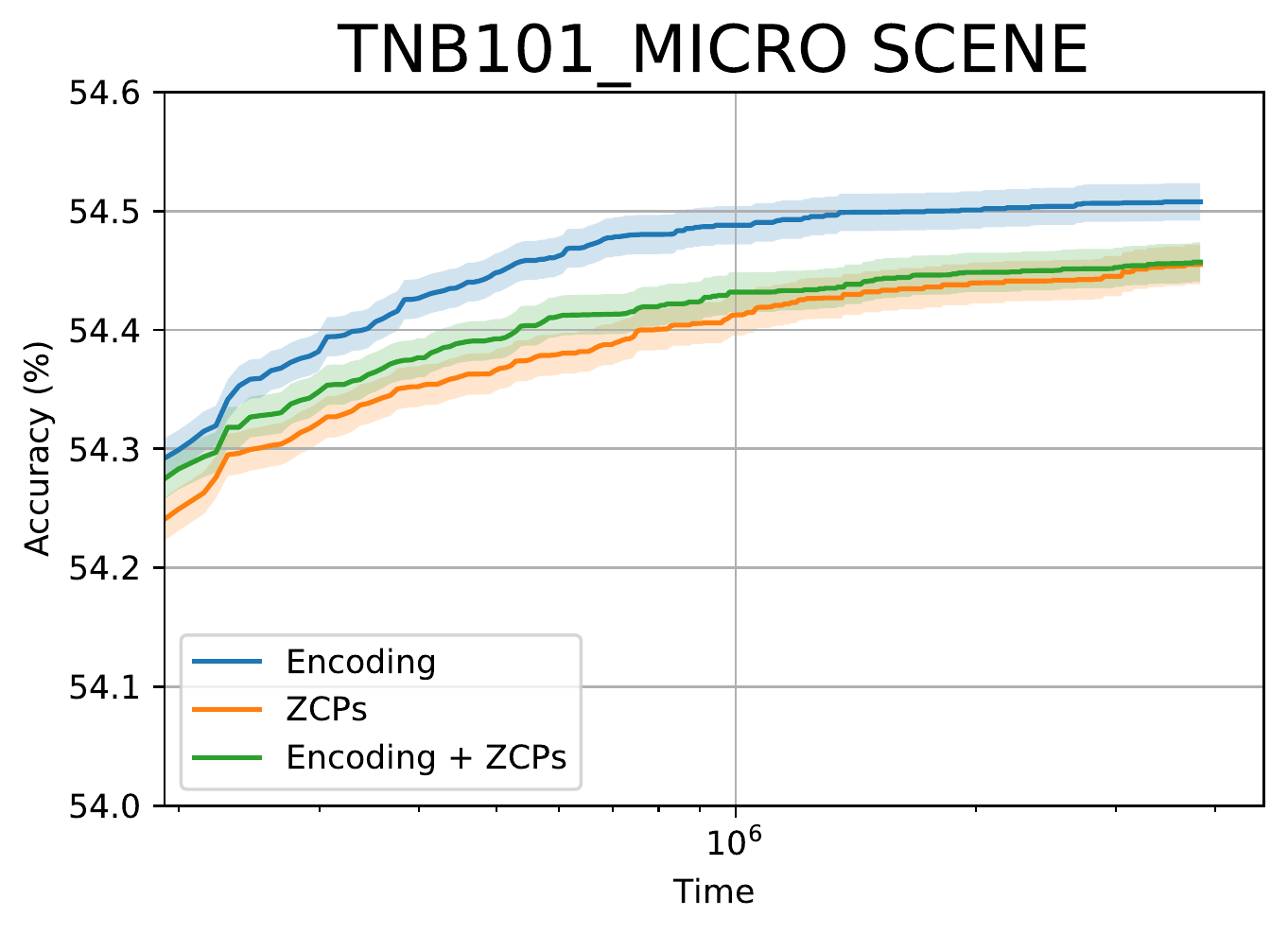}
    \includegraphics[width=.32\linewidth]{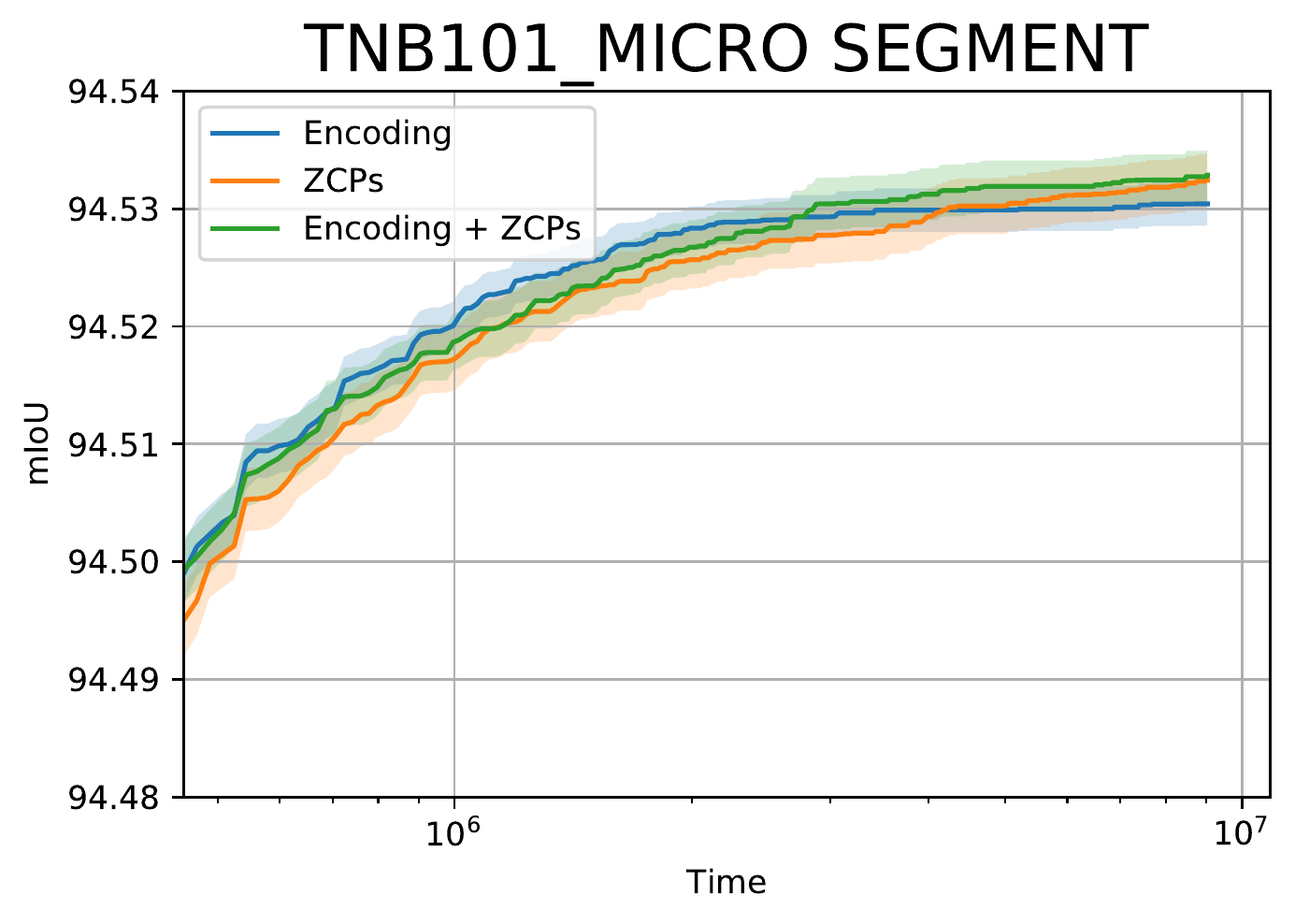}
    \includegraphics[width=.32\linewidth]{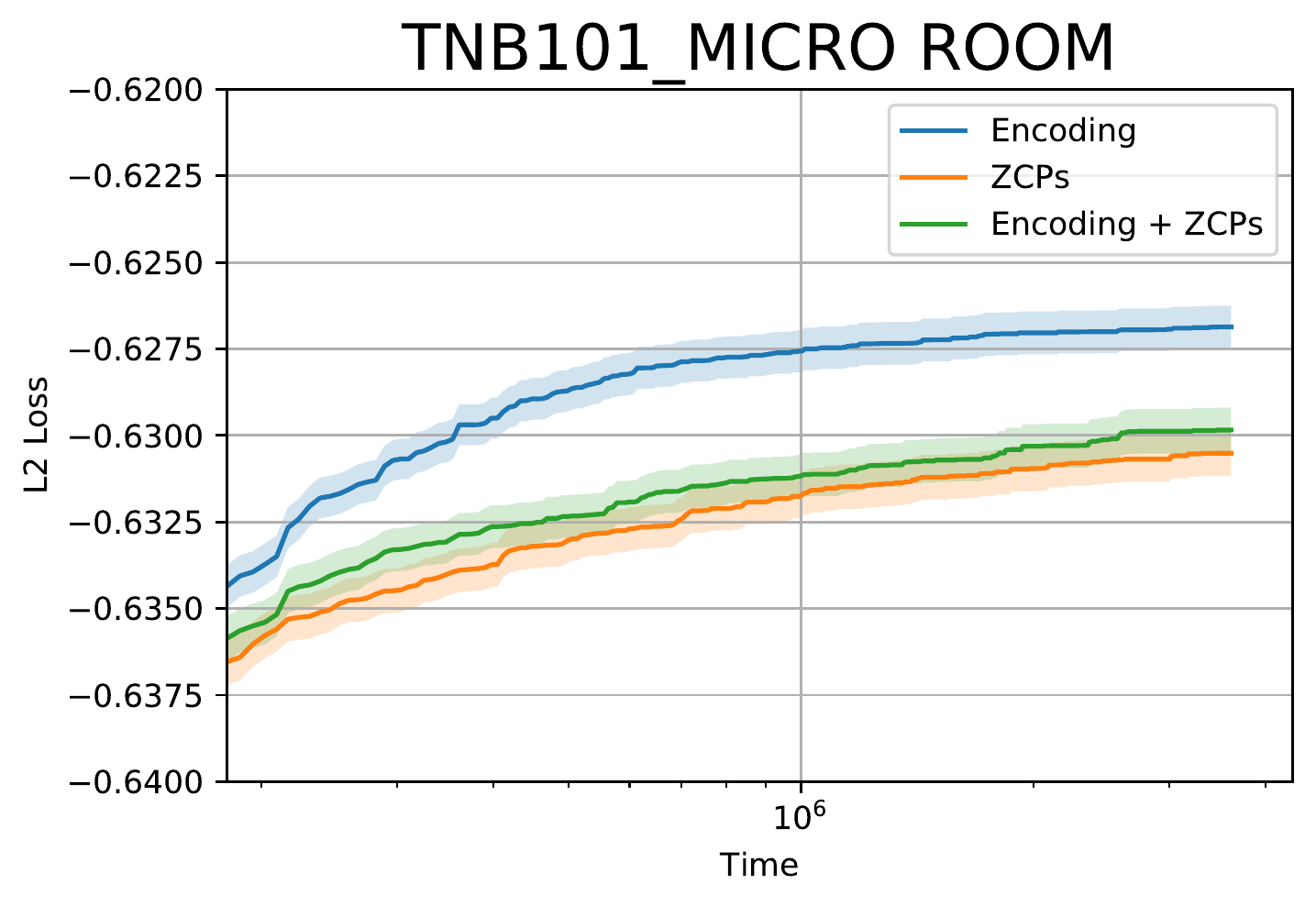}\\
    \includegraphics[width=.32\linewidth]{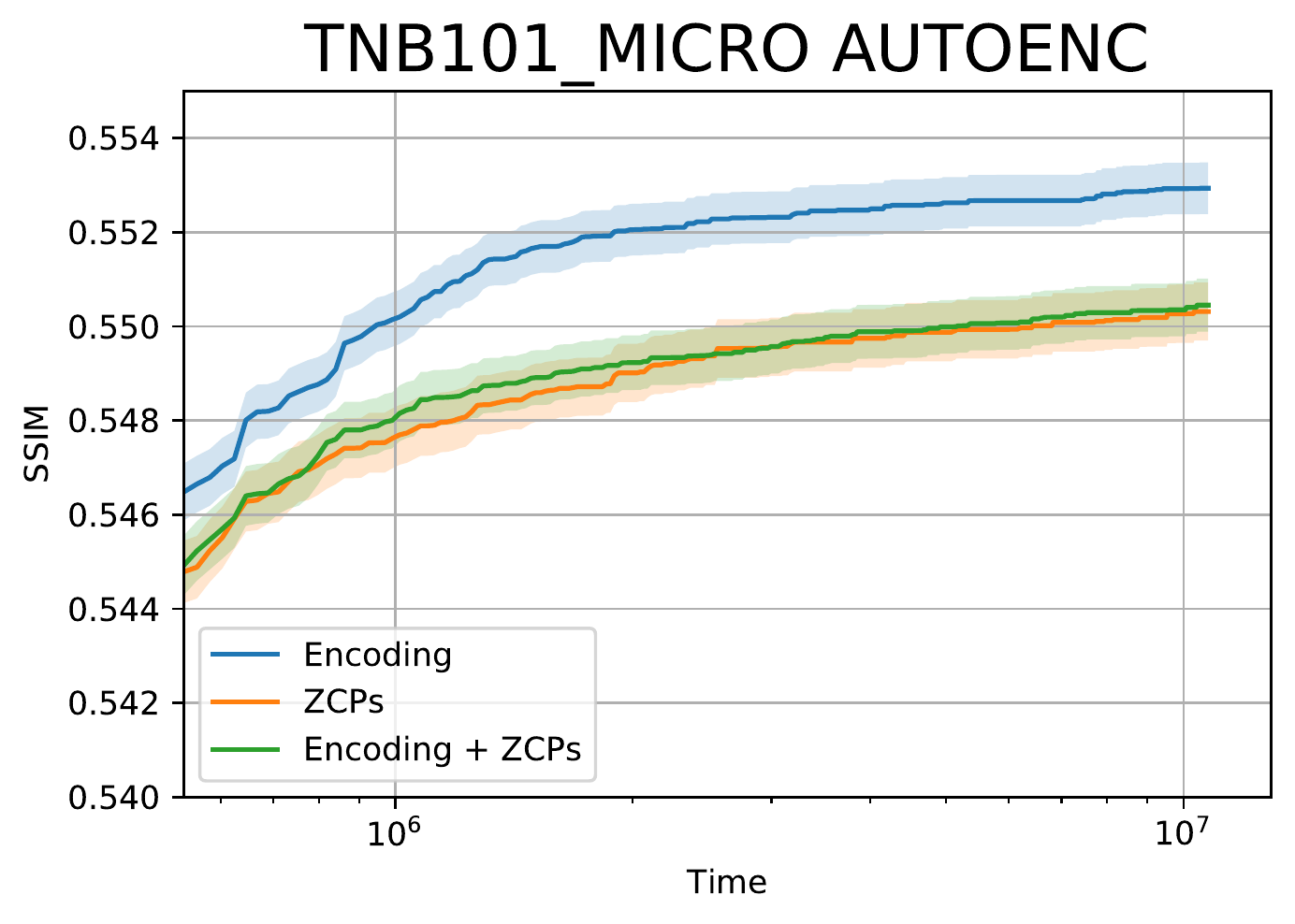}
    \caption{Performance of NPENAS with the vanilla XGBoost surrogate model vs. XGBoost using the additional ZC proxy scores (concatenated to the architecture encoding) as input.}
    \label{fig:npenas_all}
\end{figure}

Finally, we present more NAS results, extending the NAS results from Section \ref{sec:nas}.
In Figure \ref{fig:bananas_all}, we run BANANAS in the same setting as Section \ref{sec:nas}, on 11 benchmarks. We see that ZC proxies improve performance across the board.
In Figure \ref{fig:npenas_all}, we run the same experiment with NPENAS instead of BANANAS.
Note that since NPENAS requires a mutation step, we are only able to run it on complete benchmarks: NAS-Bench-201 and TransNAS-Bench-101 (in particular, not NAS-Bench-101 or NAS-Bench-301).

\section{ZC Proxy Competition} \label{app:competition}

\suite{} was used successfully in the Zero Cost NAS Competition at AutoML-Conf 2022. 
During the competition, participants developed new, better versions of ZC proxies in the \suite{} codebase.
The challenge was as follows: given $N$ models, the participant's ZC proxy will be used to rank the models for a specified task. The Kendall-Tau rank correlation is used to score the metric, averaged across three benchmarks in the test phase of the competition. 
The tasks in the development phase of the competition were NB201 with Ninapro and SVHN, NB301 with Ninapro and SVHN, and TNB101-Micro with Ninapro, SVHN, and Spherical-CIFAR100.
The tasks in the final test phase of the competition were NB101 with CIFAR10, NB201 with ImageNet16x120, NB301 with CIFAR10, TNB101-Macro with Object Classification, and TNB101-Micro with Object Classification.
The winning teams used a normalized version of \texttt{synflow}, a normalized version of \texttt{fisher}, and a product of \texttt{grad\_norm} and \texttt{params}.
For more information, see the competition homepage at \url{https://sites.google.com/view/zero-cost-nas-competition/home}.